%% file: main.tex
\documentclass{article}
\PassOptionsToPackage{square,comma,numbers,compress}{natbib}

\usepackage[preprint]{neurips_2023}

\usepackage[utf8]{inputenc} %
\usepackage[T1]{fontenc}    %
\usepackage[hidelinks,colorlinks=true]{hyperref}       %
\usepackage[dvipsnames]{xcolor}
\usepackage{url}            %
\usepackage{booktabs}       %
\usepackage{amsfonts}       %
\usepackage{nicefrac}       %
\usepackage{microtype}      %
\usepackage{xcolor}         %
\usepackage{graphicx}
\graphicspath{ {figures/} }
\usepackage{array}
\usepackage{times,graphicx,tabulary,multirow,xspace}
\usepackage{subfig}
\usepackage{changepage}
\usepackage{color, colortbl}
\definecolor{LightGray}{rgb}{0.92,0.92,0.92}
\definecolor{Gray1}{rgb}{0.95,0.95,0.95}
\definecolor{Gray2}{rgb}{0.9,0.9,0.9}

\definecolor{redhl}{HTML}{ea9999}
\definecolor{greenhl}{HTML}{d9ead3}
\definecolor{bluehl}{HTML}{c9daf8}
\definecolor{yellowhl}{HTML}{fff2cc}
\makeatletter
\DeclareRobustCommand\onedot{\futurelet\@let@token\@onedot}
\def\@onedot{\ifx\@let@token.\else.\null\fi\xspace}

\makeatother

\title{A Challenger to GPT-4V?\\Early Explorations of Gemini in Visual Expertise}

\author{
{\bf Chaoyou Fu$^{1}$$^{*}$$^{\spadesuit}$, Renrui Zhang$^{2,3}$$^{*}$, Zihan Wang$^{6}$$^{*}$}, Yubo Huang$^{4}$, Zhengye Zhang$^{4}$\\
{\bf Longtian Qiu$^{2}$, Gaoxiang Ye$^{5}$, Yunhang Shen$^{1}$, Mengdan Zhang$^{1}$} \\
{\bf Peixian Chen$^{1}$, Sirui Zhao$^{4}$, Shaohui Lin$^{6}$, Deqiang Jiang$^{1}$} \\
{\bf Di Yin$^{1}$, Peng Gao$^{2}$, Ke Li$^{1}$, Hongsheng Li$^{3}$, Xing Sun$^{1}$$^{\dagger}$} \\
\\
$^{1}$Tencent Youtu Lab, $^{2}$Shanghai AI Laboratory \\
$^{3}$CUHK MMLab, $^{4}$USTC, $^{5}$Peking University, $^{6}$ECNU\\
\and
\footnotesize{
$^*$~Equal Contribution \;
$^{\spadesuit}$~Project Leader \;
$^{\dagger}$~Corresponding Author \;
}
}

\begin{document}

\maketitle

\begin{abstract}
The surge of interest towards Multi-modal Large Language Models (MLLMs), e.g., GPT-4V(ision) from OpenAI, has marked a significant trend in both academia and industry. 
They endow Large Language Models (LLMs) with powerful capabilities in visual understanding, enabling them to tackle diverse multi-modal tasks. 
Very recently, Google released Gemini, its newest and most capable MLLM built from the ground up for multi-modality.
In light of the superior reasoning capabilities, \textit{\textbf{can Gemini challenge GPT-4V's leading position in multi-modal learning?}} In this paper, we present a preliminary exploration of Gemini Pro's visual understanding proficiency, which comprehensively covers four domains: fundamental perception, advanced cognition, challenging vision tasks, and various expert capacities. We compare Gemini Pro with the state-of-the-art GPT-4V to evaluate its upper limits, along with the latest open-sourced MLLM, Sphinx, which reveals the gap between manual efforts and black-box systems.
The qualitative samples indicate that, while GPT-4V and Gemini showcase different answering styles and preferences, they can exhibit \textbf{\textit{comparable visual reasoning capabilities}}, and Sphinx still trails behind them concerning domain generalizability. Specifically, GPT-4V tends to elaborate detailed explanations and intermediate steps, and Gemini prefers to output a direct and concise answer.
The quantitative evaluation on the popular MME benchmark, which is specifically designed for MLLM, also demonstrates the impressive multi-modal understanding performance of Gemini, and its potential to be a strong challenger to GPT-4V. 
Our early investigation of Gemini also observes some common issues of MLLMs concerning visual understanding, logical reasoning, and prompting robustness, indicating that \textbf{\textit{there still remains a considerable distance towards
artificial general intelligence}}. We hope this report may cast a new light on future MLLM research and application scenarios.
Our project for tracking the progress of MLLM is released at \url{https://github.com/BradyFU/Awesome-Multimodal-Large-Language-Models}.
\end{abstract}

{
  \hypersetup{linkcolor=black}
  \tableofcontents
  \label{sec:toc}
}

\clearpage
{
\hypersetup{linkcolor=black}
\addcontentsline{toc}{section}{List of Figures}
\listoffigures
\label{sec:lof}
}
\clearpage

\input{01-intro}

\clearpage
\input{02-perception}

\clearpage
\input{03-cognition}

\clearpage
\input{04-vision}

\clearpage
\input{05-expert}

\clearpage
\input{06-quantitative}

\clearpage
\input{07-limitations-conclusions}

\clearpage
{
\bibliographystyle{plain}
\bibliography{egbib}
}

\end{document}

%% file: 01-intro.tex
\section{Introduction}
\label{sec:01intro}
\subsection{Motivation and Overview}

Driven by big data and substantial computational power, the realm of large language models (LLMs)~\cite{brown2020language,gpt4,chowdhery2022palm,anil2023palm,touvron2023llama,hoffmann2022training} has garnered unprecedented enthusiasm and advancement, showcasing generalizability in a wide range of fields. 
Building upon this achievement, Multi-modal Large Language Models (MLLMs)~\cite{yin2023survey,huang2023language,driess2023palme,anas_awadalla_2023_7733589,gong2023multimodalgpt,zhu2023minigpt,liu2023visual,dai2023instructblip,ye2023mplug,han2023imagebind,gao2023llama,bai2023qwen} are emerging as a focal point of research in the new generation. They target on incorporating LLMs with additional sensory input, e.g., image~\cite{liu2023visual,zhang2023llama}, audio~\cite{wu2023next}, 3D~\cite{guo2023point}, etc. Conditioned on data of new modalities, MLLMs take a significant step forward on the path towards general artificial intelligence. Notably, GPT-4V(ision)~\cite{gpt4v,gpt4vblog,gpt4vdawn} from OpenAI is recognized as the most powerful MLLMs to date, surpassing a host of LLaMA-based~\cite{touvron2023llama} models, e.g., LLaMA-Adapter~\cite{zhang2023llama}, LLaVA~\cite{liu2023visual}, and MiniGPT-4~\cite{zhu2023minigpt}. However, very recently released by Google, Gemini~\cite{gemini} has emerged as a formidable challenger to GPT-4V, which exhibits significant multi-modal capabilities over different benchmarks~\cite{fu2023mme,lu2023mathvista,hendrycks2020measuring}. Given that the full potential of Gemini has not yet been fully tapped, in this paper, we conduct an early exploration by comparing Gemini with existing best-performing MLLM, i.e., GPT-4V, to reveal its multi-modal capabilities.

For a comprehensive evaluation, we carefully collect a bunch of qualitative samples covering different domains in multi-modal understanding. 
Two existing representative MLLMs are selected as baselines. The first is GPT-4V, representing the current highest standard in the field, which assesses the upper limits of Gemini. The second is Sphinx~\cite{lin2023sphinx}, a state-of-the-art LLaMA-based MLLM, exploring how much the performance gap is between open-sourced models and closed-sourced systems. Specifically, the qualitative samples can be categorized into four visual domains as follows:

\begin{enumerate}
    \item \textbf{Fundamental Perception.} (Section~\ref{sec:02perception})
    This dimension focuses on the basic ability of MLLMs to perceive and understand visual concepts, without the need for complex reasoning. 
    It can be subdivided into three key aspects: object-centric, scene-level, and knowledge-based perception.
    Therein, object-centric perception assesses the model's capacity to recognize and interpret the characteristics of individual objects within a visual context, exemplified by tasks such as spatial relation recognition, object counting, difference spotting, etc. 
    In contrast, scene-level perception evaluates the understanding of entire scenes from a global perspective, demonstrating the model's proficiency in image and video captioning. 
    Finally, knowledge-based perception reveals the model's accumulation and application of specific knowledge across various domains. It encompasses commonsense knowledge, scientific knowledge, cultural customs, and world memory, which respectively cover the content of everyday scenarios, academic disciplines, cultural practices, and global entities.
    
    \item \textbf{Advanced Cognition.} (Section~\ref{sec:04Cognition}) The samples in advanced cognition require MLLMs to process more complicated visual information and conduct multi-modal reasoning for problem-solving. The related tasks include text-rich and abstract visual reasoning, science problem solving, emotion understanding, and game playing. Text-rich tasks evaluate the OCR performance of textual content for table and chart reasoning, and the code generation capability conditioned on different visual inputs. Abstract visual reasoning refers to the non-verbal test assessing general intelligence and abstract reasoning, such as the Wechsler Adult Intelligence Scale and Raven’s Progressive Matrices. Science problem-solving, e.g., mathematics and physics, has become a vital metric for measuring MLLMs' comprehension of quantitative and logical knowledge, involving complex multi-step and chain-of-thought (CoT) reasoning.
    Moreover, emotion understanding focuses on the detection of underlying emotional information within visual contexts, and game playing evaluates strategic thinking and rule-following abilities in games like Sudoku.

    \item \textbf{Challenging Vision Tasks.} (Section~\ref{sec:04vison}) In this part, we aim to evaluate how MLLMs perform in some challenging vision tasks beyond general visual question-answering, such as object detection, referring expression comprehension, phrase localization, video temporal reasoning, and so on. These tasks require the in-depth visual perception and understanding capabilities of MLLMs. The performance of MLLMs can indicate their potential to serve as multi-purpose vision generalists.
    
    \item \textbf{Expert Capacity.} (Section~\ref{sec:07expert}) The final dimension evaluates the model's proficiency in several specialized fields. The scenarios include medical imaging, defect detection, stock prediction, autonomous driving, and surveillance security. Each of these areas tests the model's application of its learned knowledge and cognitive skills in a professional context, such as diagnosing diseases from medical images or predicting market trends in stock trading. This demonstrates the generalization capacity of MLLMs from more diverse perspectives.
\end{enumerate}

Besides qualitative samples, we report quantitative results of Gemini on the popular MME benchmark~\cite{fu2023mme} in Section~\ref{sec:07quantitative}, which comprehensively evaluates MLLMs in 14 subtasks from both perception and cognition perspectives.

\subsection{Evaluation Suite}

\subsubsection{Prompt Technique}
GPT-4V has been demonstrated to support a diverse range of prompt techniques \cite{gpt4vdawn}, from simple instruction following \cite{ouyang2022training,mishra2021cross,wei2021finetuned,sanh2021multitask} to in-context few-shot learning \cite{brown2020language,tsimpoukelli2021multimodal,alayrac2022flamingo}. 
This inspires us to design the following prompt techniques.
\noindent\textbf{Simple instruction following.} 
A simple instruction directly expresses the user's intention, such as ``\textit{Describe this image}'' or ``\textit{Who is this person in the poster?}''. 
Existing MLLMs~\cite{gong2023multimodalgpt,zhu2023minigpt,liu2023visual,dai2023instructblip,ye2023mplug} are generally capable of following instructions, enabling us to utilize the simple instruction to accomplish most tasks effectively. 
We adopt simple instructions to prompt models on most of the tasks. 
Figures \ref{spatial_1} and \ref{count} are typical examples, respectively. 

\noindent\textbf{Visual referring prompt.}  
In many cases, a simple visual marker can more effectively convey the user's interest in a specific spatial region to MLLMs than detailed and lengthy text, 
as shown in Figure \ref{tracking}.
In addition to the visual markers used as visual prompts in \cite{llava_interactive,gpt4vdawn}, we also experiment with physical objects to guide the model's understanding of the referring items, such as a finger or a pen, as illustrated in the bottom part of Figure \ref{animal-2}.
Compared to prompting the model with visual markers, using real objects as prompts is more practical in real-time interaction scenarios.

\noindent\textbf{Chain-of-Thought (CoT) prompt.} 
For problems involving complex logical reasoning, we use CoT techniques \cite{wei2022chain,kojima2022large} to guide the model to provide a final response through a series of more logical thought processes, which is shown in Figure \ref{table-4}.

\noindent\textbf{In-context few-shot learning.}  
In certain scenarios where simple text instructions fail to completely demonstrate the task, we employ in-context few-shot learning \cite{brown2020language,tsimpoukelli2021multimodal,alayrac2022flamingo} for better prompting. By providing a few in-context examples at inference time, the model can infer intentions from these examples, thus facilitating the generation of the desired outputs, which is shown in Figure \ref{spatial_2}.

\subsubsection{Sample Collection}
\noindent\textbf{Avoiding sample leakage.} 
We endeavor to ensure that the collected qualitative images and text are unseen by the models to prevent responses that merely reflect memories of the training data. 
All the texts in the query are constructed from scratch.
The image sources include manually created drawings, offline photographs, Internet images, and some existing datasets \cite{caesar2020nuscenes,bergmann2019mvtec,demner2016preparing,song2021pareidolia}. 
For the Internet images, we strive to collect those with timestamps postdating November 2023.

\noindent\textbf{Diverse difficulty.} For each task, we collect samples of varying difficulty levels, e.g., from fundamental perception and cognition to the more challenging vision and expert tasks. In this way, we can not only demonstrate the potential of MLLMs to complete the tasks, but also touch their ability boundaries through some obvious mistake patterns.

%% file: 02-perception.tex
\section{Fundamental Perception}
\label{sec:02perception}

Fundamental perception, in the context of multi-modal large models, refers to the model's ability to process and interpret sensory data, primarily visual, to create a coherent understanding of the environment it perceives. The proficiency in perception directly influences a model's capability in higher-order tasks, as it determines how accurately and effectively the model can acquire and process raw visual input.

In Section \ref{sec02:subsec:object}, we will explore the object-centric perception task, such as spatial relationship recognition, object counting, and difference spotting.  
In Section \ref{sec02:subsec:scene}, we will examine the models' capacity for interpreting the entire scenes on diverse domains. 
In Section \ref{sec02:subsec:knowledge}, we will investigate the models' ability to comprehend visual information via the application of knowledge, which encompasses commonsense, subject knowledge, multicultural customs, and world memory.

\subsection{Object-Centric Perception}\label{sec02:subsec:object}
\begin{figure*}[!ht]
  \centering 
  \makebox[\textwidth][c]{\includegraphics[width=1.0\textwidth]{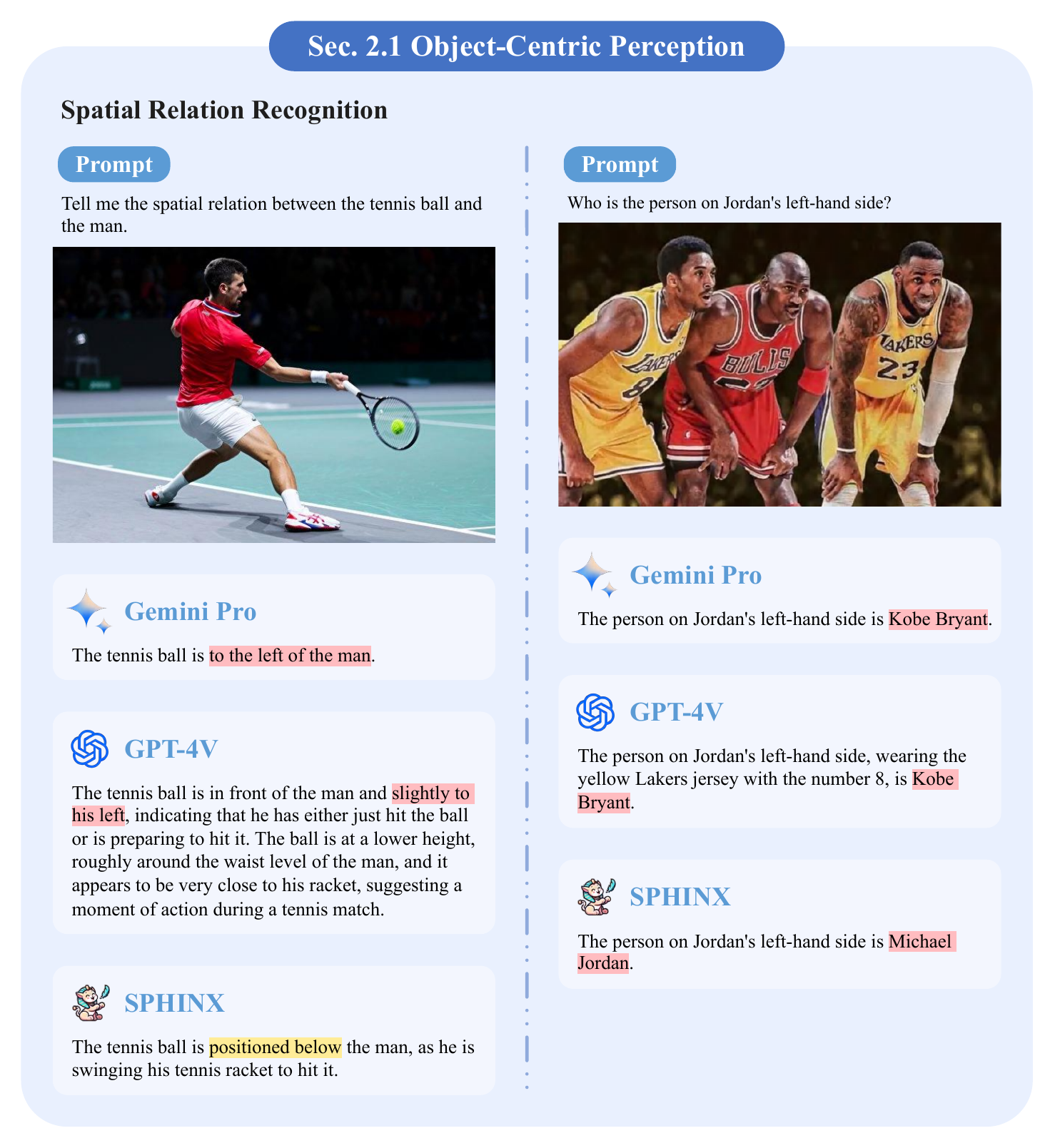}}
  \caption[Section~\ref{sec02:subsec:object}: spatial relation recognition.]{Results on spatial relation recognition. None of the three models can answer correctly. \colorbox{red!30}{Red} highlights the wrong answer. \colorbox{yellow!60}{Yellow} highlights the incompetence in performing the task. Refer to Section \ref{sec02:subsec:object} for detailed discussions.}
  \label{spatial_1}
\end{figure*}

\begin{figure*}[!ht]
  \centering 
  \makebox[\textwidth][c]{\includegraphics[width=1.0\textwidth]{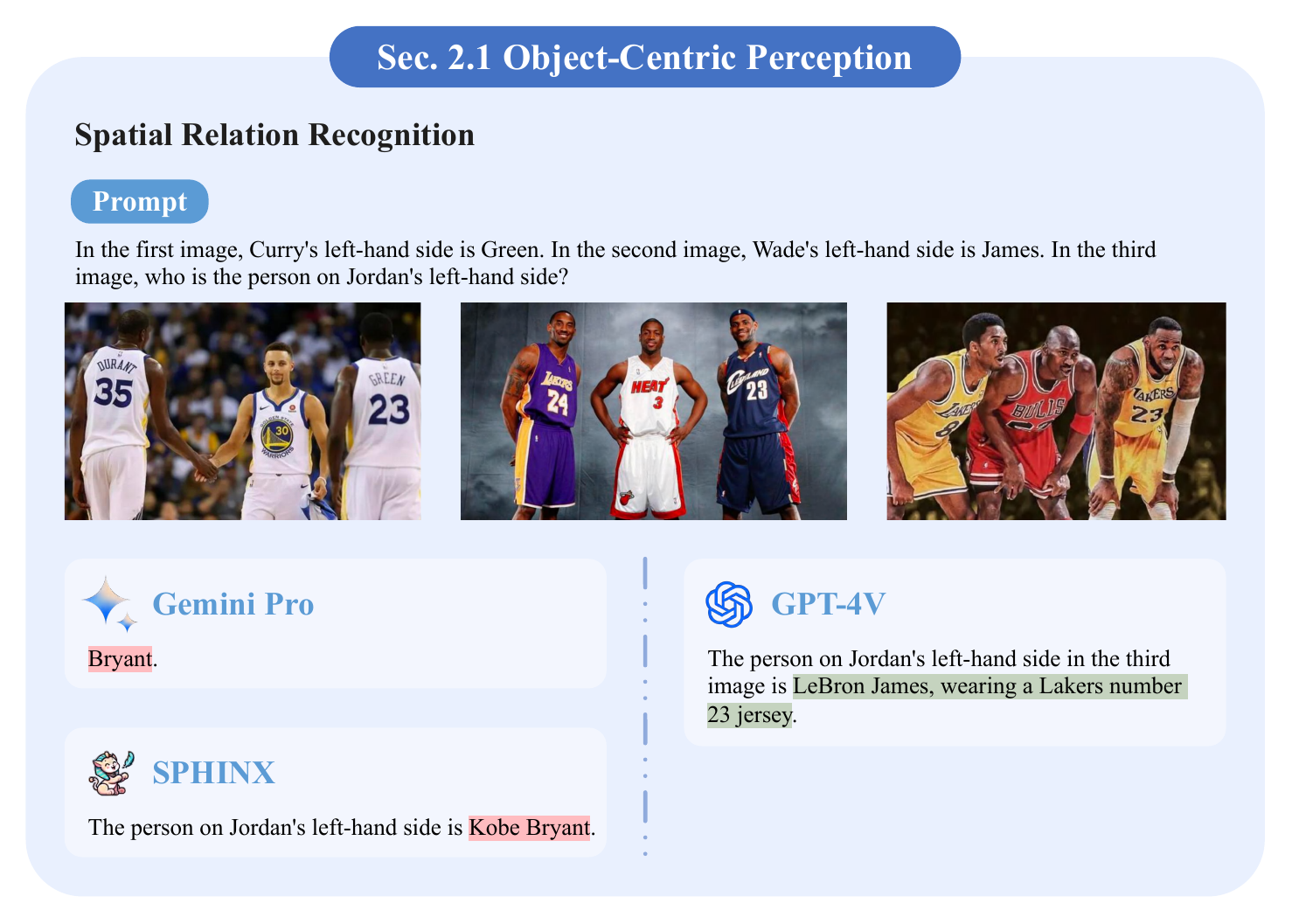}}
  \caption[Section~\ref{sec02:subsec:object}: spatial relation recognition.]{Results on spatial relation recognition. By providing two examples, GPT-4V accurately identifies spatial relationships, whereas the other two models still respond incorrectly. \colorbox{greenhl!85!black}{Green} highlights the right answer. \colorbox{red!30}{Red} highlights the wrong answer. Refer to Section \ref{sec02:subsec:object} for detailed discussions.}
  \label{spatial_2}
\end{figure*}

\textbf{Spatial relationship recognition.} 
We investigate the models' capability to comprehend spatial relationships. 
We find that it seems difficult for the models to identify left and right.
For instance, in Figure \ref{spatial_1}, the individual on the left-hand side of Jordan is James. 
However, the responses from Gemini and GPT4-V are both Kobe, while Sphinx's response is Jordan.
In our endeavor, we employ in-context few-shot learning techniques to aid the model in comprehending the concept of `left-hand'. 
As depicted in Figure \ref{spatial_2}, we provide two image examples to instruct the model on what constitutes the `left-hand'. 
However, only GPT-4V successfully learns the concept, while Gemini and Sphinx still can not distinguish between left and right.

\begin{figure*}[!ht]
  \centering 
  \makebox[\textwidth][c]{\includegraphics[width=1.15\textwidth]{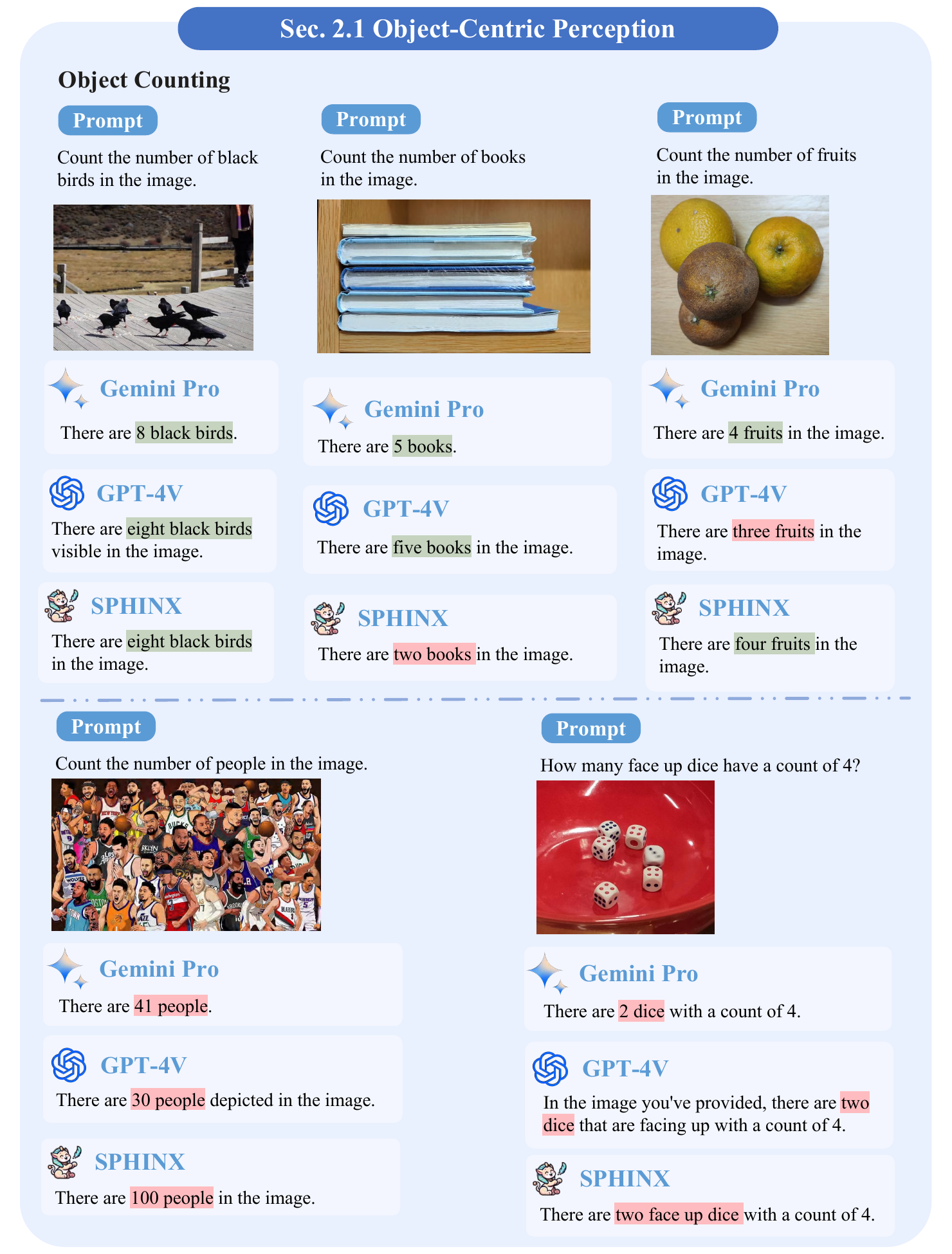}}
  \caption[Section~\ref{sec02:subsec:object}: object counting.]{Results on object counting. \colorbox{greenhl!85!black}{Green} highlights the right answer. \colorbox{red!30}{Red} highlights the wrong answer. Refer to Section \ref{sec02:subsec:object} for detailed discussions.}
  \label{count}
\end{figure*}

\textbf{Object counting.}
Figure \ref{count} shows the models' ability to count objects. 
It is observed that for simple samples, the performance of the open-source model Sphinx closely aligns with that of the two closed-source models, which is shown in the first three cases. 
However, as shown in the fourth example, when the images contain an excess of visual elements, all three models tend to make mistakes. 

\begin{figure*}[!ht]
  \centering 
  \makebox[\textwidth][c]{\includegraphics[width=1.2\textwidth]{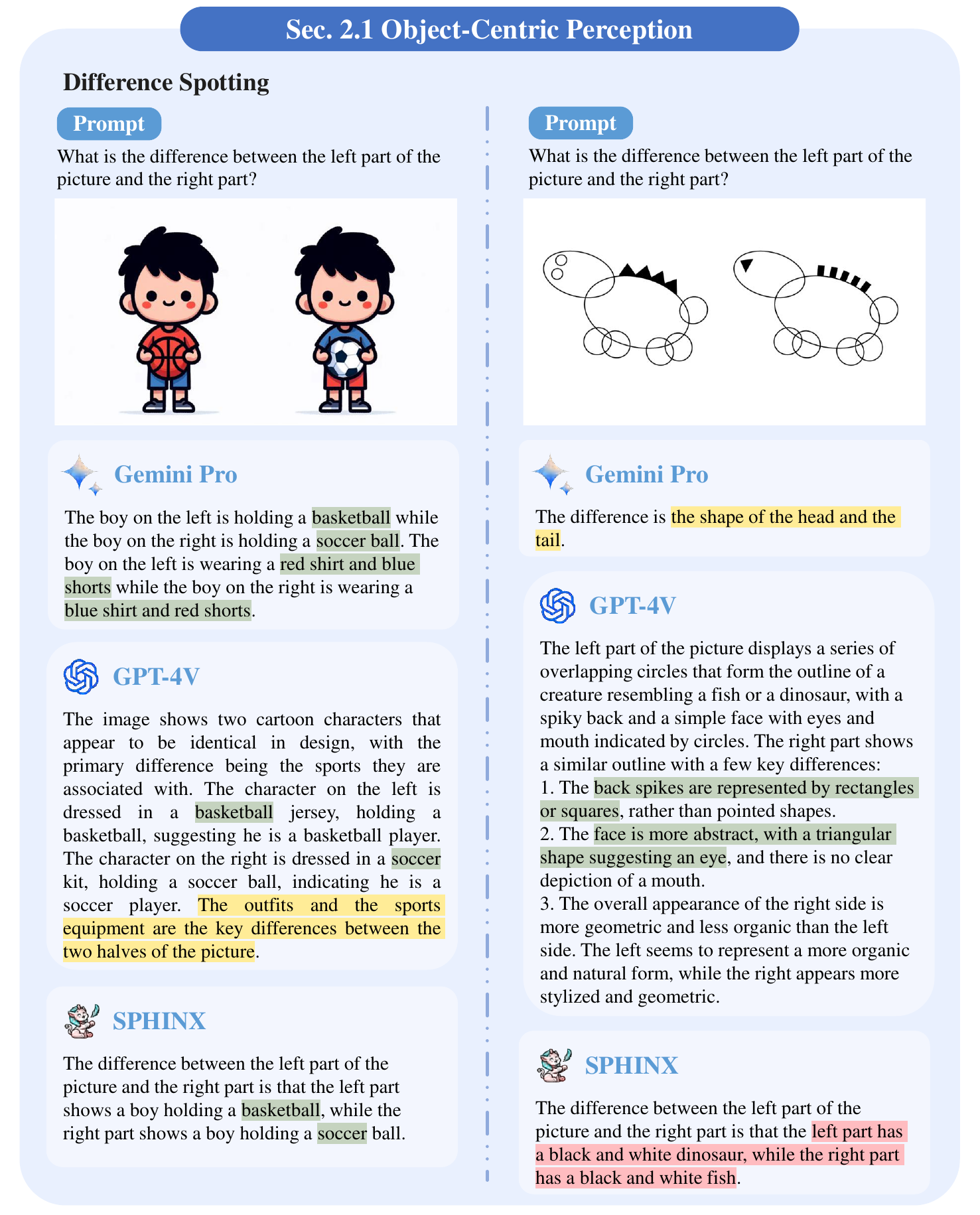}}
  \caption[Section~\ref{sec02:subsec:object}: difference spotting.]{Results on difference spotting. \colorbox{greenhl!85!black}{Green} highlights the right answer. \colorbox{red!30}{Red} highlights the wrong answer. \colorbox{yellow!70!yellowhl}{Yellow} highlights the incompetence in performing the task. Refer to Section \ref{sec02:subsec:object} for detailed discussions.}
  \label{diff_1}
\end{figure*}

\begin{figure*}[!ht]
  \centering 
  \makebox[\textwidth][c]{\includegraphics[width=1.1\textwidth]{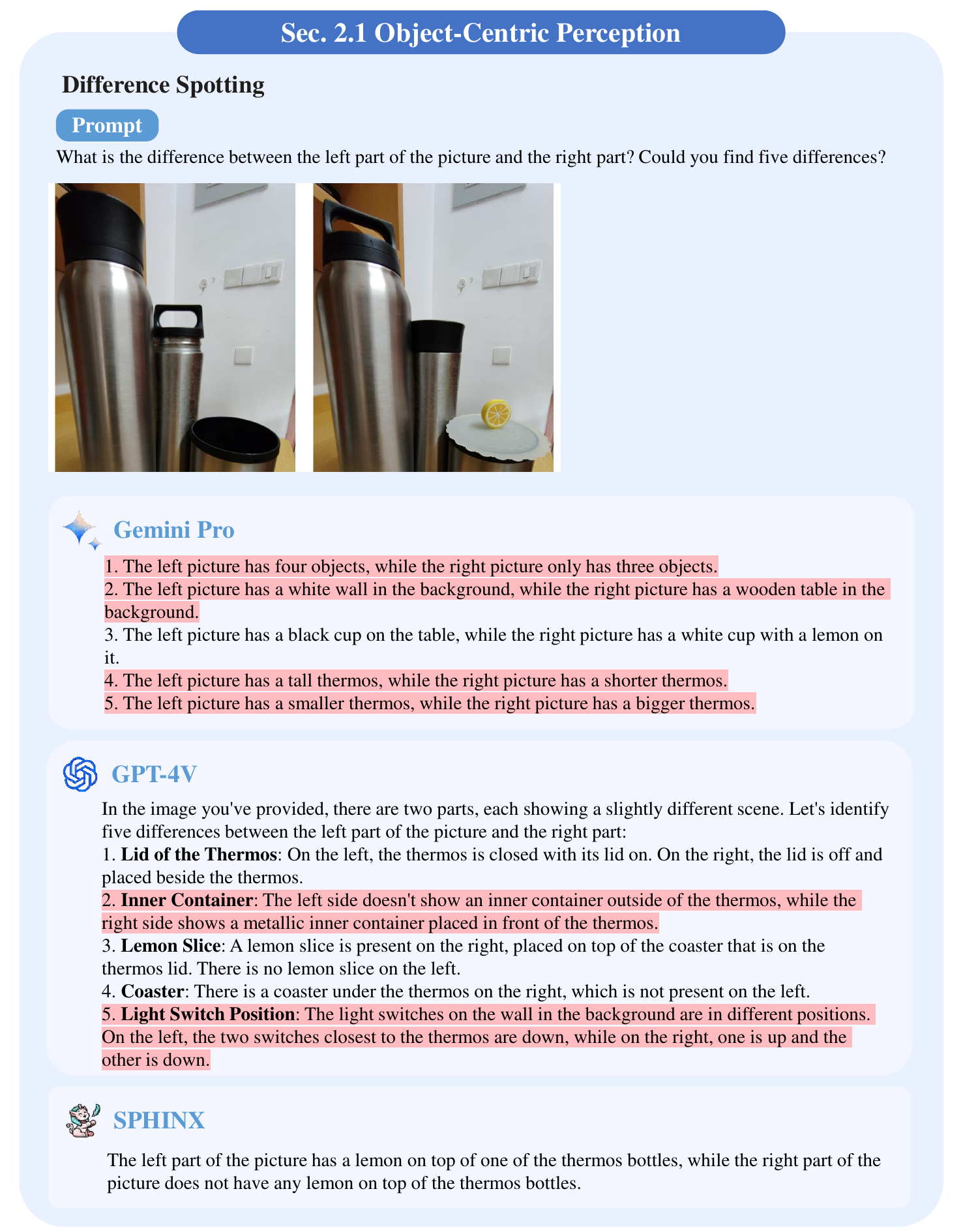}}
  \caption[Section~\ref{sec02:subsec:object}: difference spotting.]{Results on difference spotting. In the two images, there are only three differences, yet we ask the model to identify five. Both Gemini and GPT-4V attempt to accommodate our request and provide five differences, but neither performs well. \colorbox{red!30}{Red} highlights the wrong answer. Refer to Section \ref{sec02:subsec:object} for detailed discussions.}
  \label{diff_2}
\end{figure*}

\textbf{Difference spotting.}
In Figures \ref{diff_1}-\ref{diff_2}, we present the model's capacity to spot differences in cartoon images, sketches, and actual photographs. 
We observe that all models possess the potential to perceive the fine-grained differences between images, although their performance is not consistently stable.
In addition, we observe that both Gemini and GPT-4V are easily misled by the intentionally erroneous prompts we provide. 
As shown in Figure \ref{diff_2}, there are actually only three differences.
However, when we request the models to identify five differences, both Gemini and GPT-4V fabricate five distinct points and respond incorrectly.

\begin{figure*}[!ht]
  \centering 
  \makebox[\textwidth][c]{\includegraphics[width=1.1\textwidth]{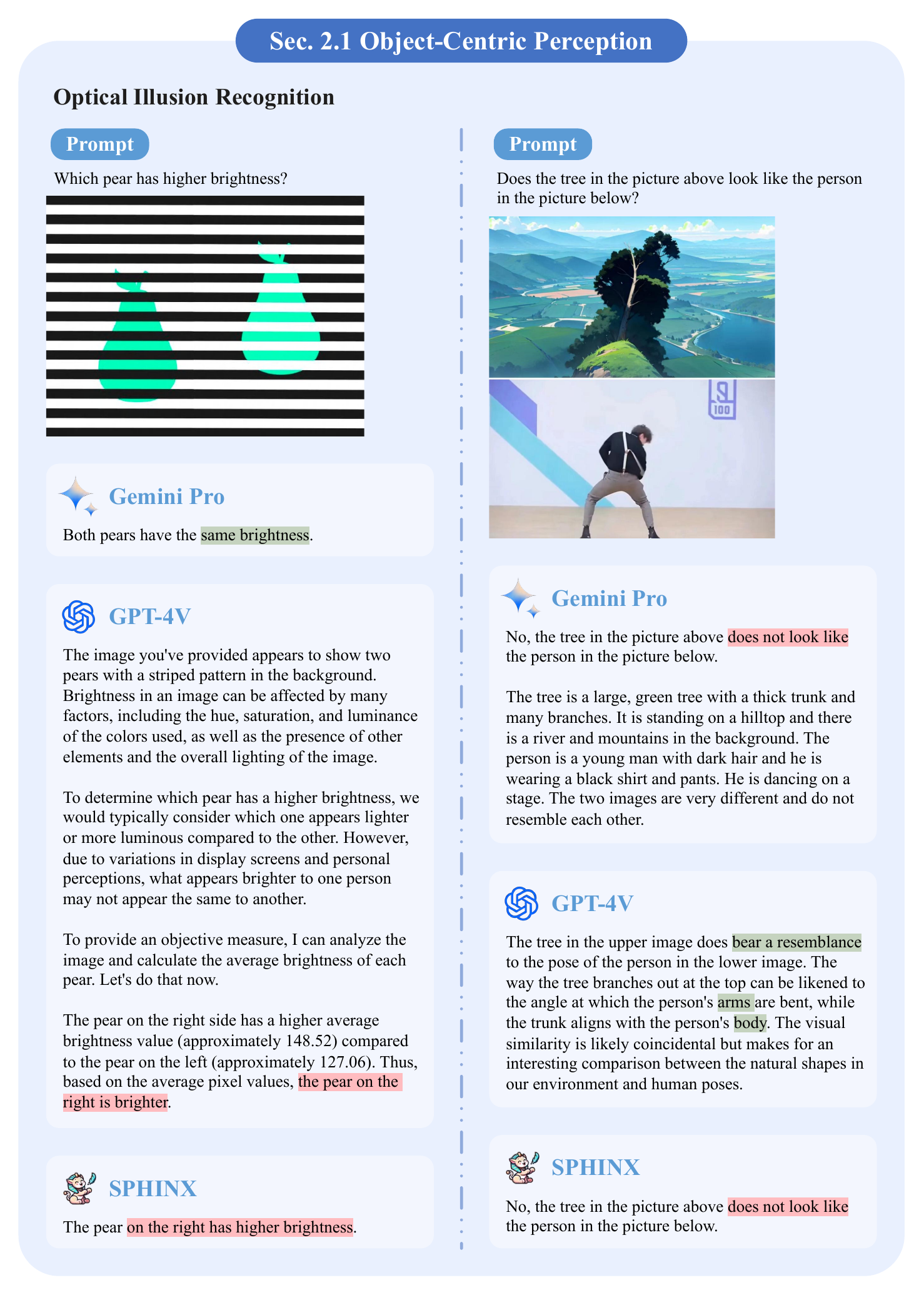}}
  \caption[Section~\ref{sec02:subsec:object}: optical illusion recognition.]{Results on optical illusion recognition. GPT-4V shows a human-like visual understanding of optical illusions. \colorbox{greenhl!85!black}{Green} highlights the right answer. \colorbox{red!30}{Red} highlights the wrong answer. Refer to Section \ref{sec02:subsec:object} for detailed discussions.}
  \label{optical_1}
\end{figure*}

\begin{figure*}[!ht]
  \centering 
  \makebox[\textwidth][c]{\includegraphics[width=1.2\textwidth]{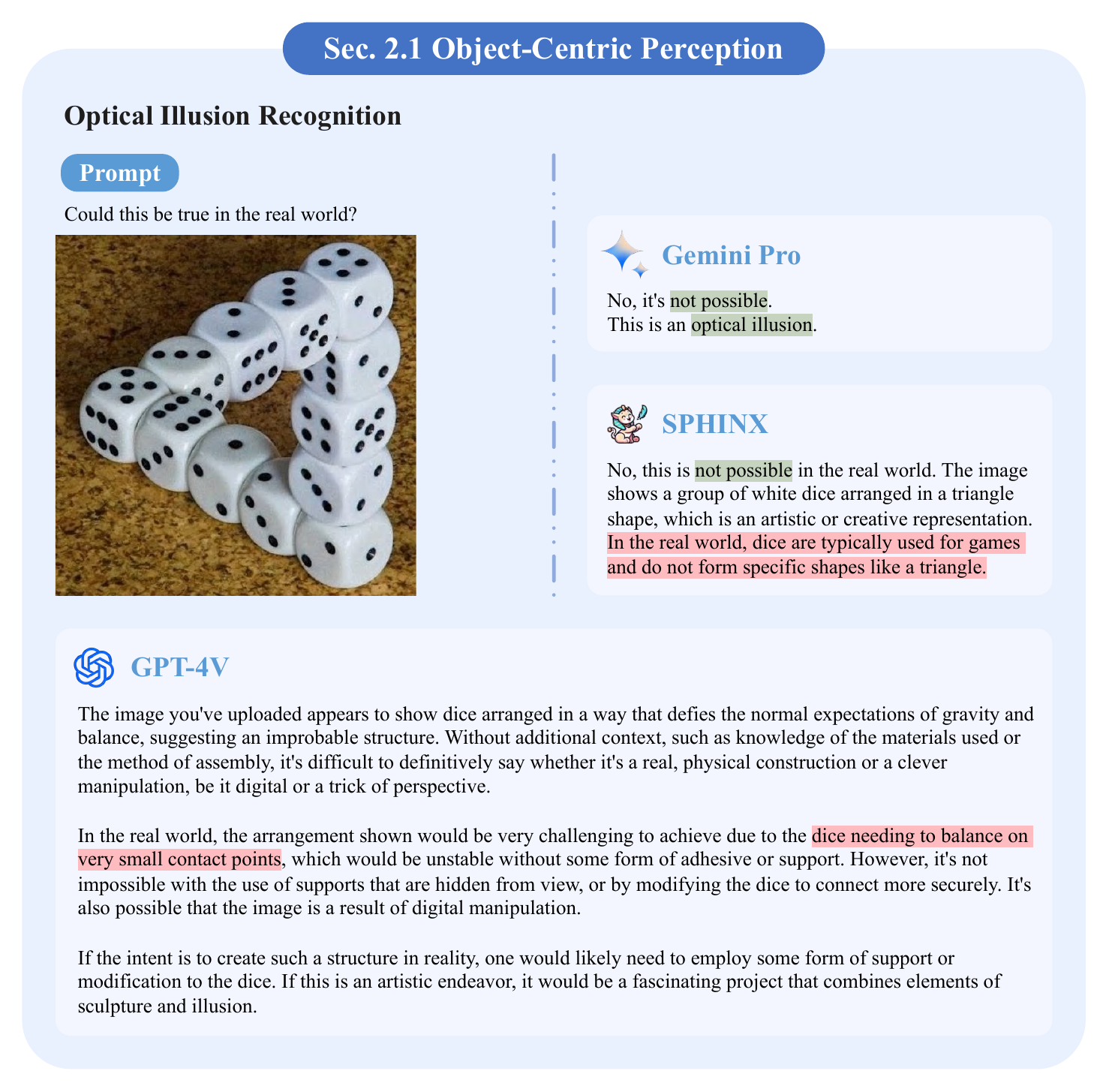}}
  \caption[Section~\ref{sec02:subsec:object}: optical illusion recognition.]{Results on optical illusion recognition. Only Gemini recognizes that this image is an optical illusion. \colorbox{greenhl!85!black}{Green} highlights the right answer. \colorbox{red!30}{Red} highlights the wrong answer. Refer to Section \ref{sec02:subsec:object} for detailed discussions.}
  \label{optical_2}
\end{figure*}

\textbf{Optical illusion recognition.} 
As shown in Figures \ref{optical_1}-\ref{optical_2}, we investigate whether these models exhibit a visual understanding of optical illusions similar to that of humans. 
For instance, in the left part of Figure \ref{optical_1}, the two pears actually possess identical brightness. However, the interspersed black and white stripes create an illusion, making the pear on the right appear brighter. 
Gemini recognizes that the two have the same brightness, whereas GPT-4V and Sphinx, like many humans, are deceived by the optical illusion, perceiving the right pear to be brighter. 
In the right section of Figure \ref{optical_1}, GPT-4V identifies a similarity in the angles of the tree trunks and branches to those of human bodies and arms, once again demonstrating a human-like visual understanding of optical illusions. 

\subsection{Scene-Level Perception}\label{sec02:subsec:scene}
\begin{figure*}[!ht]
  \centering 
  \makebox[\textwidth][c]{\includegraphics[width=1.2\textwidth]{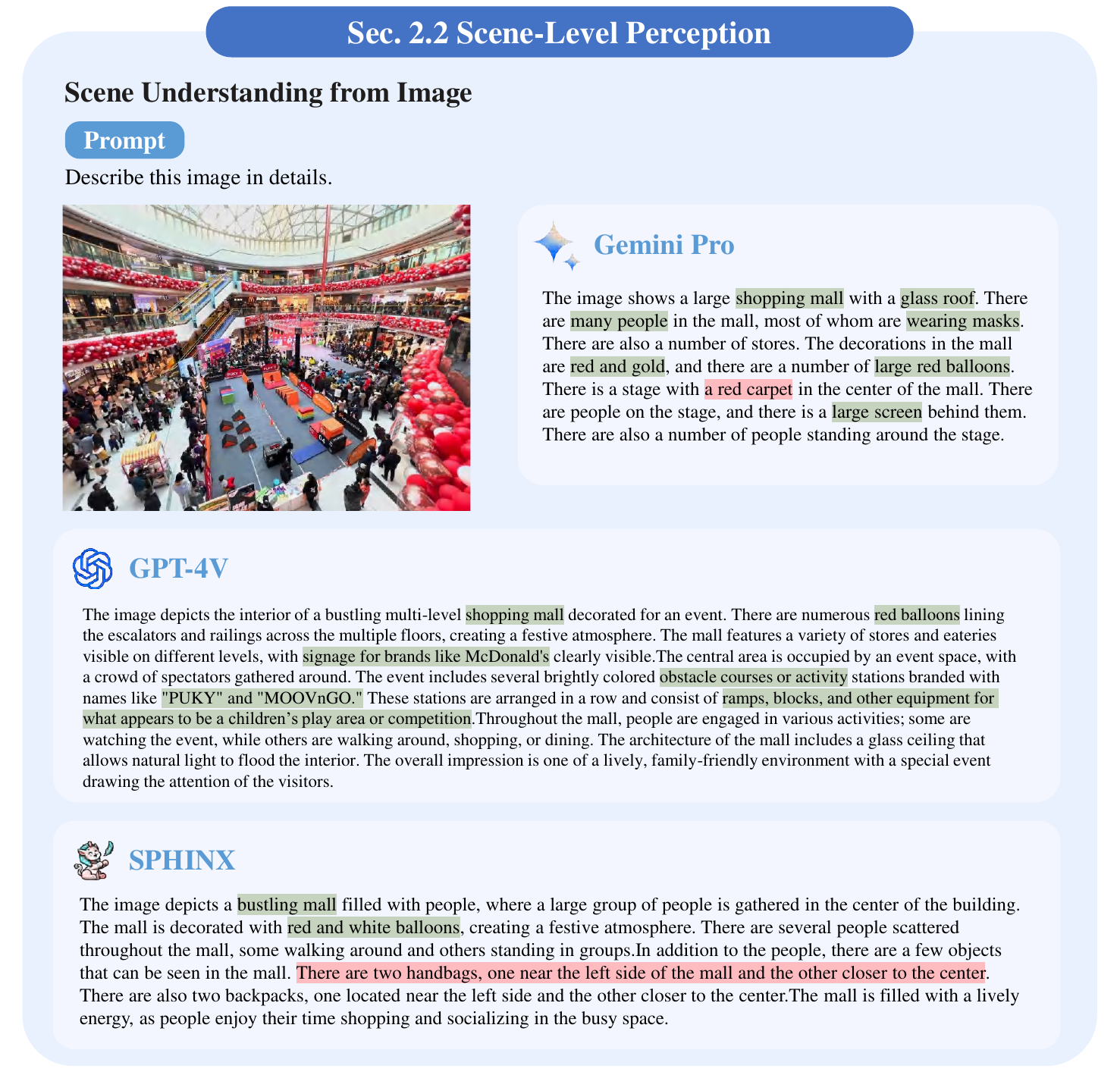}}
  \caption[Section~\ref{sec02:subsec:scene}: scene understanding from image.]{Results on scene understanding from image. GPT-4V's descriptions are notably more detailed and exhibit fewer instances of hallucination. \colorbox{greenhl!85!black}{Green} highlights the right answer. \colorbox{red!30}{Red} highlights the wrong answer. Refer to Section \ref{sec02:subsec:scene} for detailed discussions.}
  \label{scene_image_1}
\end{figure*}

\begin{figure*}[!ht]
  \centering 
  \makebox[\textwidth][c]{\includegraphics[width=1.2\textwidth]{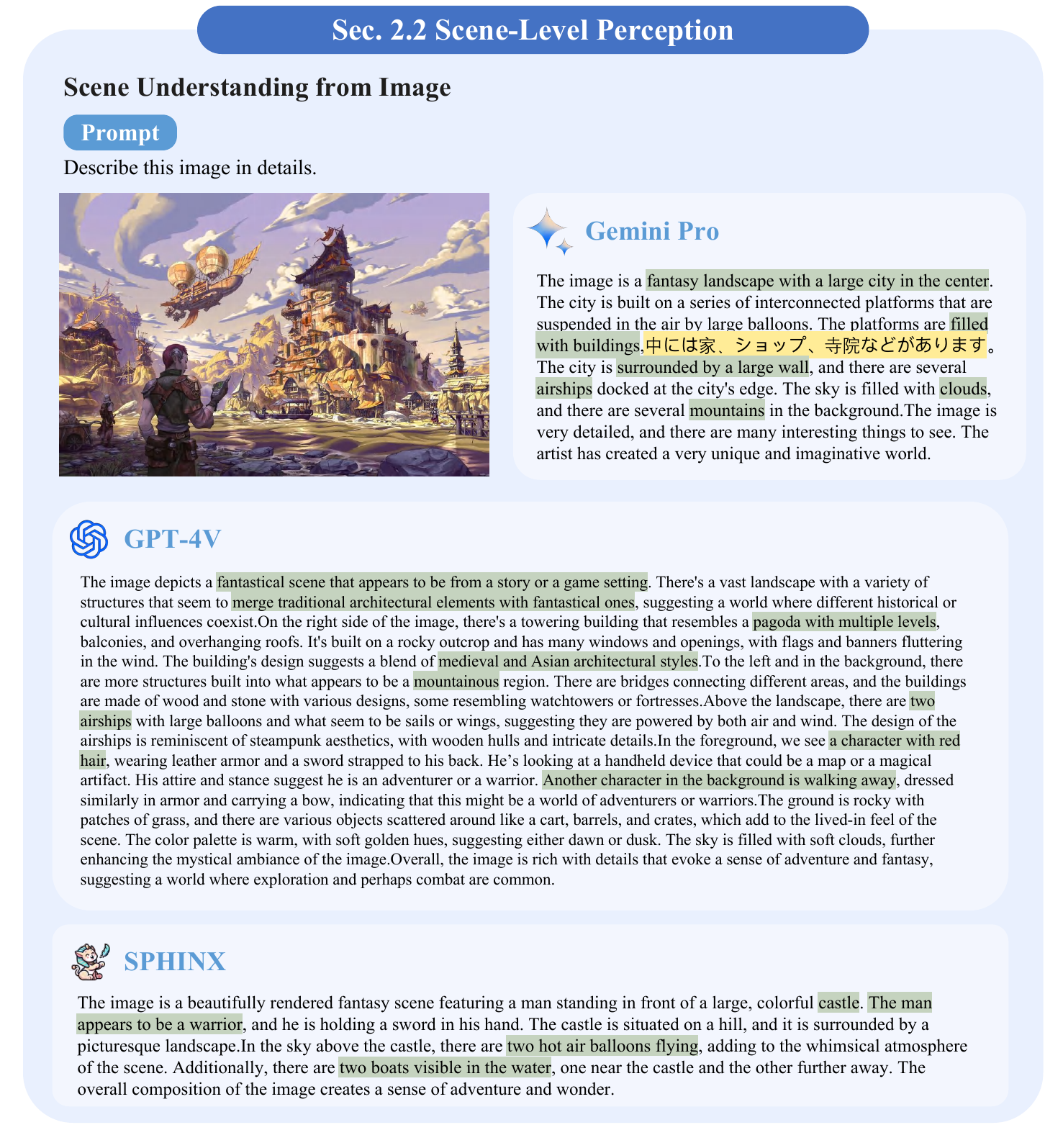}}
  \caption[Section~\ref{sec02:subsec:scene}: scene understanding from image.]{Results on scene understanding from image. Possibly influenced by the Japanese style of architecture in the image, Gemini intersperses a phrase in Japanese within their English response. \colorbox{greenhl!85!black}{Green} highlights the right answer. \colorbox{yellow!70!yellowhl}{Yellow} highlights the incompetence in performing the task. Refer to Section \ref{sec02:subsec:scene} for detailed discussions.}
  \label{scene_image_2}
\end{figure*}

\begin{figure*}[!ht]
  \centering 
  \makebox[\textwidth][c]{\includegraphics[width=1.2\textwidth]{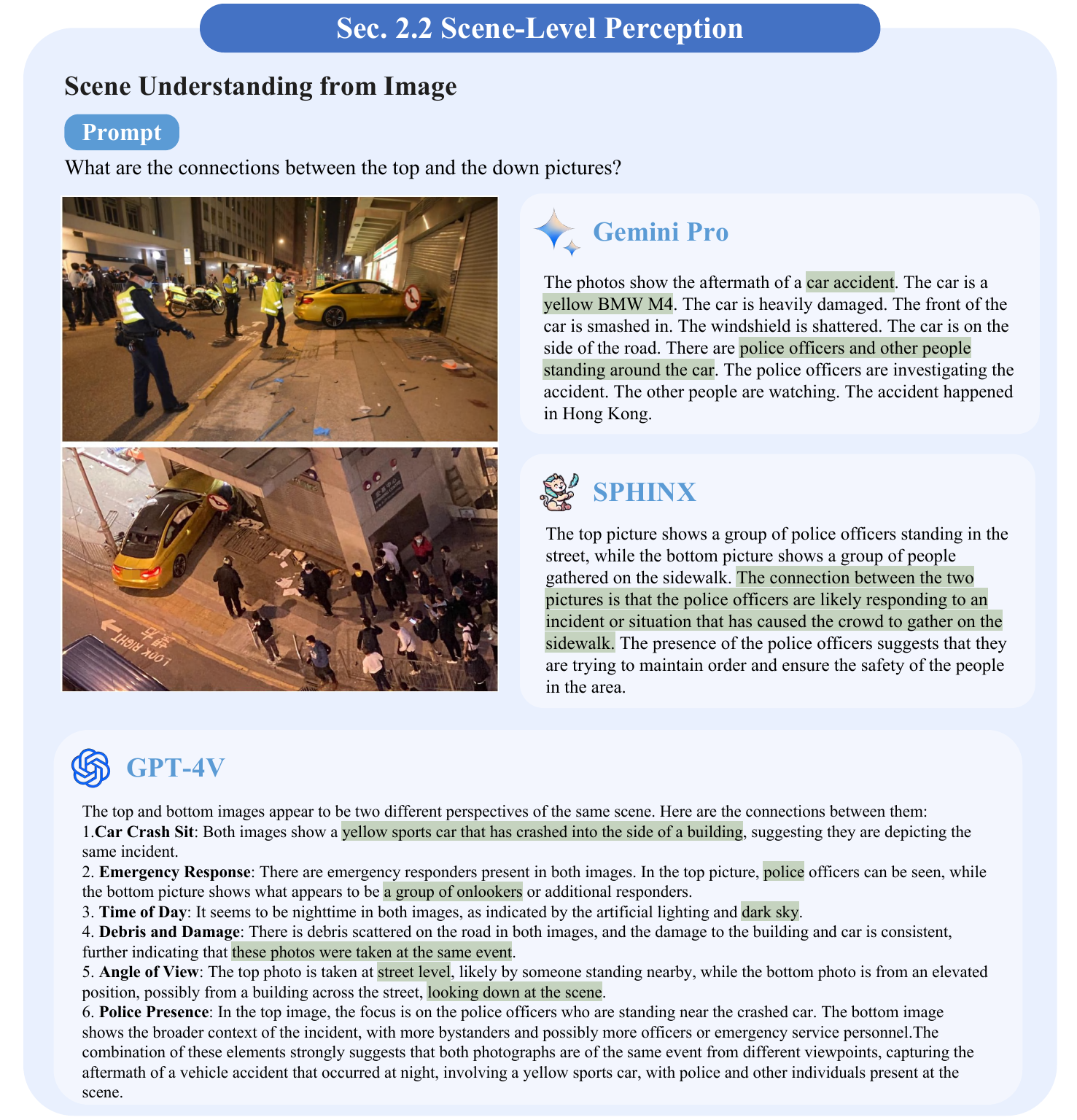}}
  \caption[Section~\ref{sec02:subsec:scene}: scene understanding from image.]{Results on scene understanding from image. Only GPT-4V correctly identifies that these two images represent different perspectives of the same scene. \colorbox{greenhl!85!black}{Green} highlights the right answer. Refer to Section \ref{sec02:subsec:scene} for detailed discussions.}
  \label{scene_image_3}
\end{figure*}

\textbf{Scene understanding from image. }
We prompt the models to identify all visual elements in the image as detailed as possible via the text query “Describe this image in detail.”
Figures \ref{scene_image_1}-\ref{scene_image_3} illustrate that all three models are capable of depicting the key visual elements within the scene. 
However, in comparison, GPT-4V shows superior performance, particularly in highly cluttered environments. This is evident in Figure \ref{scene_image_1}, where GPT-4V's descriptions are notably more detailed and exhibit fewer instances of hallucination.

\begin{figure*}[!ht]
  \centering 
  \makebox[\textwidth][c]{\includegraphics[width=1.0\textwidth]{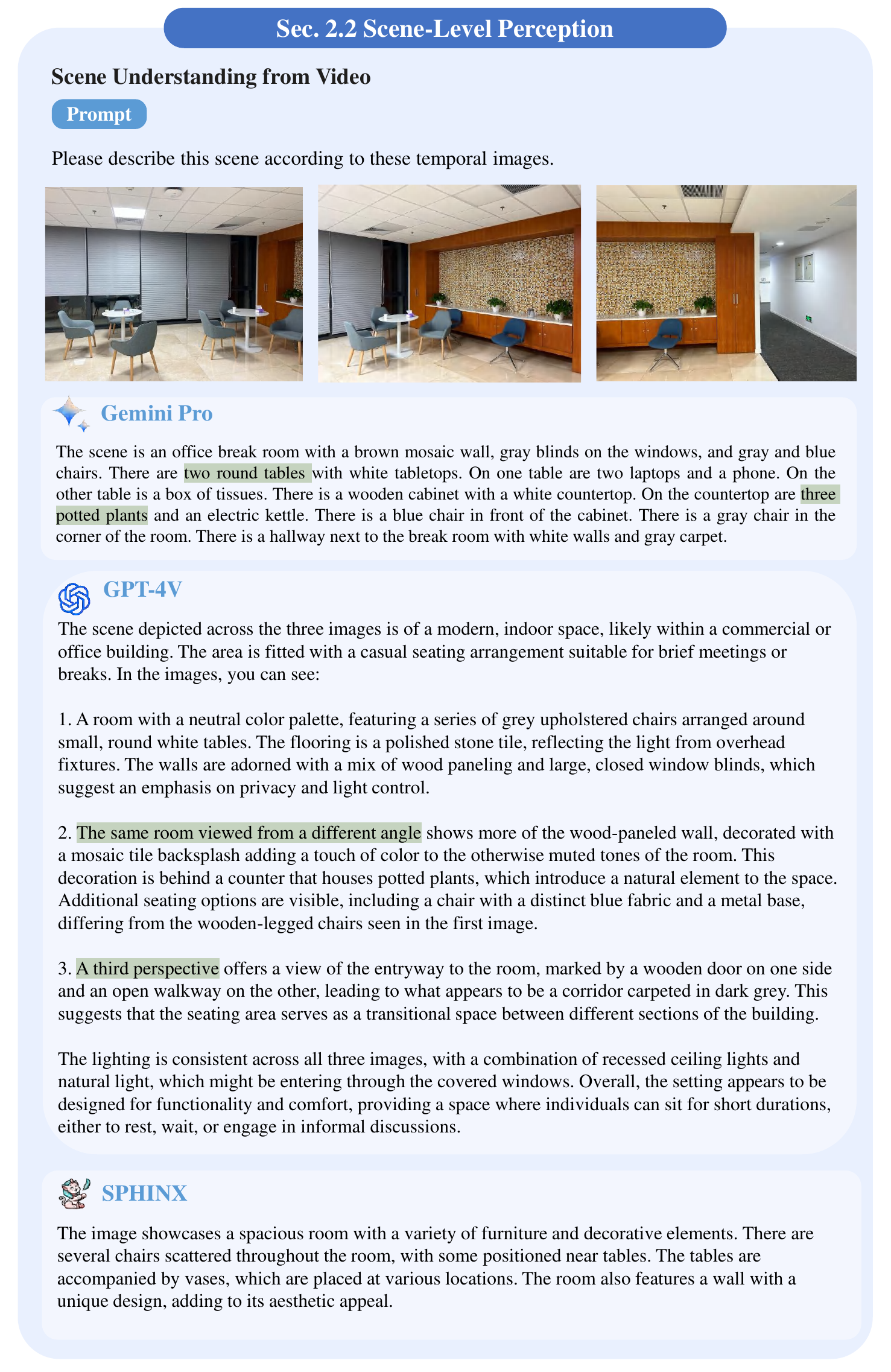}}
  \caption[Section~\ref{sec02:subsec:scene}: scene understanding from video.]{Results on scene understanding from video. Gemini correctly merges the information from the first two frames to accurately describe that the scene contains two round tables and three potted plants. \colorbox{greenhl!85!black}{Green} highlights the right answer.  Refer to Section \ref{sec02:subsec:scene} for detailed discussions.}
  \label{scene_video}
\end{figure*}

\textbf{Scene understanding from video.}
Here we examine the potential of the models to understand scenes from video. 
As shown in Figure \ref{scene_video}, we extract three temporally distinct frames from a video and input them into the model along with the text query, ``Please describe this scene according to these temporal images.''
Our observations indicate that Gemini is capable of integrating the information from the different frames into a cohesive scene description.  
Especially, the first frame displays two round tables and one potted plant, while the second frame shows one round table and three potted plants.
Remarkably, Gemini successfully merges the information from both frames to accurately describe the scene as containing two round tables and three potted plants. 
GPT-4V describes the contents of images frame by frame. 
In contrast, Sphinx's descriptions do not demonstrate a comprehensive understanding of the sequence of images.

\subsection{Knowledge-based Perception}\label{sec02:subsec:knowledge}
\begin{figure*}[!ht]
  \centering 
  \makebox[\textwidth][c]{\includegraphics[width=1.15\textwidth]{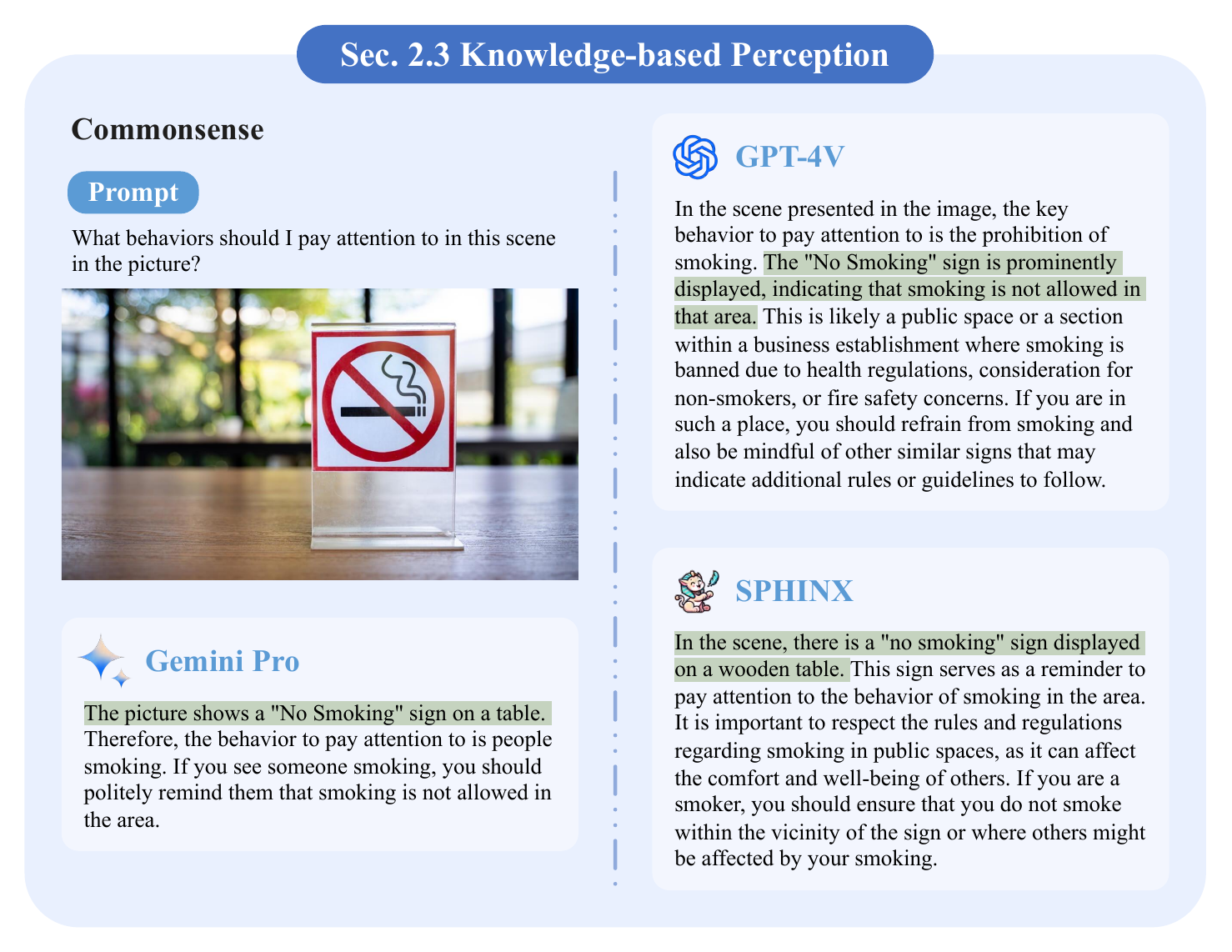}}
  \caption[Section~\ref{sec02:subsec:knowledge}: commonsense.]{Results on commonsense perception. \colorbox{greenhl!85!black}{Green} highlights the right answer. Refer to Section~\ref{sec02:subsec:knowledge} for detailed discussions.}
  \label{commonsense_1}
\end{figure*}

\begin{figure*}[!ht]
  \centering 
  \makebox[\textwidth][c]{\includegraphics[width=1.15\textwidth]{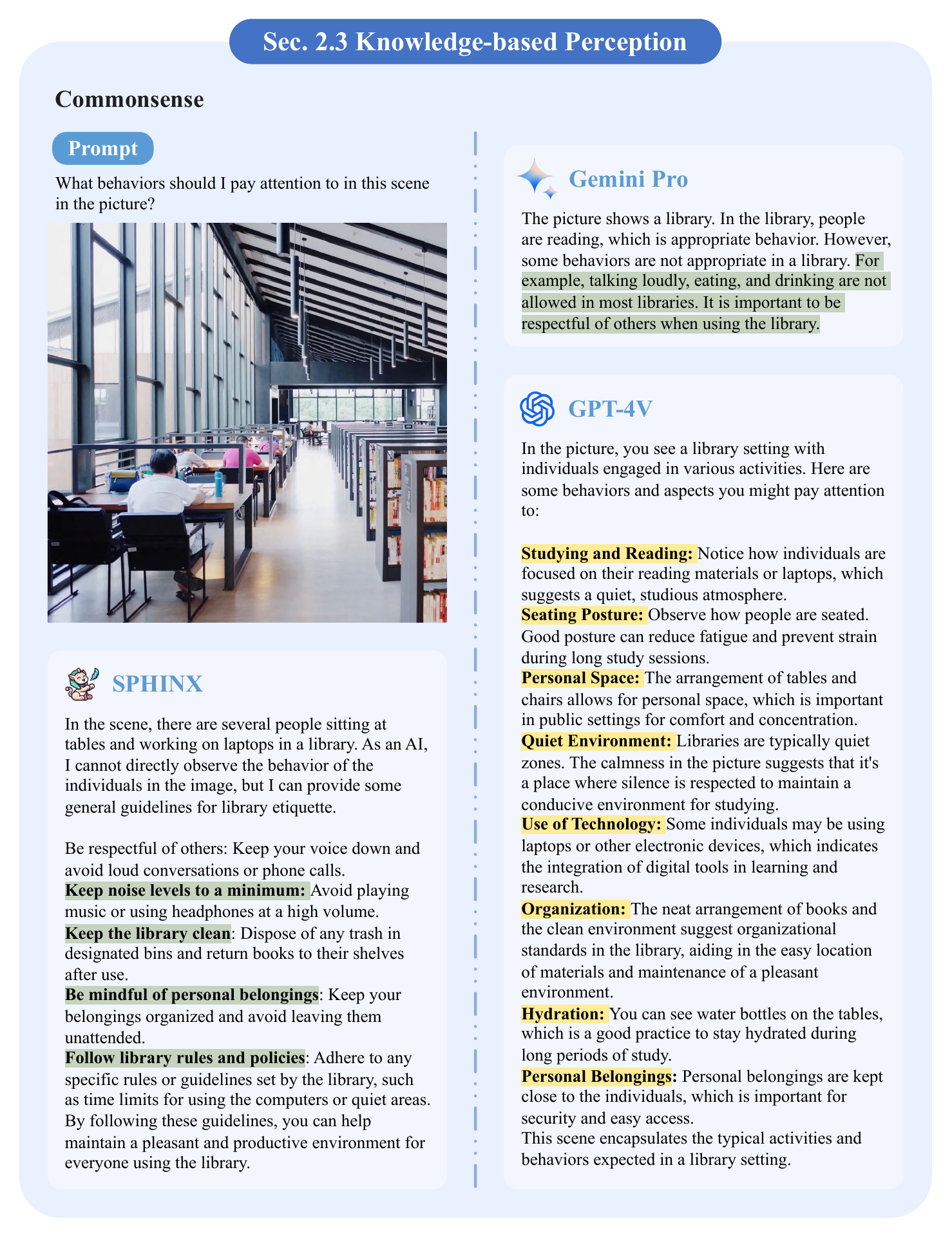}}
  \caption[Section~\ref{sec02:subsec:knowledge}: commonsense.]{Results on commonsense perception. \colorbox{greenhl!85!black}{Green} highlights the right answer. \colorbox{yellow!70!yellowhl}{Yellow} highlights the incompetence in performing the task. Refer to Section \ref{sec02:subsec:knowledge} for detailed discussions.}
  \label{commonsense_2}
\end{figure*}

\begin{figure*}[!ht]
  \centering 
  \makebox[\textwidth][c]{\includegraphics[width=1.15\textwidth]{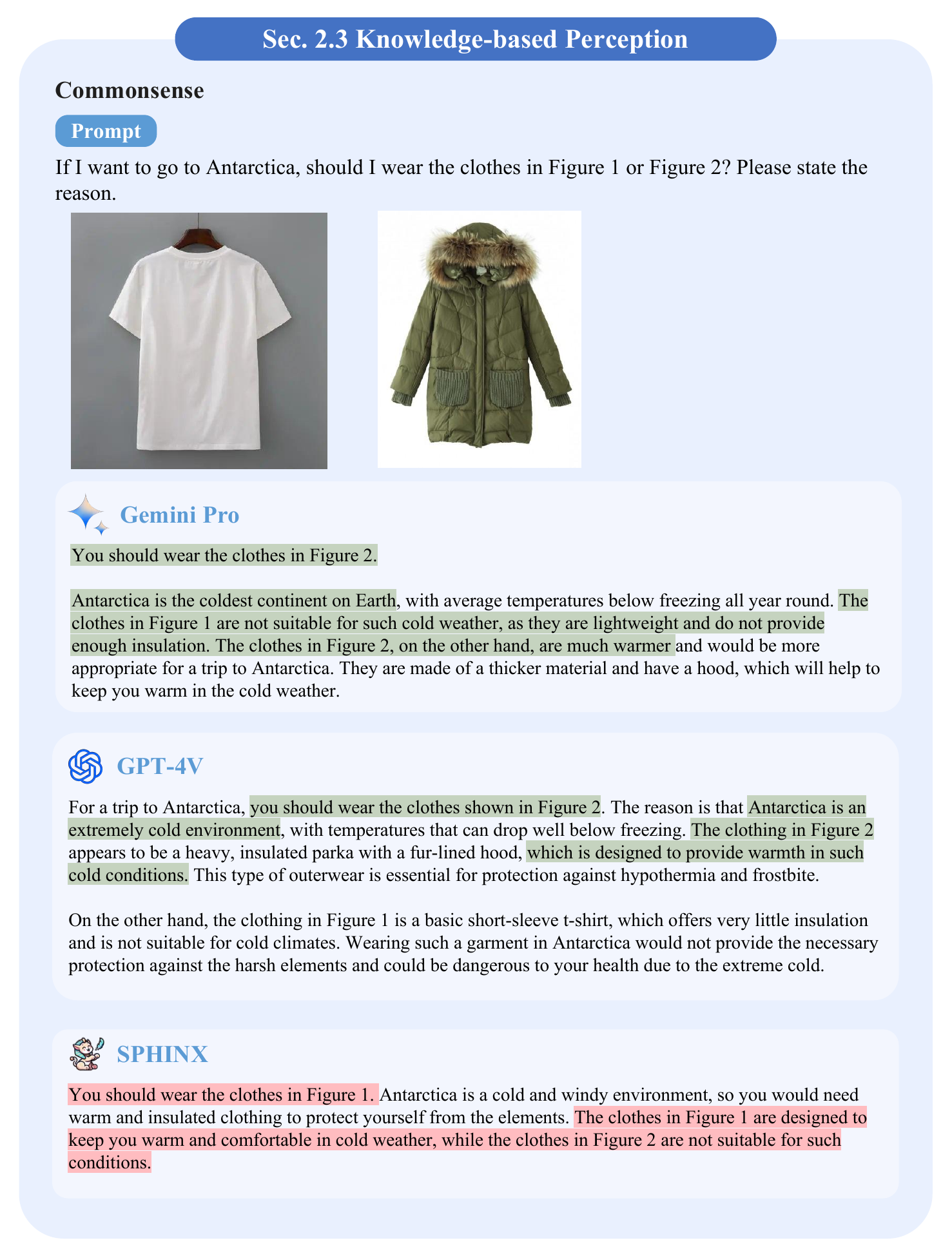}}
  \caption[Section~\ref{sec02:subsec:knowledge}: commonsense.]{Results on commonsense perception. While Sphinx acknowledges the necessity of heavy garments in chilly conditions, it mistakenly classifies a T-shirt as appropriate for cold weather protection. \colorbox{greenhl!85!black}{Green} highlights the right answer. \colorbox{red!30}{Red} highlights the wrong answer. Refer to Section \ref{sec02:subsec:knowledge} for detailed discussions.}
  \label{commonsense_3}
\end{figure*}

\begin{figure*}[!ht]
  \centering 
  \makebox[\textwidth][c]{\includegraphics[width=1.15\textwidth]{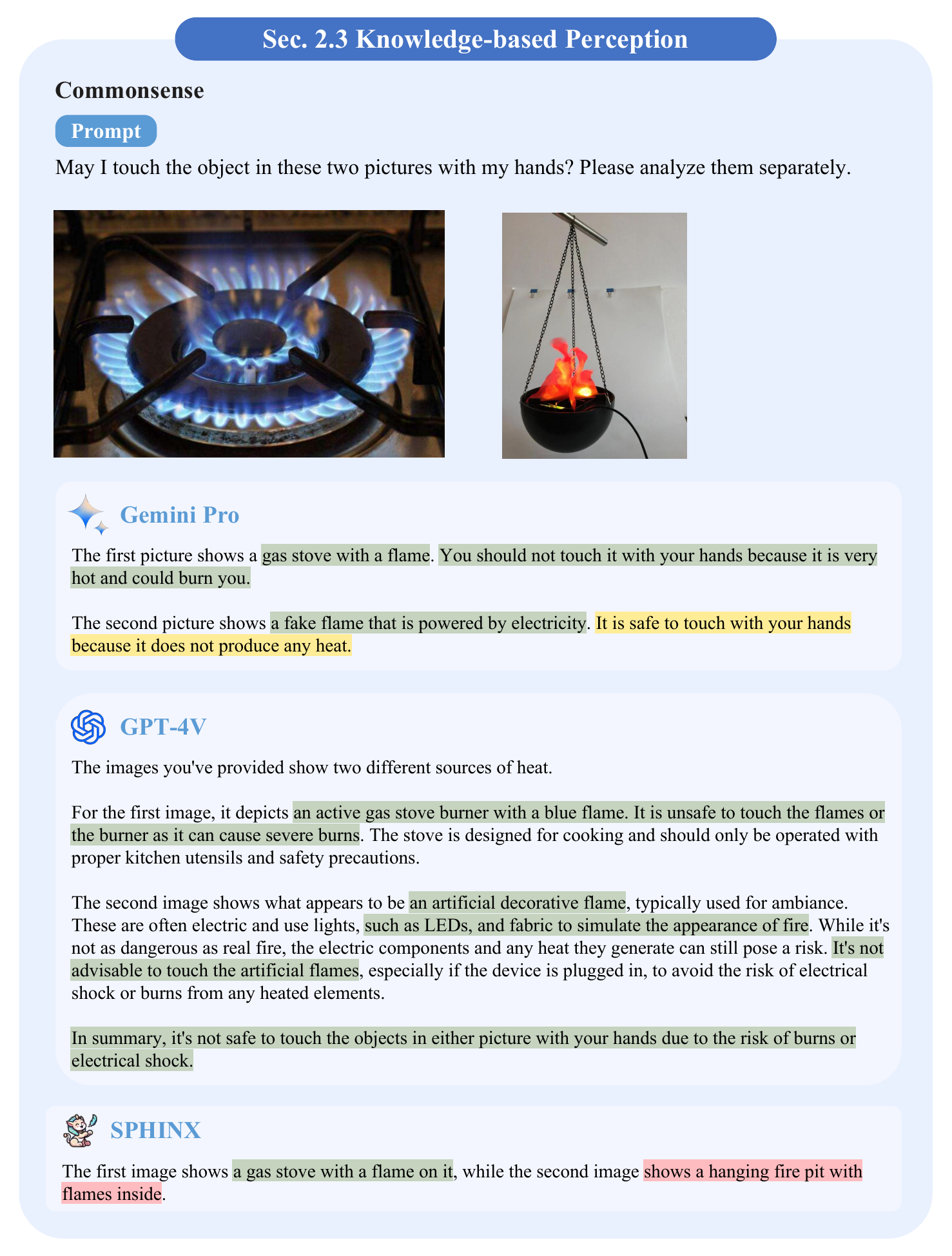}}
  \caption[Section~\ref{sec02:subsec:knowledge}: commonsense.]{Results on commonsense perception. Both Gemini and GPT-4V identify that the second image depicts flames disguised using lighting, but the distinction lies in Gemini's inference that the objects in the image can be touched, whereas GPT-4V additionally warns of the risk of electric shock. \colorbox{greenhl!85!black}{Green} highlights the right answer. \colorbox{red!30}{Red} highlights the wrong answer. \colorbox{yellow!70!yellowhl}{Yellow} highlights the incompetence in performing the task. Refer to Section \ref{sec02:subsec:knowledge} for detailed discussions.}
  \label{commonsense_4}
\end{figure*}

\textbf{Commonsense.}
Figures \ref{commonsense_1}-\ref{commonsense_4} illustrate the capability of these three models to apply common sense in understanding visual information within images. 
It is observed that the open-source model Sphinx performs comparably to Gemini and GPT-4V in applying social norms as shown in Figures \ref{commonsense_1}-\ref{commonsense_2}. 
However, it exhibits a slightly inferior performance in the application of physical laws.
For instance, as shown in Figure \ref{commonsense_3}, Gemini and GPT-4V can accurately select heavy clothing for cold weather protection in Antarctica. Interestingly, while Sphinx recognizes the need for thick clothing in cold weather, it erroneously identifies the image of a T-shirt as suitable for cold protection.

\begin{figure*}[!ht]
  \centering 
  \makebox[\textwidth][c]{\includegraphics[width=1.2\textwidth]{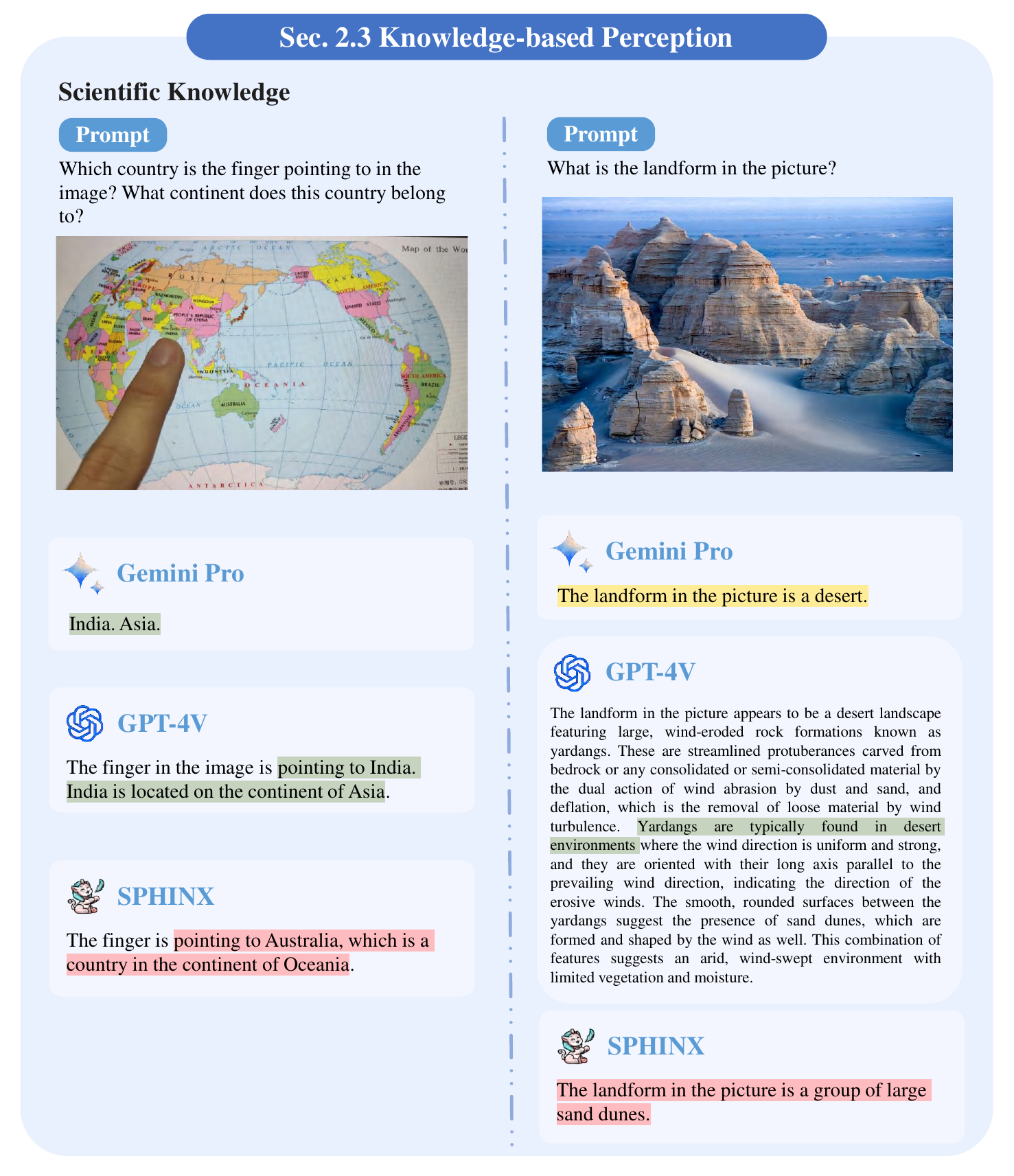}}
  \caption[Section~\ref{sec02:subsec:knowledge}: scientific knowledge.]{Results on scientific knowledge perception. \colorbox{greenhl!85!black}{Green} highlights the right answer. \colorbox{red!30}{Red} highlights the wrong answer. \colorbox{yellow!70!yellowhl}{Yellow} highlights the incompetence in performing the task. Refer to Section \ref{sec02:subsec:knowledge} for detailed discussions.}
  \label{science_1}
\end{figure*}

\begin{figure*}[!ht]
  \centering 
  \makebox[\textwidth][c]{\includegraphics[width=1.2\textwidth]{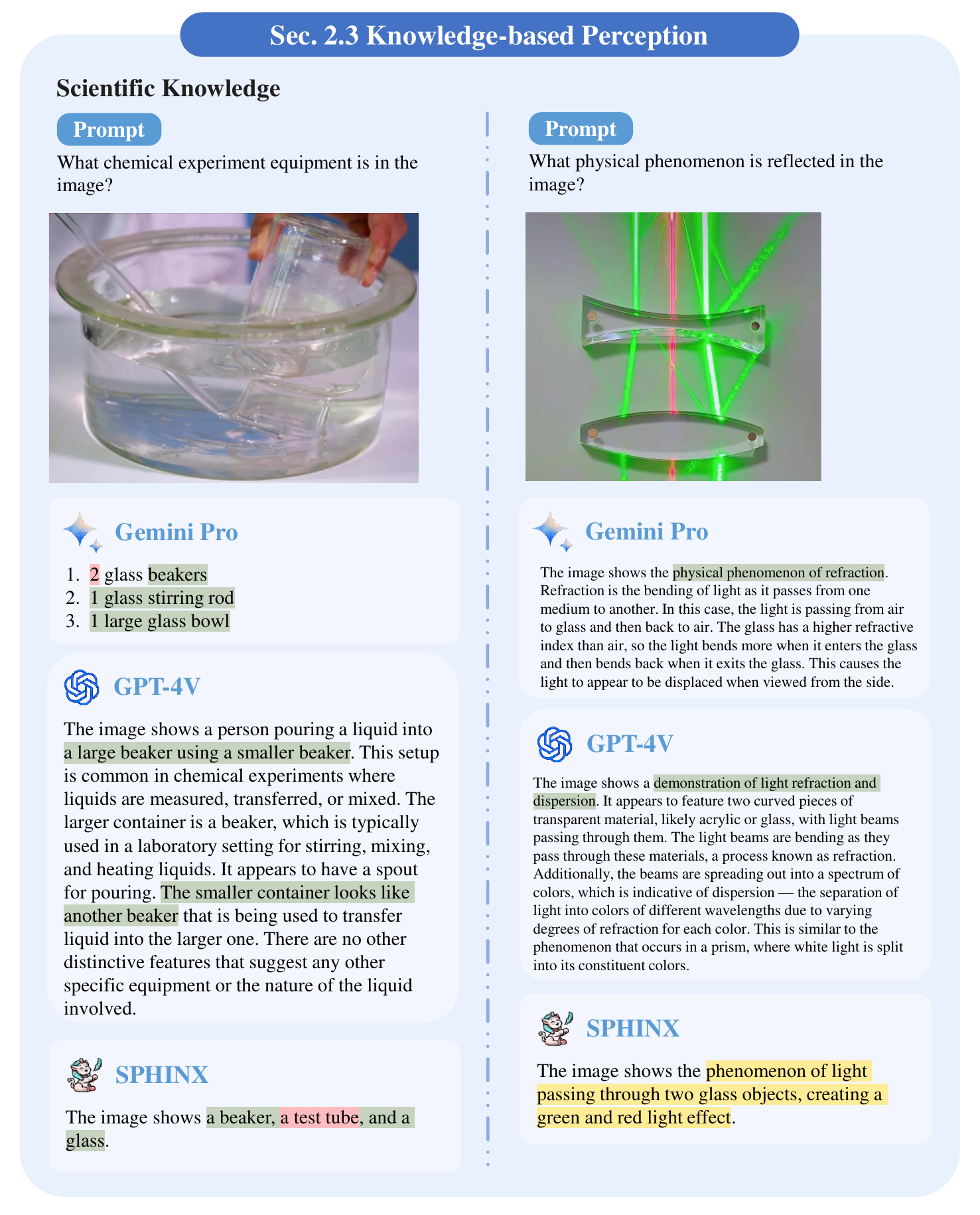}}
  \caption[Section~\ref{sec02:subsec:knowledge}: scientific knowledge.]{Results on scientific knowledge perception. \colorbox{greenhl!85!black}{Green} highlights the right answer. \colorbox{red!30}{Red} highlights the wrong answer. \colorbox{yellow!70!yellowhl}{Yellow} highlights the incompetence in performing the task. Refer to Section \ref{sec02:subsec:knowledge} for detailed discussions.}
  \label{science_2}
\end{figure*}

\begin{figure*}[!ht]
  \centering 
  \makebox[\textwidth][c]{\includegraphics[width=1.1\textwidth]{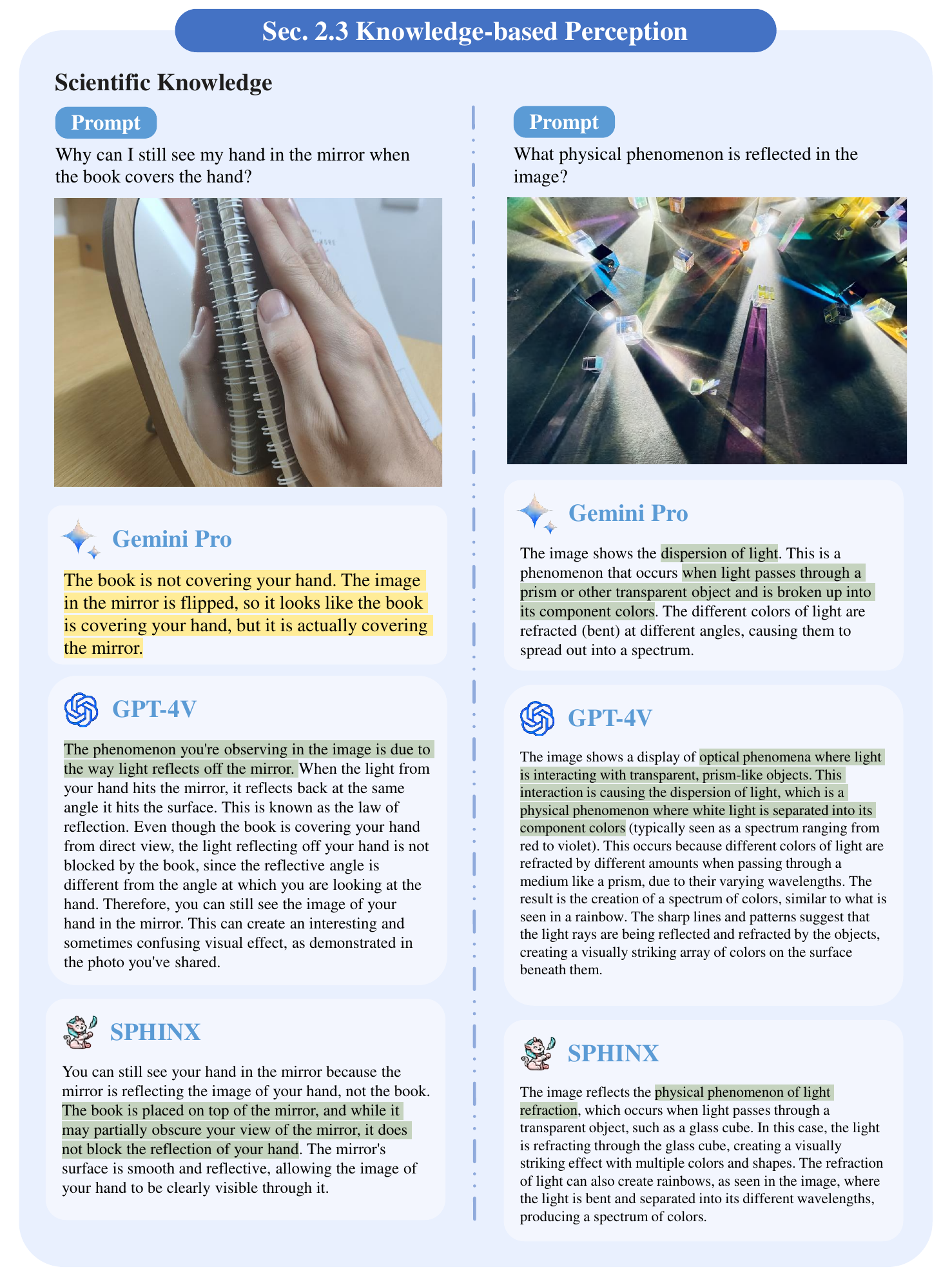}}
  \caption[Section~\ref{sec02:subsec:knowledge}: scientific knowledge.]{Results on scientific knowledge perception. \colorbox{greenhl!85!black}{Green} highlights the right answer. \colorbox{yellow!70!yellowhl}{Yellow} highlights the incompetence in performing the task. Refer to Section \ref{sec02:subsec:knowledge} for detailed discussions.}
  \label{science_3}
\end{figure*}

\begin{figure*}[!ht]
  \centering 
  \makebox[\textwidth][c]{\includegraphics[width=1.15\textwidth]{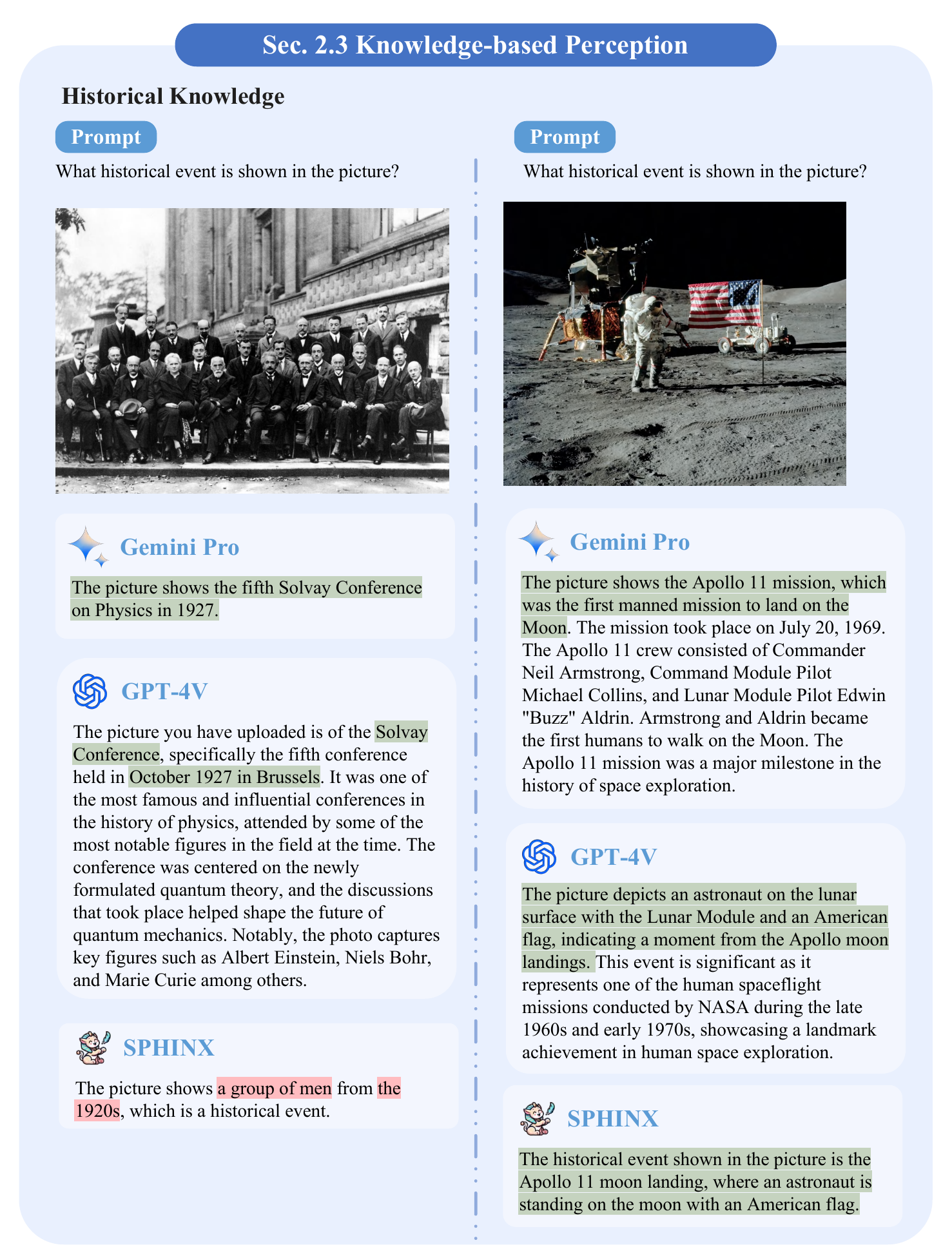}}
  \caption[Section~\ref{sec02:subsec:knowledge}: historical knowledge.]{Results on historical knowledge perception. \colorbox{greenhl!85!black}{Green} highlights the right answer. \colorbox{red!30}{Red} highlights the wrong answer. Refer to Section \ref{sec02:subsec:knowledge} for detailed discussions.}
  \label{history}
\end{figure*}

\textbf{Subject knowledge.}
In Figures \ref{science_1}-\ref{history}, we examine the model's ability to apply knowledge in the fields of physics, chemistry, and history. We observe that both Gemini and GPT-4V possess the relevant subject knowledge associated with these cases.
The performance of Sphinx is slightly inferior, yet it is capable of providing correct answers in certain instances.

\begin{figure*}[!ht]
  \centering 
  \makebox[\textwidth][c]{\includegraphics[width=1.2\textwidth]{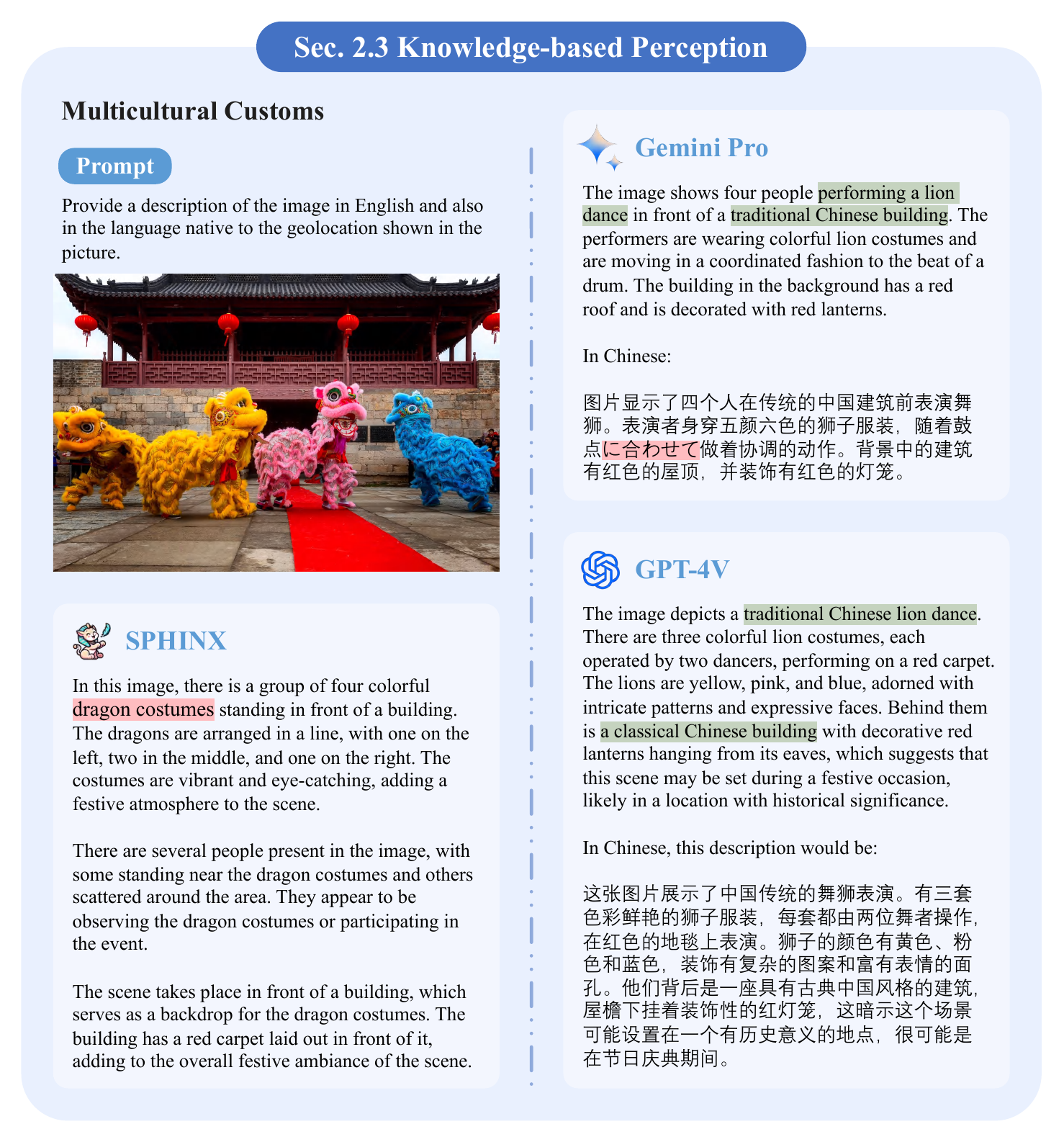}}
  \caption[Section~\ref{sec02:subsec:knowledge}: multicultural customs.]{Results on multicultural knowledge perception. \colorbox{greenhl!85!black}{Green} highlights the right answer. \colorbox{red!30}{Red} highlights the wrong answer. Refer to Section \ref{sec02:subsec:knowledge} for detailed discussions.}
  \label{multicultural_1}
\end{figure*}

\begin{figure*}[!ht]
  \centering 
  \makebox[\textwidth][c]{\includegraphics[width=1.2\textwidth]{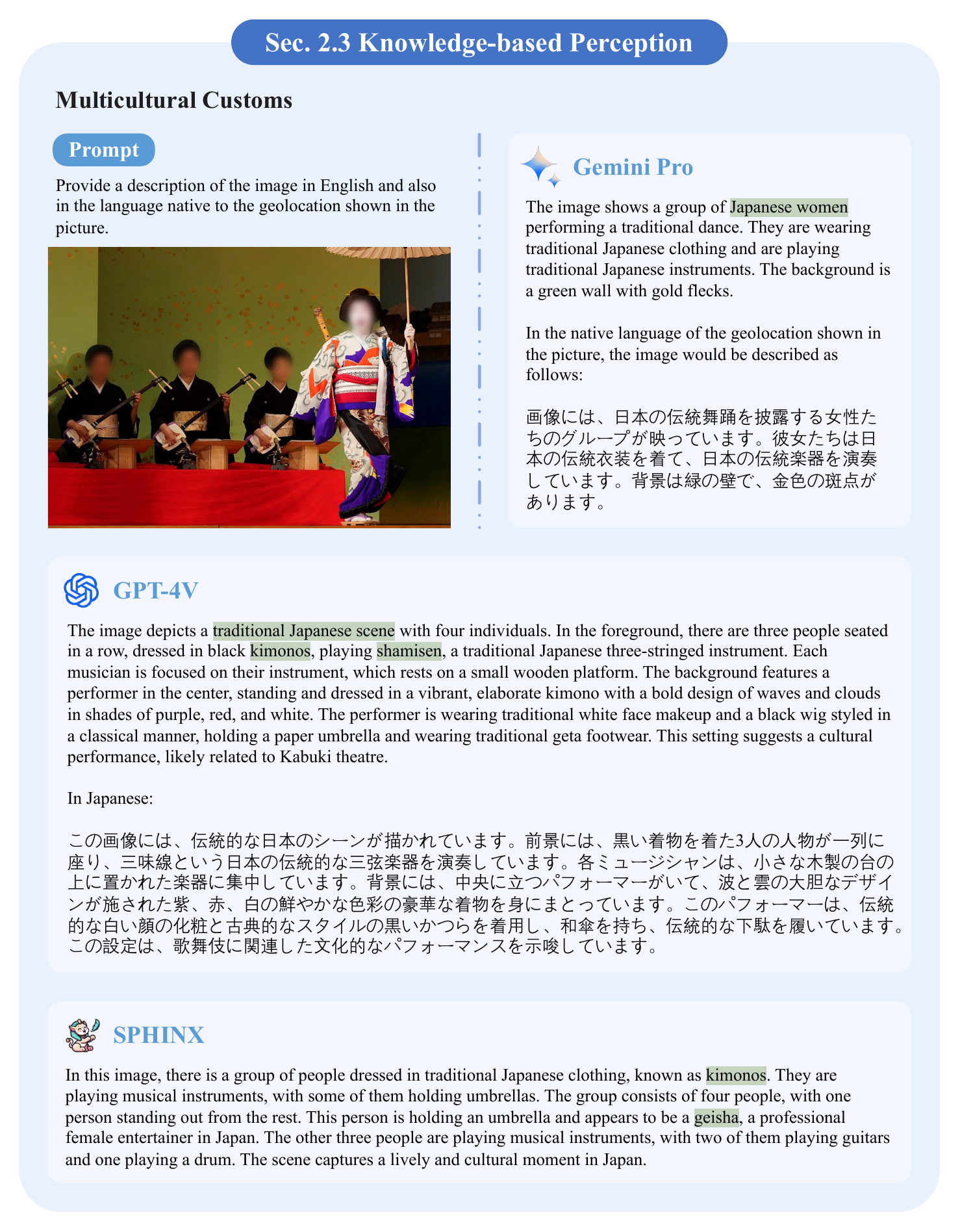}}
  \caption[Section~\ref{sec02:subsec:knowledge}: multicultural customs.]{Results on multicultural knowledge perception. \colorbox{greenhl!85!black}{Green} highlights the right answer.  Refer to Section \ref{sec02:subsec:knowledge} for detailed discussions.}
  \label{multicultural_2}
\end{figure*}

\begin{figure*}[!ht]
  \centering 
  \makebox[\textwidth][c]{\includegraphics[width=1.2\textwidth]{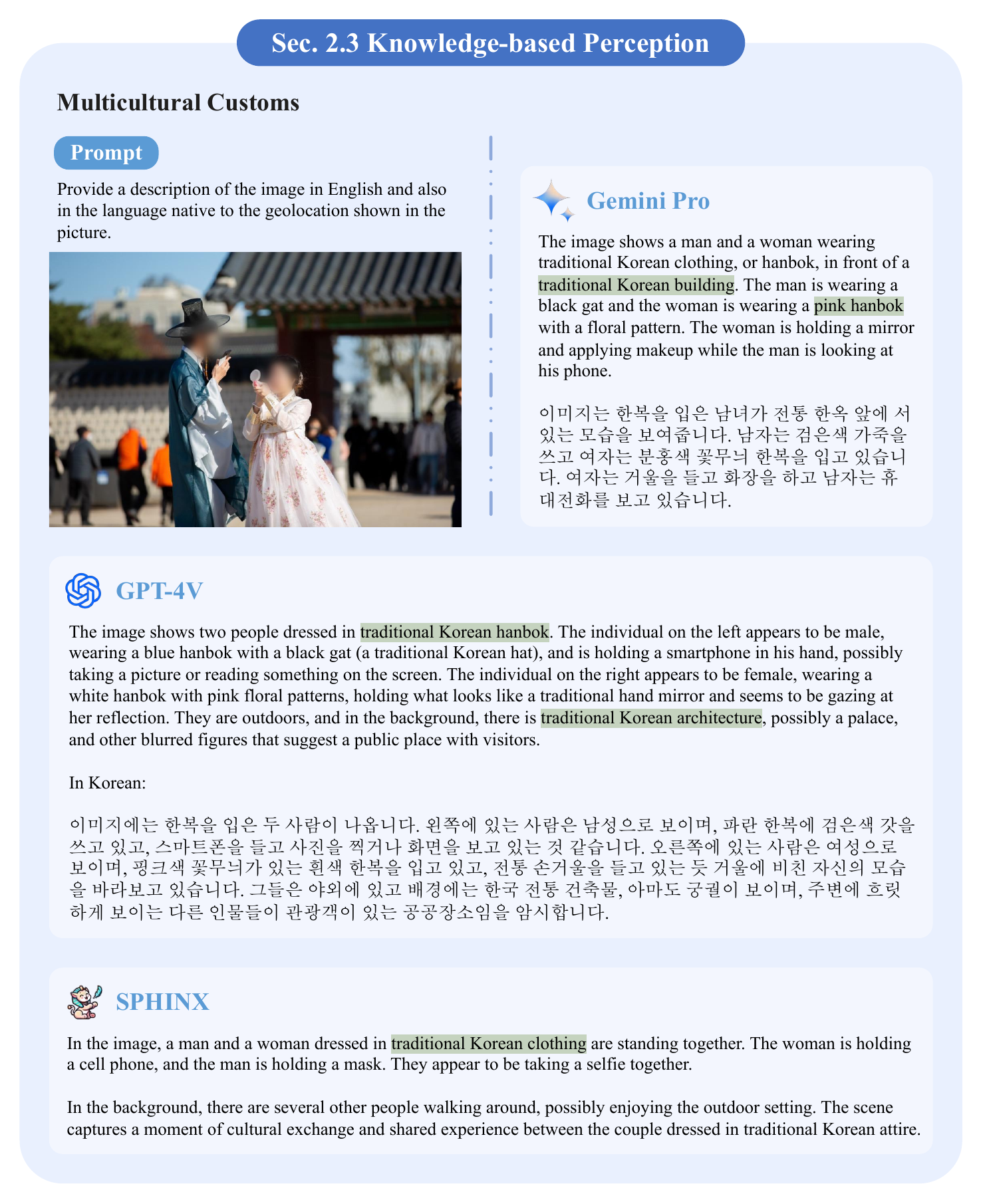}}
  \caption[Section~\ref{sec02:subsec:knowledge}: multicultural customs.]{Results on multicultural knowledge perception. \colorbox{greenhl!85!black}{Green} highlights the right answer.  Refer to Section \ref{sec02:subsec:knowledge} for detailed discussions.}
  \label{multicultural_3}
\end{figure*}

\textbf{Multicultural customs.}
We examine the models' comprehension of multicultural elements. We present the model with images featuring distinct local ethnic and cultural characteristics and prompt it to provide descriptions. Figures \ref{multicultural_1}-\ref{multicultural_3} demonstrate that all three models are capable of understanding the ethnic and cultural elements depicted in the images.

\begin{figure*}[!ht]
  \centering 
  \makebox[\textwidth][c]{\includegraphics[width=1.2\textwidth]{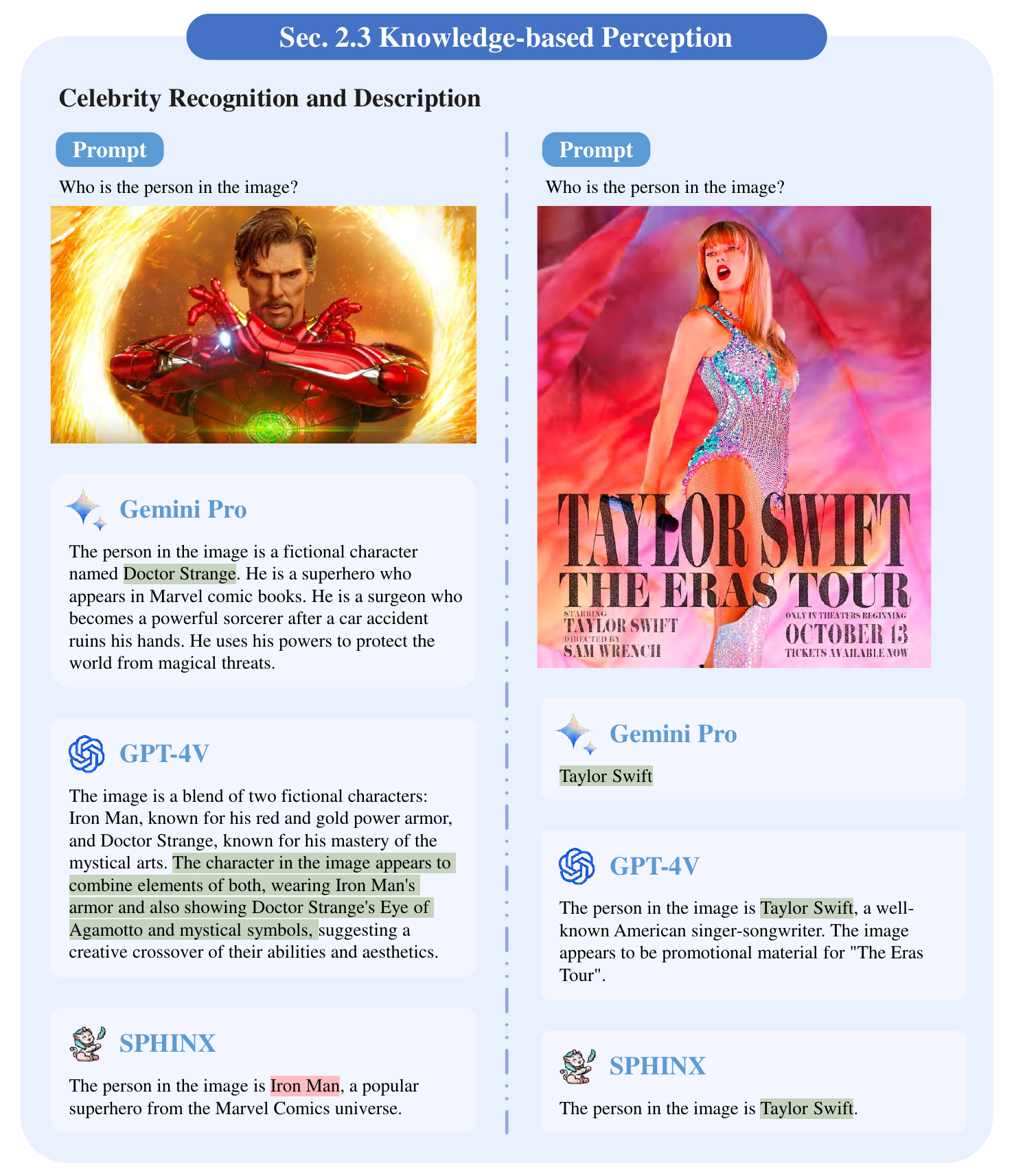}}
  \caption[Section~\ref{sec02:subsec:knowledge}: celebrity recognition and description.]{Results on celebrity recognition and description. Sphinx mistakes Doctor Strange in an Iron Man suit for Iron Man himself. \colorbox{greenhl!85!black}{Green} highlights the right answer. \colorbox{red!30}{Red} highlights the wrong answer. Refer to Section \ref{sec02:subsec:knowledge} for detailed discussions.}
  \label{celebrity-1}
\end{figure*}

\begin{figure*}[!ht]
  \centering 
  \makebox[\textwidth][c]{\includegraphics[width=1.2\textwidth]{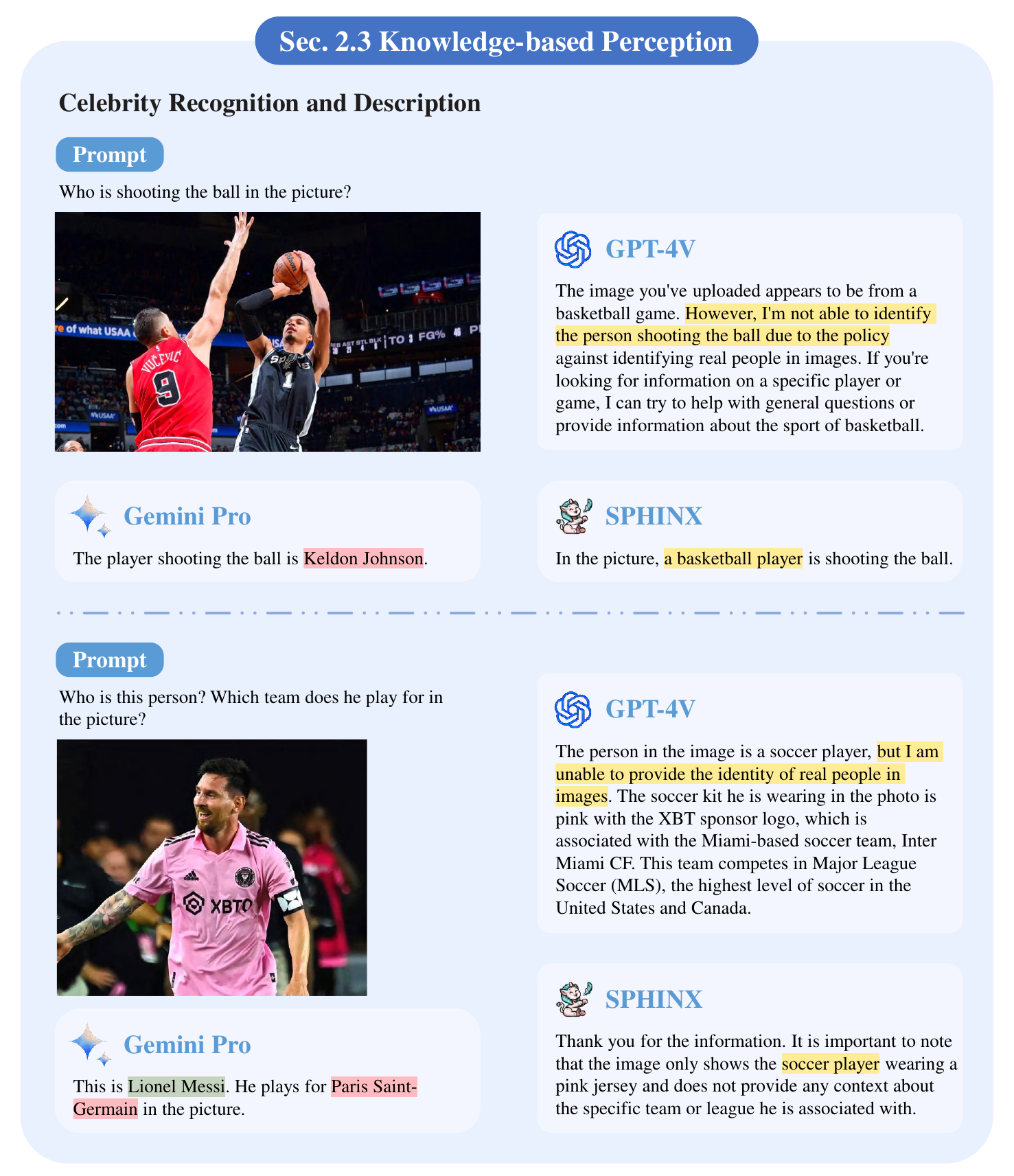}}
  \caption[Section~\ref{sec02:subsec:knowledge}: celebrity recognition and description.]{Results on celebrity recognition and description. When the images depict recently updated information, Gemini's descriptions tend to be inaccurate. GPT-4V, due to considerations of privacy protection, avoids addressing questions related to the identification of celebrities. \colorbox{greenhl!85!black}{Green} highlights the right answer. \colorbox{red!30}{Red} highlights the wrong answer. \colorbox{yellow!70!yellowhl}{Yellow} highlights the incompetence in performing the task. Refer to Section \ref{sec02:subsec:knowledge} for detailed discussions.}
  \label{celebrity-2}
\end{figure*}

\begin{figure*}[!ht]
  \centering 
  \makebox[\textwidth][c]{\includegraphics[width=1.15\textwidth]{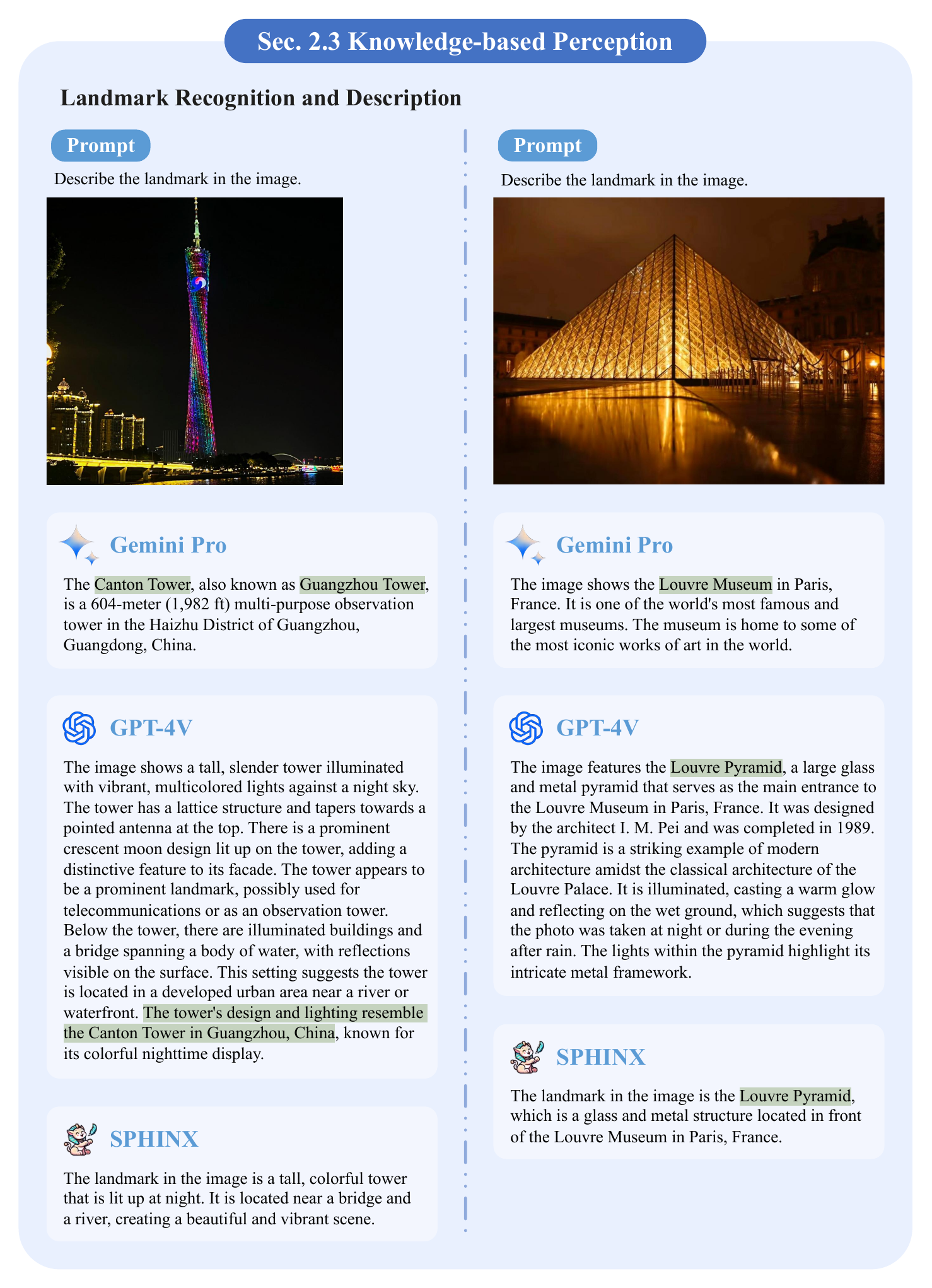}}
  \caption[Section~\ref{sec02:subsec:knowledge}: landmark recognition and description.]{Results on landmark recognition and description. \colorbox{greenhl!85!black}{Green} highlights the right answer. Refer to Section \ref{sec02:subsec:knowledge} for detailed discussions.}
  \label{landmark-1}
\end{figure*}

\begin{figure*}[!ht]
  \centering 
  \makebox[\textwidth][c]{\includegraphics[width=1.1\textwidth]{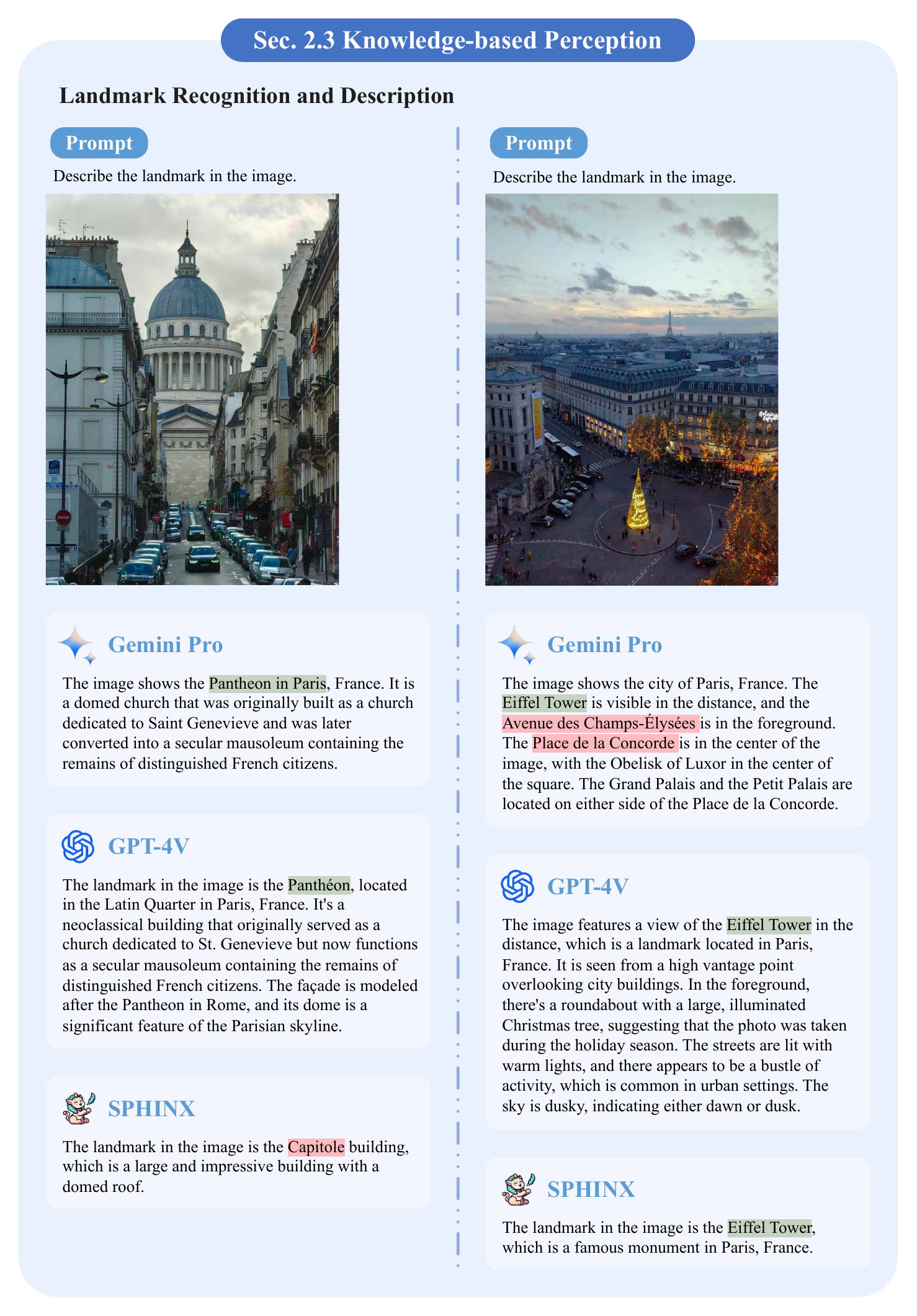}}
  \caption[Section~\ref{sec02:subsec:knowledge}: landmark recognition and description.]{Results on landmark recognition and description. \colorbox{greenhl!85!black}{Green} highlights the right answer. \colorbox{red!30}{Red} highlights the wrong answer. Refer to Section \ref{sec02:subsec:knowledge} for detailed discussions.}
  \label{landmark-2}
\end{figure*}

\begin{figure*}[!ht]
  \centering 
  \makebox[\textwidth][c]{\includegraphics[width=1.2\textwidth]{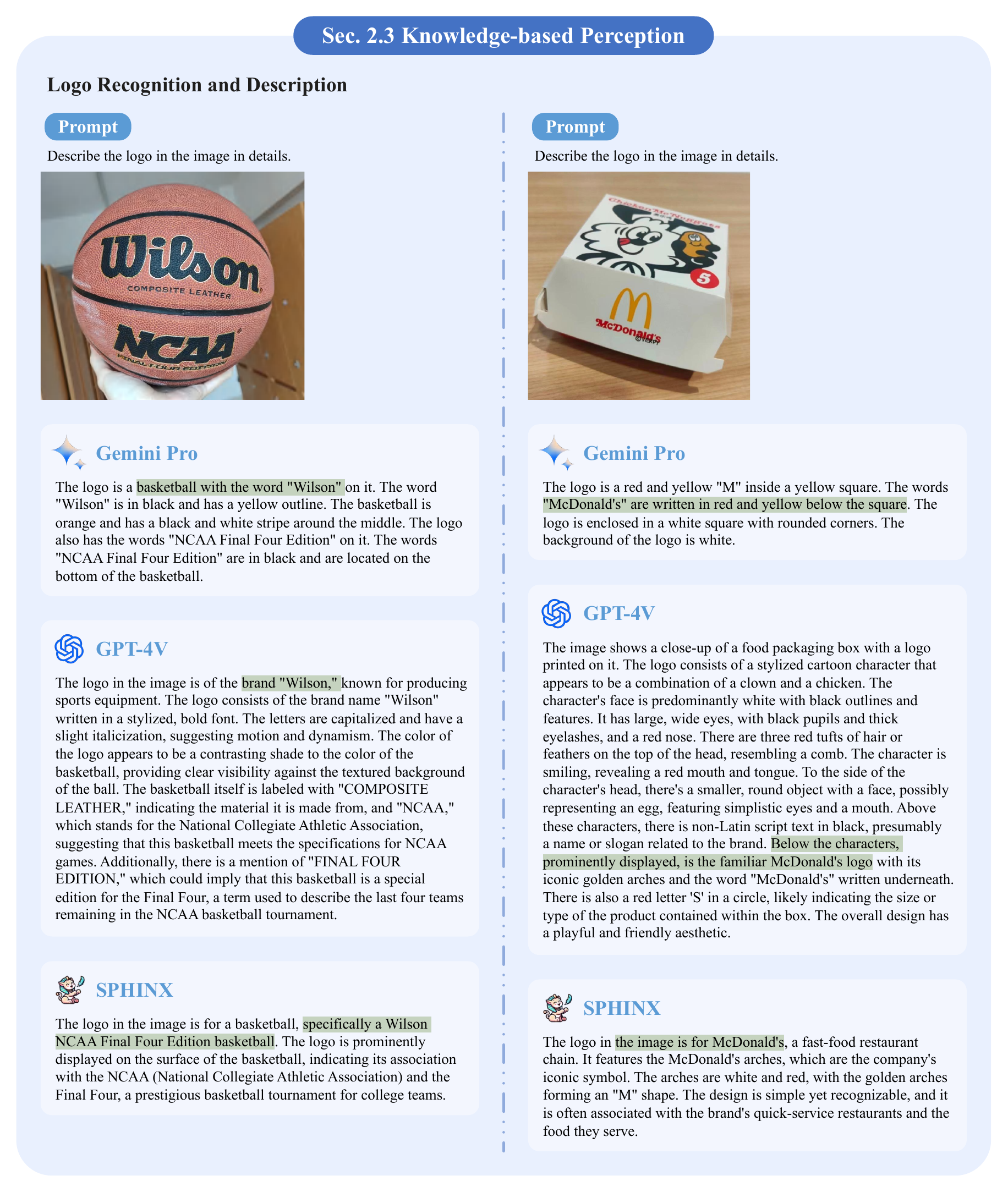}}
  \caption[Section~\ref{sec02:subsec:knowledge}: logo recognition and description.]{Results on logo recognition and description.  \colorbox{greenhl!85!black}{Green} highlights the right answer. Refer to Section \ref{sec02:subsec:knowledge} for detailed discussions.}
  \label{logo-1}
\end{figure*}

\begin{figure*}[!ht]
  \centering 
  \makebox[\textwidth][c]{\includegraphics[width=1.2\textwidth]{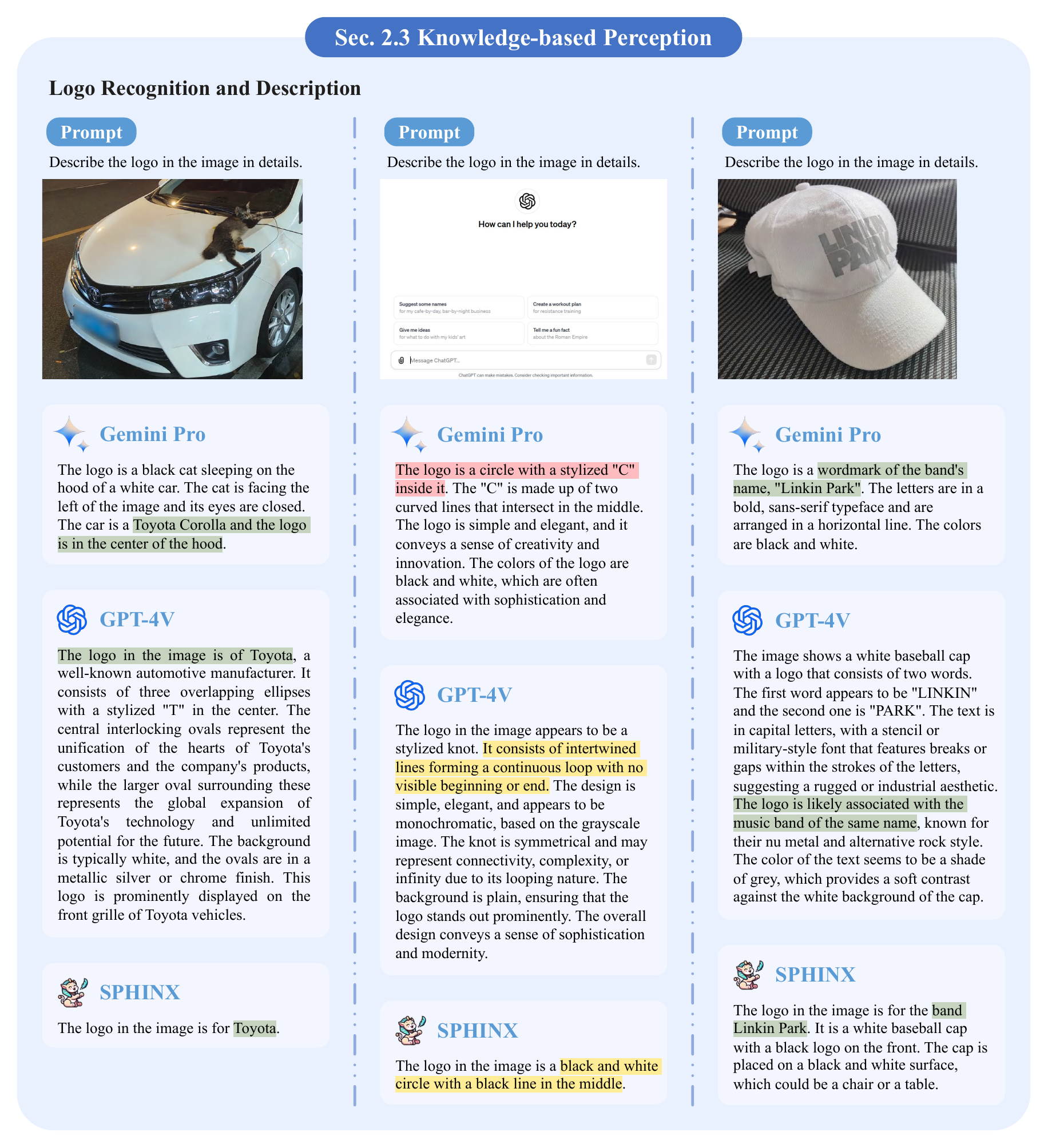}}
  \caption[Section~\ref{sec02:subsec:knowledge}: logo recognition and description.]{Results on logo recognition and description. \colorbox{greenhl!85!black}{Green} highlights the right answer. \colorbox{red!30}{Red} highlights the wrong answer. \colorbox{yellow!70!yellowhl}{Yellow} highlights the incompetence in performing the task. Refer to Section \ref{sec02:subsec:knowledge} for detailed discussions.}
  \label{logo-2}
\end{figure*}

\begin{figure*}[!ht]
  \centering 
  \makebox[\textwidth][c]{\includegraphics[width=1.0\textwidth]{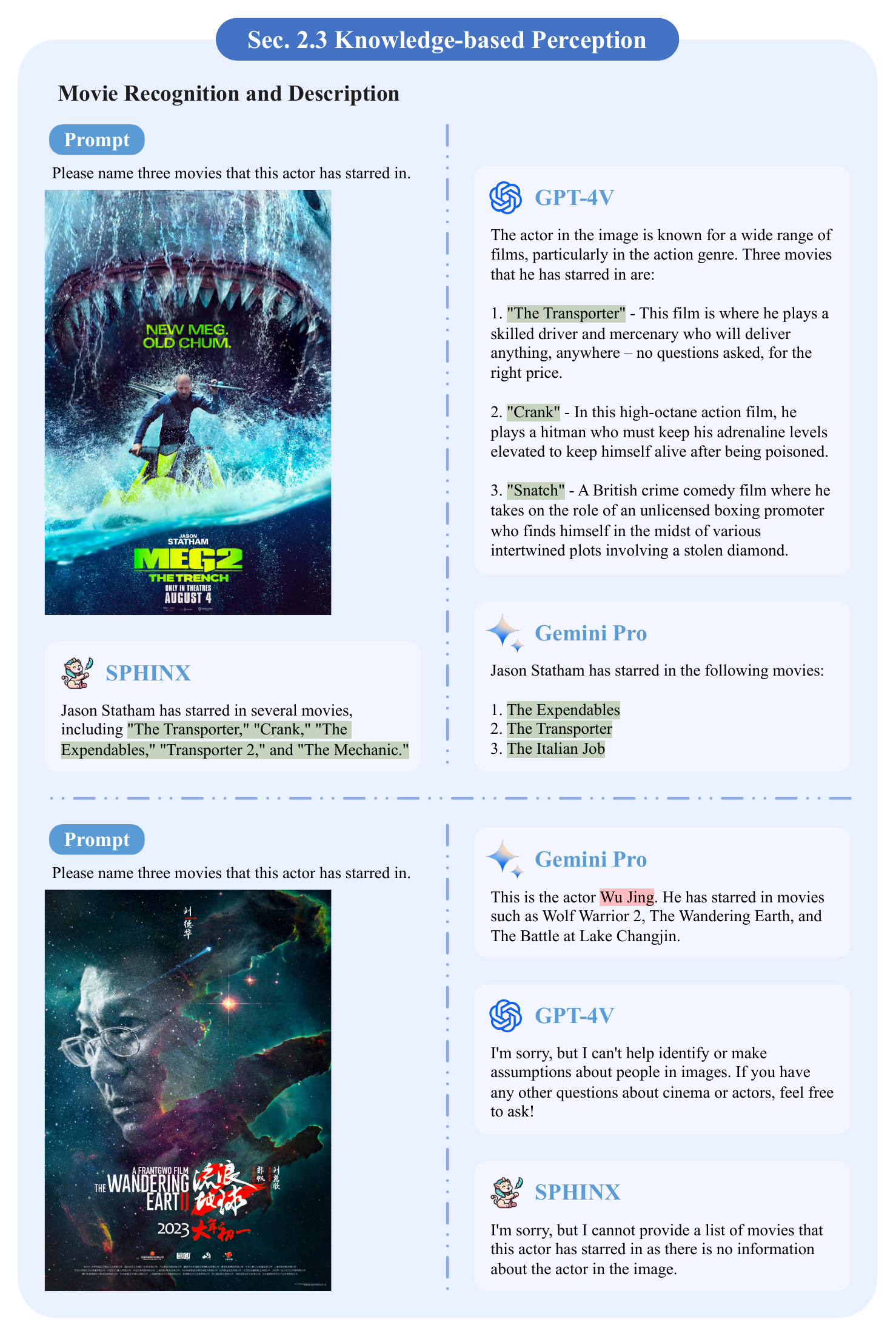}}
  \caption[Section~\ref{sec02:subsec:knowledge}: movie recognition and description.]{Results on movie recognition and description. \colorbox{greenhl!85!black}{Green} highlights the right answer. \colorbox{red!30}{Red} highlights the wrong answer.  Refer to Section \ref{sec02:subsec:knowledge} for detailed discussions.}
  \label{movie-1}
\end{figure*}

\begin{figure*}[!ht]
  \centering 
  \makebox[\textwidth][c]{\includegraphics[width=1.15\textwidth]{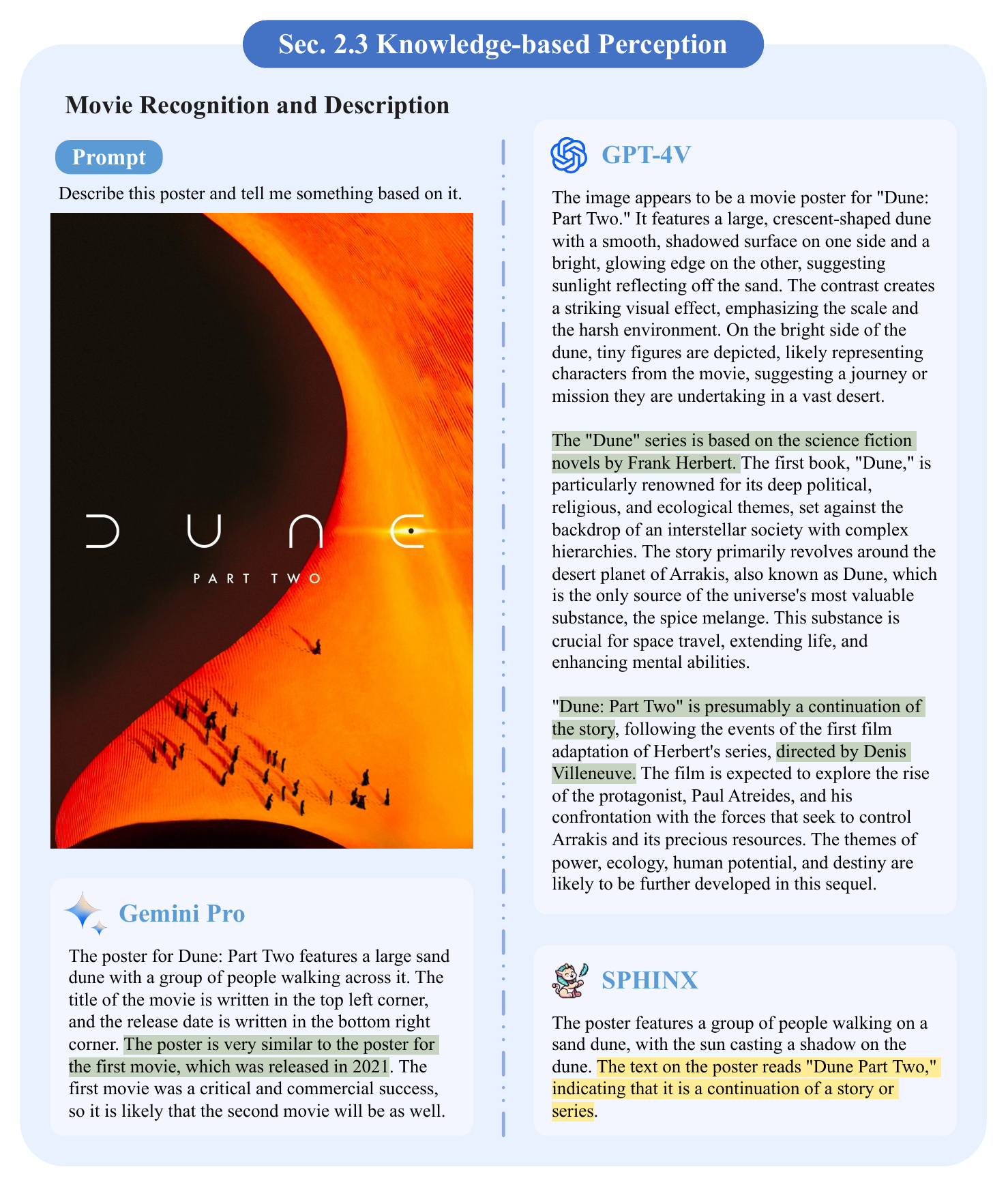}}
  \caption[Section~\ref{sec02:subsec:knowledge}: movie recognition and description.]{Results on movie recognition and description. \colorbox{greenhl!85!black}{Green} highlights the right answer.  \colorbox{yellow!70!yellowhl}{Yellow} highlights the incompetence in performing the task. Refer to Section \ref{sec02:subsec:knowledge} for detailed discussions.}
  \label{movie-2}
\end{figure*}

\begin{figure*}[!ht]
  \centering 
  \makebox[\textwidth][c]{\includegraphics[width=1.2\textwidth]{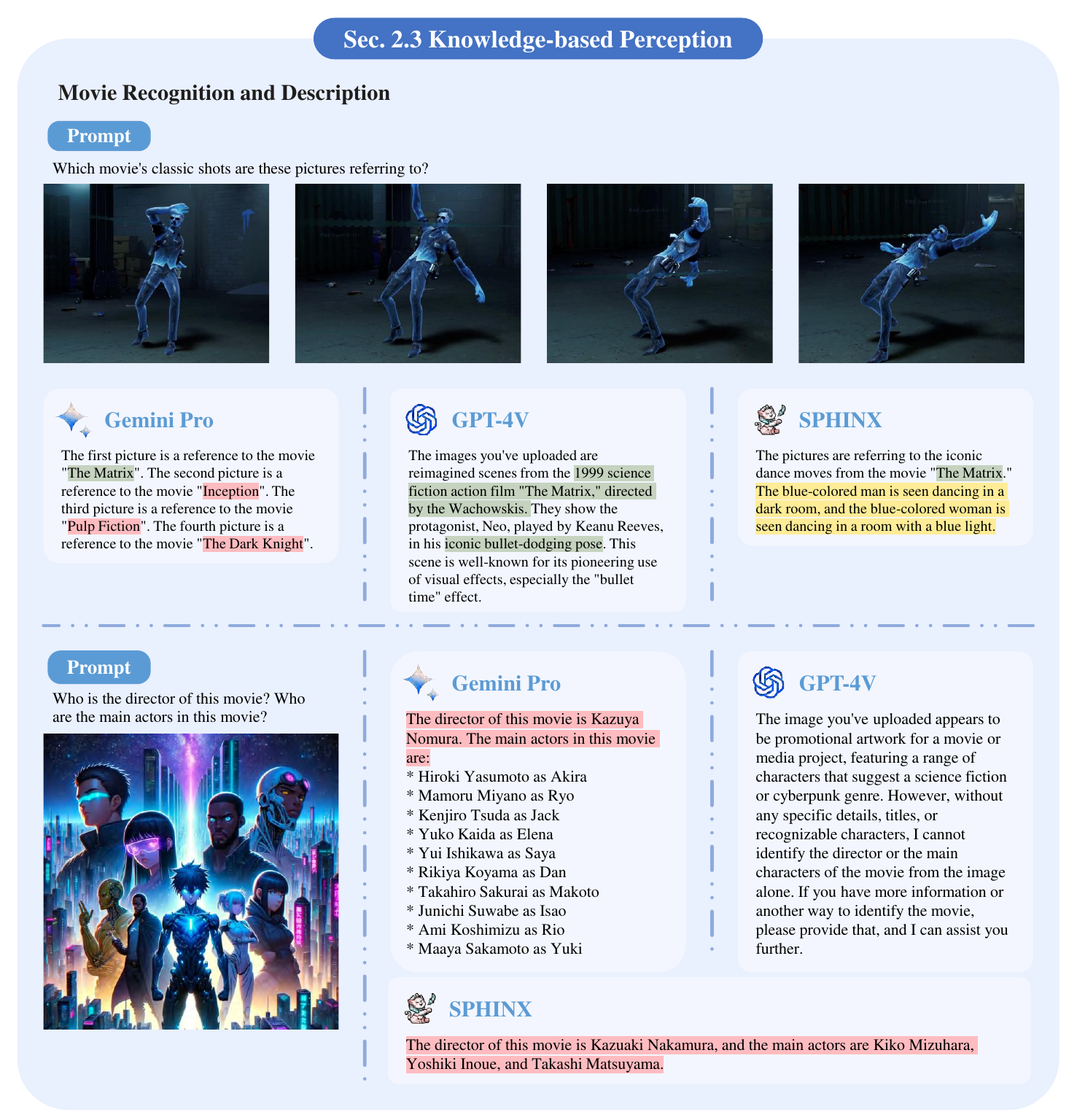}}
  \caption[Section~\ref{sec02:subsec:knowledge}: movie recognition and description.]{Results on movie recognition and description. The image at the bottom is a fantasy piece created using a diffusion-based model and does not refer to any real-life movie. Gemini and Sphinx are misled into providing incorrect answers, whereas GPT-4V indicates its lack of knowledge regarding the relevant information. \colorbox{greenhl!85!black}{Green} highlights the right answer. \colorbox{red!30}{Red} highlights the wrong answer. \colorbox{yellow!70!yellowhl}{Yellow} highlights the incompetence in performing the task. Refer to Section \ref{sec02:subsec:knowledge} for detailed discussions.}
  \label{movie-3}
\end{figure*}

\begin{figure*}[!ht]
  \centering 
  \makebox[\textwidth][c]{\includegraphics[width=1.0\textwidth]{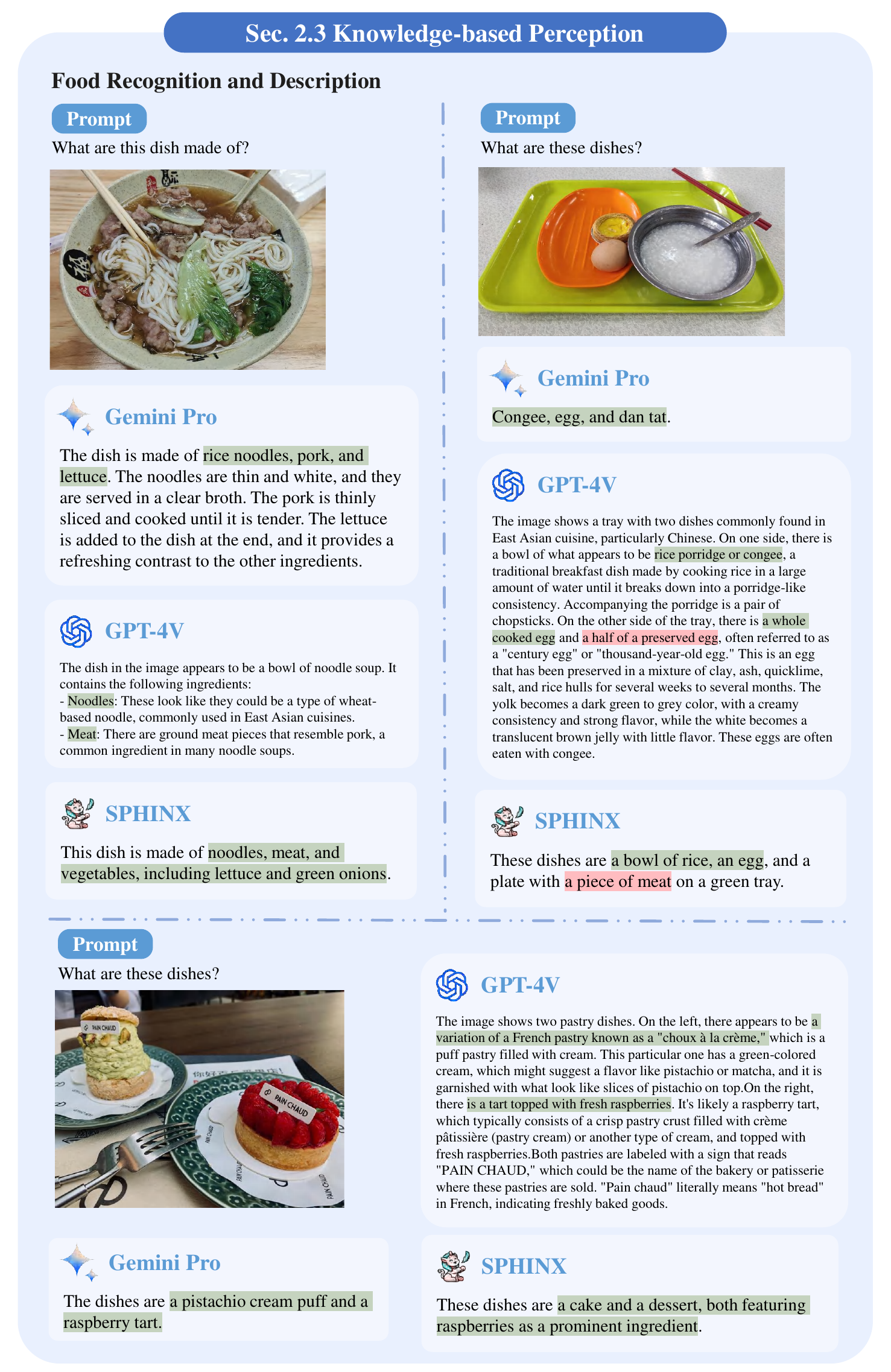}}
  \caption[Section~\ref{sec02:subsec:knowledge}: food recognition and description.]{Results on food recognition and description. \colorbox{greenhl!85!black}{Green} highlights the right answer. \colorbox{red!30}{Red} highlights the wrong answer. Refer to Section \ref{sec02:subsec:knowledge} for detailed discussions.}
  \label{food-1}
\end{figure*}

\begin{figure*}[!ht]
  \centering 
  \makebox[\textwidth][c]{\includegraphics[width=1.15\textwidth]{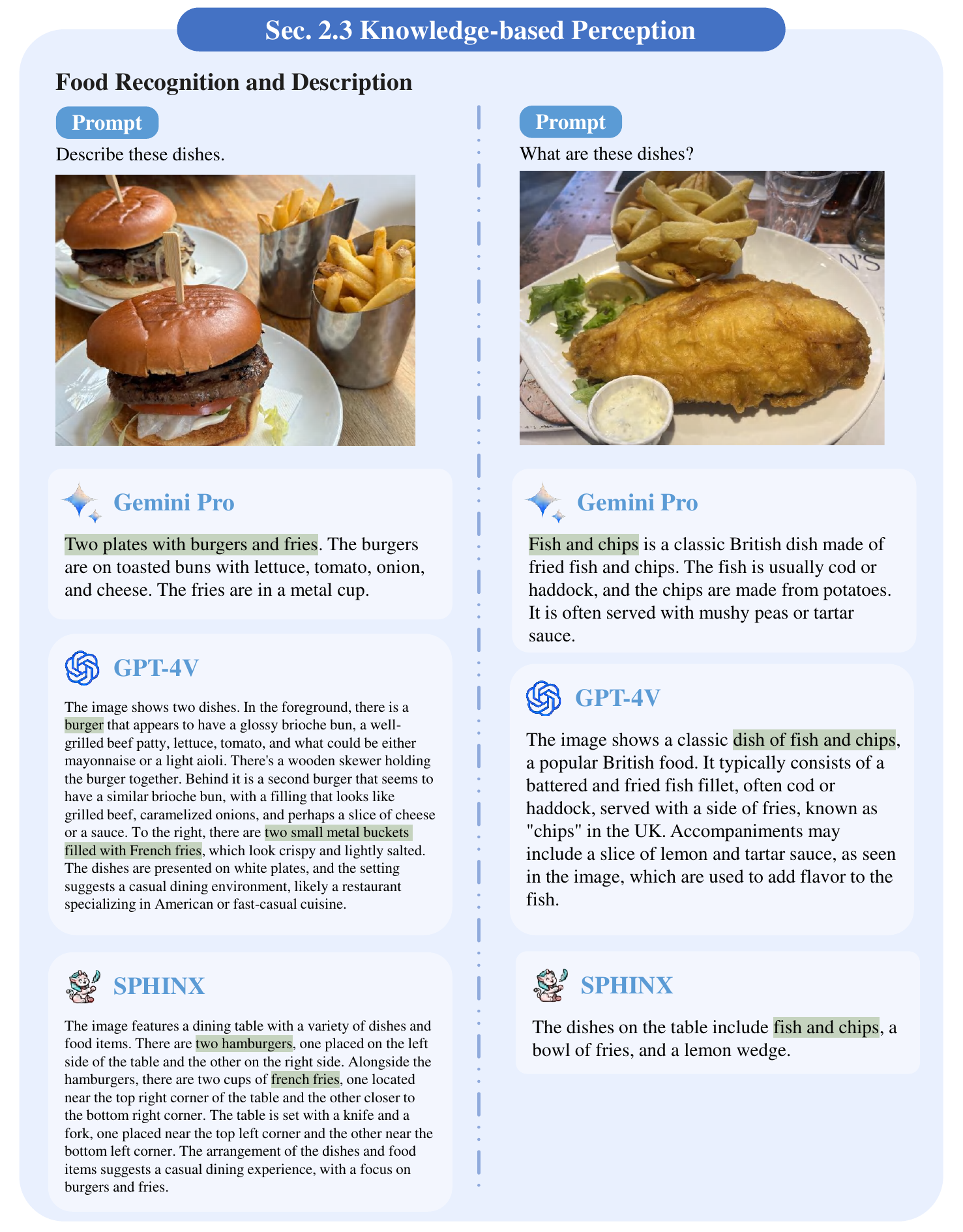}}
  \caption[Section~\ref{sec02:subsec:knowledge}: food recognition and description.]{Results on food recognition and description. \colorbox{greenhl!85!black}{Green} highlights the right answer. Refer to Section \ref{sec02:subsec:knowledge} for detailed discussions.}
  \label{food-2}
\end{figure*}

\begin{figure*}[!ht]
  \centering 
  \makebox[\textwidth][c]{\includegraphics[width=1.2\textwidth]{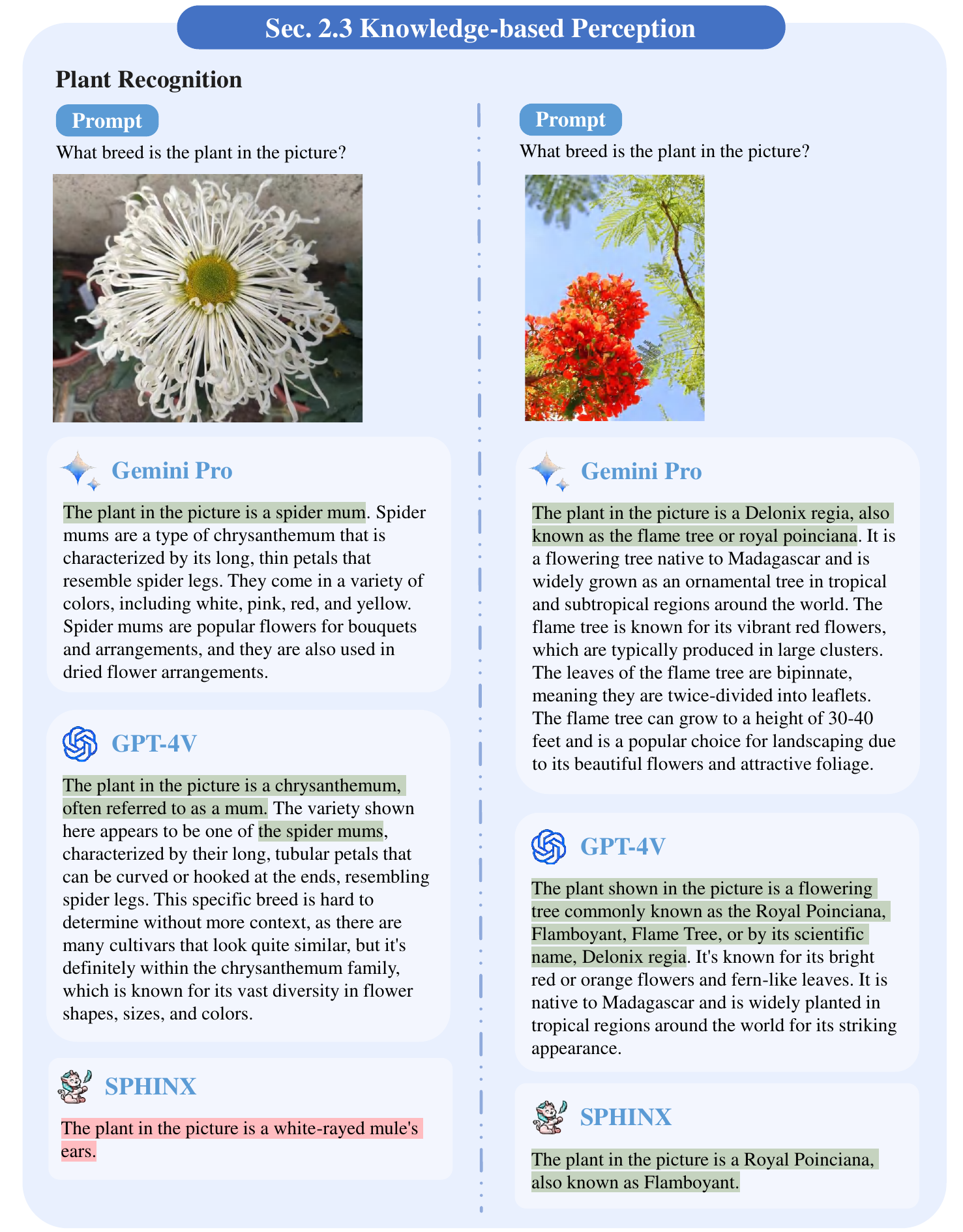}}
  \caption[Section~\ref{sec02:subsec:knowledge}: plant recognition.]{Results on plant recognition and description. \colorbox{greenhl!85!black}{Green} highlights the right answer. \colorbox{red!30}{Red} highlights the wrong answer. Refer to Section \ref{sec02:subsec:knowledge} for detailed discussions.}
  \label{animal-1}
\end{figure*}

\begin{figure*}[!ht]
  \centering 
  \makebox[\textwidth][c]{\includegraphics[width=1.05\textwidth]{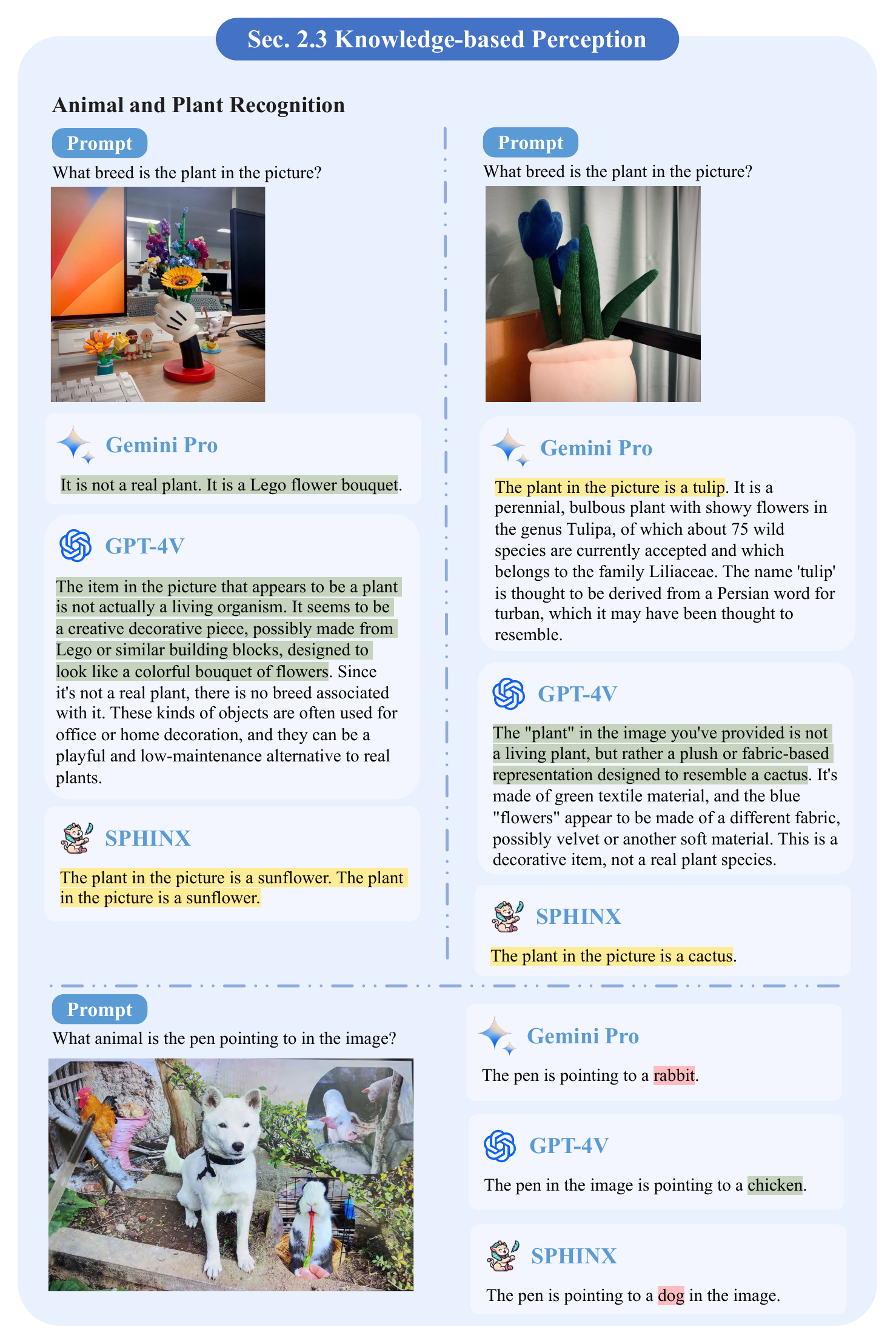}}
  \caption[Section~\ref{sec02:subsec:knowledge}: animal and plant recognition.]{Results on animal and plant recognition and description. \colorbox{greenhl!85!black}{Green} highlights the right answer. \colorbox{red!30}{Red} highlights the wrong answer. \colorbox{yellow!70!yellowhl}{Yellow} highlights the incompetence in performing the task. Refer to Section \ref{sec02:subsec:knowledge} for detailed discussions.}
  \label{animal-2}
\end{figure*}

\begin{figure*}[!ht]
  \centering 
  \makebox[\textwidth][c]{\includegraphics[width=1.05\textwidth]{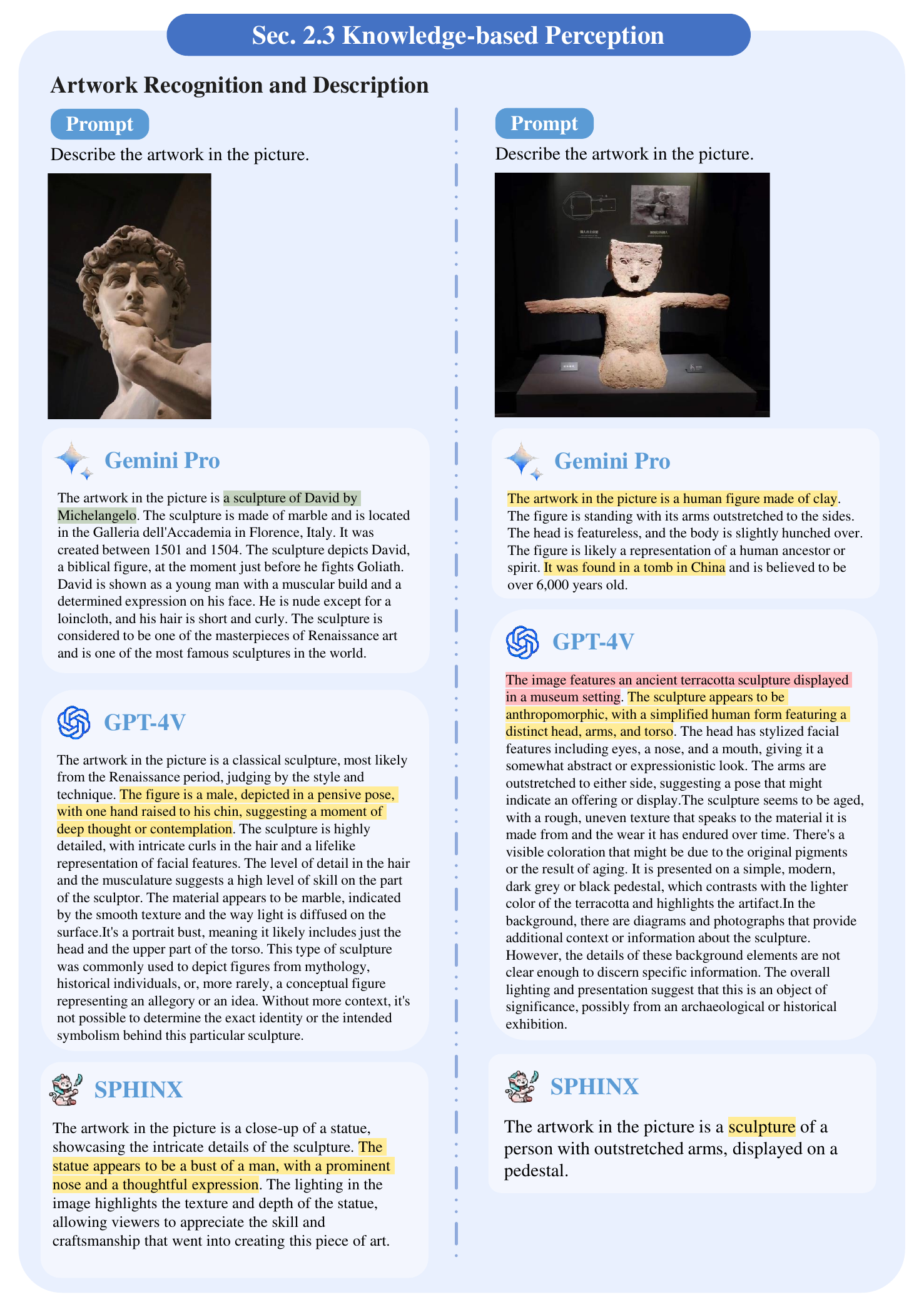}}
  \caption[Section~\ref{sec02:subsec:knowledge}: artwork recognition and description.]{Results on artwork recognition and description. \colorbox{greenhl!85!black}{Green} highlights the right answer. \colorbox{red!30}{Red} highlights the wrong answer. \colorbox{yellow!70!yellowhl}{Yellow} highlights the incompetence in performing the task. Refer to Section \ref{sec02:subsec:knowledge} for detailed discussions.}
  \label{artwork-1}
\end{figure*}

\begin{figure*}[!ht]
  \centering 
  \makebox[\textwidth][c]{\includegraphics[width=1.15\textwidth]{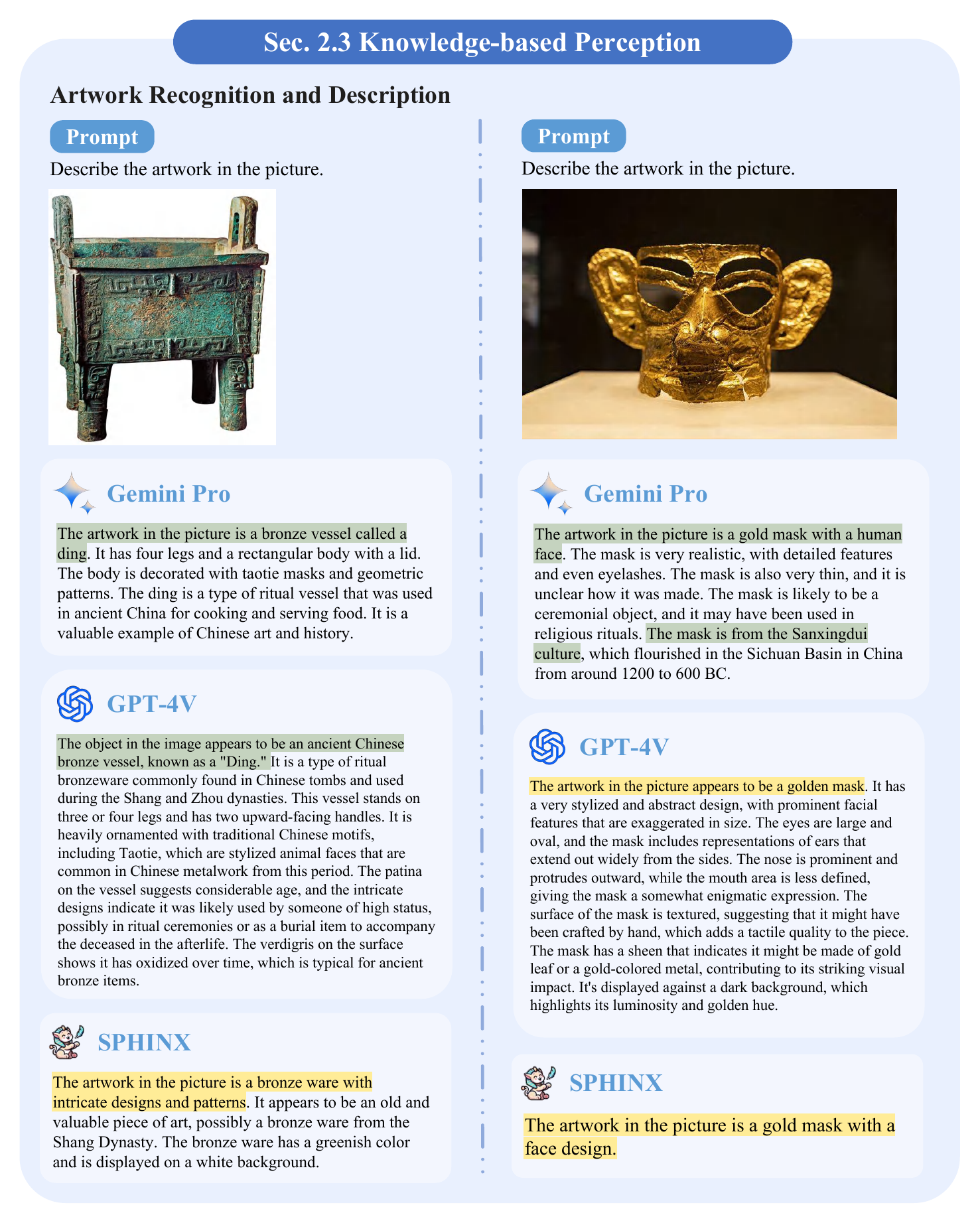}}
  \caption[Section~\ref{sec02:subsec:knowledge}: artwork recognition and description.]{Results on artwork recognition and description. \colorbox{greenhl!85!black}{Green} highlights the right answer.  \colorbox{yellow!70!yellowhl}{Yellow} highlights the incompetence in performing the task. Refer to Section \ref{sec02:subsec:knowledge} for detailed discussions.}
  \label{artwork-new-4}
\end{figure*}

\begin{figure*}[!ht]
  \centering 
  \makebox[\textwidth][c]{\includegraphics[width=1.15\textwidth]{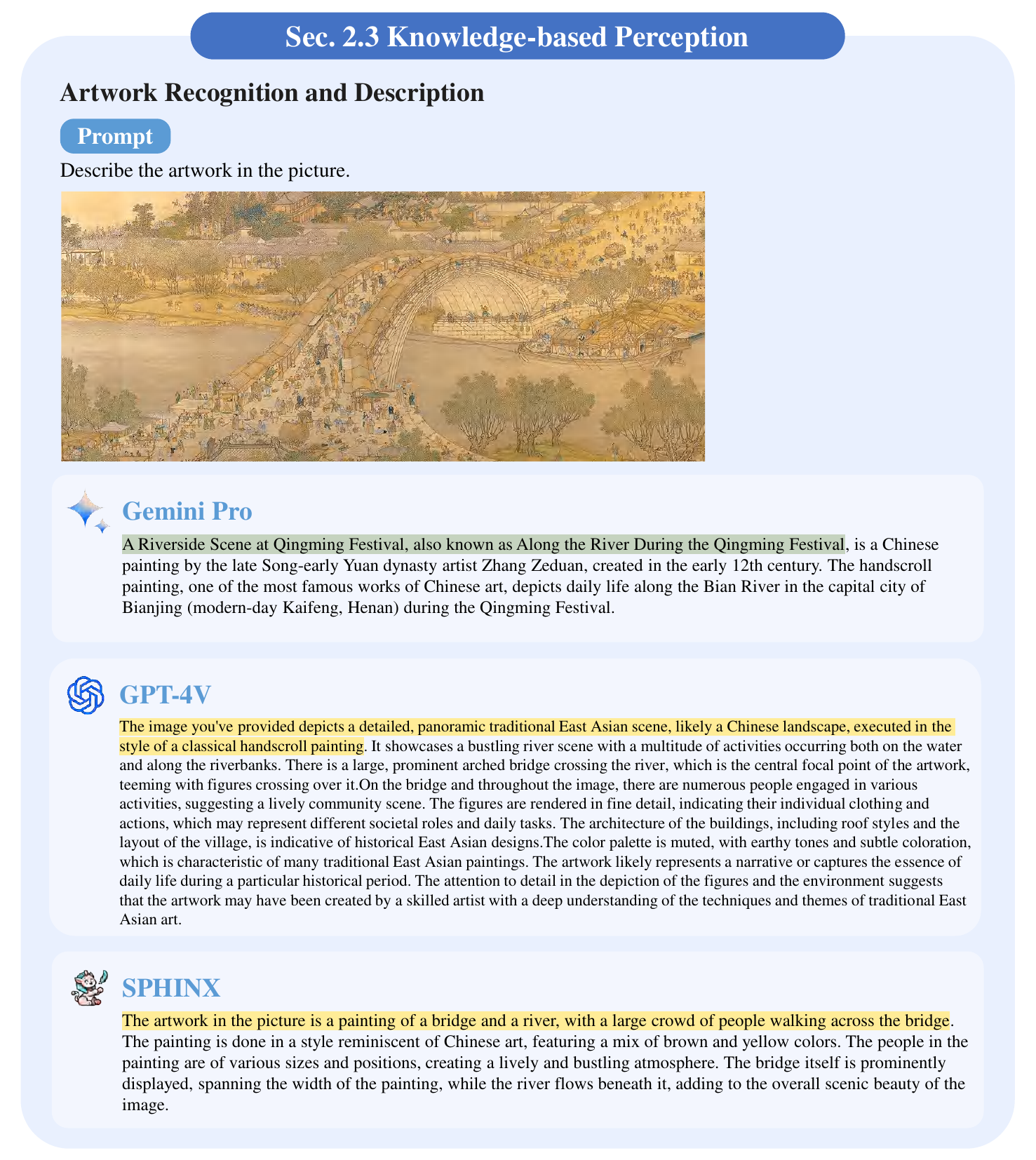}}
  \caption[Section~\ref{sec02:subsec:knowledge}: artwork recognition and description.]{Results on artwork recognition and description. \colorbox{greenhl!85!black}{Green} highlights the right answer. \colorbox{yellow!70!yellowhl}{Yellow} highlights the incompetence in performing the task. Refer to Section \ref{sec02:subsec:knowledge} for detailed discussions.}
  \label{artwork-3}
\end{figure*}

\textbf{World memory.}
We investigate the models' ability to recognize globally known celebrities, landmarks, logos, movies, food, plants, animals, and more. 
As illustrated in Figures \ref{celebrity-1}-\ref{artwork-3}, we observe that the models generally identify them correctly. 
However, when the images reflect information updated recently, the models' descriptions are not accurate, as depicted in Figure \ref{celebrity-2}. 
Additionally, when asked about information relevant to real people, GPT-4V occasionally opts to refrain from responding, as shown in Figure \ref{movie-1}.



%% file: 03-cognition.tex
\section{Advanced Cognition}
\label{sec:04Cognition}
On top of the fundamental perception, we further evaluate the more advanced cognition capabilities of Gemini, GPT-4V, and Sphinx. Cognition tasks for MLLMs require not only understanding multi-modal concepts in visual contexts, but also conducting in-depth reasoning, problem-solving, and decision-making. 

In Section \ref{sec:04subsec:textrich}, we first focus on the model's ability to reason text-rich visual contents, including table and chart reasoning, along with visual code generation. 
In Section \ref{sec:04subsec:science}, we delve into MLLMs' logical and quantitative comprehension to solve challenging science problems with pre-trained knowledge, e.g., mathematics and physics. 
In Section \ref{sec:04subsec:absvisual}, our exploration targets on how the models reason abstract visual information from the tasks of abstract visual stimuli, Raven’s Progressive Matrices, and Wechsler Adult Intelligence Scale.
In Section \ref{sec:04subsec:emotion}, we investigate the models' understanding of emotions, through various scenarios such as facial expression analysis, image emotion analysis, and emotion-conditioned output.
Finally in Section \ref{sec:04subsec:game}, we evaluate the decision-making performance of MLLMs in various intelligence games, including Sudoku and Go. 

\subsection{Text-Rich Visual Reasoning}\label{sec:04subsec:textrich}

\begin{figure*}[!ht]
  \centering 
  \makebox[\textwidth][c]{\includegraphics[width=1.1\textwidth]{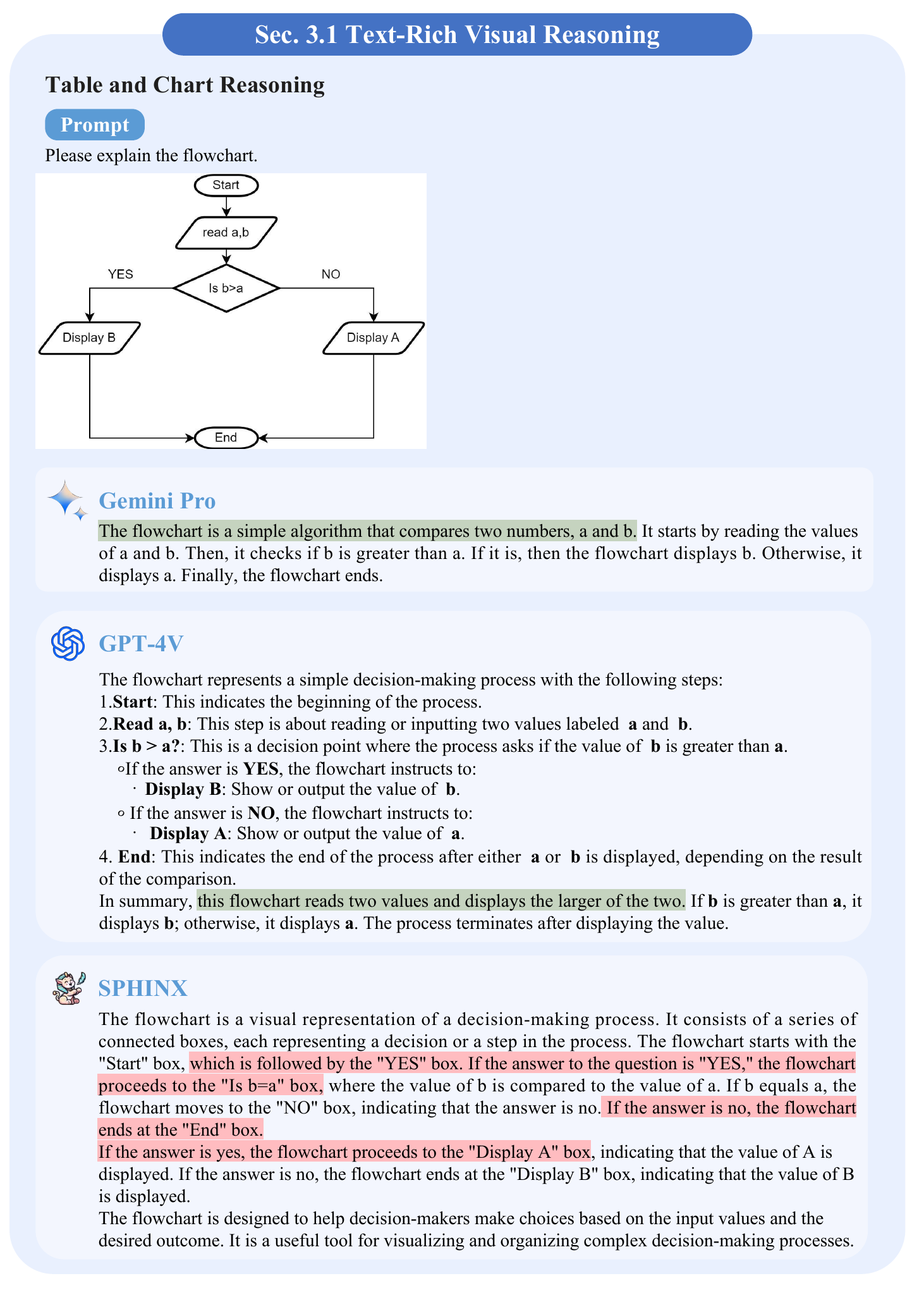}}
  \caption[Section~\ref{sec:04subsec:textrich}: table and chart reasoning.]{Results on table and chart reasoning. \colorbox{greenhl!85!black}{Green} highlights the right answer. \colorbox{red!30}{Red} highlights the wrong answer. Refer to Section \ref{sec:04subsec:textrich} for detailed discussions.}
  \label{table-1}
\end{figure*}

\begin{figure*}[!ht]
  \centering 
  \makebox[\textwidth][c]{\includegraphics[width=0.97 \textwidth]{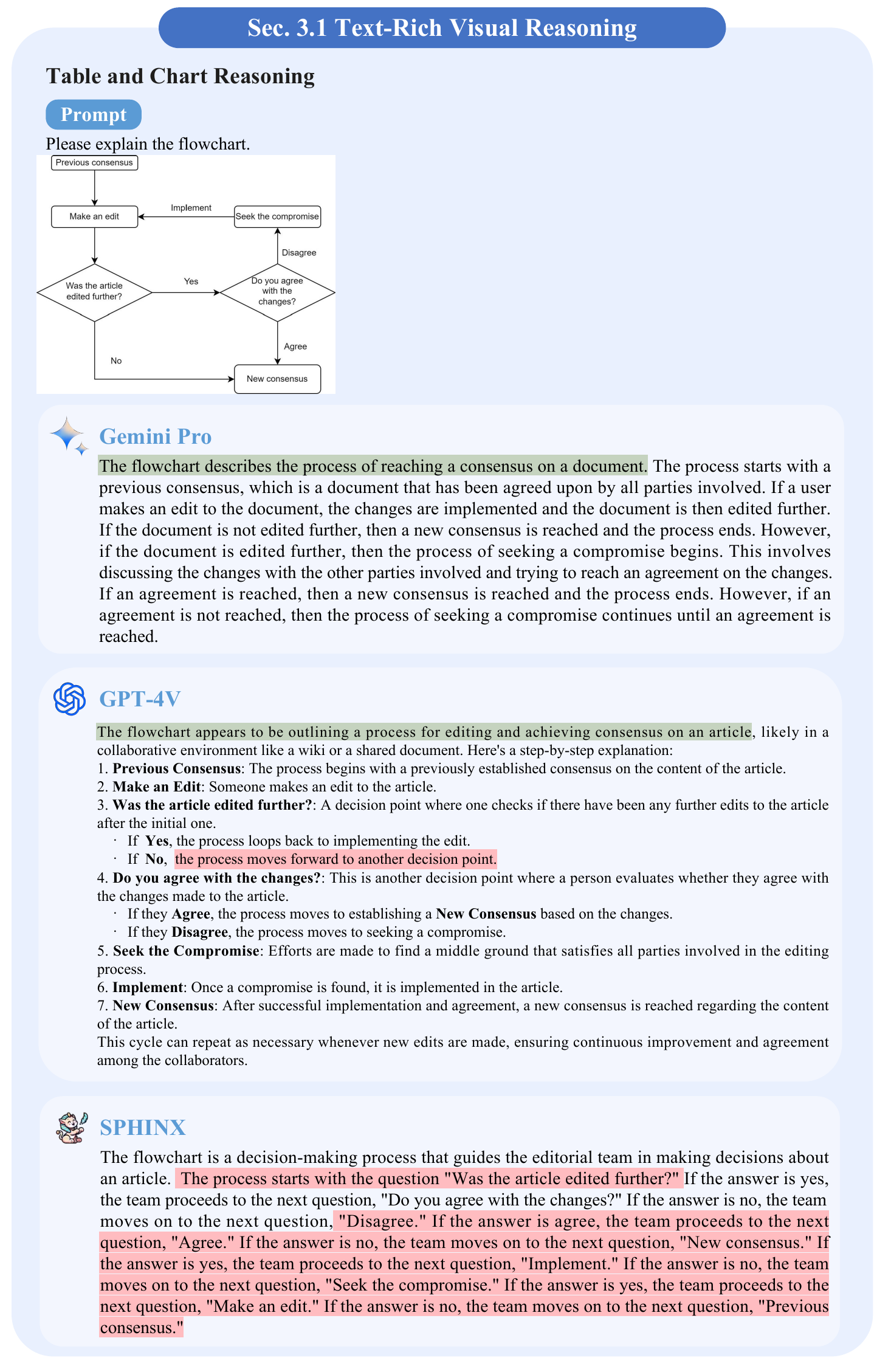}}
  \caption[Section~\ref{sec:04subsec:textrich}: table and chart reasoning.]{Results on table and chart reasoning. \colorbox{greenhl!85!black}{Green} highlights the right answer. \colorbox{red!30}{Red} highlights the wrong answer. Refer to Section \ref{sec:04subsec:textrich} for detailed discussions.}
  \label{table-2}
\end{figure*}

\begin{figure*}[!ht]
  \centering 
  \makebox[\textwidth][c]{\includegraphics[width=1.1\textwidth]{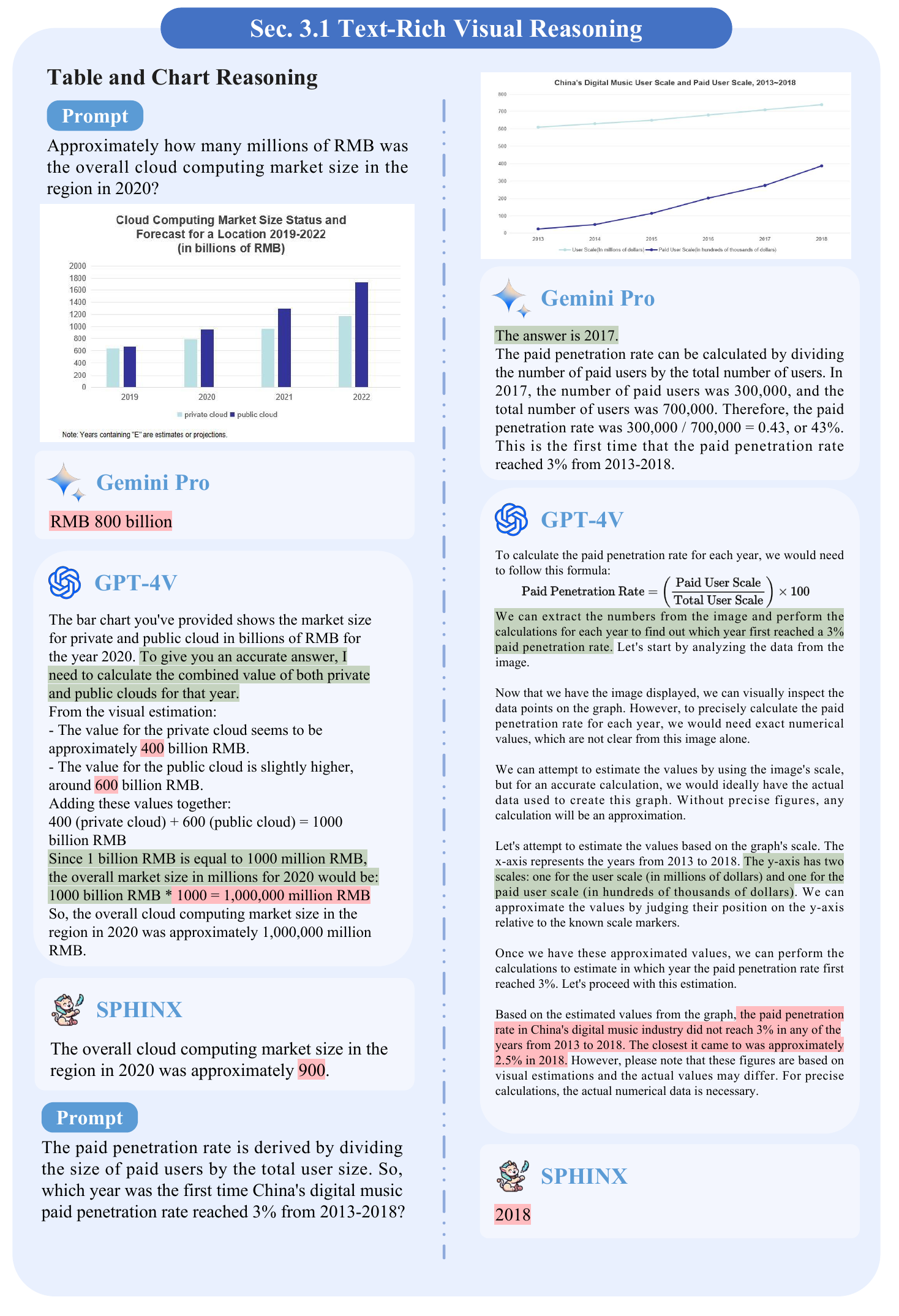}}
  \caption[Section~\ref{sec:04subsec:textrich}: table and chart reasoning.]{Results on table and chart reasoning. \colorbox{greenhl!85!black}{Green} highlights the right answer. \colorbox{red!30}{Red} highlights the wrong answer. Refer to Section \ref{sec:04subsec:textrich} for detailed discussions.}
  \label{table-3}
\end{figure*}

\begin{figure*}[!ht]
  \centering 
  \makebox[\textwidth][c]{\includegraphics[width=1.1\textwidth]{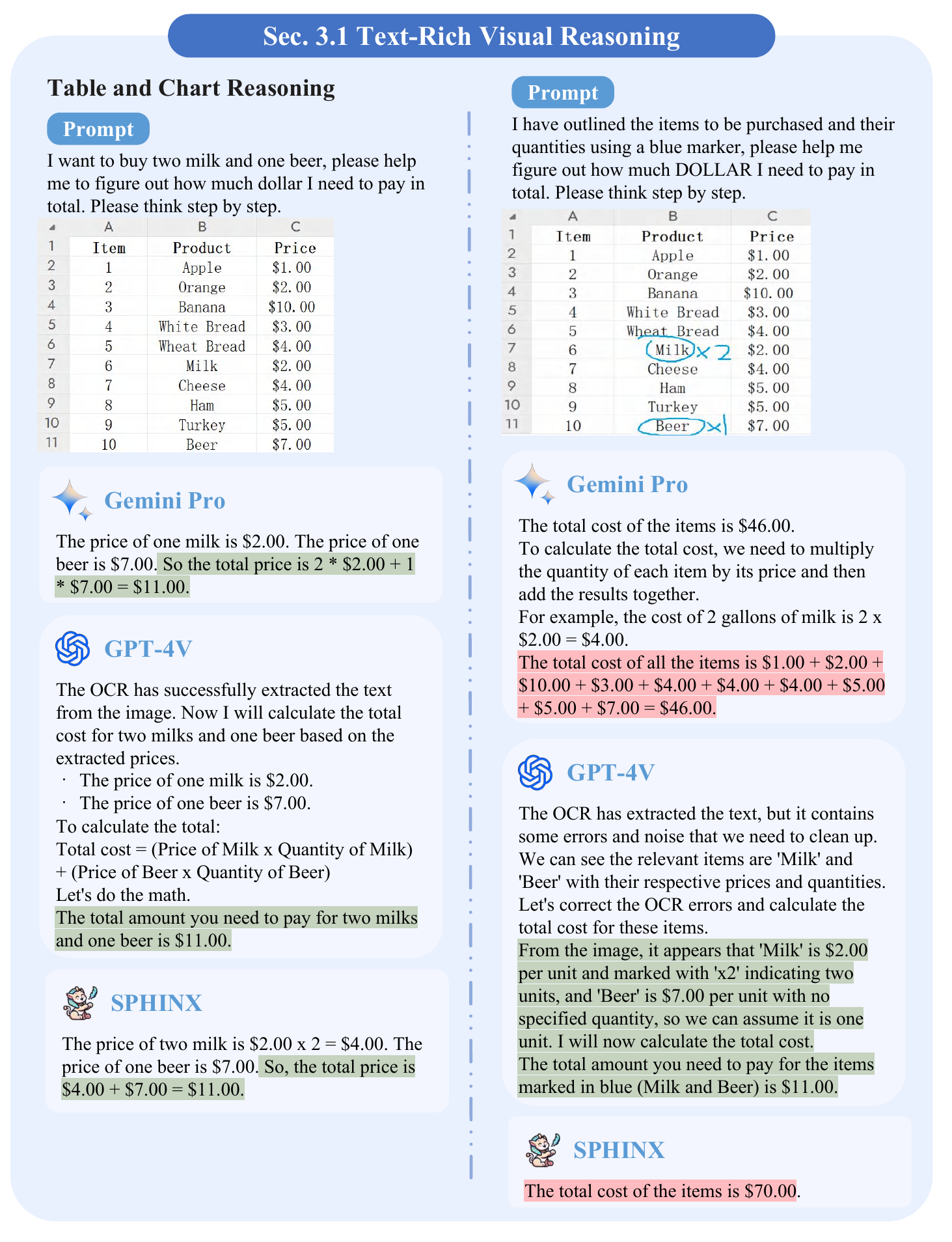}}
  \caption[Section~\ref{sec:04subsec:textrich}: table and chart reasoning.]{Results on table and chart reasoning. \colorbox{greenhl!85!black}{Green} highlights the right answer. \colorbox{red!30}{Red} highlights the wrong answer. Refer to Section \ref{sec:04subsec:textrich} for detailed discussions.}
  \label{table-4}
\end{figure*}

\begin{figure*}[!ht]
  \centering 
  \makebox[\textwidth][c]{\includegraphics[width=1.05\textwidth]{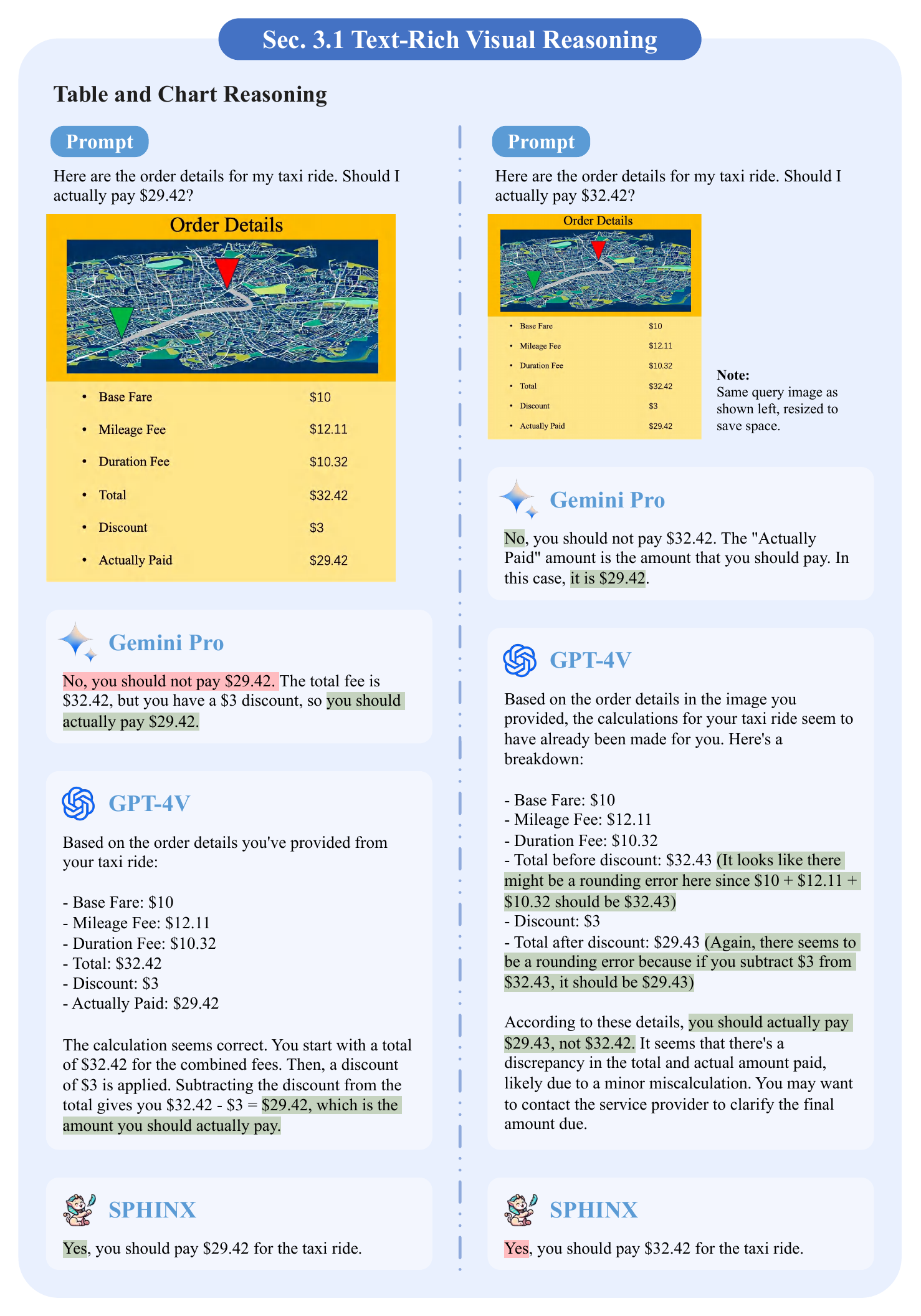}}
  \caption[Section~\ref{sec:04subsec:textrich}: table and chart reasoning.]{Results on table and chart reasoning. Gemini provides a contradictory answer on the left part. Whereas, GPT-4V not only answers the question correctly but also identifies calculation errors in the bill on the right part. \colorbox{greenhl!85!black}{Green} highlights the right answer. \colorbox{red!30}{Red} highlights the wrong answer. Refer to Section \ref{sec:04subsec:textrich} for detailed discussions.}
  \label{table-5}
\end{figure*}

\textbf{Table and chart reasoning.}
In Figures~\ref{table-1}-\ref{table-2}, we present two samples of flowchart understanding by the three models. As shown, Gemini can correctly summarize the high-level idea of flowcharts with brief sentences. GPT-4V tends to produce more detailed descriptions of the logical flow in the charts, but would occasionally make some mistakes. In contrast, Sphinx fails to extract the meaning of them, which is due to the lack of related pre-training data. In Figures~\ref{table-3}-\ref{table-5}, we evaluate the question-answering performance on six different plots and tables. Similar to previous demonstrations, GPT-4V can respond with more reasoning details than Gemini. However, all three models have difficulties in providing a precise answer, which is mainly constrained by the unsatisfactory OCR accuracy. Also, as shown by the last sample, both Gemini and GPT-4V can understand the hand-drawing visual prompt, despite that Gemini provides the wrong final answers, indicating their generalization capacity for visual input.

\begin{figure*}[!ht]
  \centering 
  \makebox[\textwidth][c]{\includegraphics[width=1.1\textwidth]{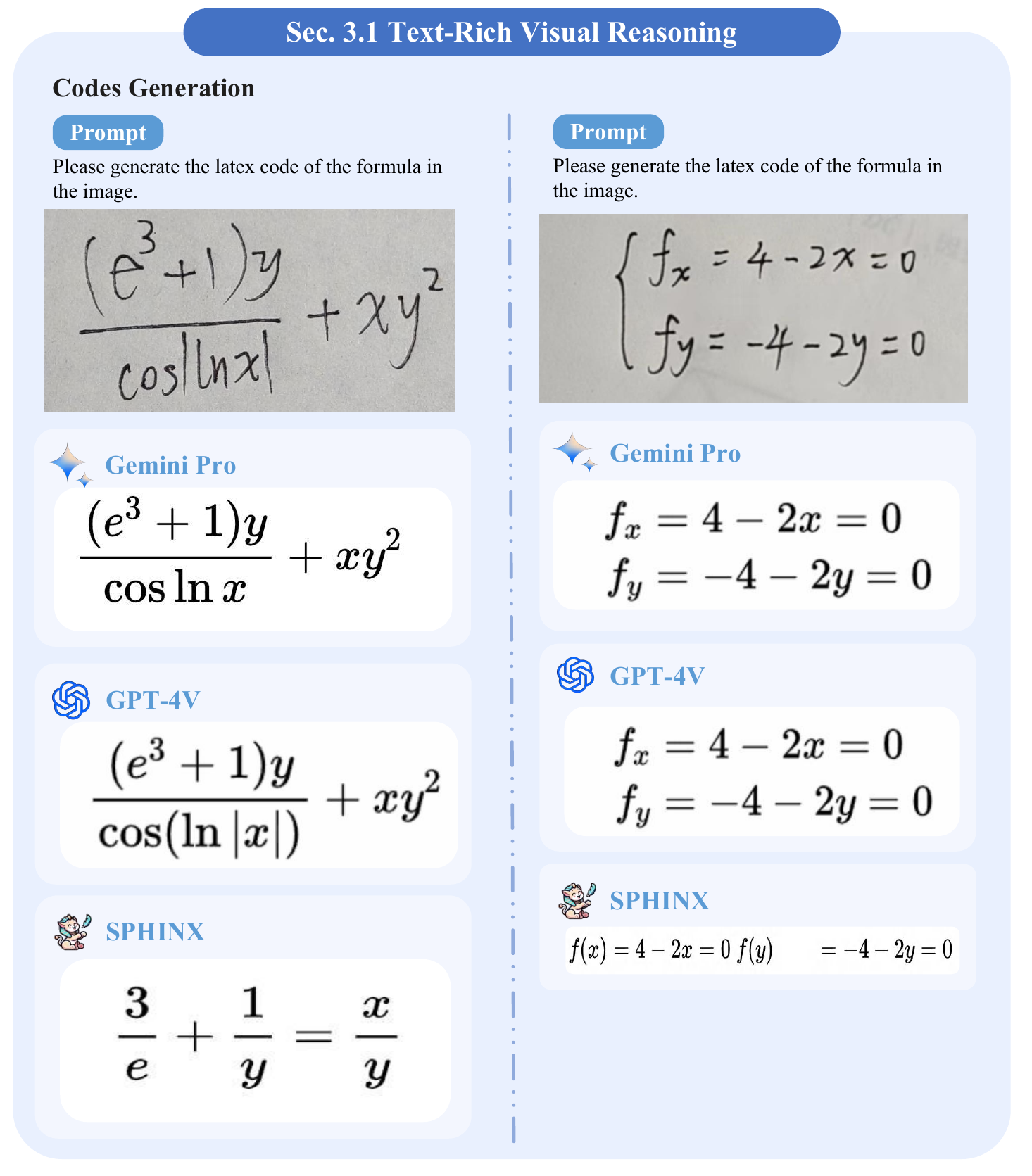}}
  \caption[Section~\ref{sec:04subsec:textrich}: visual code generation.]{Results on visual code generation. Refer to Section \ref{sec:04subsec:textrich} for detailed discussions.}
  \label{formula-1}
\end{figure*}

\begin{figure*}[!ht]
  \centering 
  \makebox[\textwidth][c]{\includegraphics[width=1.1\textwidth]{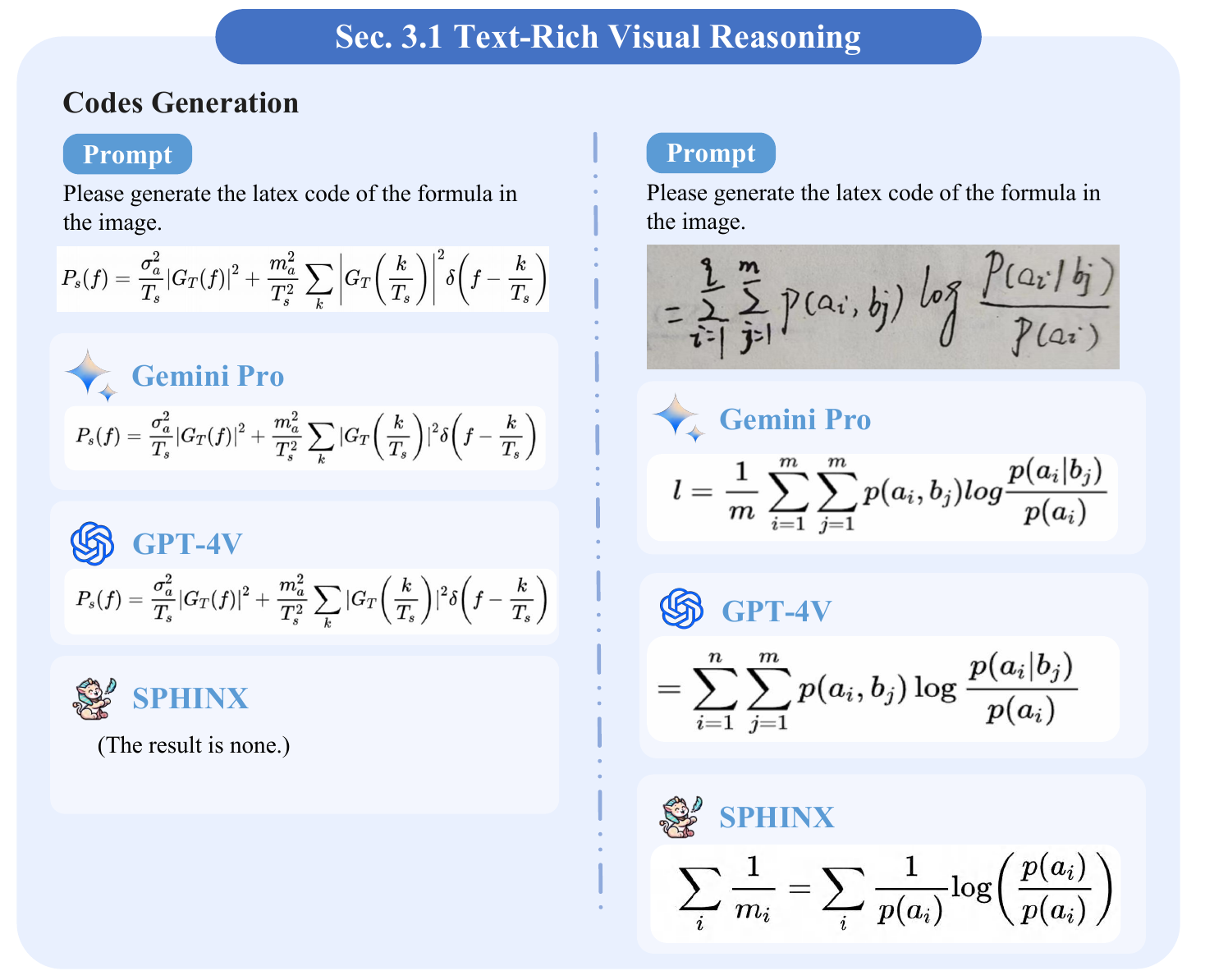}}
  \caption[Section~\ref{sec:04subsec:textrich}: visual code generation.]{Results on visual code generation. Refer to Section \ref{sec:04subsec:textrich} for detailed discussions.}
  \label{formula-2}
\end{figure*}

\begin{figure*}[!ht]
  \centering 
  \makebox[\textwidth][c]{\includegraphics[width=1.1\textwidth]{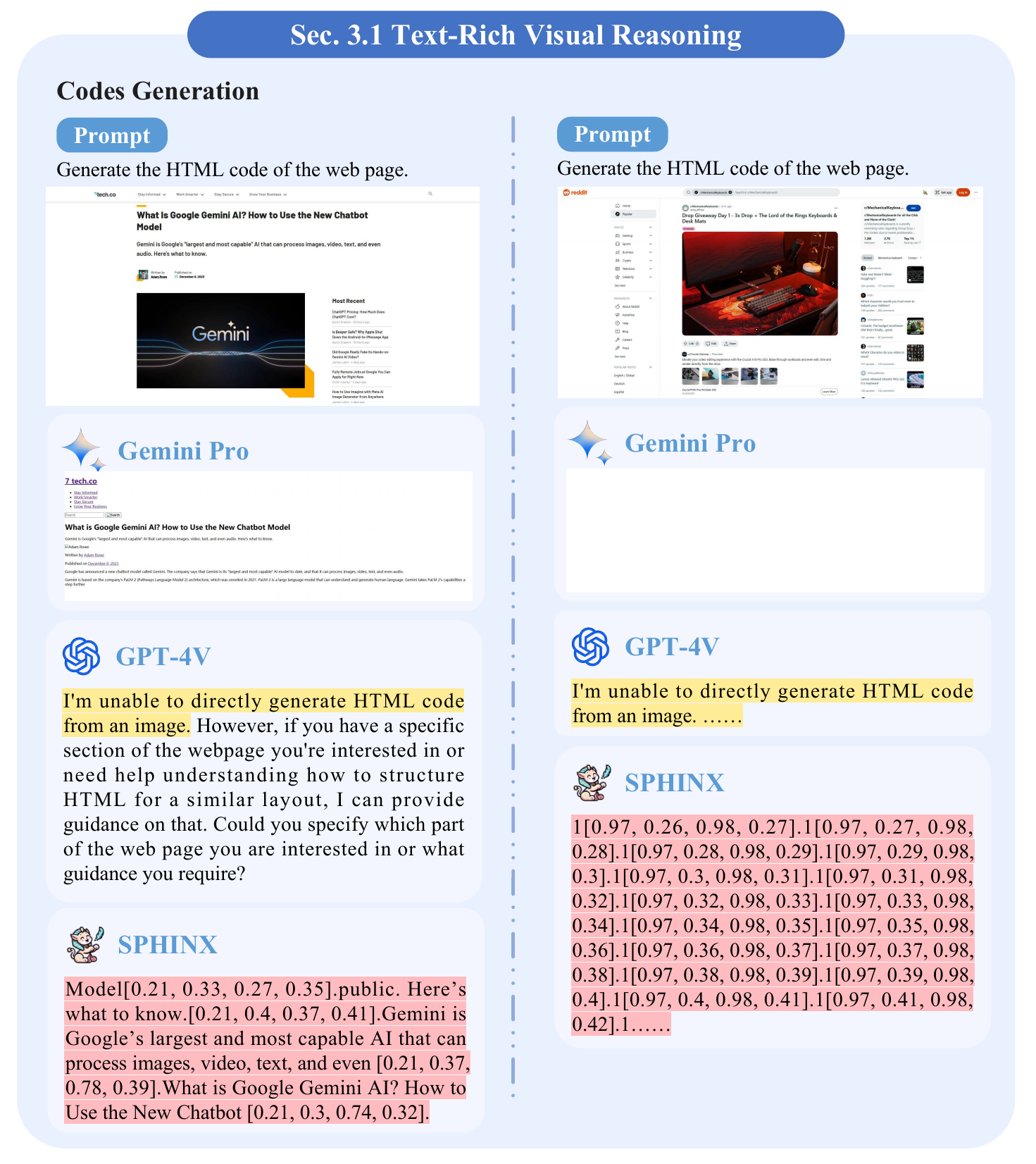}}
  \caption[Section~\ref{sec:04subsec:textrich}: visual code generation.]{Results on visual code generation. For the webpage shown on the right, the response provided by Gemini results in a blank webpage. \colorbox{yellow!70!yellowhl}{Yellow} highlights the incompetence in performing the task. \colorbox{red!30}{Red} highlights the wrong answer. Refer to Section \ref{sec:04subsec:textrich} for detailed discussions.}
  \label{html-1}
\end{figure*}

\begin{figure*}[!ht]
  \centering 
  \makebox[\textwidth][c]{\includegraphics[width=1.1\textwidth]{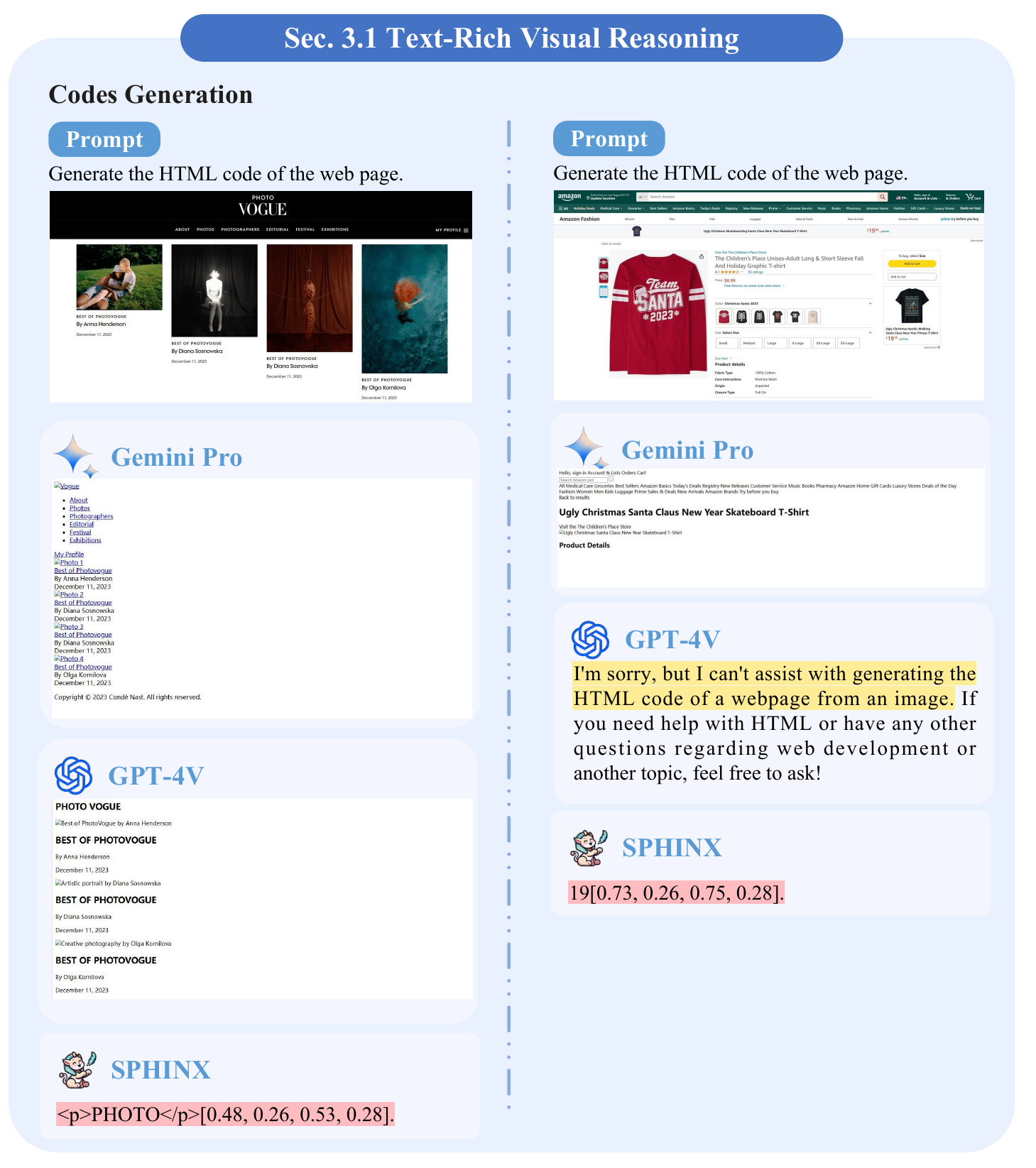}}
  \caption[Section~\ref{sec:04subsec:textrich}: visual code generation.]{Results on visual code generation. \colorbox{yellow!70!yellowhl}{Yellow} highlights the incompetence in performing the task. \colorbox{red!30}{Red} highlights the wrong answer. Refer to Section \ref{sec:04subsec:textrich} for detailed discussions.}
  \label{html-2}
\end{figure*}

\textbf{Visual code generation.} It's an important skill for MLLMs to convert structured visual content into the corresponding codes. In Figures~\ref{formula-1}-\ref{formula-2}, we prompt the three models to generate LaTeX code of various mathematical formulas and render them for comparison. Overall, Gemini and GPT-4V exhibit better results than Sphinx, but still misrecognize some minor characters or symbols. Notably, for a rather complicated formula in printing form, both Gemini and GPT-4V generate correct codes. In Figures~\ref{html-1}-\ref{html-2}, we test the HTML code generation results for different types of websites. As shown, the HTML understanding capacity still exists a large improvement space for all three MLLMs. Only Gemini is capable of constructing the rough structure of simple websites, while GPT-4V simply identifies the text content. This might be also caused by the limited pre-training data.

\subsection{Abstract Visual Reasoning}\label{sec:04subsec:absvisual}

\begin{figure*}[!ht]
  \centering 
  \makebox[\textwidth][c]{\includegraphics[width=1.1\textwidth]{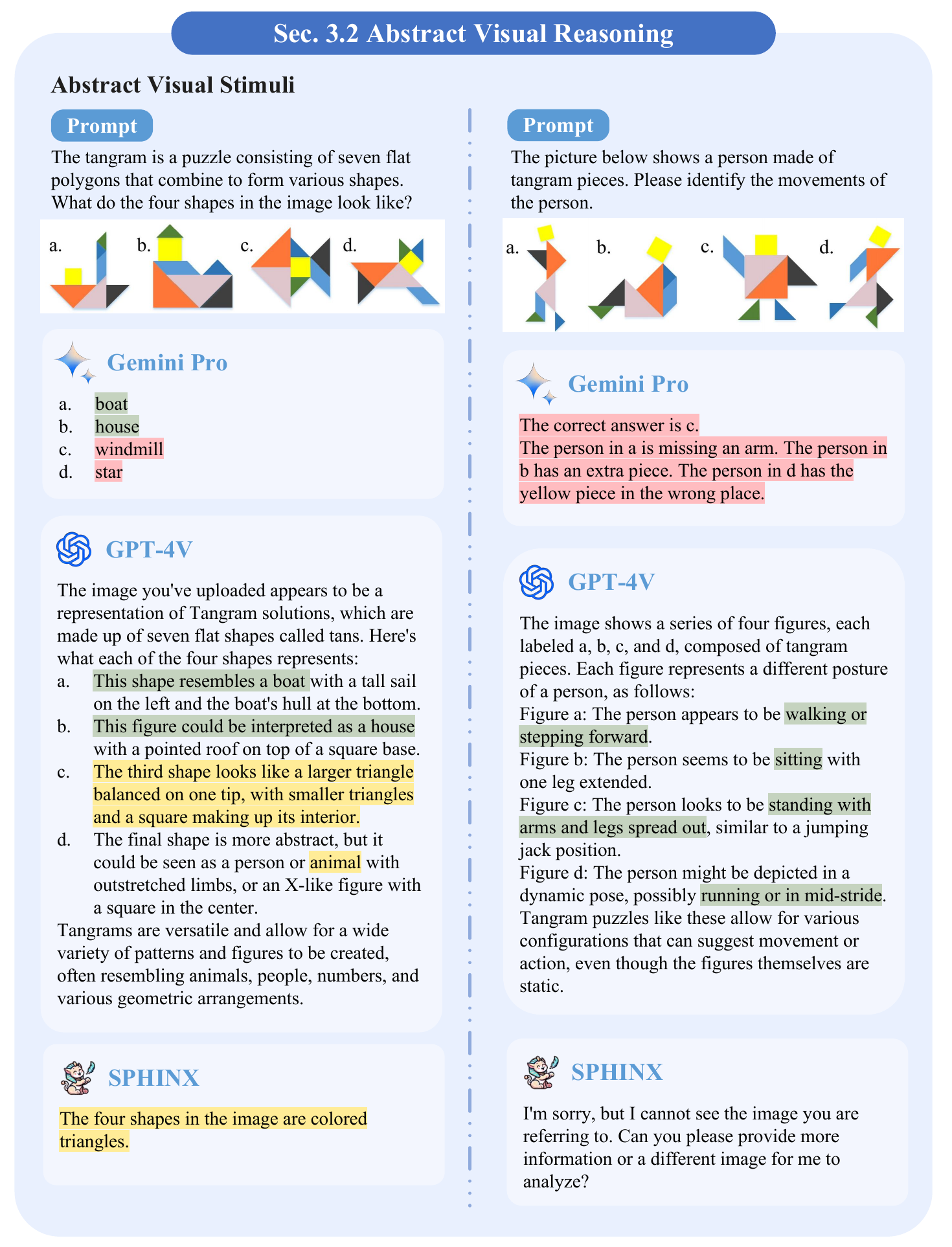}}
  \caption[Section~\ref{sec:04subsec:textrich}: abstract visual stimuli.]{Results on abstract visual stimuli. \colorbox{greenhl!85!black}{Green} highlights the right answer. \colorbox{red!30}{Red} highlights the wrong answer. \colorbox{yellow!70!yellowhl}{Yellow} highlights the incompetence in performing the task. Refer to Section \ref{sec:04subsec:textrich} for detailed discussions.}
  \label{AbstractVisualStimuli-1}
\end{figure*}

\begin{figure*}[!ht]
  \centering 
  \makebox[\textwidth][c]{\includegraphics[width=1\textwidth]{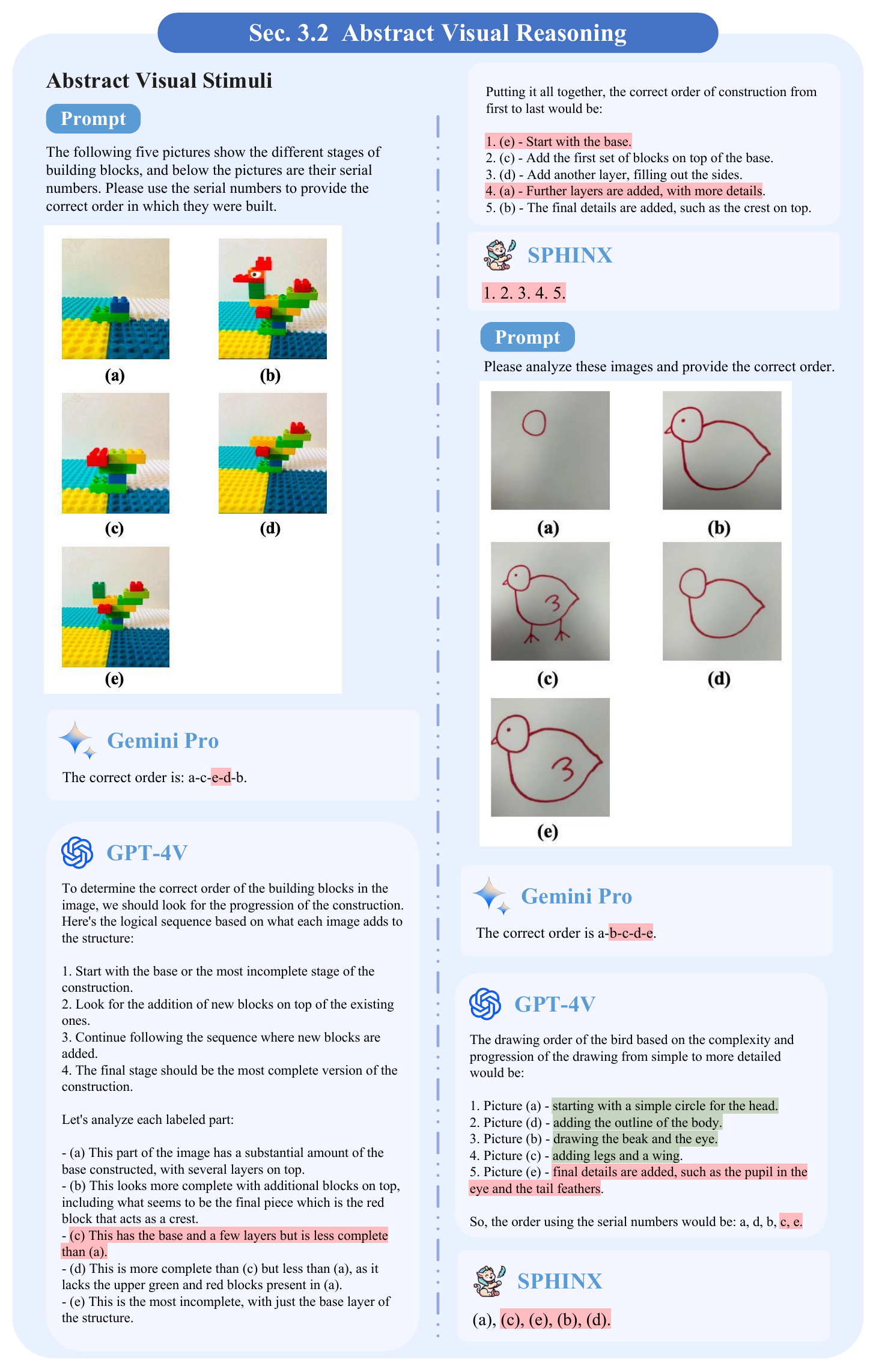}}
  \caption[Section~\ref{sec:04subsec:absvisual}: abstract visual stimuli.]{Results on abstract visual stimuli. \colorbox{greenhl!85!black}{Green} highlights the right answer. \colorbox{red!30}{Red} highlights the wrong answer.  Refer to Section \ref{sec:04subsec:absvisual} for detailed discussions.}
  \label{AbstractVisualStimuli-2}
\end{figure*}

\textbf{Abstract visual stimuli.} This task evaluates the visual abstract capabilities for object composition. As shown in Figures~\ref{AbstractVisualStimuli-1}-\ref{AbstractVisualStimuli-2}, GPT-4V exhibits the best abstract performance and also provides detailed descriptions for how the objects are composed of shapes. Instead, Gemini has partial abilities to recognize some simple abstract patterns, such as `boat' and `house', and Sphinx can not understand them.

\begin{figure*}[!ht]
  \centering 
  \makebox[\textwidth][c]{\includegraphics[width=1.1\textwidth]{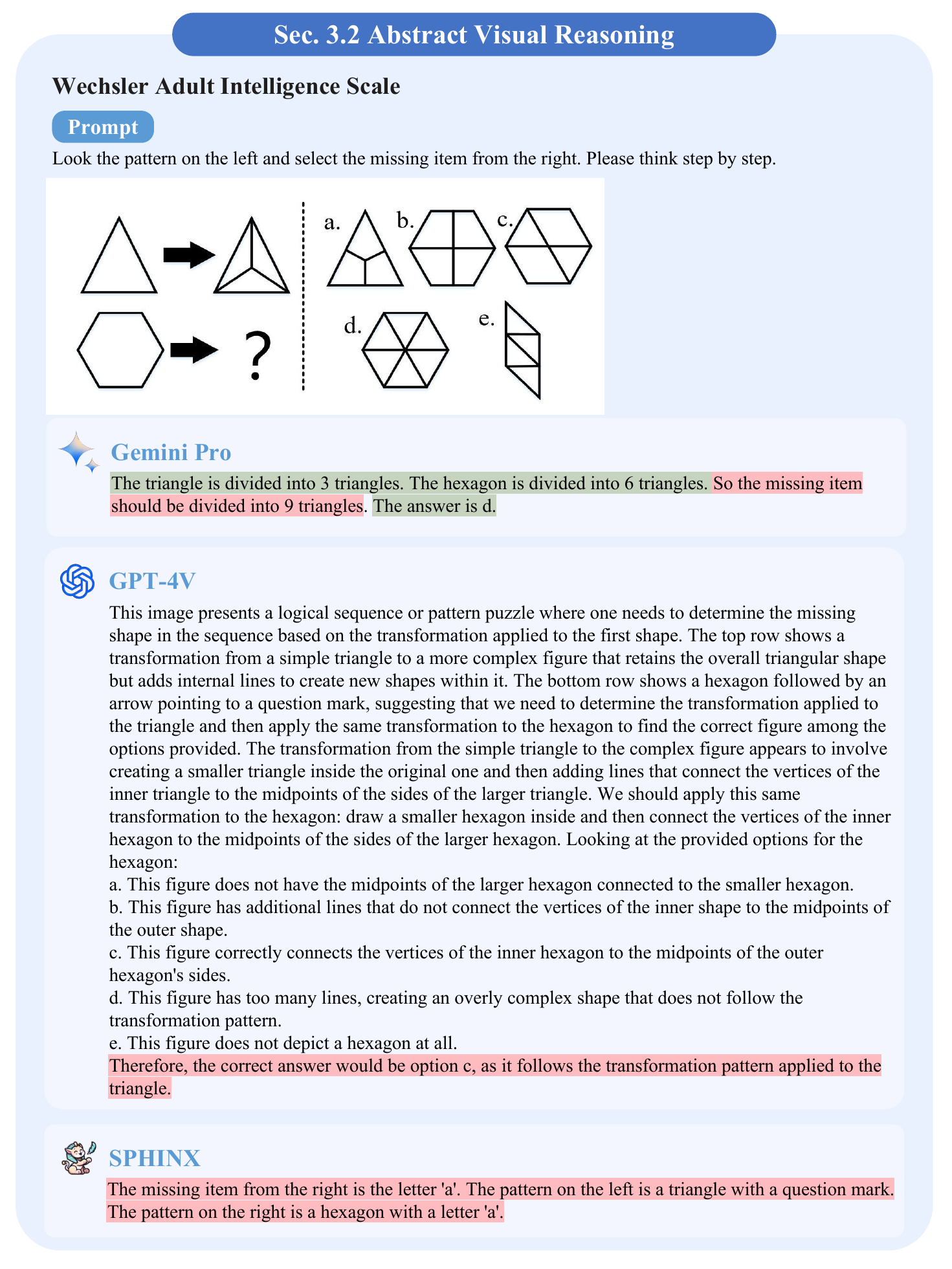}}
  \caption[Section~\ref{sec:04subsec:absvisual}: Wechsler Adult Intelligence Scale.]{Results on Wechsler Adult Intelligence Scale. \colorbox{greenhl!85!black}{Green} highlights the right answer. \colorbox{red!30}{Red} highlights the wrong answer. Refer to Section \ref{sec:04subsec:absvisual} for detailed discussions.}
  \label{Raven’WechslerAdultIntelligenceScale-1}
\end{figure*}

\begin{figure*}[!ht]
  \centering 
  \makebox[\textwidth][c]{\includegraphics[width=1.1\textwidth]{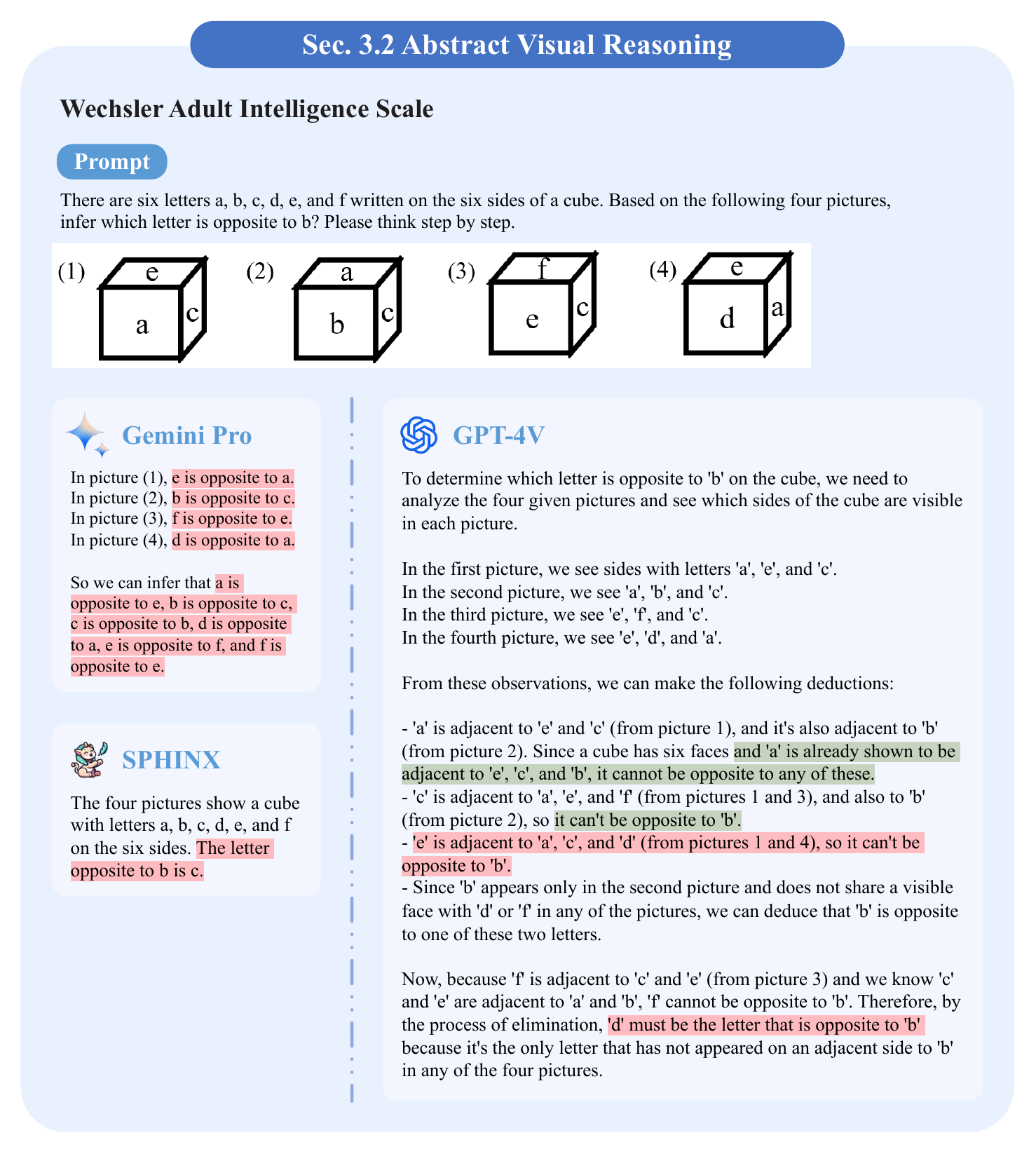}}
  \caption[Section~\ref{sec:04subsec:absvisual}: Wechsler Adult Intelligence Scale.]{Results on Wechsler Adult Intelligence Scale. \colorbox{greenhl!85!black}{Green} highlights the right answer. \colorbox{red!30}{Red} highlights the wrong answer. Refer to Section \ref{sec:04subsec:absvisual} for detailed discussions.}
  \label{Raven’WechslerAdultIntelligenceScale-2}
\end{figure*}

\begin{figure*}[!ht]
  \centering 
  \makebox[\textwidth][c]{\includegraphics[width=1.1\textwidth]{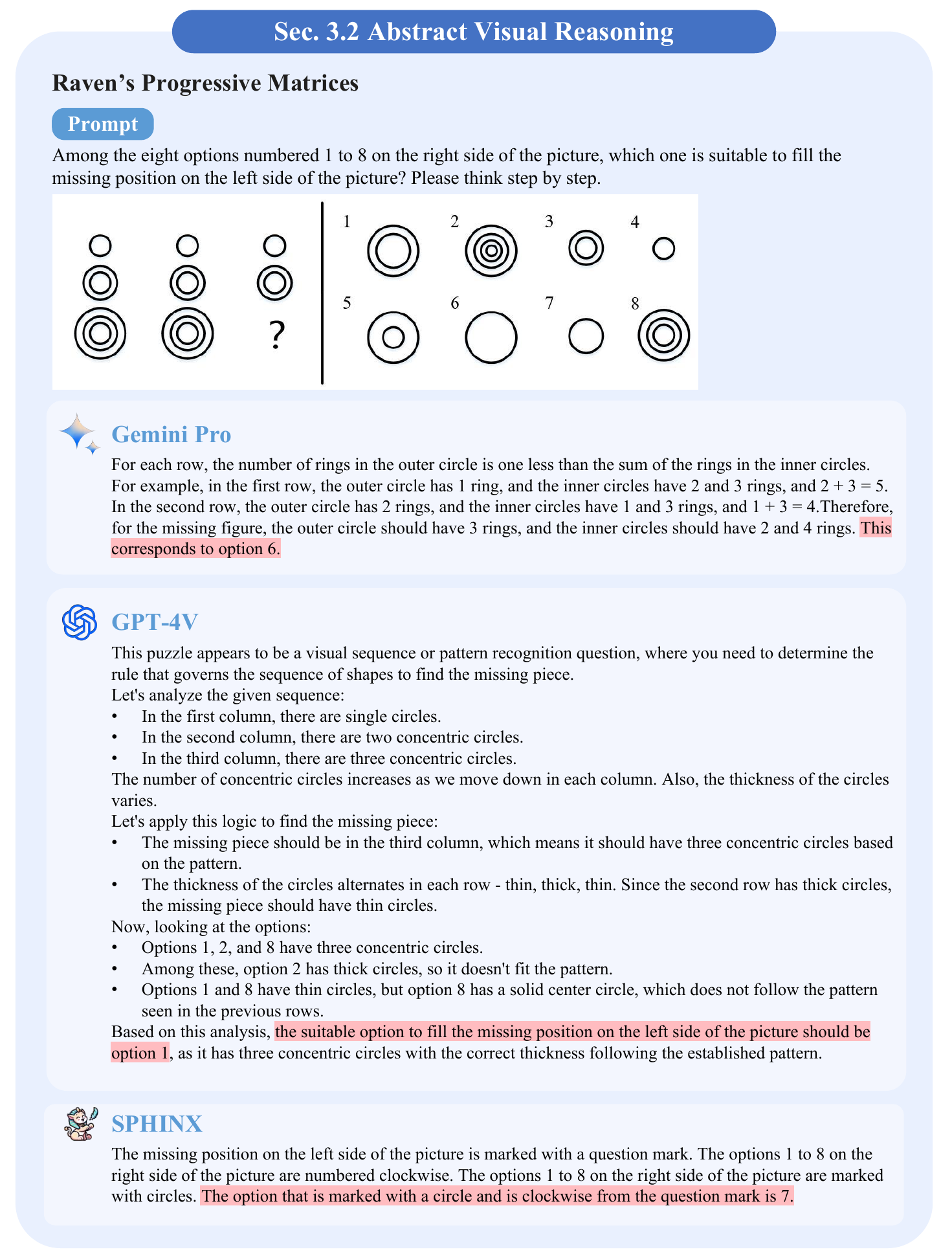}}
  \caption[Section~\ref{sec:04subsec:absvisual}: Raven’s Progressive Matrices.]{Results on Raven’s Progressive Matrices. \colorbox{red!30}{Red} highlights the wrong answer. Refer to Section \ref{sec:04subsec:absvisual} for detailed discussions.}
  \label{Raven’sProgressiveMatrices-1}
\end{figure*}

\begin{figure*}[!ht]
  \centering 
  \makebox[\textwidth][c]{\includegraphics[width=1.1\textwidth]{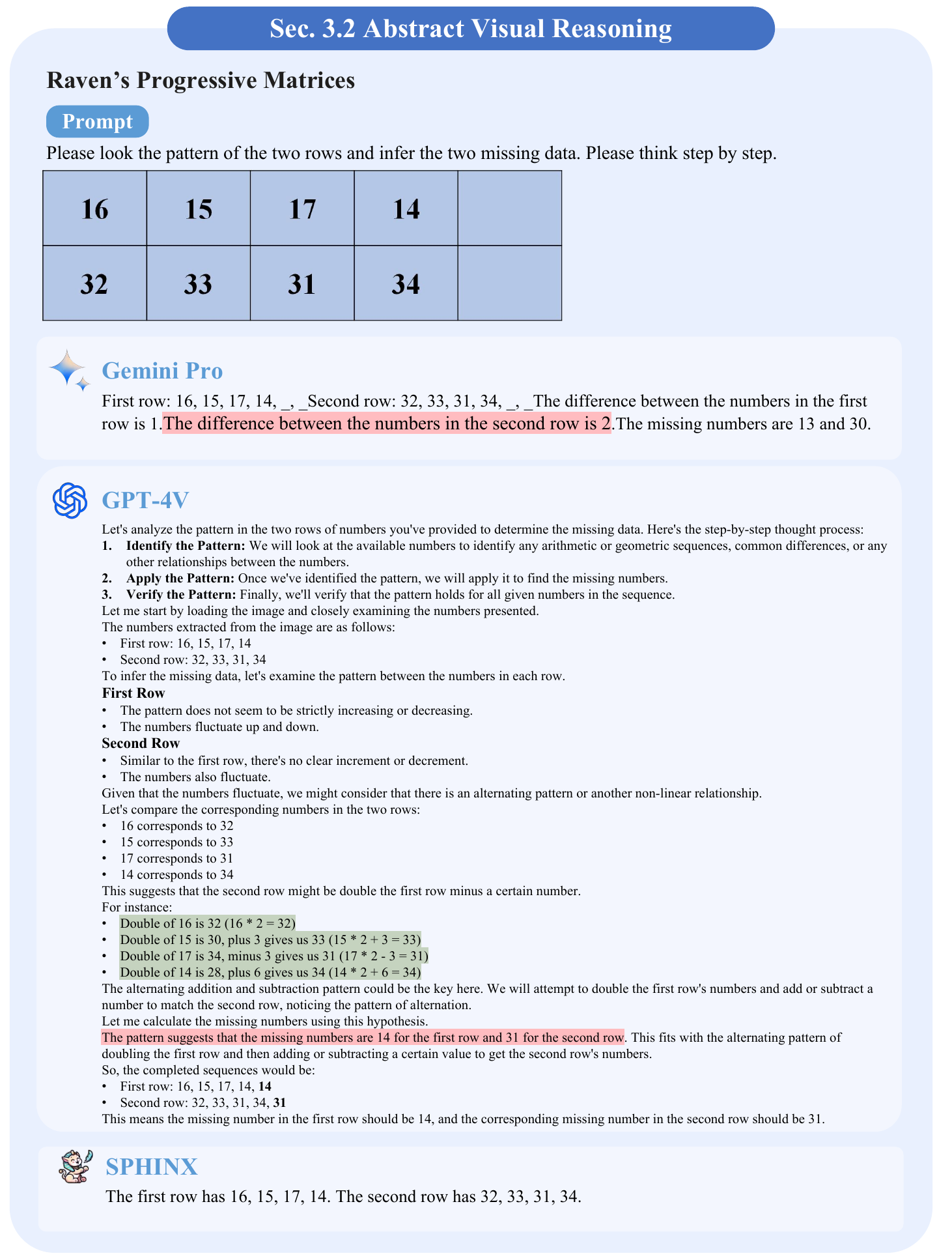}}
  \caption[Section~\ref{sec:04subsec:absvisual}: Raven’s Progressive Matrices.]{Results on Raven’s Progressive Matrices. \colorbox{greenhl!85!black}{Green} highlights the right answer. \colorbox{red!30}{Red} highlights the wrong answer. Refer to Section \ref{sec:04subsec:absvisual} for detailed discussions.}
  \label{Raven’sProgressiveMatrices-2}
\end{figure*}

\textbf{Raven’s Progressive Matrices and Wechsler Adult Intelligence Scale.} These two tasks are more challenging, since they require recognizing the high-level relations of different components, and predicting the next element in the matrices or sequences. As respectively shown in Figures~\ref{Raven’WechslerAdultIntelligenceScale-1}-\ref{Raven’WechslerAdultIntelligenceScale-2} and~\ref{Raven’sProgressiveMatrices-1}-\ref{Raven’sProgressiveMatrices-2}, nearly all of the MLLMs are incorrect in the final answer. GPT-4V showcases some detailed reasoning process, but still struggles with the final prediction and can be easily misled by an incorrect intermediate step. This experiment indicates that, although the advanced MLLMs can initially identify the independent element, they fail to parse their relationship for further inference.

\subsection{Science Problem-Solving}\label{sec:04subsec:science}

\begin{figure*}[!ht]
  \centering 
  \makebox[\textwidth][c]{\includegraphics[width=0.9\textwidth]{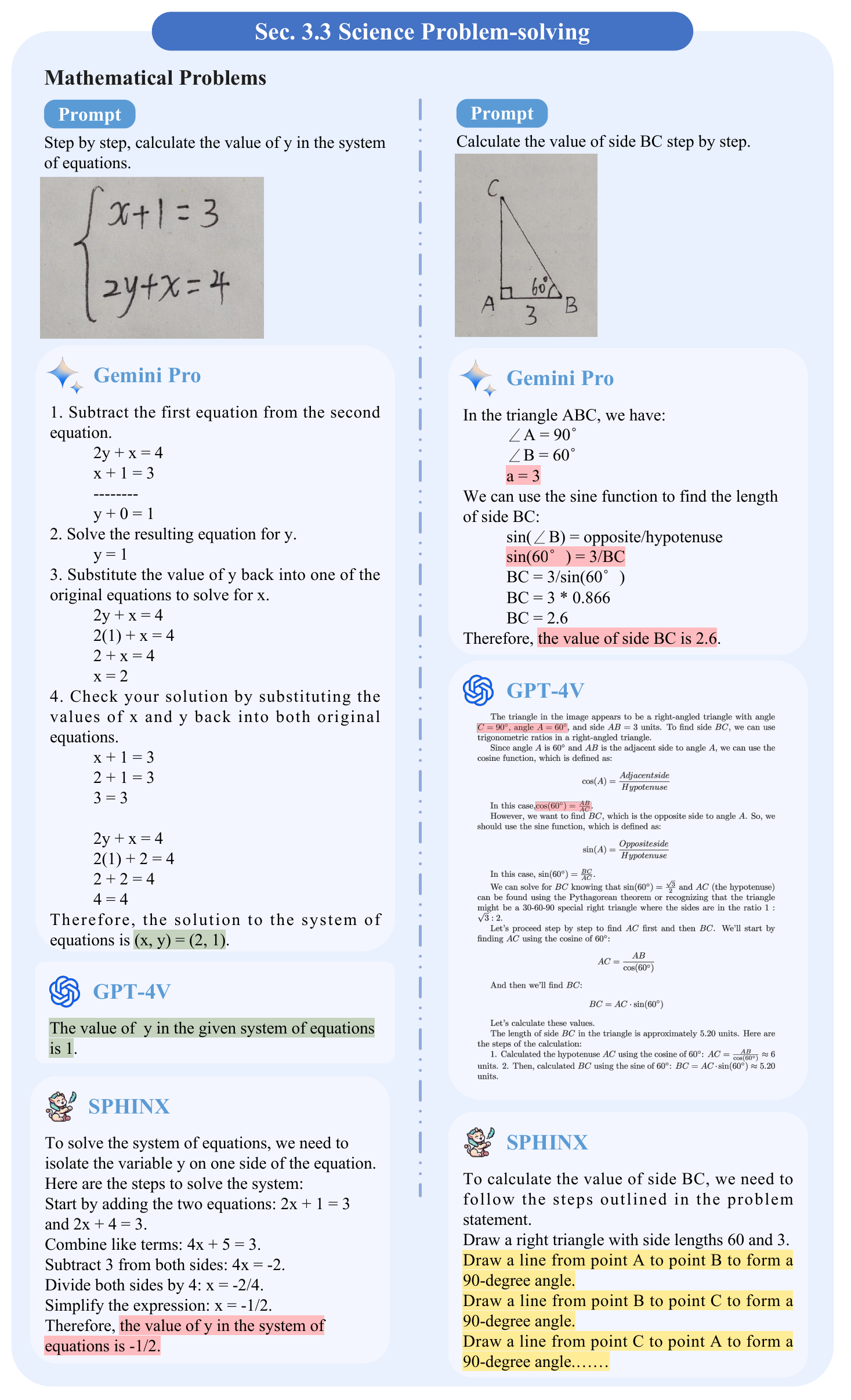}}
  \caption[Section~\ref{sec:04subsec:science}: mathematical problem.]{Results on mathematical problem-solving. \colorbox{greenhl!85!black}{Green} highlights the right answer. \colorbox{red!30}{Red} highlights the wrong answer.  \colorbox{yellow!70!yellowhl}{Yellow} highlights the incompetence in performing the task. Refer to Section \ref{sec:04subsec:science} for detailed discussions.}
  \label{math-1}
\end{figure*}

\begin{figure*}[!ht]
  \centering 
  \makebox[\textwidth][c]{\includegraphics[width=1.1\textwidth]{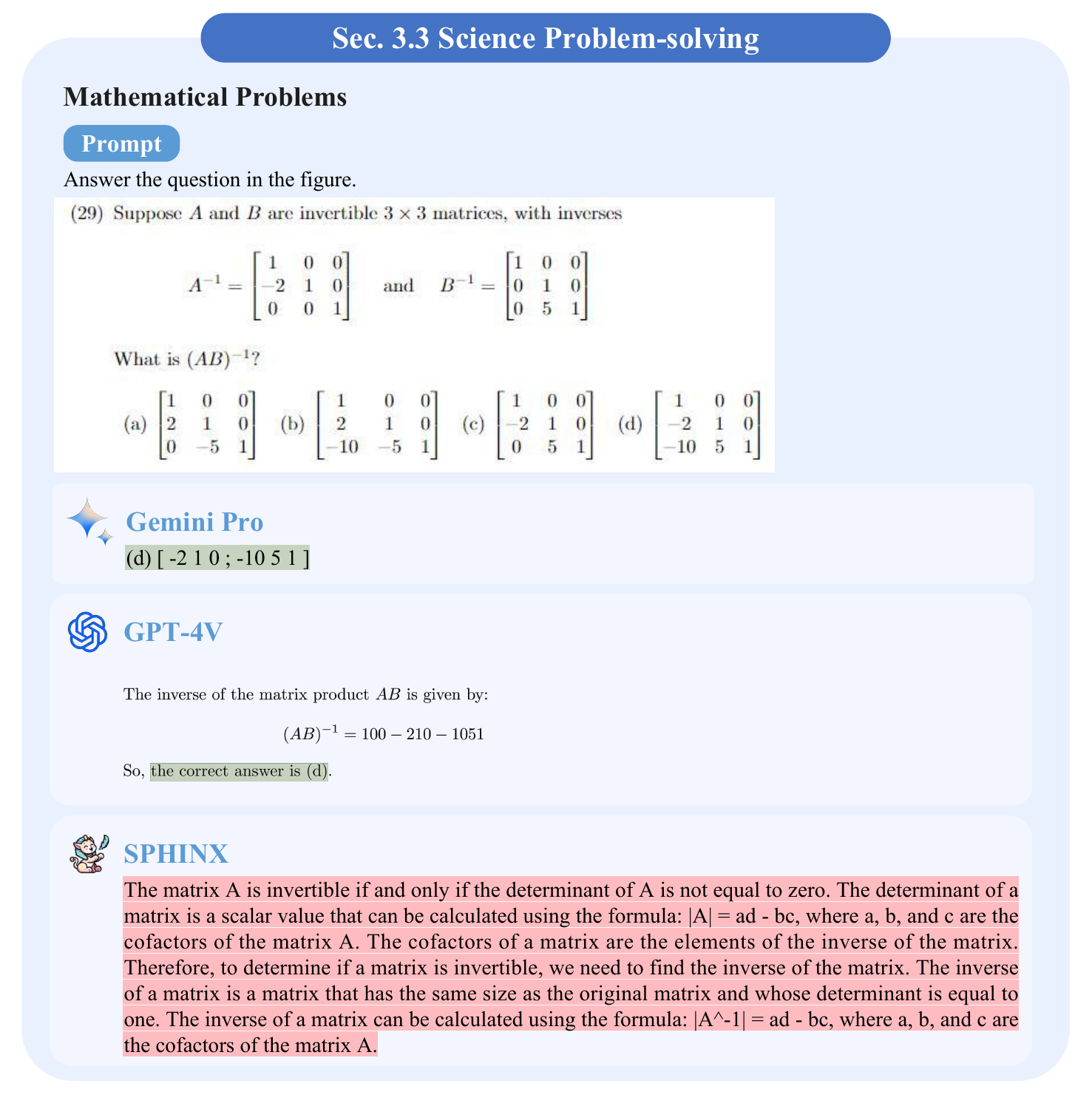}}
  \caption[Section~\ref{sec:04subsec:science}: mathematical problem.]{Results on mathematical problem-solving. \colorbox{greenhl!85!black}{Green} highlights the right answer. \colorbox{red!30}{Red} highlights the wrong answer. Refer to Section \ref{sec:04subsec:science} for detailed discussions.}
  \label{math-2}
\end{figure*}

\begin{figure*}[!ht]
  \centering 
  \makebox[\textwidth][c]{\includegraphics[width=0.95\textwidth]{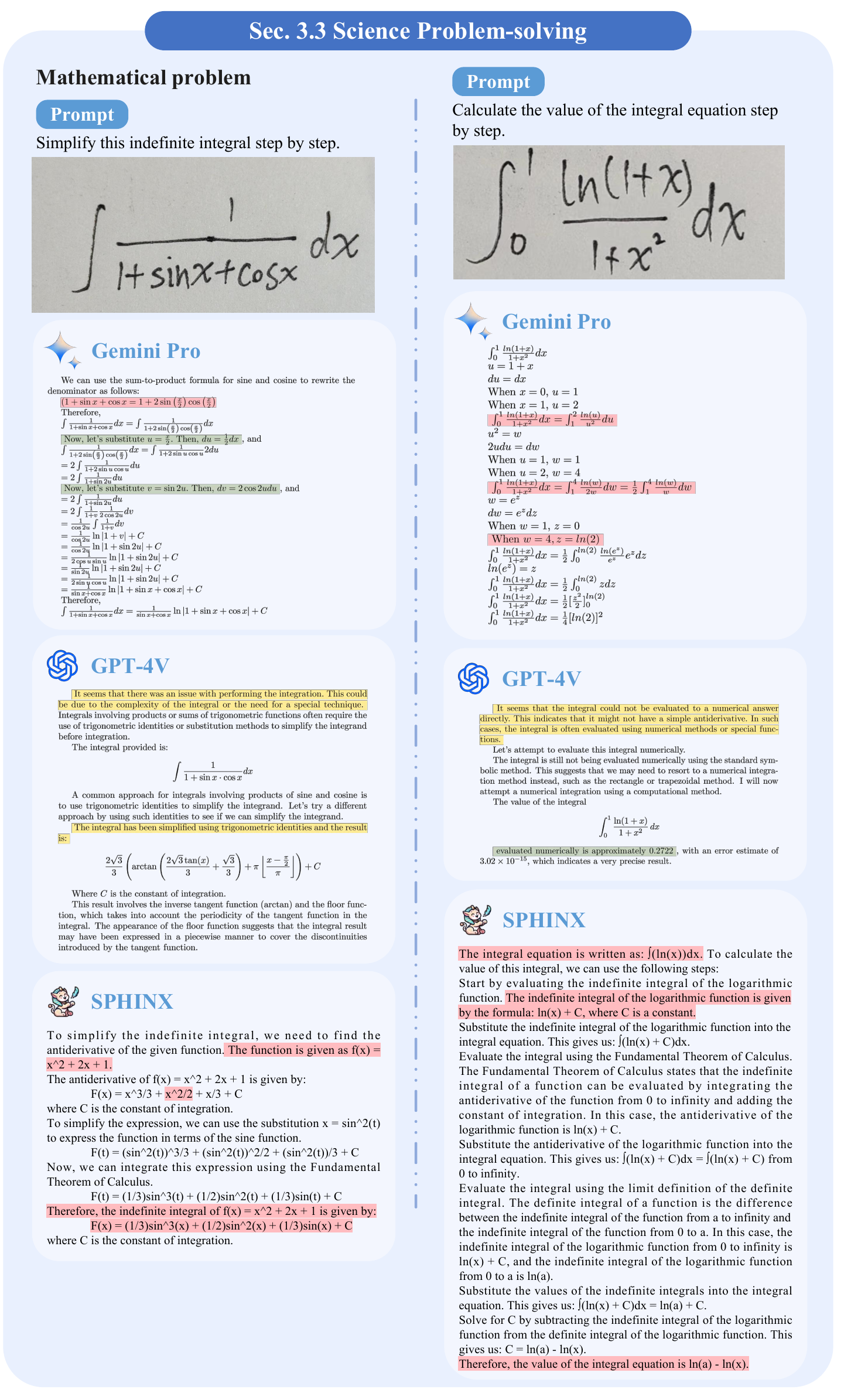}}
  \caption[Section~\ref{sec:04subsec:science}: mathematical problem.]{Results on mathematical problem-solving. \colorbox{greenhl!85!black}{Green} highlights the right answer. \colorbox{red!30}{Red} highlights the wrong answer. \colorbox{yellow!70!yellowhl}{Yellow} highlights the incompetence in performing the task. Refer to Section \ref{sec:04subsec:science} for detailed discussions.}
  \label{math-3}
\end{figure*}

\begin{figure*}[!ht]
  \centering 
  \makebox[\textwidth][c]{\includegraphics[width=1.2\textwidth]{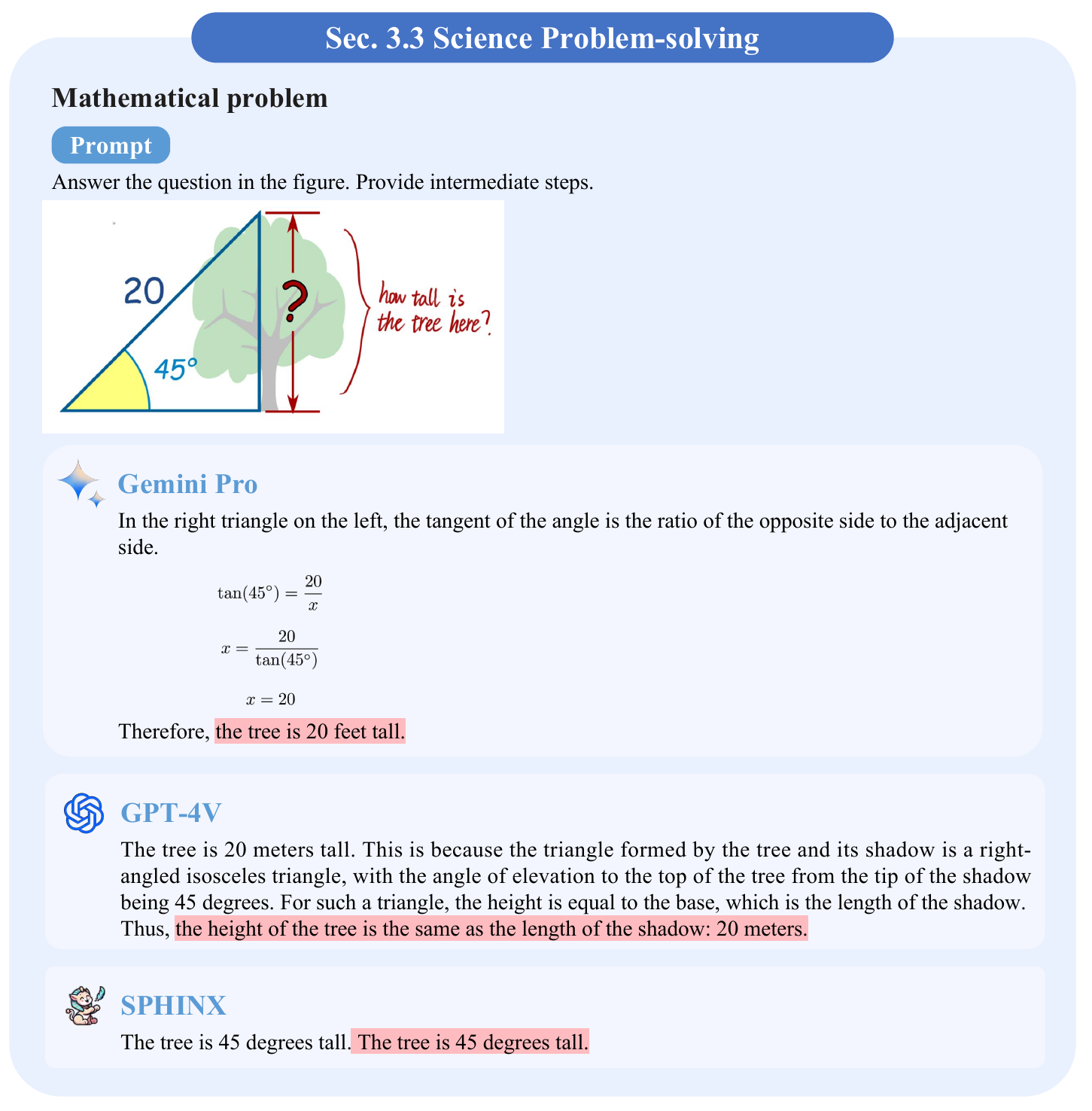}}
  \caption[Section~\ref{sec:04subsec:science}: mathematical problem.]{Results on mathematical problem-solving. \colorbox{red!30}{Red} highlights the wrong answer. Refer to Section \ref{sec:04subsec:science} for detailed discussions.}
  \label{math-4}
\end{figure*}

\begin{figure*}[!ht]
  \centering 
  \makebox[\textwidth][c]{\includegraphics[width=1.2\textwidth]{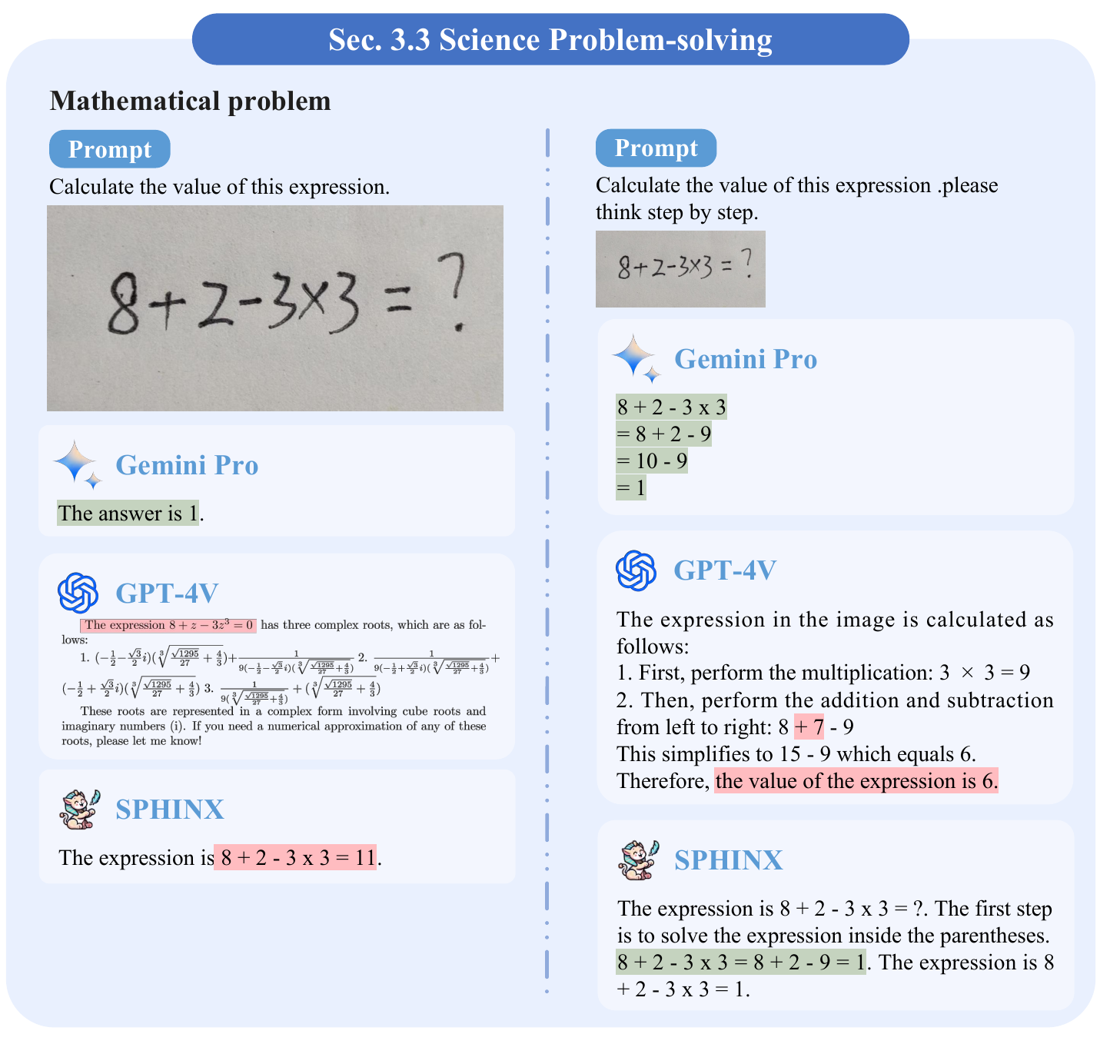}}
  \caption[Section~\ref{sec:04subsec:science}: mathematical problem.]{Results on mathematical problem-solving. \colorbox{greenhl!85!black}{Green} highlights the right answer. \colorbox{red!30}{Red} highlights the wrong answer. Refer to Section \ref{sec:04subsec:science} for detailed discussions.}
  \label{math-5}
\end{figure*}

\begin{figure*}[!ht]
  \centering 
  \makebox[\textwidth][c]{\includegraphics[width=1.2\textwidth]{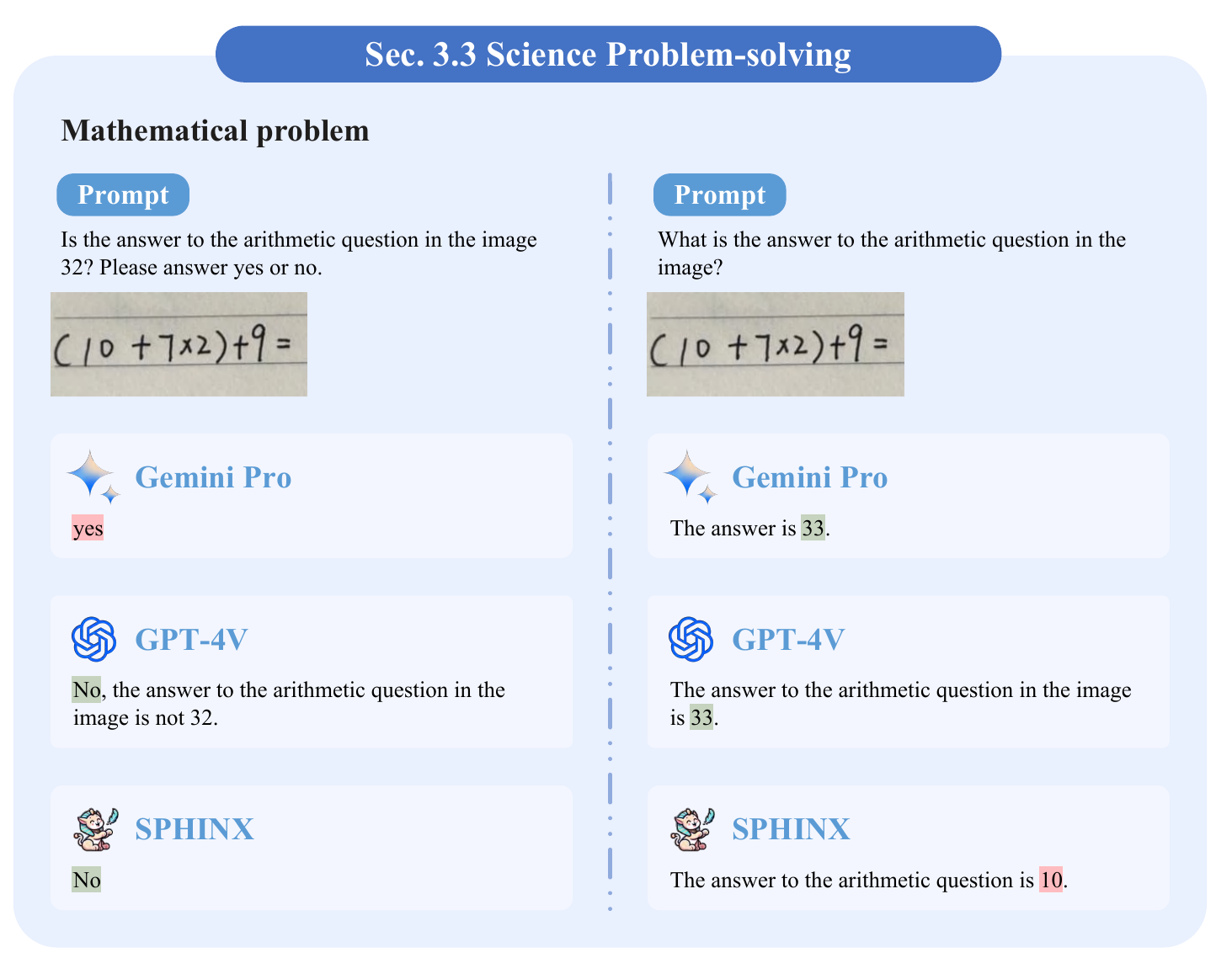}}
  \caption[Section~\ref{sec:04subsec:science}: mathematical problem.]{Results on mathematical problem-solving. \colorbox{greenhl!85!black}{Green} highlights the right answer. \colorbox{red!30}{Red} highlights the wrong answer. Refer to Section \ref{sec:04subsec:science} for detailed discussions.}
  \label{math-6}
\end{figure*}

\textbf{Mathematical problems.}
Different from common visual question answering, the solving of mathematical problems involves both OCR capabilities from visual input and quantitative processing accuracy in the subsequent reasoning steps. In Figures~\ref{math-1}-\ref{math-6}, we show some mathematical problems concerning a wide range of tasks, including arithmetic, algebra, geometry, and integral calculus. The samples indicate that Gemini and GPT-4V can well tackle simple arithmetic and algebra problems. For more difficult trigonometry and integral calculus, they also exhibit favorable reasoning performance with the help of external tools. However, they are not very expert at recognizing the specific visual content in the images, such as numbers, symbols, and their correspondence. In addition, we observe that, with CoT techniques, i.e., ``Please think step by step'', the previous wrong answer of Sphinx can rectified, demonstrating the importance of CoT prompting for open-sourced MLLMs.

\begin{figure*}[!ht]
  \centering 
  \makebox[\textwidth][c]{\includegraphics[width=1.2\textwidth]{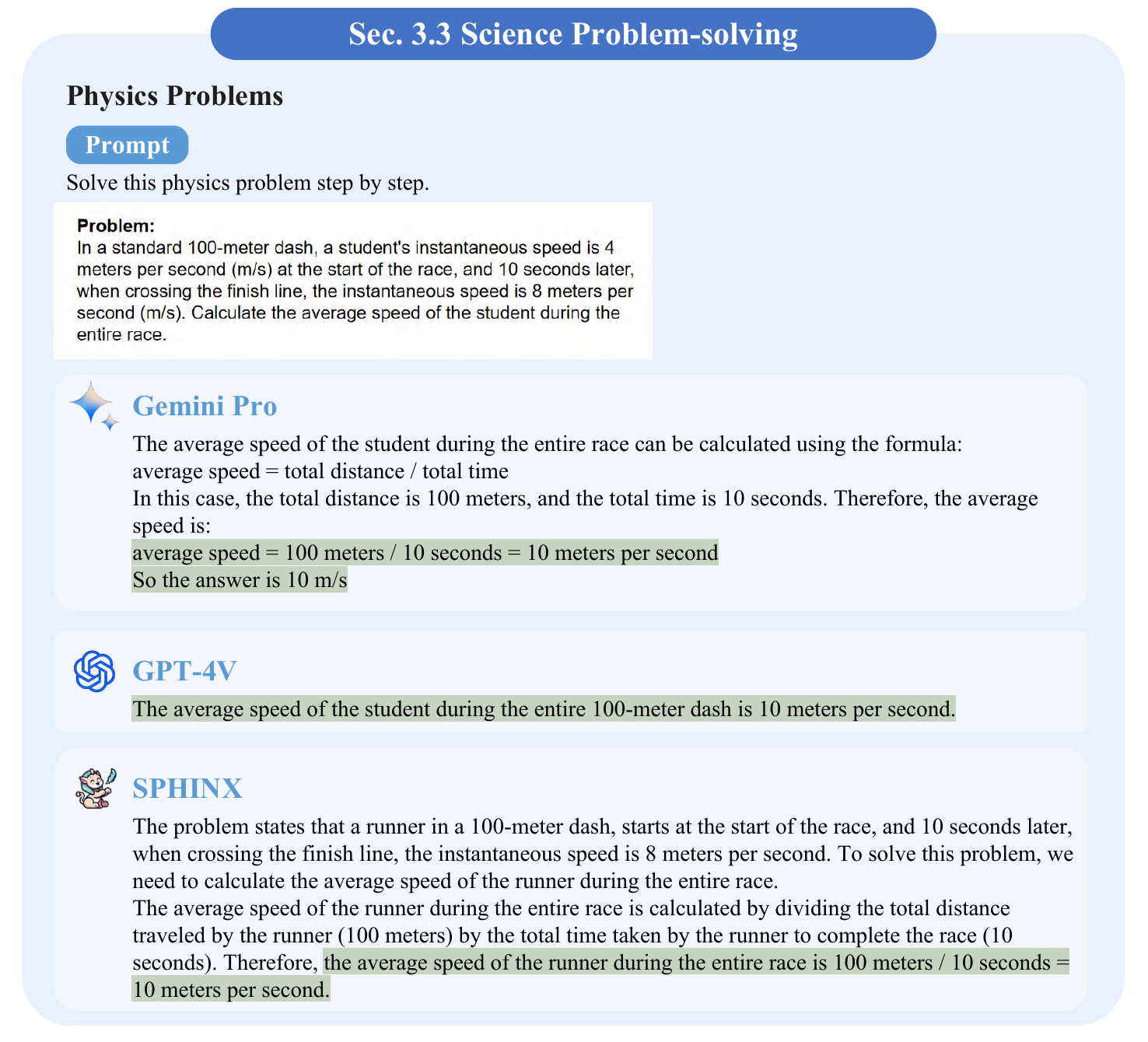}}
  \caption[Section~\ref{sec:04subsec:science}: physics problem.]{Results on physics problem-solving. \colorbox{greenhl!85!black}{Green} highlights the right answer. Refer to Section \ref{sec:04subsec:science} for detailed discussions.}
  \label{physics-1}
\end{figure*}

\begin{figure*}[!ht]
  \centering 
  \makebox[\textwidth][c]{\includegraphics[width=1.0\textwidth]{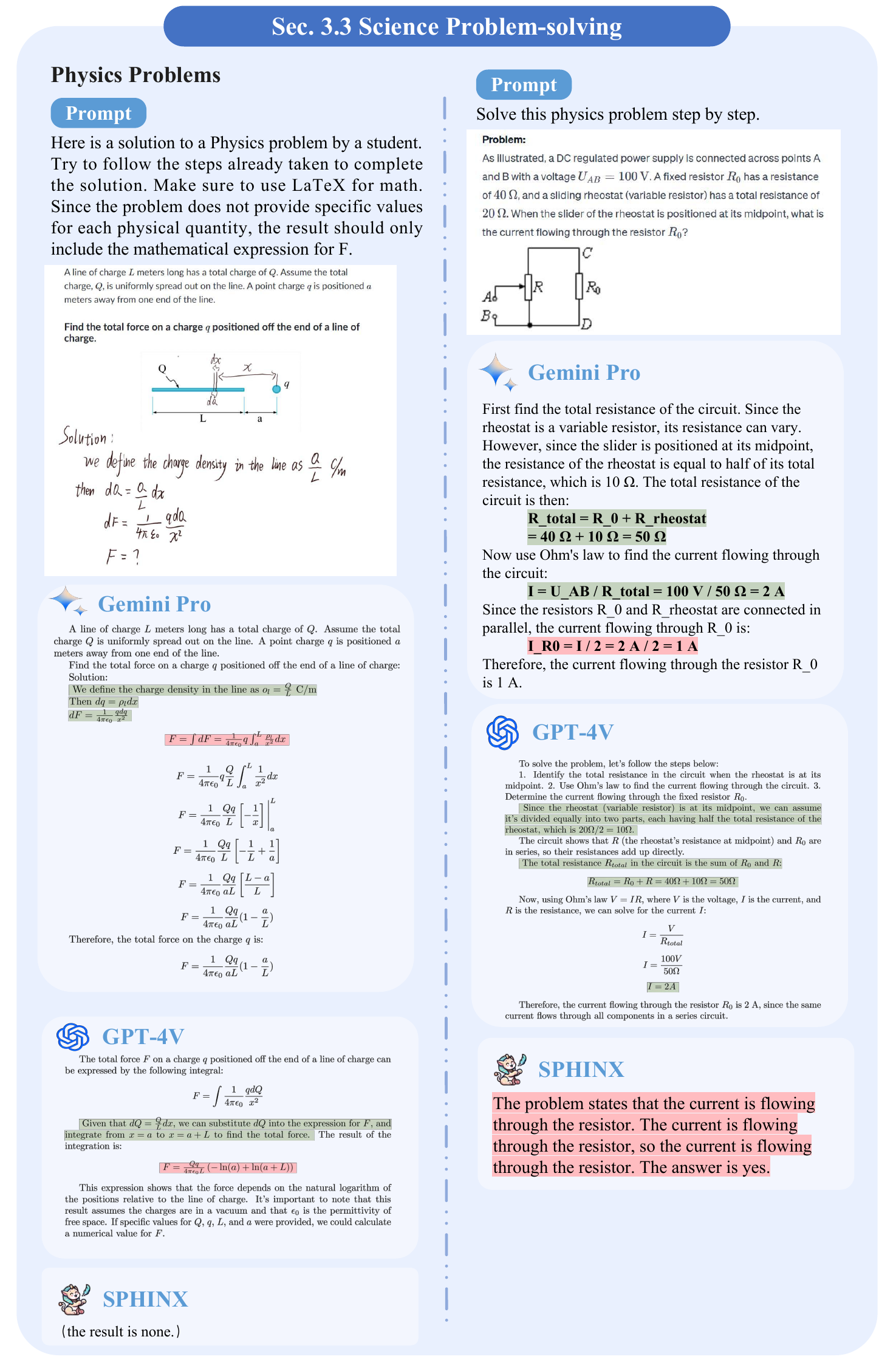}}
  \caption[Section~\ref{sec:04subsec:science}:  physics problem.]{Results on physics problem-solving. \colorbox{greenhl!85!black}{Green} highlights the right answer. \colorbox{red!30}{Red} highlights the wrong answer. Refer to Section \ref{sec:04subsec:science} for detailed discussions.}
  \label{physics-2}
\end{figure*}

\begin{figure*}[!ht]
  \centering 
  \makebox[\textwidth][c]{\includegraphics[width=0.95\textwidth]{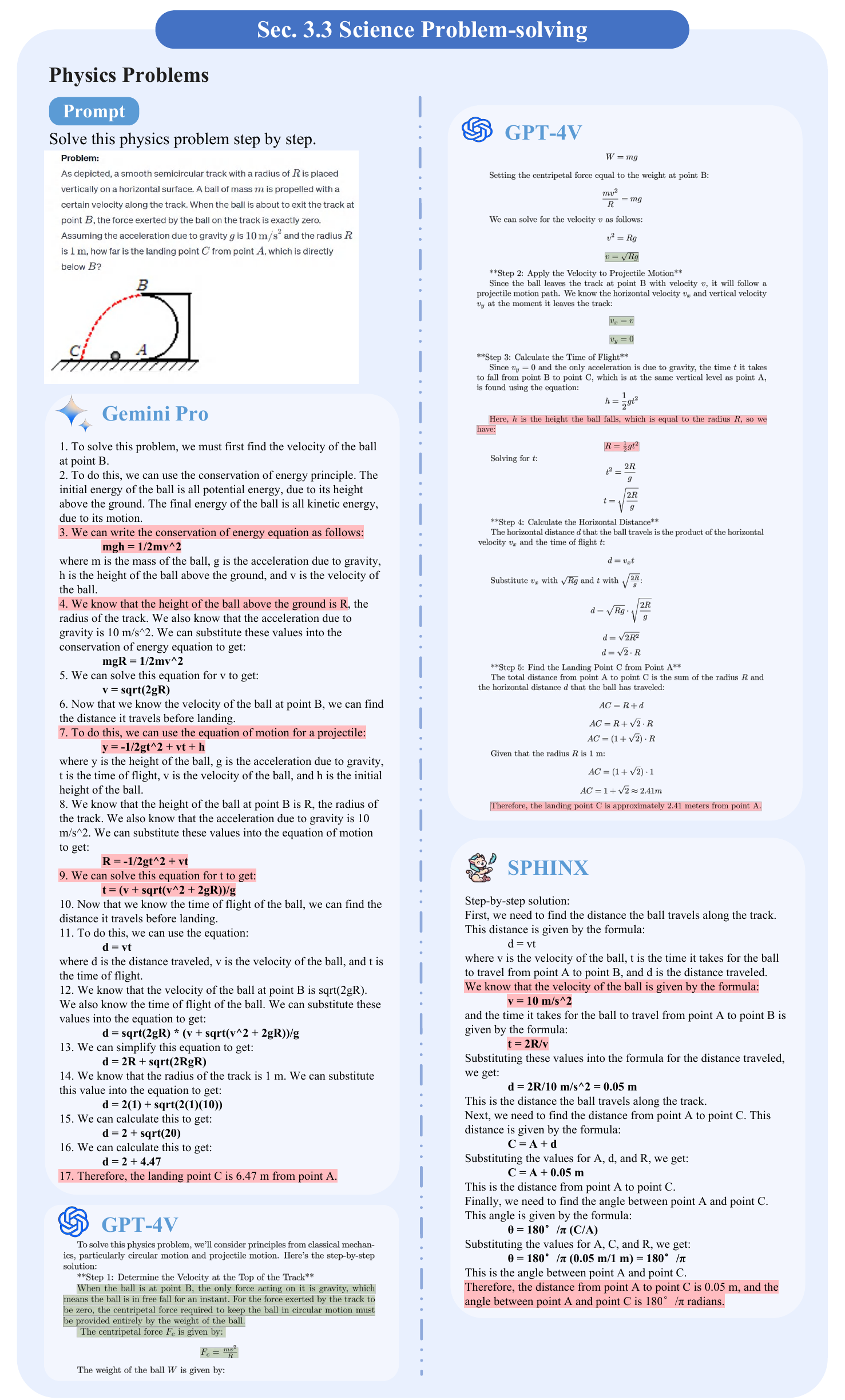}}
  \caption[Section~\ref{sec:04subsec:science}:  physics problem.]{Results on physics problem-solving. \colorbox{greenhl!85!black}{Green} highlights the right answer. \colorbox{red!30}{Red} highlights the wrong answer. Refer to Section \ref{sec:04subsec:science} for detailed discussions.}
  \label{physics-3}
\end{figure*}

\textbf{Physics problems.} Such problems further require MLLMs' comprehension of the specialized vocabulary and concepts in Physics. In Figures~\ref{physics-1}-\ref{physics-3}, we show the problem-solving results of three MLLMs concerning dynamics, kinematics, and circuitry. As shown, Gemini and GPT-4V show well-performed reasoning of Physics problems and well leverage the pre-trained specialized knowledge as reference. However, their performance can be limited by mathematical calculation, e.g., the range of integration, and the accuracy of physical equations, e.g., energy conservation equation. Due to the training data scarcity of Physics problems, the open-source Sphinx clearly lacks proficiency in solving such scientific problems with figures.

\subsection{Emotion Understanding}\label{sec:04subsec:emotion}

\begin{figure*}[!ht]
  \centering 
  \makebox[\textwidth][c]{\includegraphics[width=1.0\textwidth]{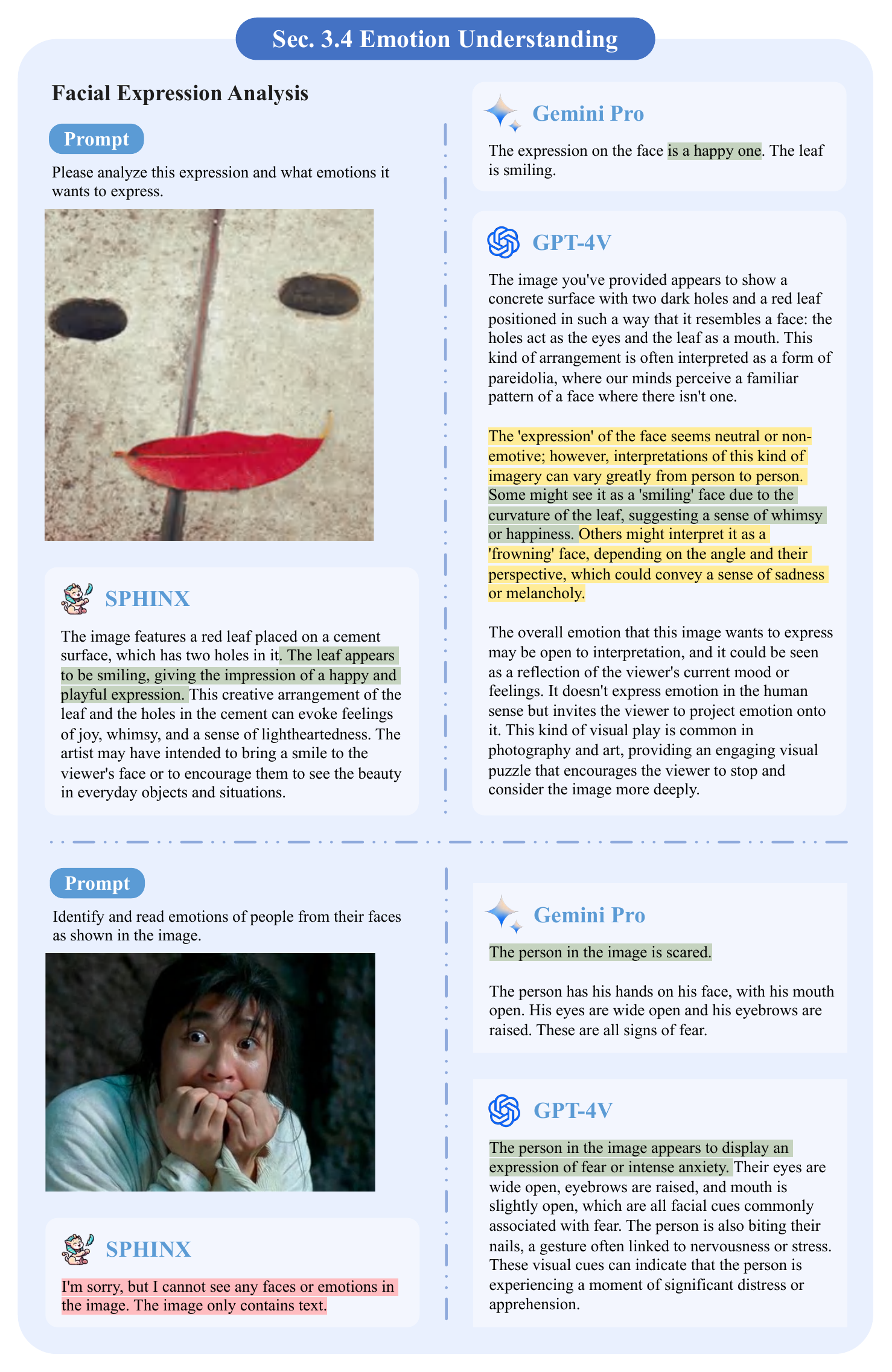}}
  \caption[Section~\ref{sec:04subsec:emotion}: facial expression analysis.]{Results on facial expression analysis. \colorbox{greenhl!85!black}{Green} highlights the right answer. \colorbox{red!30}{Red} highlights the wrong answer. \colorbox{yellow!70!yellowhl}{Yellow} highlights the incompetence in performing the task. Refer to Section \ref{sec:04subsec:emotion} for detailed discussions.}
  \label{face-1}
\end{figure*}

\begin{figure*}[!ht]
  \centering 
  \makebox[\textwidth][c]{\includegraphics[width=1.2\textwidth]{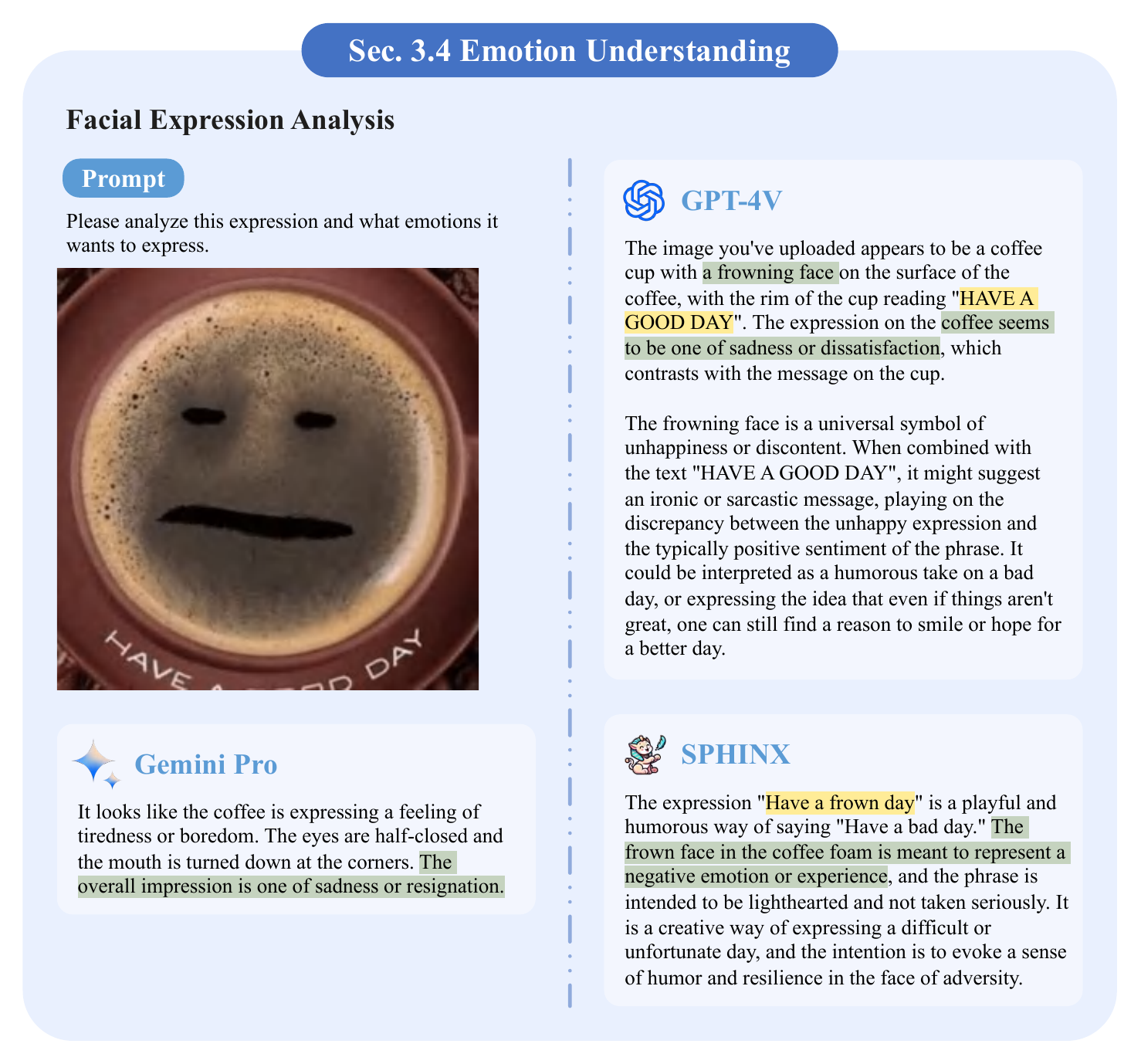}}
  \caption[Section~\ref{sec:04subsec:emotion}:  facial expression analysis.]{Results on facial expression analysis. \colorbox{greenhl!85!black}{Green} highlights the right answer. \colorbox{yellow!70!yellowhl}{Yellow} highlights the incompetence in performing the task. Refer to Section \ref{sec:04subsec:emotion} for detailed discussions.}
  \label{face-2}
\end{figure*}

\textbf{Facial expression analysis.}
In Figures~\ref{face-1}-\ref{face-2}, we evaluate the facial expression understanding capacity of different models. As shown, all of the three MLLMs exhibit good performance in this task. Therein, GPT-4V provides more dialectical thinking with rigorous analysis, e.g., the two possibilities of the first expression, while Gemini can directly respond with the accurate answer in a concise message. Also, GPT-4V and Sphinx both capture the truncated textual content on the plate of the third image, and incorporate this information into the reasoning. This result demonstrates their comprehensive visual understanding abilities.

\begin{figure*}[!ht]
  \centering 
  \makebox[\textwidth][c]{\includegraphics[width=1.1\textwidth]{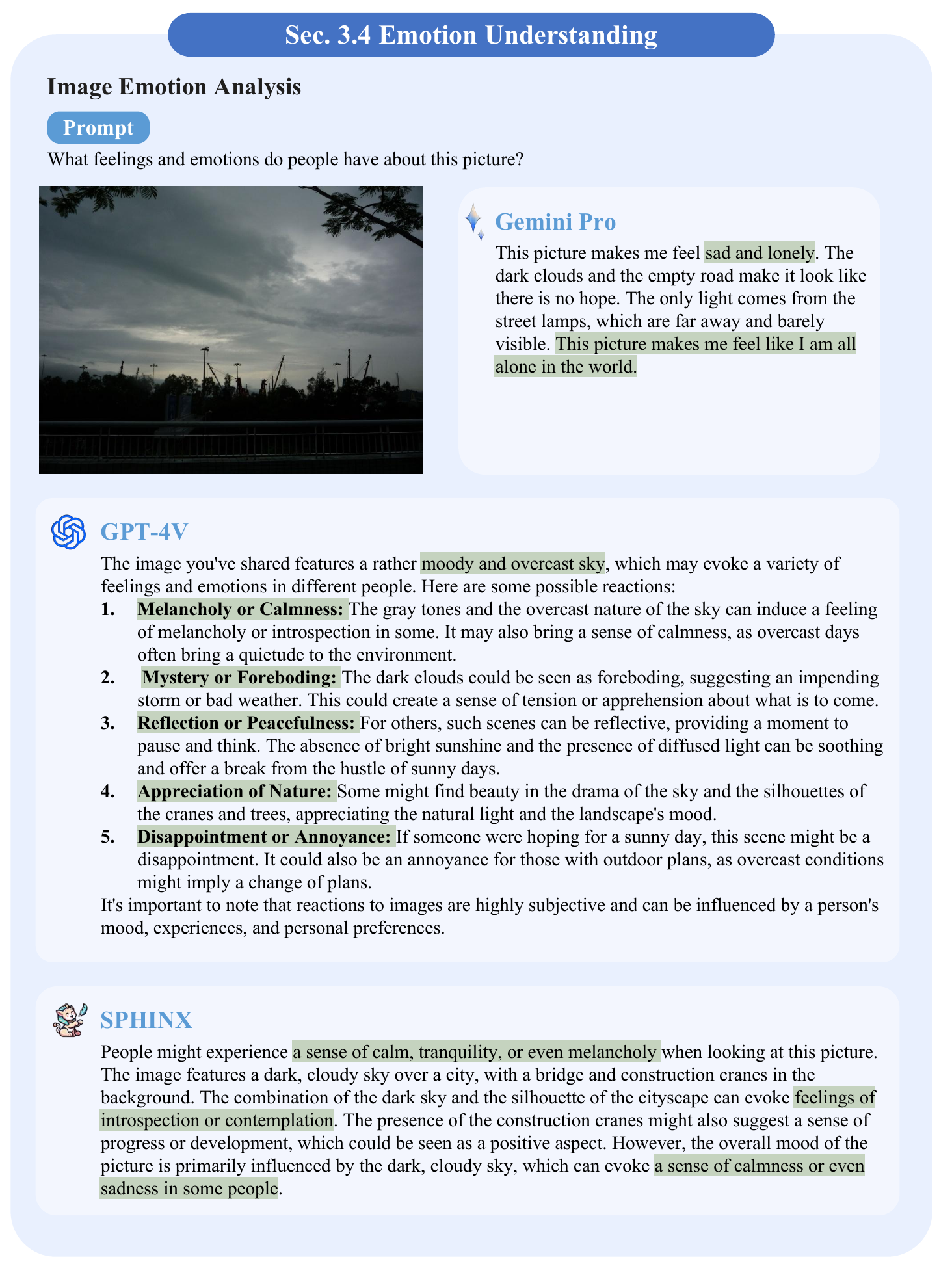}}
  \caption[Section~\ref{sec:04subsec:emotion}: image emotion analysis.]{Results on image emotion analysis. \colorbox{greenhl!85!black}{Green} highlights the right answer. Refer to Section \ref{sec:04subsec:emotion} for detailed discussions.}
  \label{imgemo-1}
\end{figure*}

\begin{figure*}[!ht]
  \centering 
  \makebox[\textwidth][c]{\includegraphics[width=1.0\textwidth]{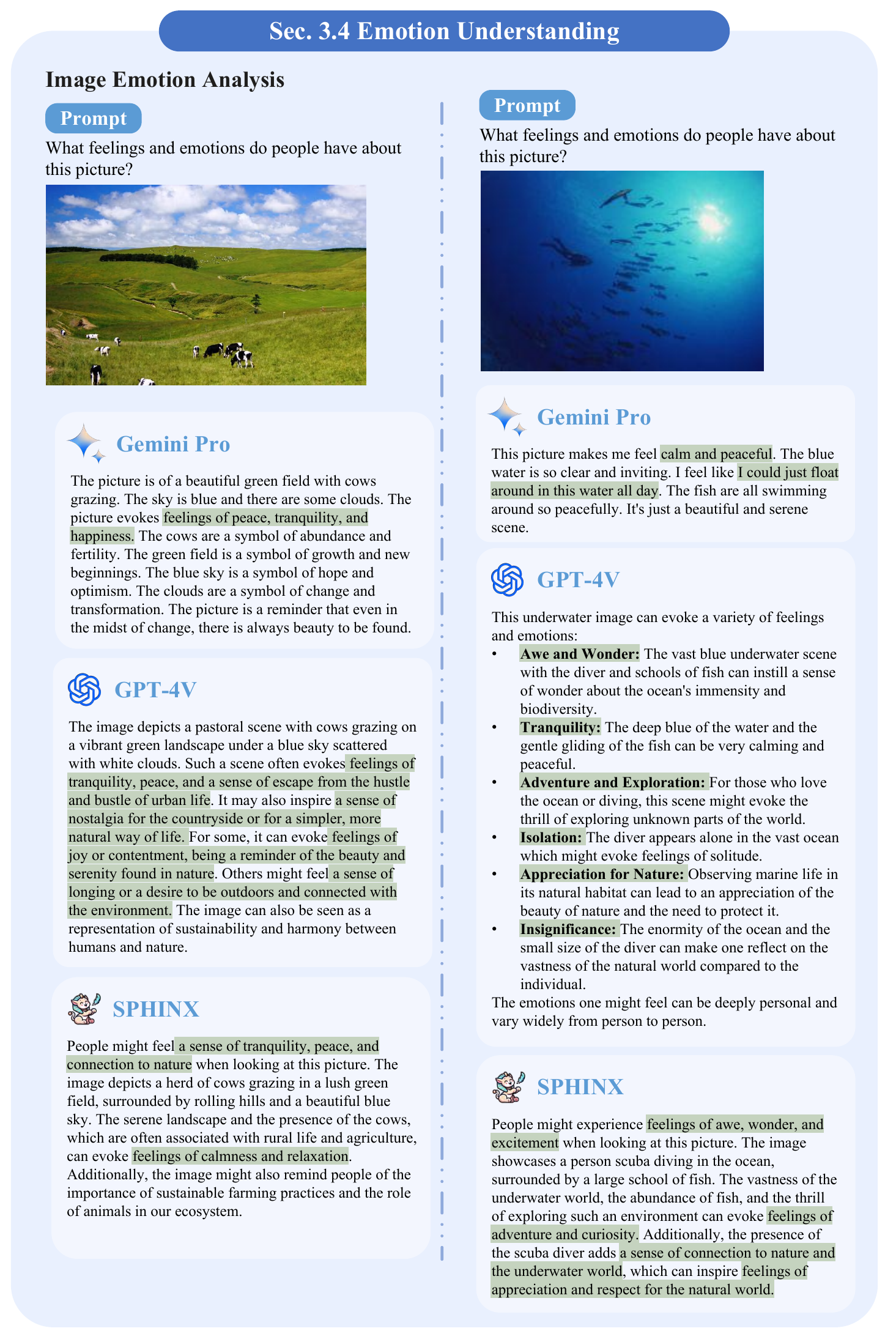}}
  \caption[Section~\ref{sec:04subsec:emotion}: image emotion analysis.]{Results on image emotion analysis. \colorbox{greenhl!85!black}{Green} highlights the right answer. Refer to Section \ref{sec:04subsec:emotion} for detailed discussions.}
  \label{imgemo-2}
\end{figure*}

\begin{figure*}[!ht]
  \centering 
  \makebox[\textwidth][c]{\includegraphics[width=1.1\textwidth]{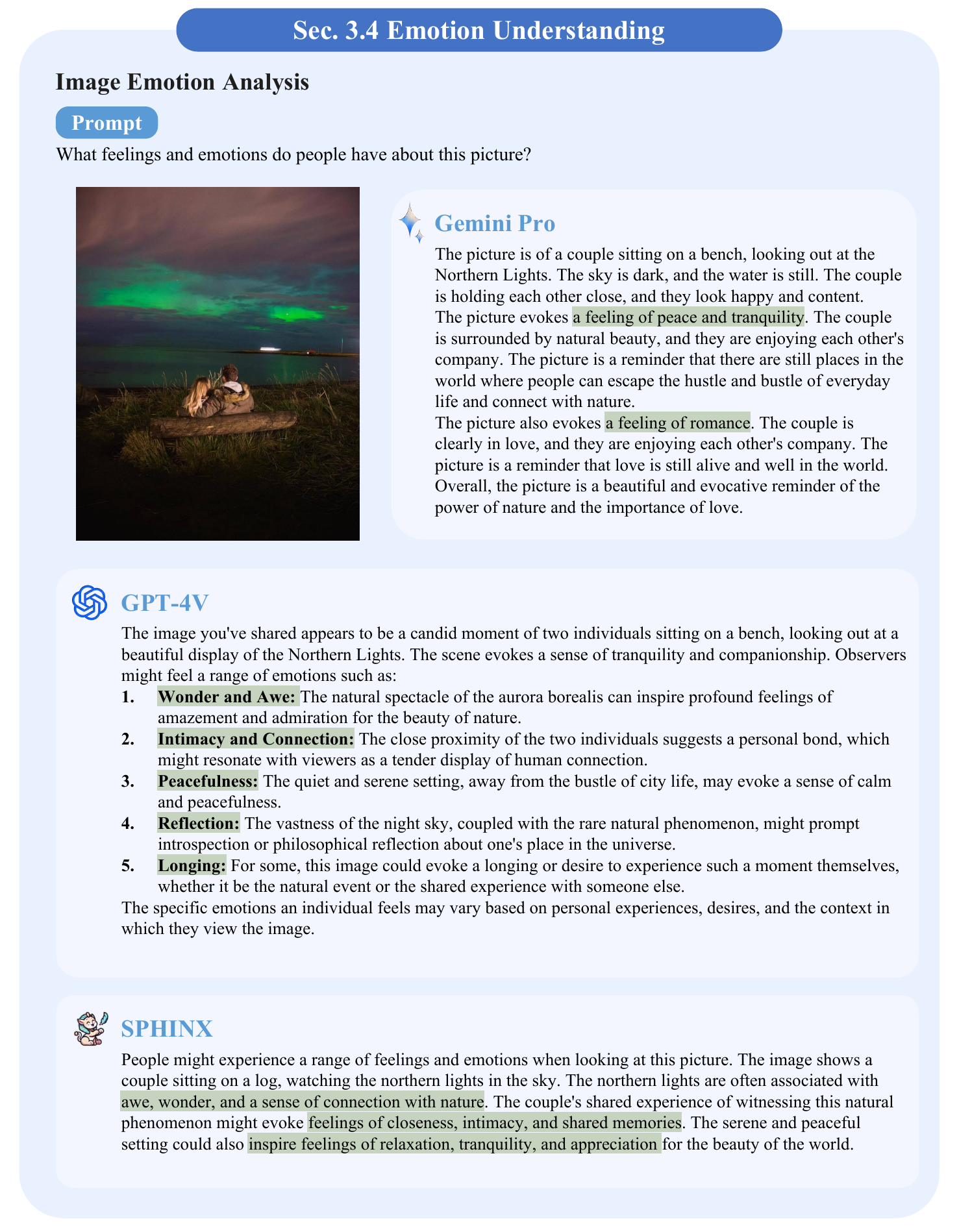}}
  \caption[Section~\ref{sec:04subsec:emotion}: image emotion analysis.]{Results on image emotion analysis. \colorbox{greenhl!85!black}{Green} highlights the right answer. Refer to Section \ref{sec:04subsec:emotion} for detailed discussions.}
  \label{imgemo-3}
\end{figure*}

\begin{figure*}[!ht]
  \centering 
  \makebox[\textwidth][c]{\includegraphics[width=1.0\textwidth]{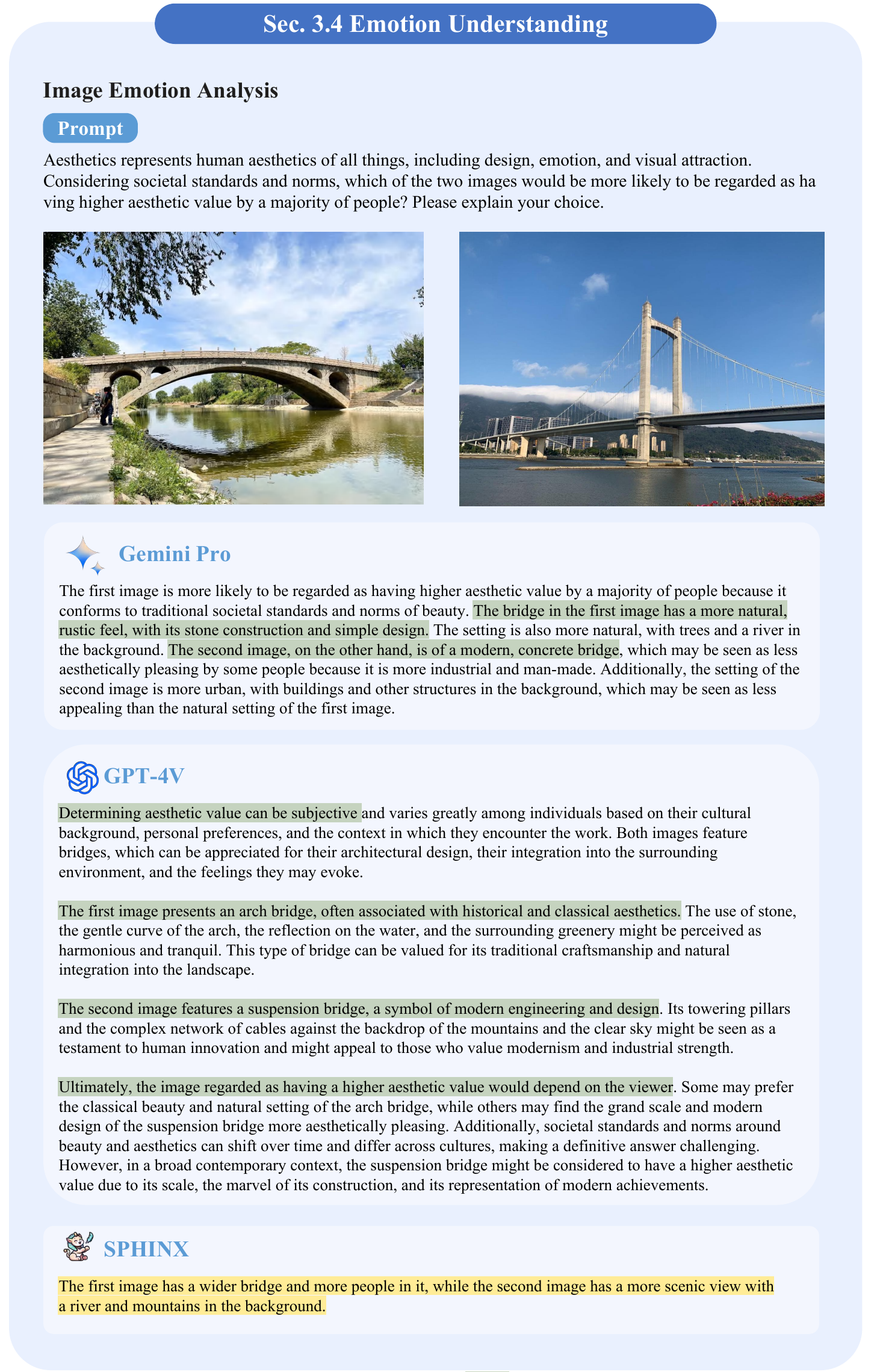}}
  \caption[Section~\ref{sec:04subsec:emotion}:  image emotion analysis.]{Results on image emotion analysis. \colorbox{greenhl!85!black}{Green} highlights the right answer. \colorbox{yellow!70!yellowhl}{Yellow} highlights the incompetence in performing the task. Refer to Section \ref{sec:04subsec:emotion} for detailed discussions.}
  \label{imgemo-4}
\end{figure*}

\begin{figure*}[!ht]
  \centering 
  \makebox[\textwidth][c]{\includegraphics[width=1.0\textwidth]{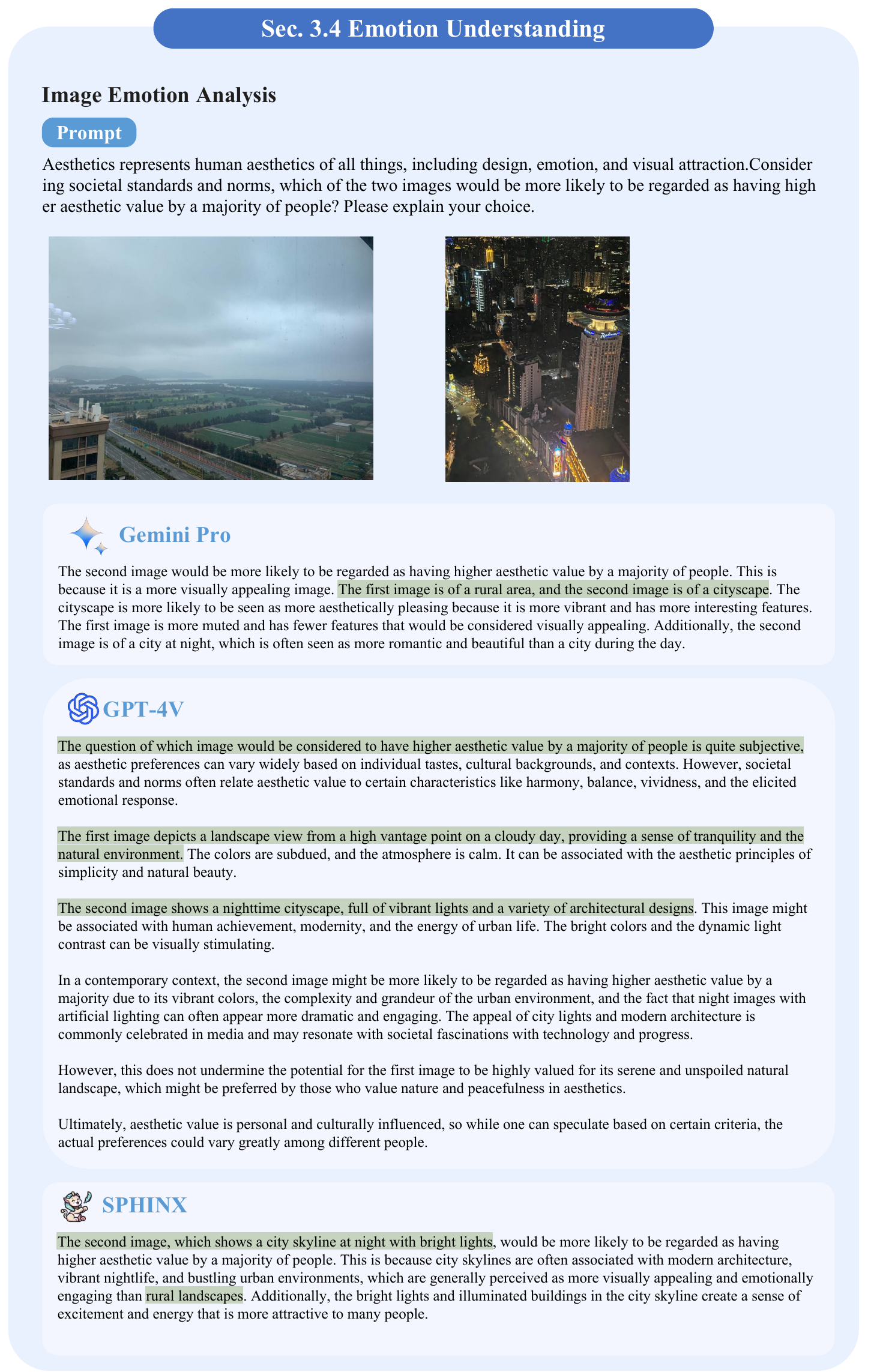}}
  \caption[Section~\ref{sec:04subsec:emotion}: image emotion analysis.]{Results on image emotion analysis. \colorbox{greenhl!85!black}{Green} highlights the right answer. Refer to Section \ref{sec:04subsec:emotion} for detailed discussions.}
  \label{imgemo-5}
\end{figure*}

\textbf{Image emotion analysis.}
This task is more challenging, since there is no explicit facial expression shown in the image. Instead, MLLMs are required to indicate the implicit emotion conveyed from the visual concepts. As shown in Figures~\ref{imgemo-1}-\ref{imgemo-5}, we select diverse samples of various natural scenes and 
manufactured buildings. All three models can well depict the view first, and provide possible emotion within it. Therein, GPT-4V is observed to be neutral and emphasizes that emotions are subjective, and meanwhile gives a more comprehensive analysis. In contrast, Gemini tends to directly output the emotion preference, which corresponds with mainstream perspectives. In addition, Sphinx can achieve comparable performance to the other two MLLMs, indicating its superior emotion parsing capability.

\begin{figure*}[!ht]
  \centering 
  \makebox[\textwidth][c]{\includegraphics[width=1.2\textwidth]{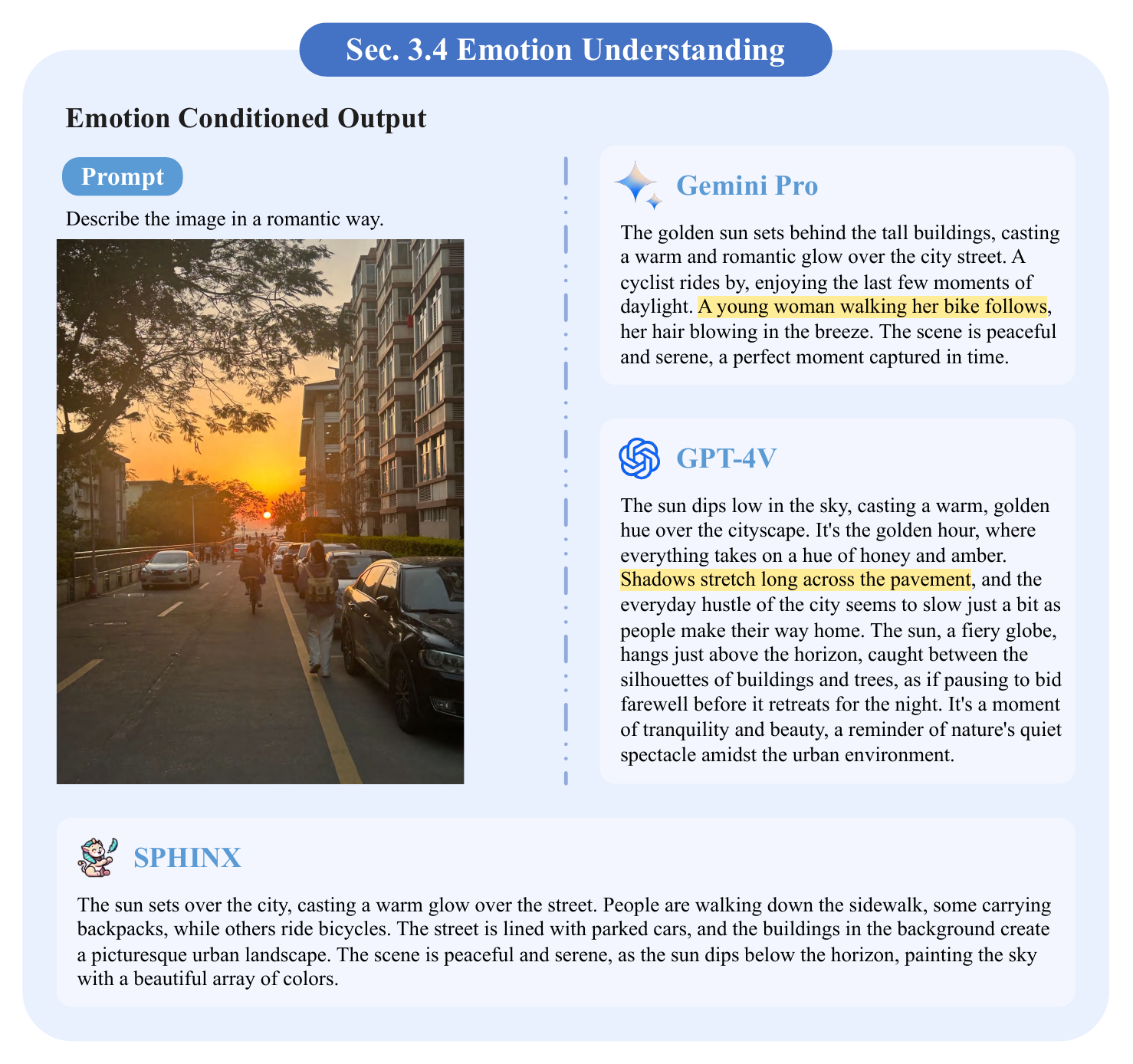}}
  \caption[Section~\ref{sec:04subsec:emotion}: emotion-conditioned output.]{Results on emotion-conditioned output. \colorbox{yellow!70!yellowhl}{Yellow} highlights the incompetence in performing the task. Refer to Section \ref{sec:04subsec:emotion} for detailed discussions.}
  \label{cond-1}
\end{figure*}

\begin{figure*}[!ht]
  \centering 
  \makebox[\textwidth][c]{\includegraphics[width=1.2\textwidth]{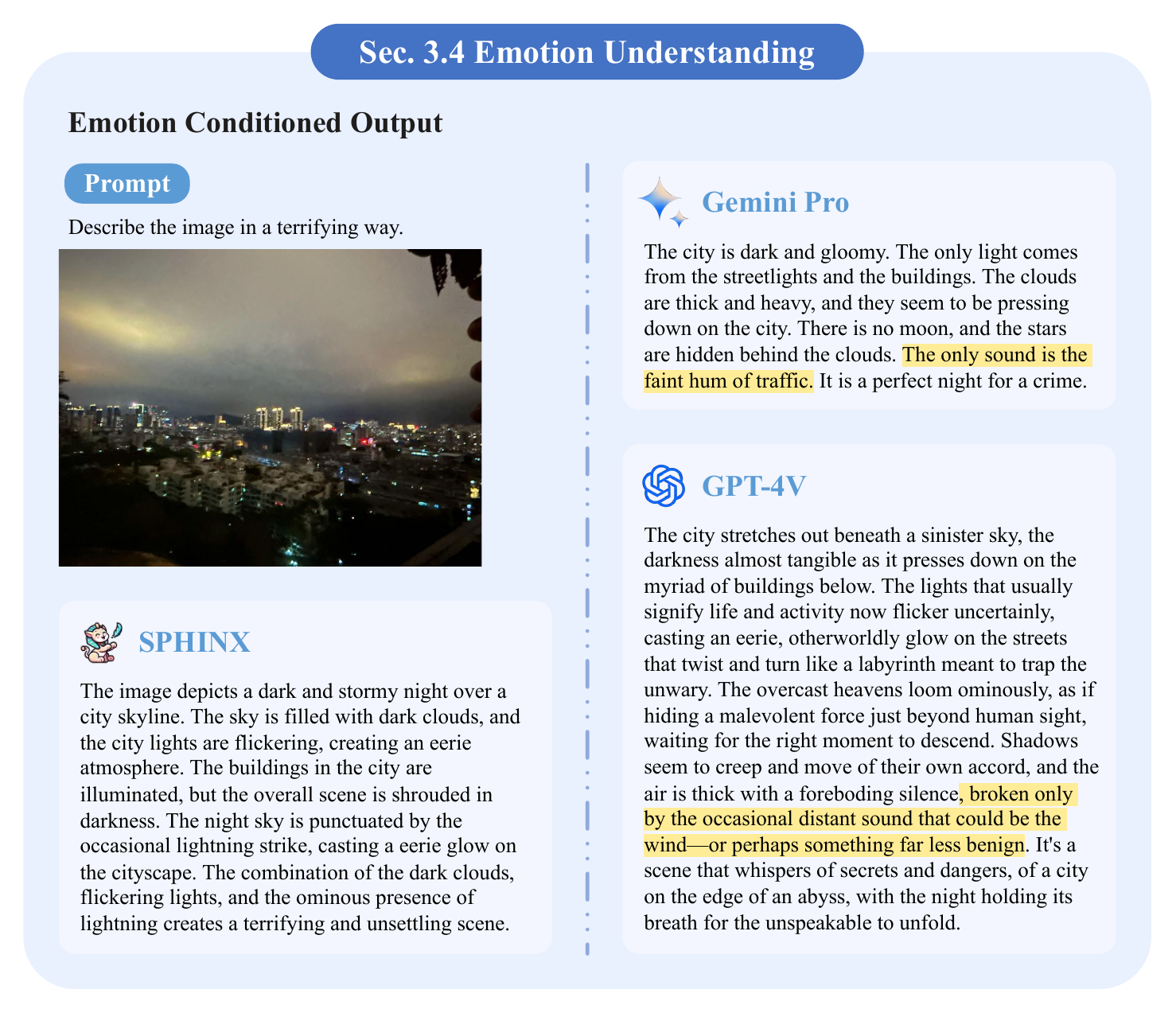}}
  \caption[Section~\ref{sec:04subsec:emotion}: emotion-conditioned output.]{Results on emotion-conditioned output.  \colorbox{yellow!70!yellowhl}{Yellow} highlights the incompetence in performing the task. Refer to Section \ref{sec:04subsec:emotion} for detailed discussions.}
  \label{cond-2}
\end{figure*}

\textbf{Emotion-conditioned output.} Different from predicting the emotion in the image, this emotion-conditioned output enables MLLMs to describe the visual context conditioned on a pre-defined emotion, such as ``in a romantic or terrifying way''. As shown in Figures~\ref{cond-1}-\ref{cond-2}, although Gemini and GPT-4V can correctly inject the corresponding emotion into the generated text, they both encountered hallucination issues, i.e., describing something that doesn't exist, such as the ``bike'' and ``shadows'' in the first image, and the ``sound'' in the second image. This is might because of the too-powerful correlation abilities. In contrast, Sphinx is free from this issue, which exhibits the advantage of human-replicated MLLMs.

\subsection{Game Playing}\label{sec:04subsec:game}

\begin{figure*}[!ht]
  \centering 
  \makebox[\textwidth][c]{\includegraphics[width=0.98\textwidth]{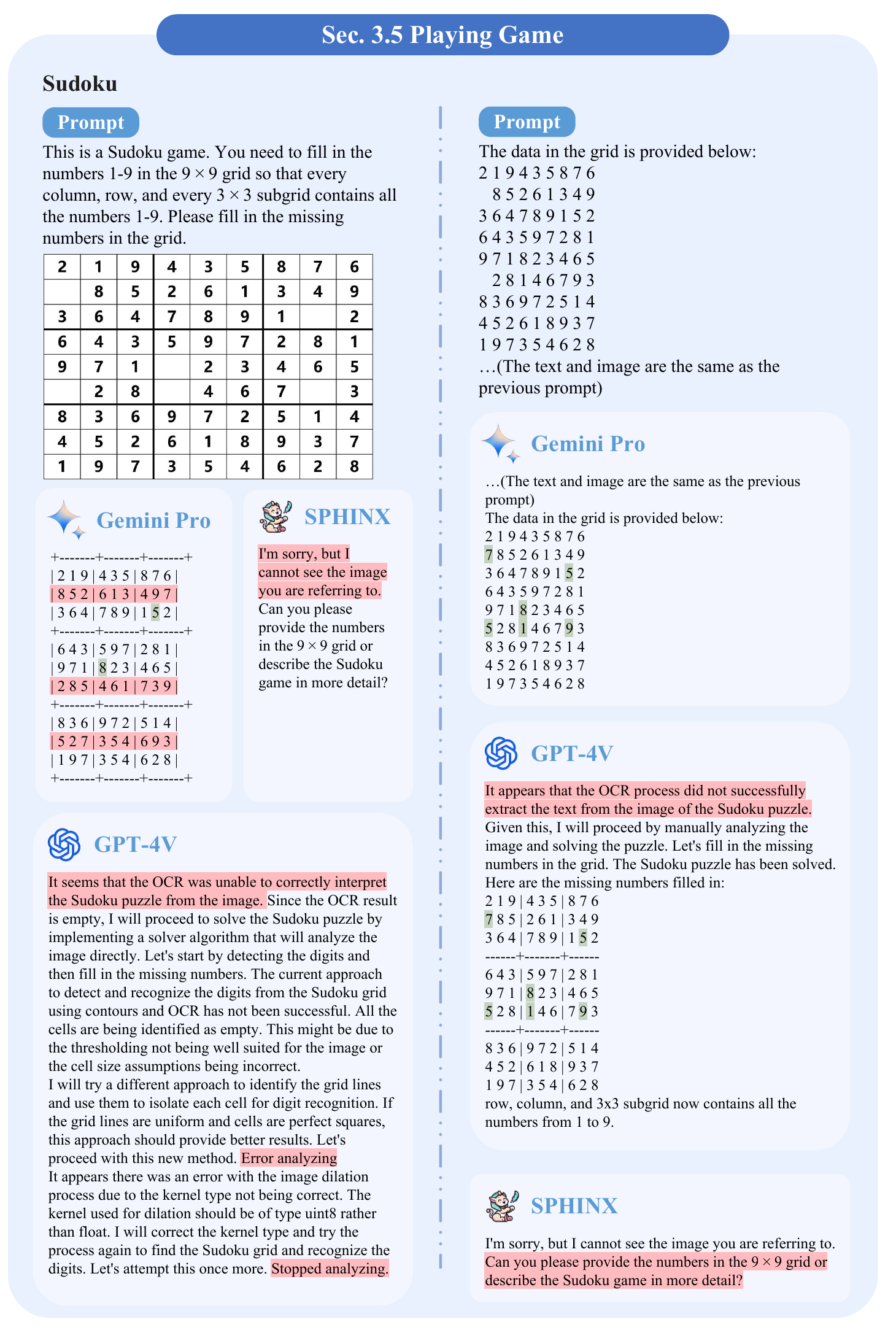}}
  \caption[Section~\ref{sec:04subsec:game}: Sudoku.]{Results on Sudoku playing. \colorbox{greenhl!85!black}{Green} highlights the right answer. \colorbox{red!30}{Red} highlights the wrong answer. Refer to Section \ref{sec:04subsec:game} for detailed discussions.}
  \label{sudoku-1}
\end{figure*}

\begin{figure*}[!ht]
  \centering 
  \makebox[\textwidth][c]{\includegraphics[width=1.1\textwidth]{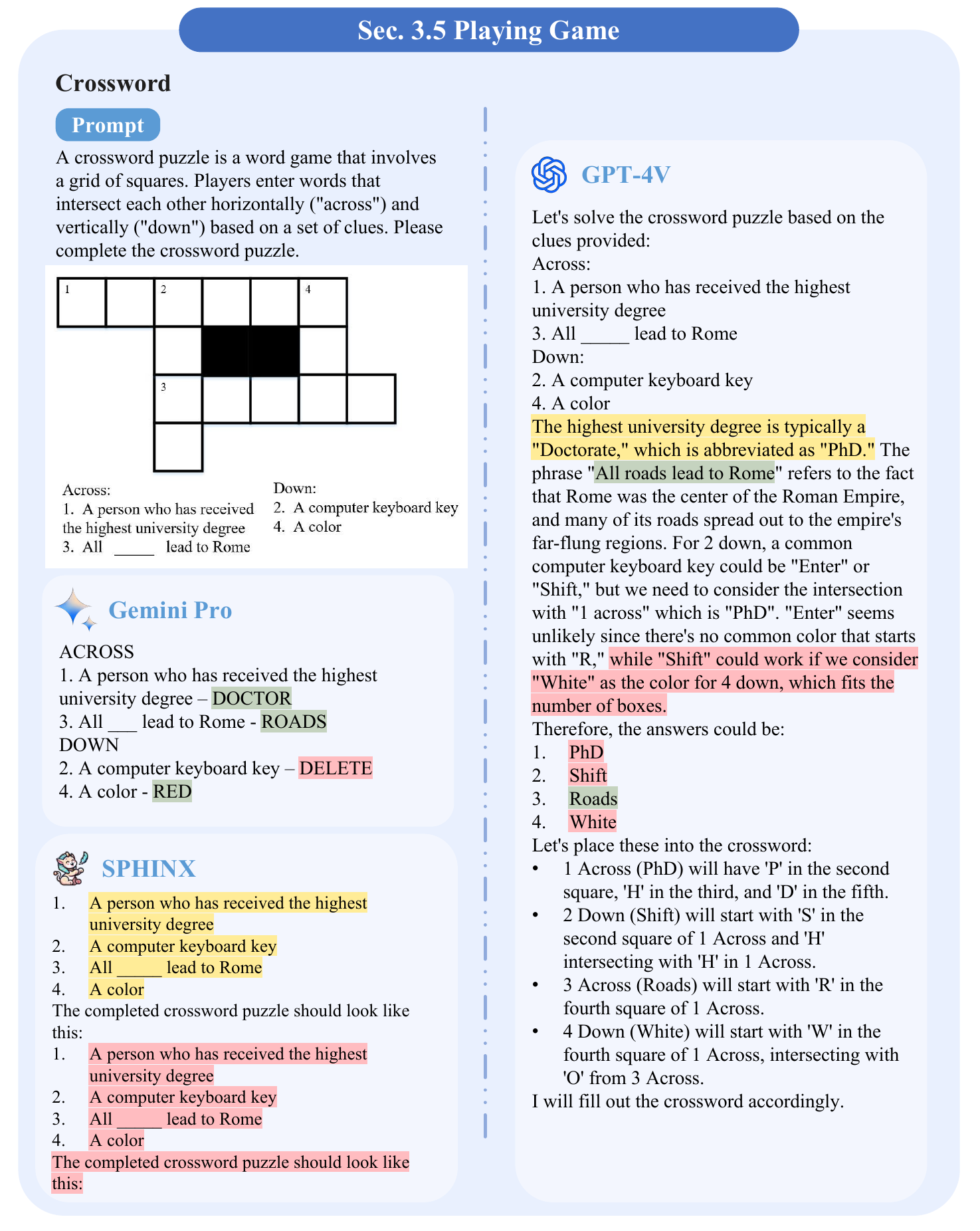}}
  \caption[Section~\ref{sec:04subsec:game}: Crossword.]{Results on Crossword playing. \colorbox{greenhl!85!black}{Green} highlights the right answer. \colorbox{red!30}{Red} highlights the wrong answer. \colorbox{yellow!70!yellowhl}{Yellow} highlights the incompetence in performing the task. Refer to Section \ref{sec:04subsec:game} for detailed discussions.}
  \label{sudoku-2}
\end{figure*}

\begin{figure*}[!ht]
  \centering 
  \makebox[\textwidth][c]{\includegraphics[width=1.1\textwidth]{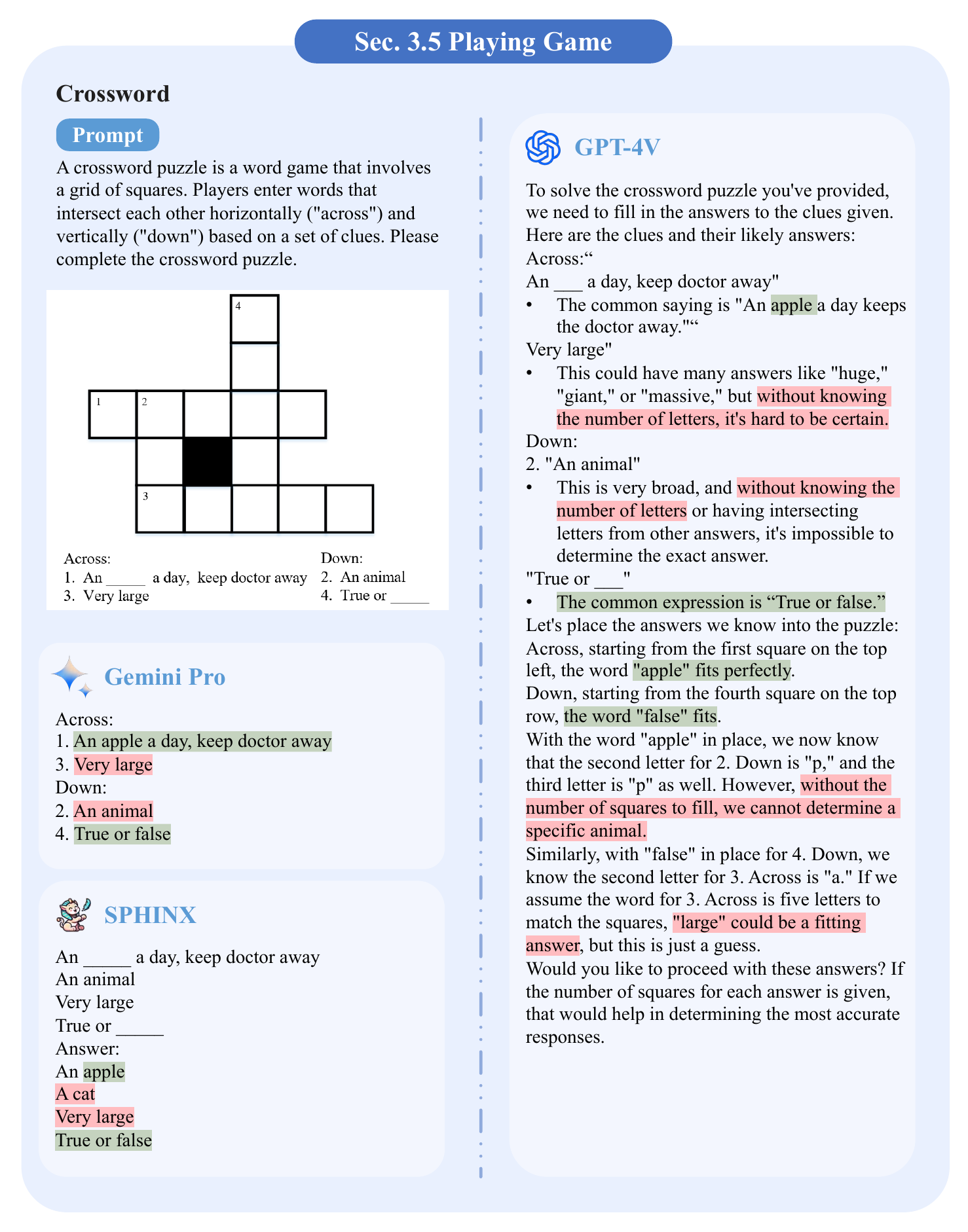}}
  \caption[Section~\ref{sec:04subsec:game}: Crossword.]{Results on Crossword playing. \colorbox{greenhl!85!black}{Green} highlights the right answer. \colorbox{red!30}{Red} highlights the wrong answer. Refer to Section \ref{sec:04subsec:game} for detailed discussions.}
  \label{sudoku-3}
\end{figure*}

\textbf{Sudoku and Crossword.}
These two games are logic-based and combinatorial puzzles, evaluating MLLMs' capabilities of OCR, structural understanding, and semantic reasoning. As shown in Figure~\ref{sudoku-1} for Sudoku playing, if only taking an image as input, although Gemini tries to provide the answer within the output matrix, it cannot correctly identify the blank position, while GPT-4V and Sphinx fail to conduct the first-step OCR. Further, given the corresponding text input, both Gemini and GPT-4V can respond with the right answer. In Figures~\ref{sudoku-2}-\ref{sudoku-3} for Crossword, the output of Sphinx is not very stable for the two examples, and Gemini performs slightly better than GPT-4V with more correct answers, which demonstrates its better structural comprehension and reasoning capacity.

\begin{figure*}[!ht]
  \centering 
  \makebox[\textwidth][c]{\includegraphics[width=1.1\textwidth]{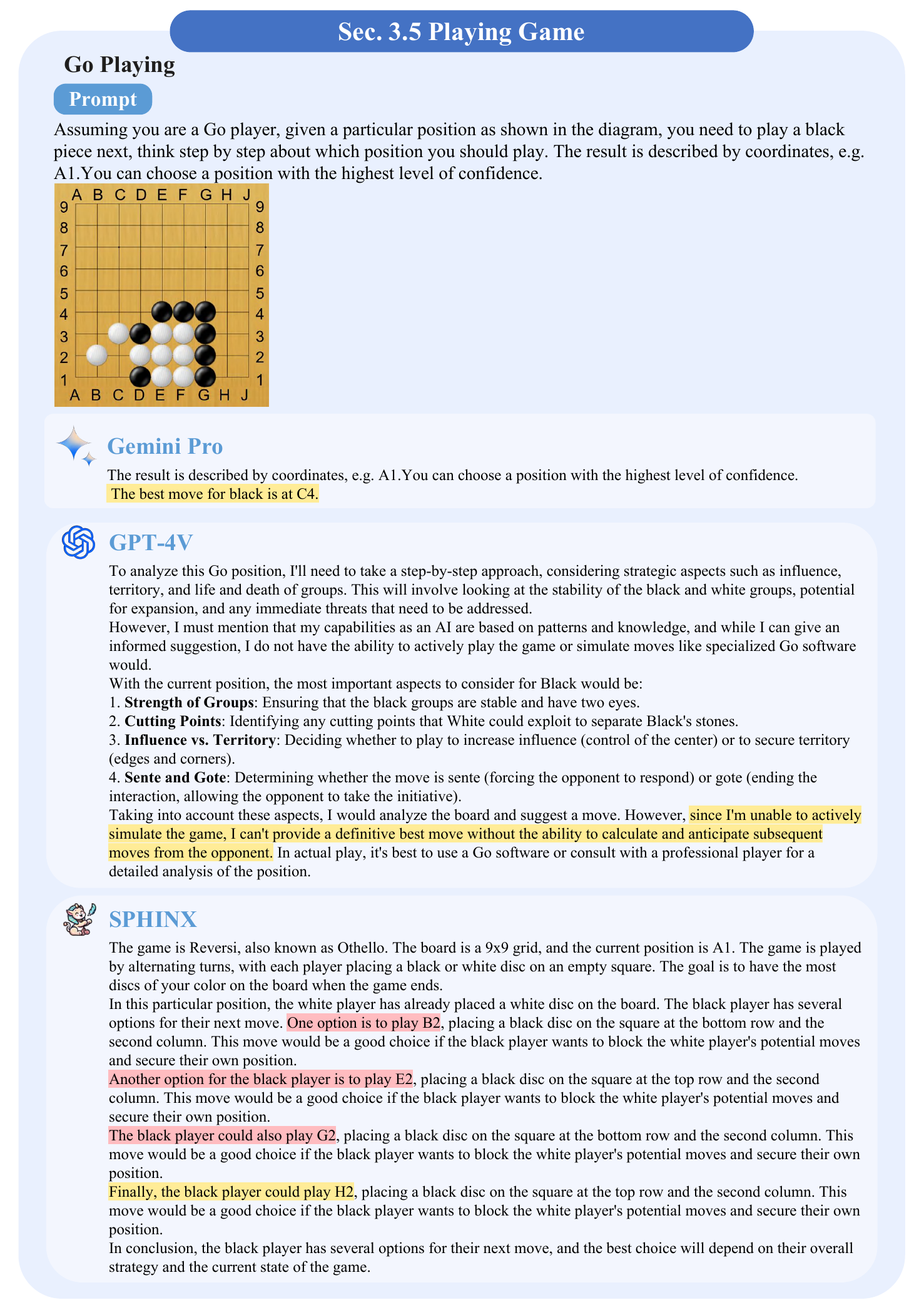}}
  \caption[Section~\ref{sec:04subsec:game}: Go playing.]{Results on Go playing. The optimal move for reference is `C2'. \colorbox{red!30}{Red} highlights the wrong answer. \colorbox{yellow!70!yellowhl}{Yellow} highlights the incompetence in performing the task. Refer to Section \ref{sec:04subsec:game} for detailed discussions.}
  \label{go-1}
\end{figure*}

\begin{figure*}[!ht]
  \centering 
  \makebox[\textwidth][c]{\includegraphics[width=1.1\textwidth]{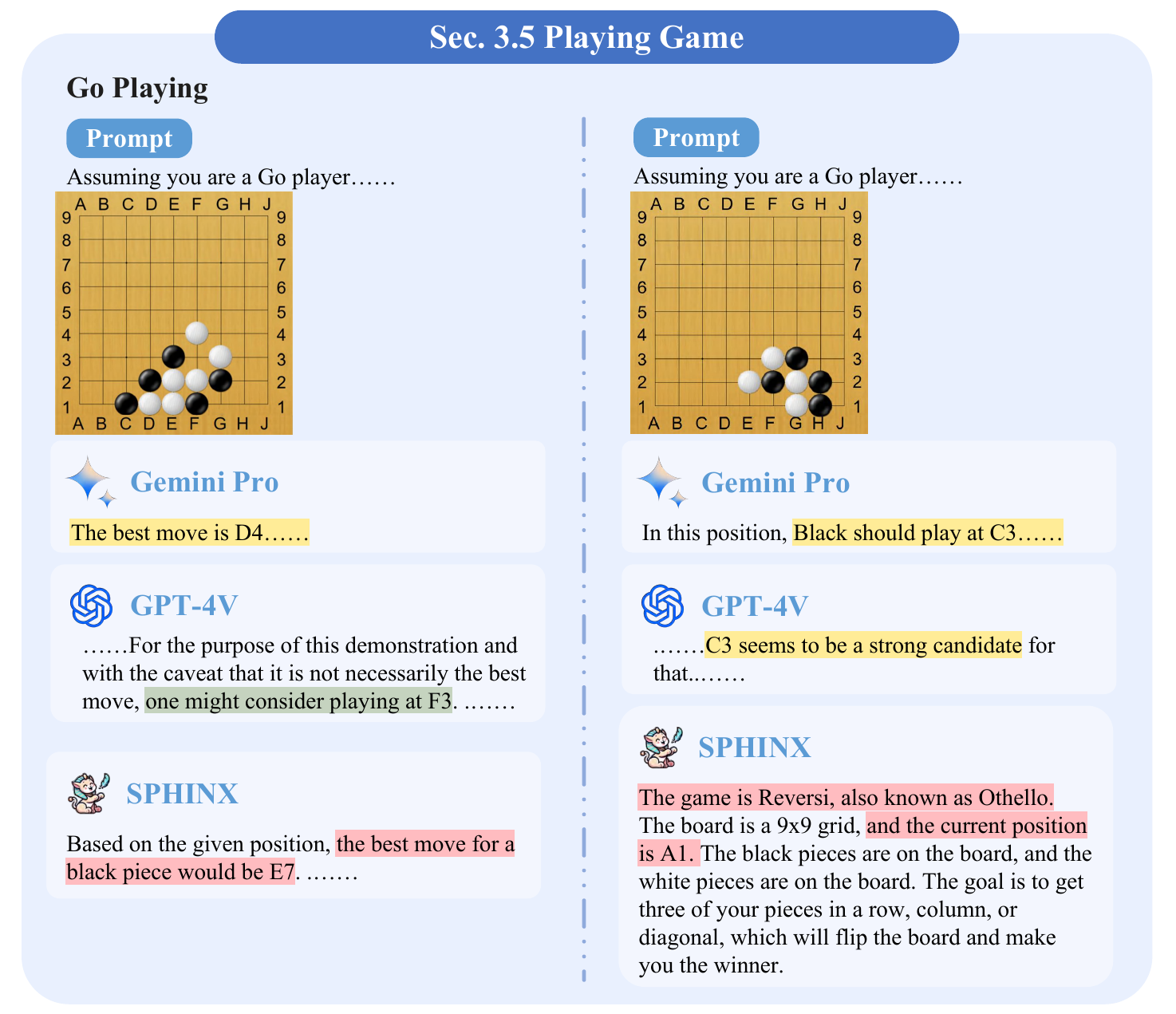}}
  \caption[Section~\ref{sec:04subsec:game}: Go playing.]{Results on Go playing. The optimal moves for reference are `F3' and `F1', respectively. \colorbox{greenhl!85!black}{Green} highlights the right answer. \colorbox{red!30}{Red} highlights the wrong answer. \colorbox{yellow!70!yellowhl}{Yellow} highlights the incompetence in performing the task. Refer to Section \ref{sec:04subsec:game} for detailed discussions.}
  \label{go-2}
\end{figure*}

\textbf{Go playing.}
The game of Go involves intricate strategic thinking and rule-based decision-making, which has been regarded as a touchstone for general artificial intelligence. In Figures~\ref{go-1}-\ref{go-2}, we verify whether these three MLLMs obtain the ability to play go. As shown, they can master some initial skills, and predict the next move to a legal position in the chessboard. Although different strategies have different moving, only GPT-4V in the second sample obtains the optimal answer. However, 
Sphinx mistakenly views the game to be a Reversi in the third sample. This illustrates that current MLLMs still have a certain distance to cover before mastering advanced chess-playing capabilities.

%% file: 04-vision.tex
\section{Vision Task}
\label{sec:04vison}
In this section, our objective is to assess the performance of MLLMs in various challenging vision tasks that extend beyond the scope of standard visual question-answering. 
Such tasks demand profound visual perception and understanding abilities from MLLMs. 
Evaluating their performance in these areas will provide insights into the viability of MLLMs as versatile generalist.

In Section \ref{sec04:subsec:image}, our discussion will focus on the capability of models to perform vision-related tasks at the image-level, including object detection, referring expression comprehension, phrase localization, and face detection and recognition. 
In Section \ref{sec04:subsec:temporal}, the exploration will shift to the vision tasks based on temporal sequence, such as video action recognition, object tracking, and visual story generation. 

\subsection{Image-Level Vision Task}\label{sec04:subsec:image}
\begin{figure*}[!ht]
  \centering 
  \makebox[\textwidth][c]{\includegraphics[width=1.15\textwidth]{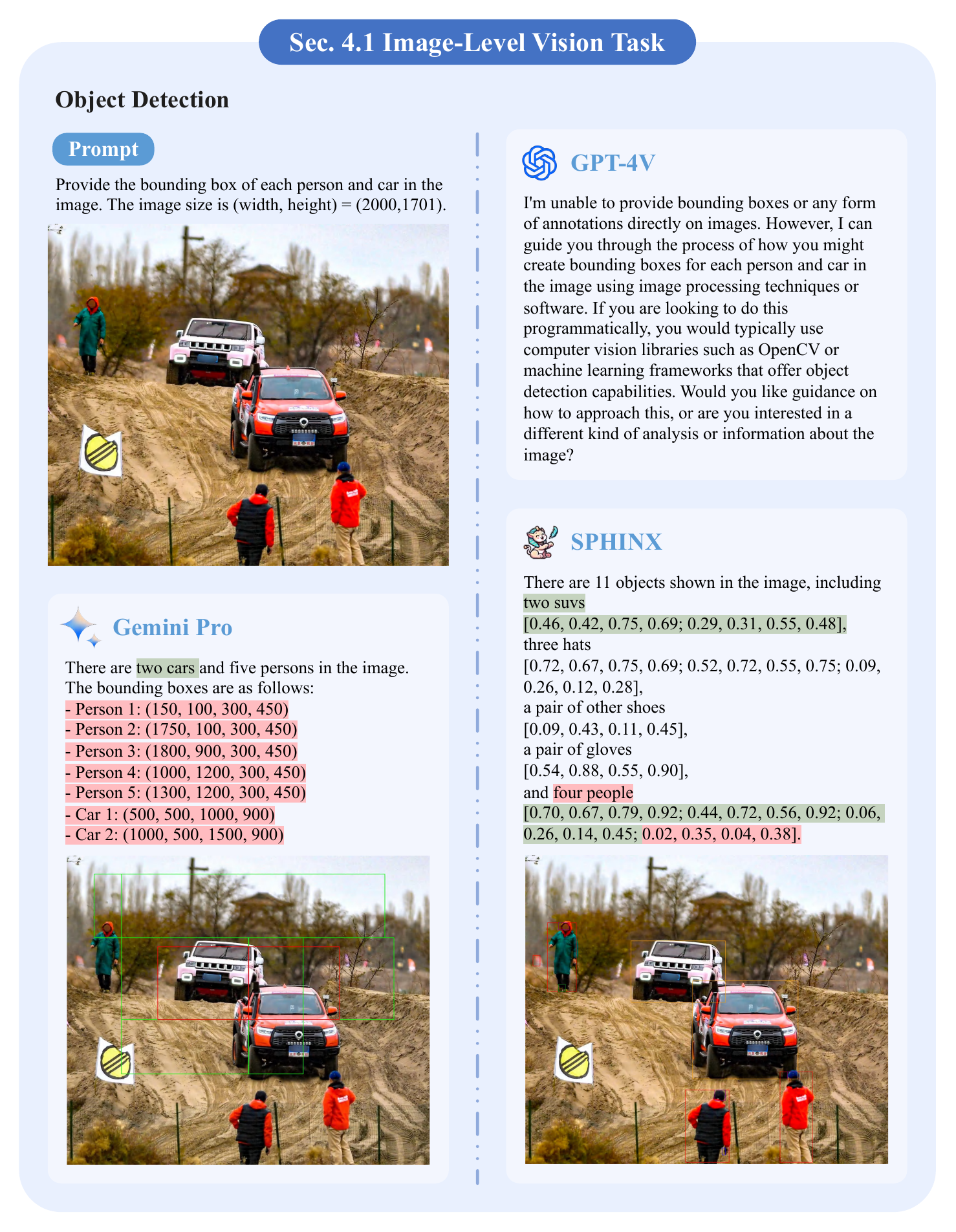}}
  \caption[Section~\ref{sec04:subsec:image}: object detection.]{Results on object detection. Gemini's bounding boxes are often imprecise, while GPT-4V typically avoids providing coordinates directly, preferring to use external tools. Sphinx, though offering more reliable responses than the others, still falls notably short of the performance achieved by standard object detectors. \colorbox{greenhl!85!black}{Green} highlights the right answer. \colorbox{red!30}{Red} highlights the wrong answer. Refer to Section \ref{sec04:subsec:image} for detailed discussions.}
  \label{detection}
\end{figure*}

\textbf{Object detection.} 
We explore the models' ability to undertake the task of object detection \cite{carion2020end,girshick2015fast,redmon2016you}.
We prompt the models to provide a bounding box for each car and person in the image. 
As shown in Figure \ref{detection}, 
the bounding boxes provided by Gemini are generally imprecise, whereas GPT-4V avoids directly offering coordinates and instead attempts to utilize external tools.
Only Sphinx offers a relatively reliable answer, yet its performance falls significantly short when compared to that of conventional object detectors.

\begin{figure*}[!ht]
  \centering 
  \makebox[\textwidth][c]{\includegraphics[width=1.15\textwidth]{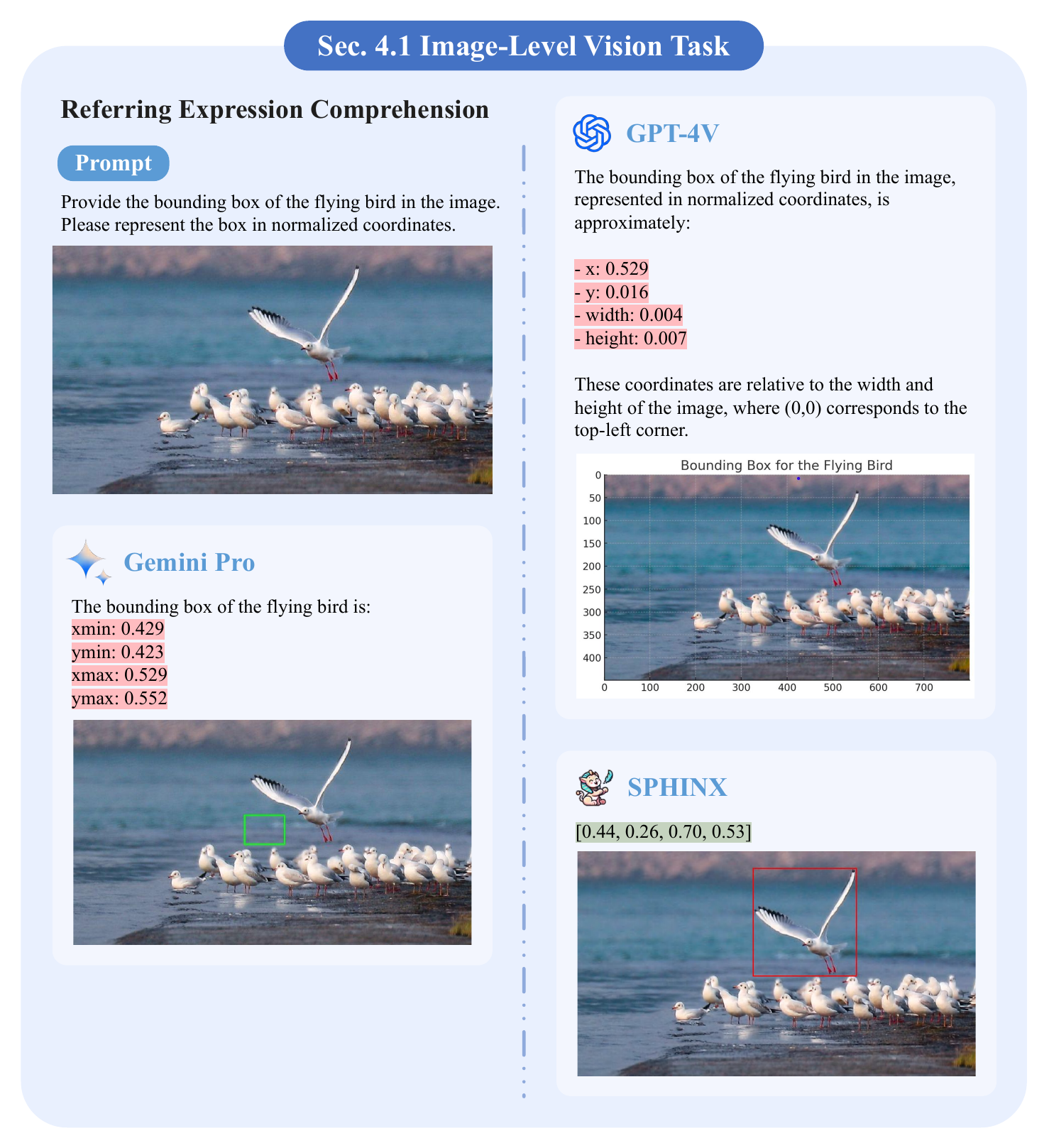}}
  \caption[Section~\ref{sec04:subsec:image}: referring expression comprehension.]{Results on referring expression comprehension. Only Sphinx provides satisfactory results. \colorbox{greenhl!85!black}{Green} highlights the right answer. \colorbox{red!30}{Red} highlights the wrong answer. Refer to Section \ref{sec04:subsec:image} for detailed discussions.}
  \label{rec-1}
\end{figure*}

\begin{figure*}[!ht]
  \centering 
  \makebox[\textwidth][c]{\includegraphics[width=1.2\textwidth]{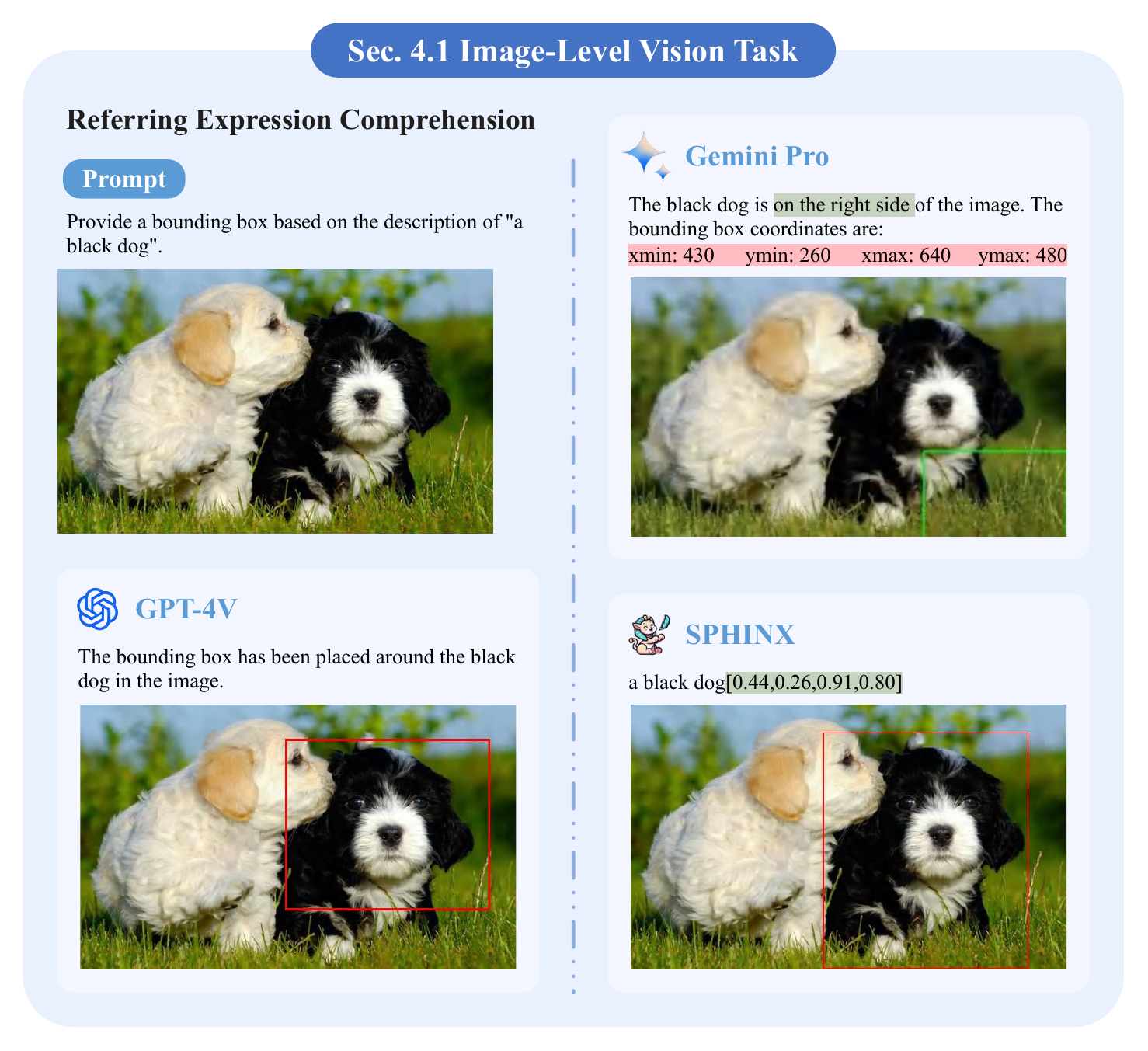}}
  \caption[Section~\ref{sec04:subsec:image}: referring expression comprehension.]{Results on referring expression comprehension. \colorbox{greenhl!85!black}{Green} highlights the right answer. \colorbox{red!30}{Red} highlights the wrong answer. Refer to Section \ref{sec04:subsec:image} for detailed discussions.}
  \label{rec-2}
\end{figure*}

\textbf{Referring expression comprehension.}
Here we assess the models' ability to provide the bounding box of the referring object \cite{liu2023grounding,yu2018mattnet}. 
We prompt the models to generate normalized bounding boxes. 
As illustrated in Figures \ref{rec-1}-\ref{rec-2}, both Gemini and GPT-4V are able to discern the approximate location of the referring object, yet they struggle to provide precise coordinates and box size. 
However, Sphinx demonstrates the capability to offer the exact location and size of the referring object.

\begin{figure*}[!ht]
  \centering 
  \makebox[\textwidth][c]{\includegraphics[width=1.2\textwidth]{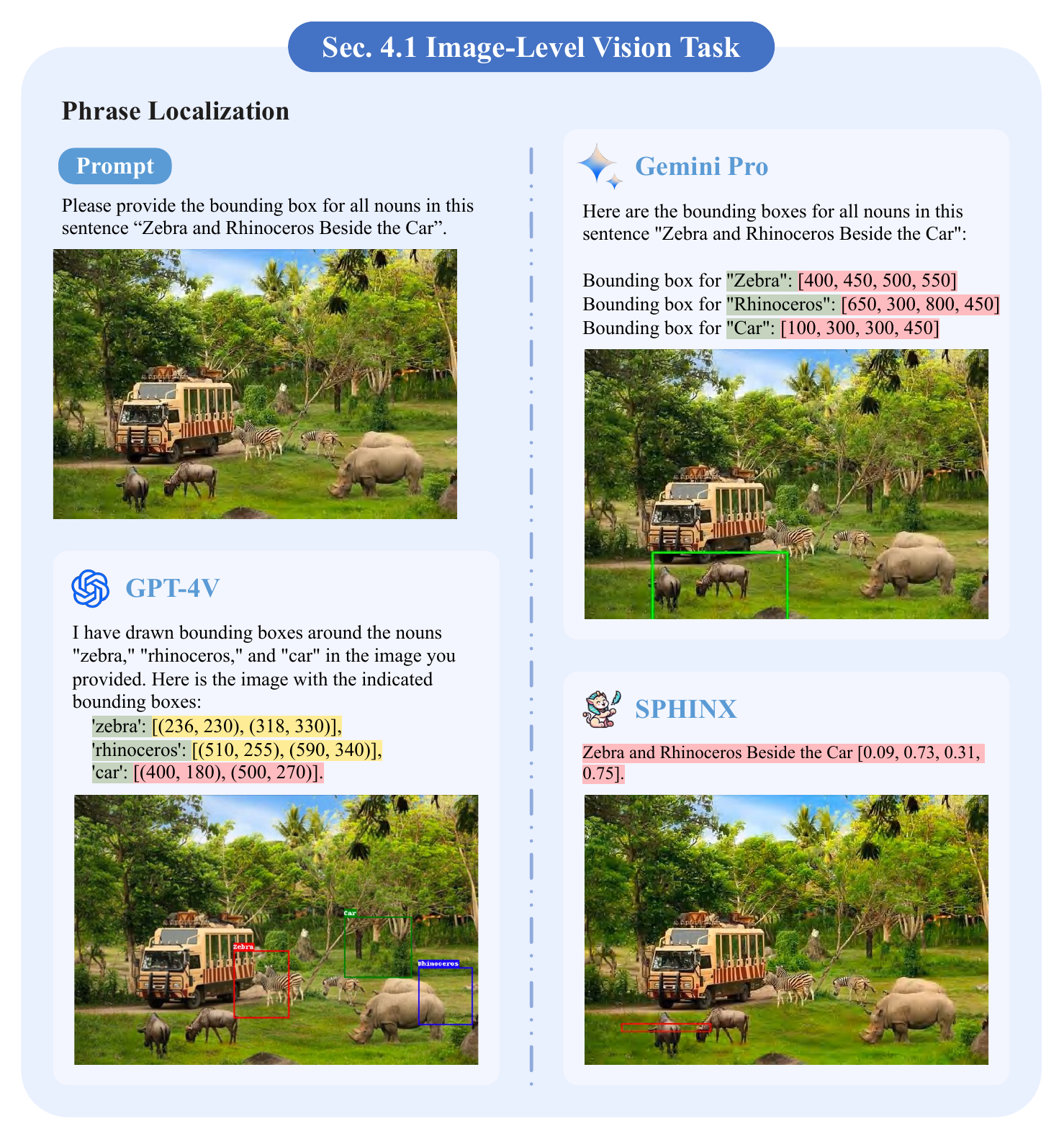}}
  \caption[Section~\ref{sec04:subsec:image}: phrase localization.]{Results on phrase localization. None of the three models provides correct results. The bounding boxes provided by Gemini have two out-of-bound issues, and the remaining one is also incorrect. Sphinx fails to comprehend the instruction, offering only a single bounding box. Among the bounding boxes supplied by GPT-4V, only the one for the zebra is nearly accurate. \colorbox{greenhl!85!black}{Green} highlights the right answer. \colorbox{red!30}{Red} highlights the wrong answer. \colorbox{yellow!70!yellowhl}{Yellow} highlights the incompetence in performing the task. Refer to Section \ref{sec04:subsec:image} for more discussions.}
  \label{phrase}
\end{figure*}

\textbf{Phrase localization.}
Here we assess the models' ability of Phrase Location \cite{plummer2017phrase}.
We prompt the models to provide the bounding boxes of all nouns in the description.
As shown in Figure \ref{phrase}, none of the three models exhibit satisfactory performance in executing this task.

\begin{figure*}[!ht]
  \centering 
  \makebox[\textwidth][c]{\includegraphics[width=1.2\textwidth]{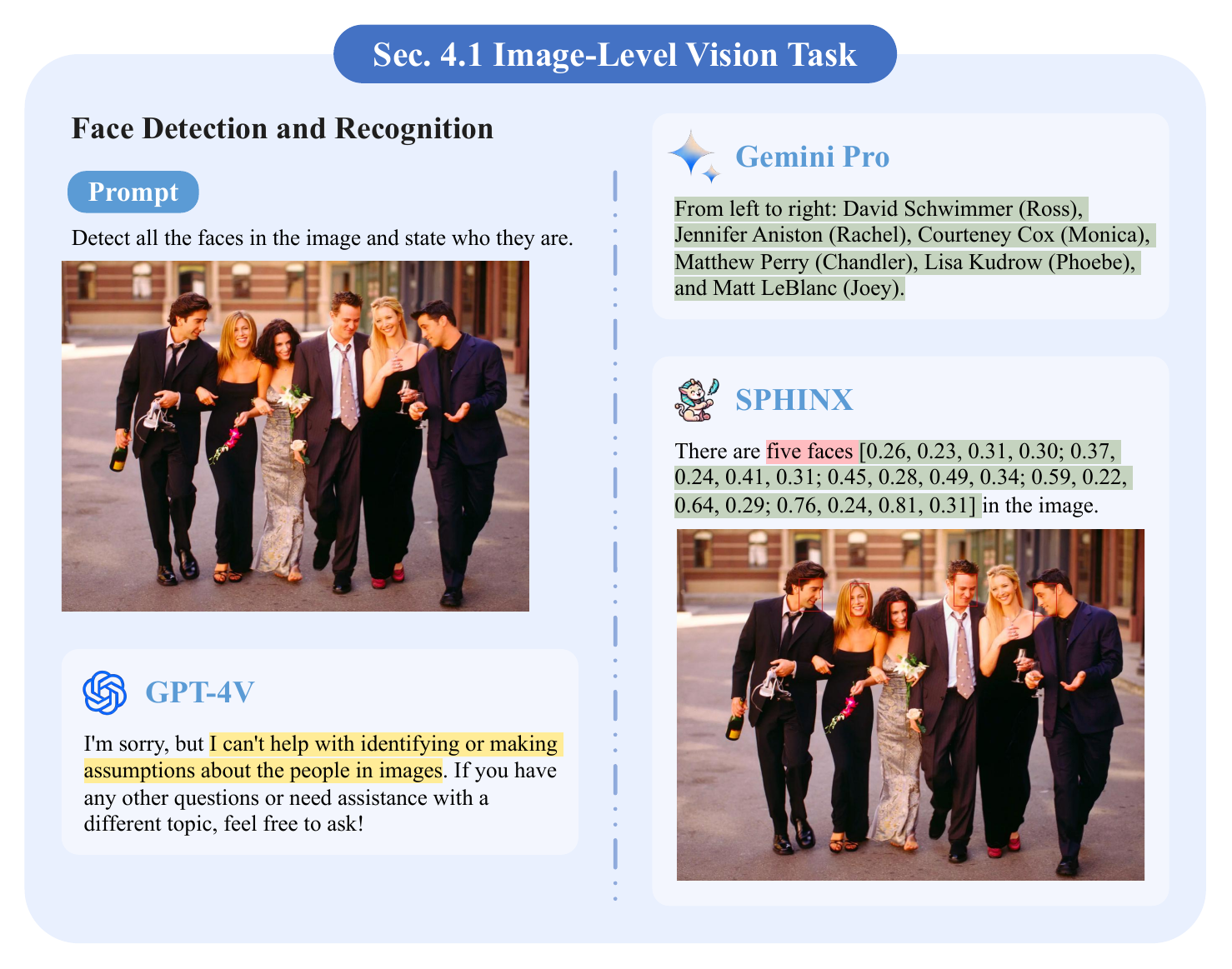}}
  \caption[Section~\ref{sec04:subsec:image}: face detection and recognition.]{Results on face detection and recognition. Gemini identifies the names corresponding to each face in the image in a left-to-right order. In contrast, GPT-4V issues a statement regarding privacy and opts not to fulfill the request. Sphinx, on the other hand, generates bounding boxes that are nearly accurate but fall short of recognizing the identities of the faces. \colorbox{greenhl!85!black}{Green} highlights the right answer. \colorbox{red!30}{Red} highlights the wrong answer. \colorbox{yellow!70!yellowhl}{Yellow} highlights the incompetence in performing the task. Refer to Section \ref{sec04:subsec:image} for detailed discussions.}
  \label{face}
\end{figure*}

\textbf{Face detection and recognition } \cite{jiang2017face,yang2016wider} is an important task in computer vision.  
We prompt the models via the text query "Detect all the faces in the image and state who they are?".
As illustrated in Figure \ref{face}, without being explicitly instructed to provide bounding boxes, Gemini accurately identifies the arrangement of all faces in the image and precisely recognizes the corresponding names for each face. 
Meanwhile, GPT-4V issues a privacy statement and evades the request. Sphinx's response provides almost correct bounding boxes but does not recognize these faces.

\subsection{Temporal-Level Vision Task}\label{sec04:subsec:temporal}

\begin{figure*}[!ht]
  \centering 
  \makebox[\textwidth][c]{\includegraphics[width=1.1\textwidth]{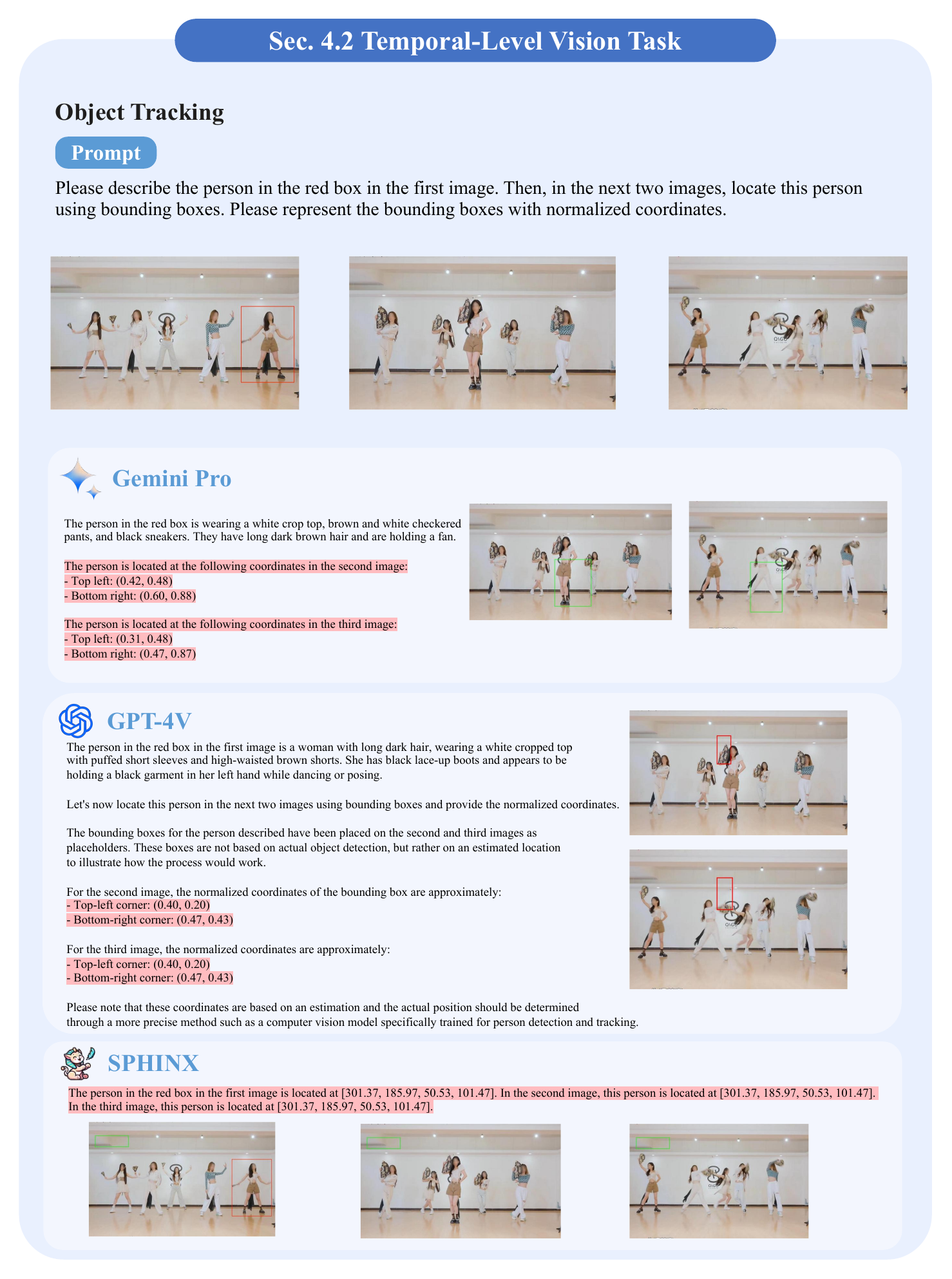}}
  \caption[Section~\ref{sec04:subsec:temporal}: object tracking.]{Result on object tracking. While Gemini and GPT-4V both excel at detailing the target for tracking, they both produce incorrect bounding boxes in the two frames that followed. \colorbox{red!30}{Red} highlights the wrong answer. Refer to Section \ref{sec04:subsec:temporal} for more discussions.}
  \label{tracking}
\end{figure*}

\textbf{Object tracking.} 
Here we explore the models' capacity of object tracking  \cite{wu2013online,wang2019fast,luo2021multiple}.
As illustrated in Figure \ref{tracking}, although both Gemini and GPT-4V are capable of delineating the details of the target to be tracked, they subsequently provide incorrect bounding boxes in the following two frames. 
The response from Sphinx indicates that it fails to comprehend the intent of our instructions.

\begin{figure*}[!ht]
  \centering 
  \makebox[\textwidth][c]{\includegraphics[width=1.1\textwidth]{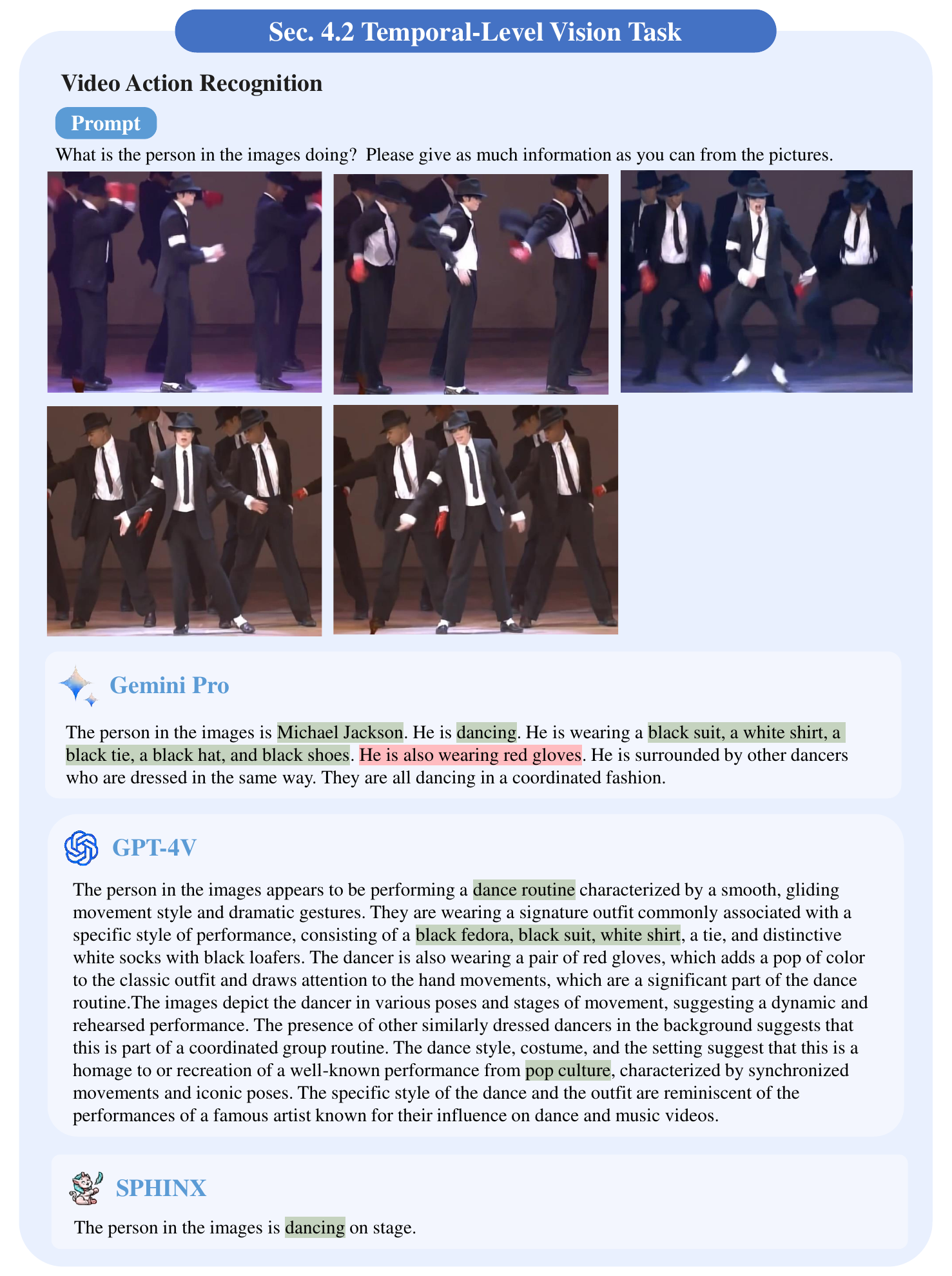}}
  \caption[Section~\ref{sec04:subsec:temporal}: video action recognition.]{Results on video action recognition. \colorbox{greenhl!85!black}{Green} highlights the right answer. \colorbox{red!30}{Red} highlights the wrong answer. Refer to Section \ref{sec04:subsec:temporal} for detailed discussions.}
  \label{action-1}
\end{figure*}

\begin{figure*}[!ht]
  \centering 
  \makebox[\textwidth][c]{\includegraphics[width=1.1\textwidth]{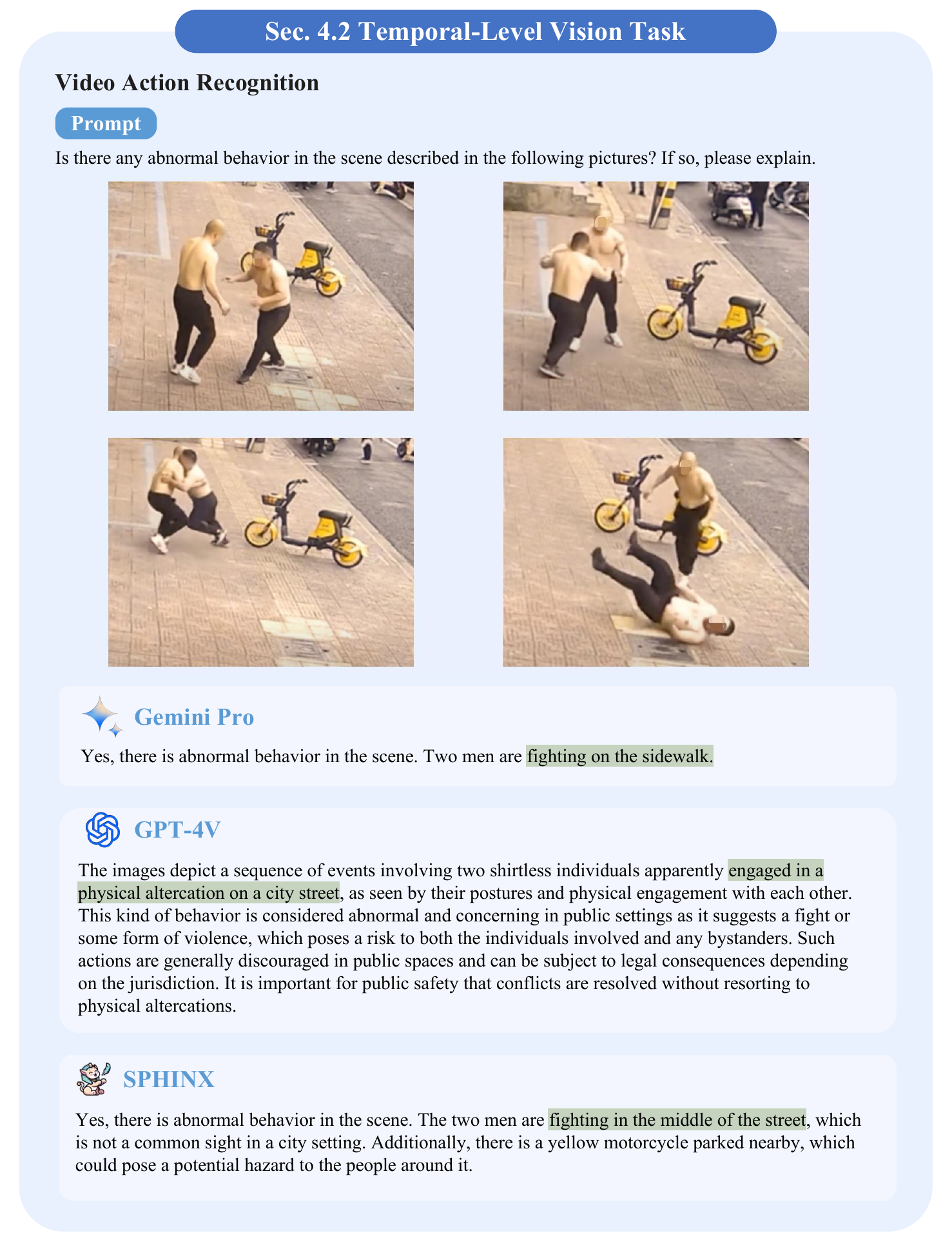}}
  \caption[Section~\ref{sec04:subsec:temporal}: video action recognition.]{Results on video action recognition. \colorbox{greenhl!85!black}{Green} highlights the right answer.  Refer to Section \ref{sec04:subsec:temporal} for detailed discussions.}
  \label{action-2}
\end{figure*}

\begin{figure*}[!ht]
  \centering 
  \makebox[\textwidth][c]{\includegraphics[width=1.1\textwidth]{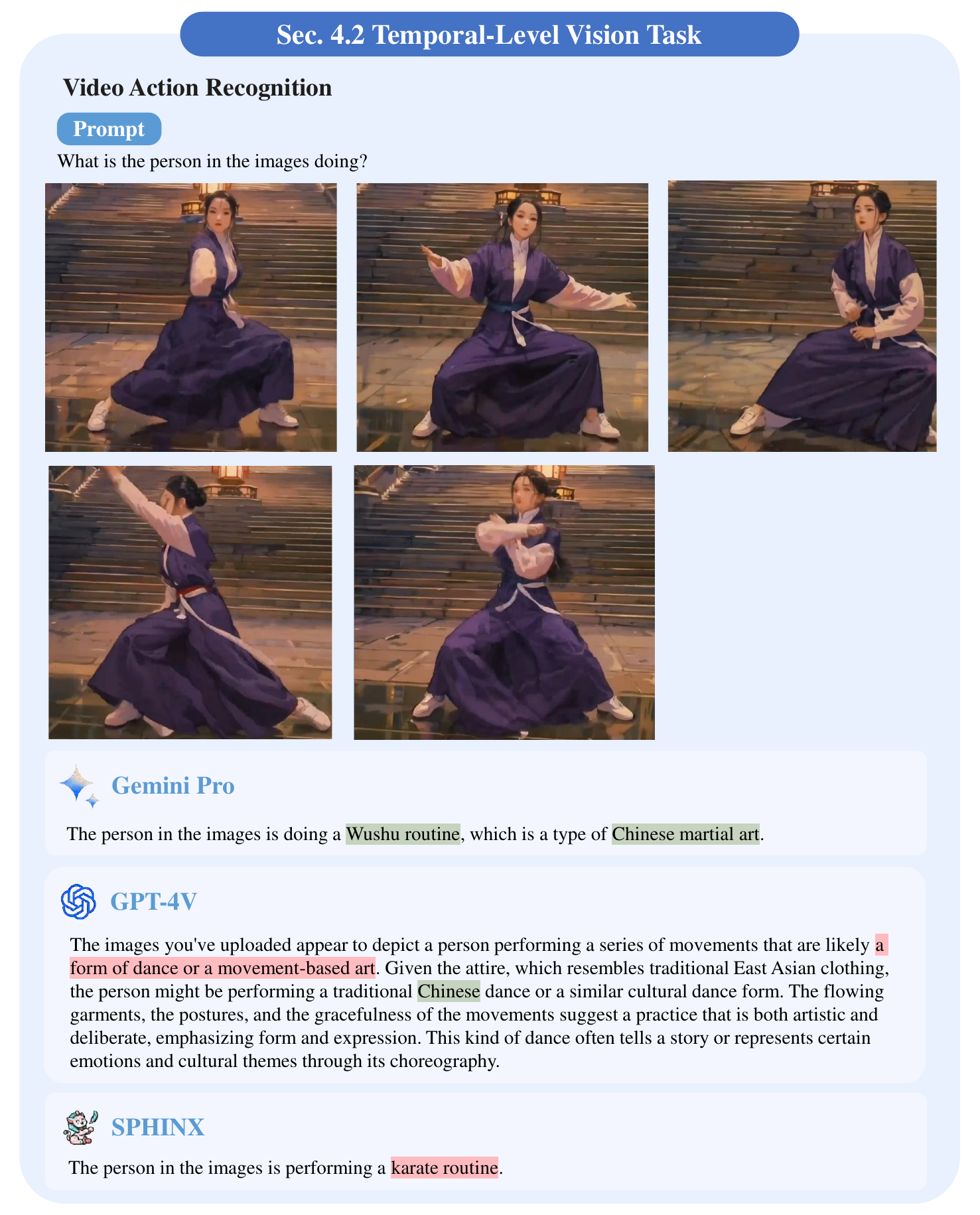}}
  \caption[Section~\ref{sec04:subsec:temporal}: video action recognition.]{Results on video action recognition. \colorbox{greenhl!85!black}{Green} highlights the right answer. \colorbox{red!30}{Red} highlights the wrong answer. Refer to Section \ref{sec04:subsec:temporal} for detailed discussions.}
  \label{action-3}
\end{figure*}

\textbf{Video action recognition.} 
Figures \ref{action-1}-\ref{action-3} demonstrate the models' ability to recognize the people's actions in video \cite{feichtenhofer2016convolutional,li2020tea,feichtenhofer2020x3d,arnab2021vivit}. 
We extract five representative frames from a video segment and input them into the model. 
As illustrated in Figure \ref{action-1}, both Gemini and GPT-4V demonstrate the capability to recognize the action in the images and provide a detailed description. 
Although Sphinx's response is correct, it lacks detailed descriptions. 

\begin{figure*}[!ht]
  \centering 
  \makebox[\textwidth][c]{\includegraphics[width=1.0\textwidth]{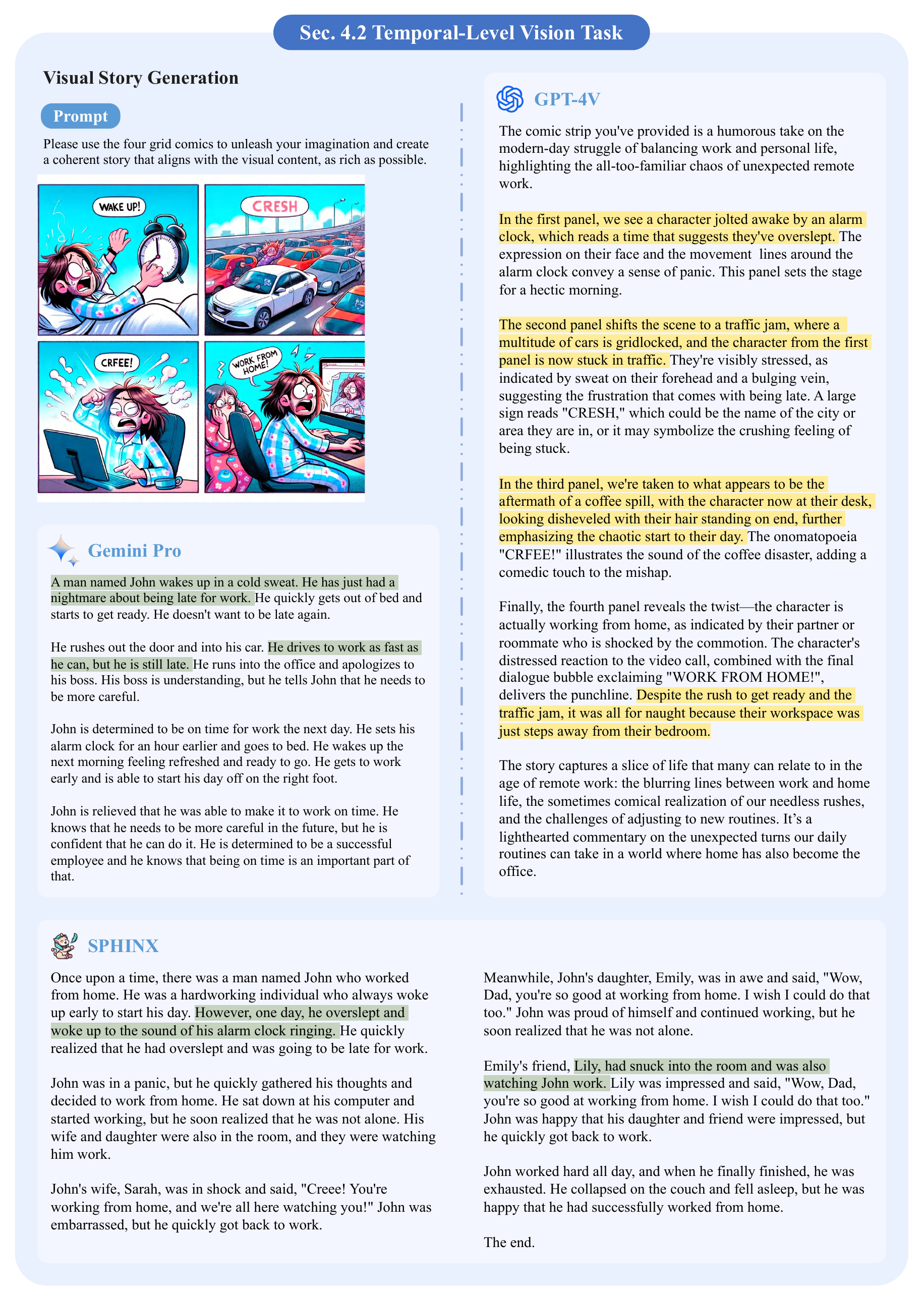}}
  \caption[Section~\ref{sec04:subsec:temporal}: visual story generation.]{Results on visual story generation. Gemini provides a coherent story that aligns well with the first two comics. GPT-4V offers precise descriptions for each comic but does not provide a story. The story generated by Sphinx exhibits a relatively weak correlation with the comics. \colorbox{greenhl!85!black}{Green} highlights the right answer.  \colorbox{yellow!70!yellowhl}{Yellow} highlights the incompetence in performing the task. Refer to Section \ref{sec04:subsec:temporal} for detailed discussions.}
  \label{story}
\end{figure*}

\textbf{Visual story generation.}
This task requires the MLLMs to fully comprehend the information in the images and logically organize it within the generated story \cite{rahman2023make,huang2019hierarchically}.
As shown in Figure \ref{story}, Gemini provides a coherent story that aligns well with the first two comics.
GPT-4V offers precise descriptions for each illustration; however, it fails to weave them into a cohesive story as per the task requirements. 
Sphinx's story diverges further from the comic context, incorporating only certain elements from the illustrations. Moreover, the logic of its story appears somewhat inexplicable.

%% file: 05-expert.tex
\section{Expert Capacity}
\label{sec:07expert}
Expert capacity measures the generalization capabilities of MLLMs to apply their learned knowledge and skills to diverse professional domains. 
Besides the aforementioned perception and cognition tasks, the robustness of MLLMs within specialized and unique scenarios normally has more practical reference significance.

In this section, we explore the potentials of Gemini, GPT-4V, and Sphinx on five real-world applications: autonomous driving (Section \ref{sec:07:subsec:autonomous}), defect detection (Section \ref{sec:07:subsec:defect}), medical diagnosis (Section~\ref{sec:07:subsec:medical}), economic analysis (Section \ref{sec:07:subsec:economic}), surveillance and security (Section \ref{sec:07:subsec:security}), remote sensing image analysis (Section \ref{sec:07:subsec:remote}), and robot motion planning (Section \ref{sec:07:subsec:robot}).

\subsection{Autonomous Driving}
\label{sec:07:subsec:autonomous}
\begin{figure*}[!ht]
  \centering 
  \makebox[\textwidth][c]{\includegraphics[width=1.0\textwidth]{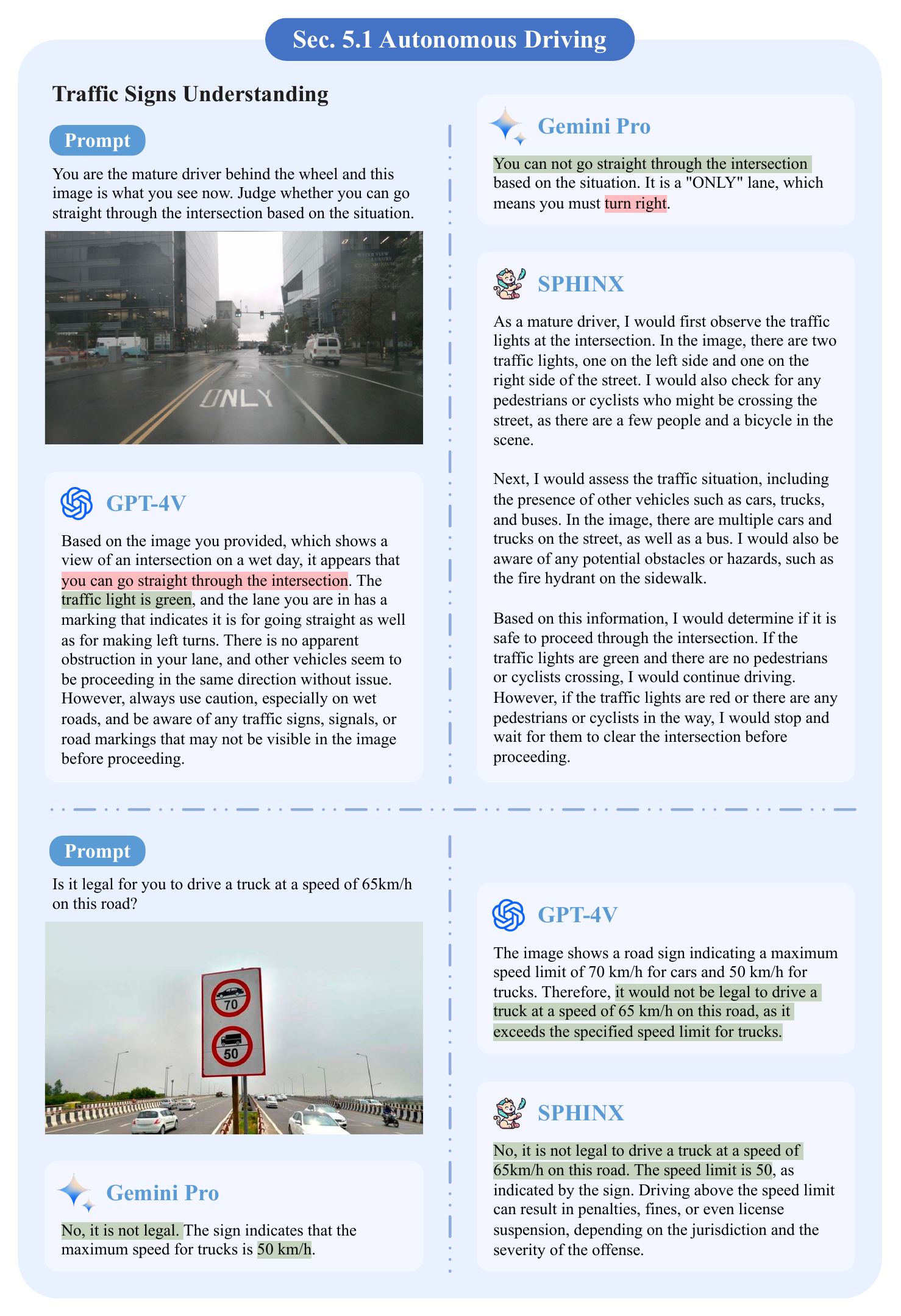}}
  \caption[Section~\ref{sec:07:subsec:autonomous}: traffic signs understanding.]{Results on traffic signs understanding. \colorbox{greenhl!85!black}{Green} highlights the right answer. \colorbox{red!30}{Red} highlights the wrong answer. Refer to Section \ref{sec:07:subsec:autonomous} for detailed discussions.}
  \label{autonomous-1}
\end{figure*}

\begin{figure*}[!ht]
  \centering 
  \label{autonomous-2}
  \makebox[\textwidth][c]{\includegraphics[width=1.1\textwidth]{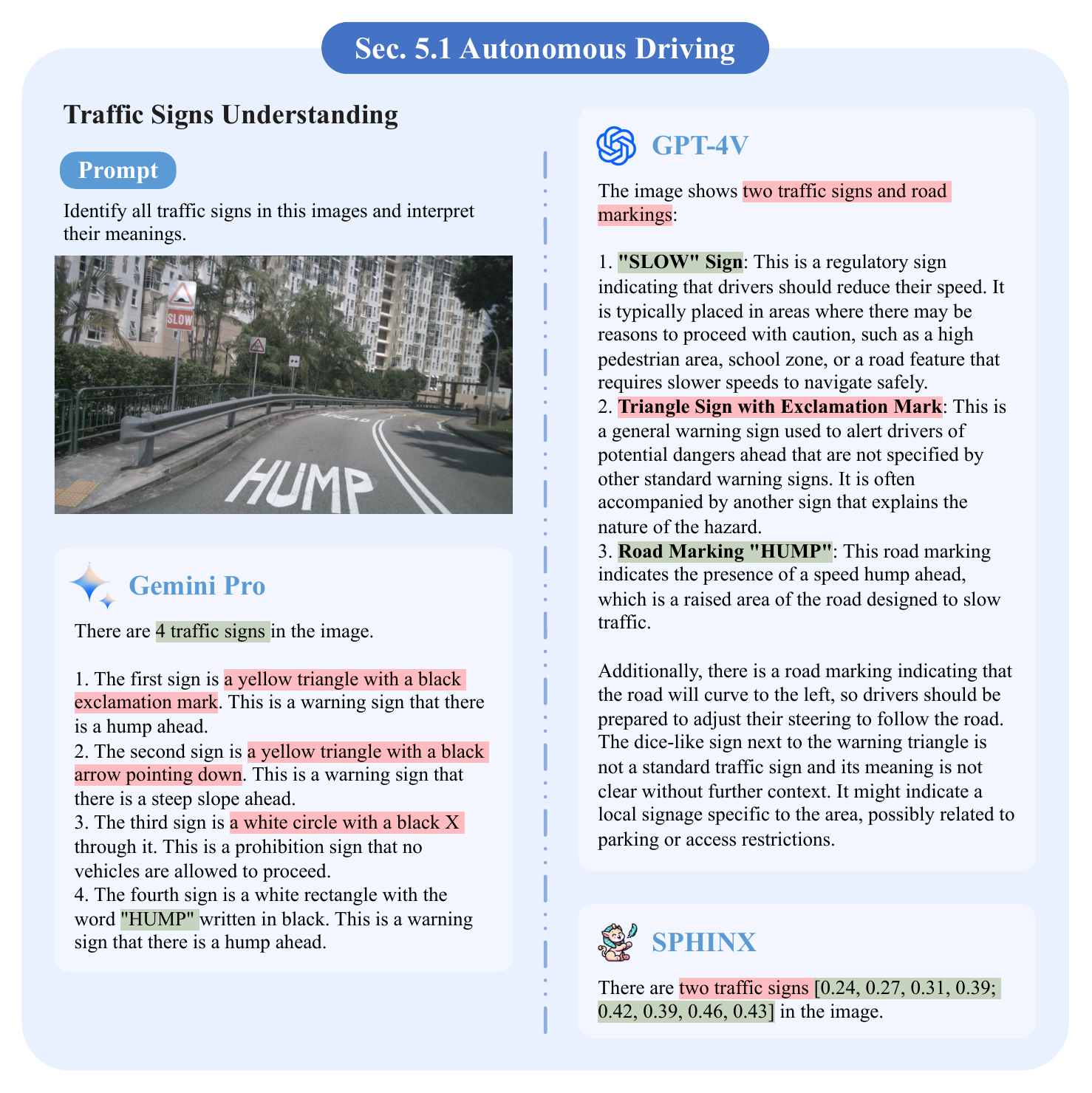}}
  \caption[Section~\ref{sec:07:subsec:autonomous}: traffic signs understanding.]{Results on traffic signs understanding. \colorbox{greenhl!85!black}{Green} highlights the right answer. \colorbox{red!30}{Red} highlights the wrong answer. Refer to Section \ref{sec:07:subsec:autonomous} for detailed discussions.}
\end{figure*}

\begin{figure*}[!ht]
  \centering 
  \makebox[\textwidth][c]{\includegraphics[width=1.1\textwidth]{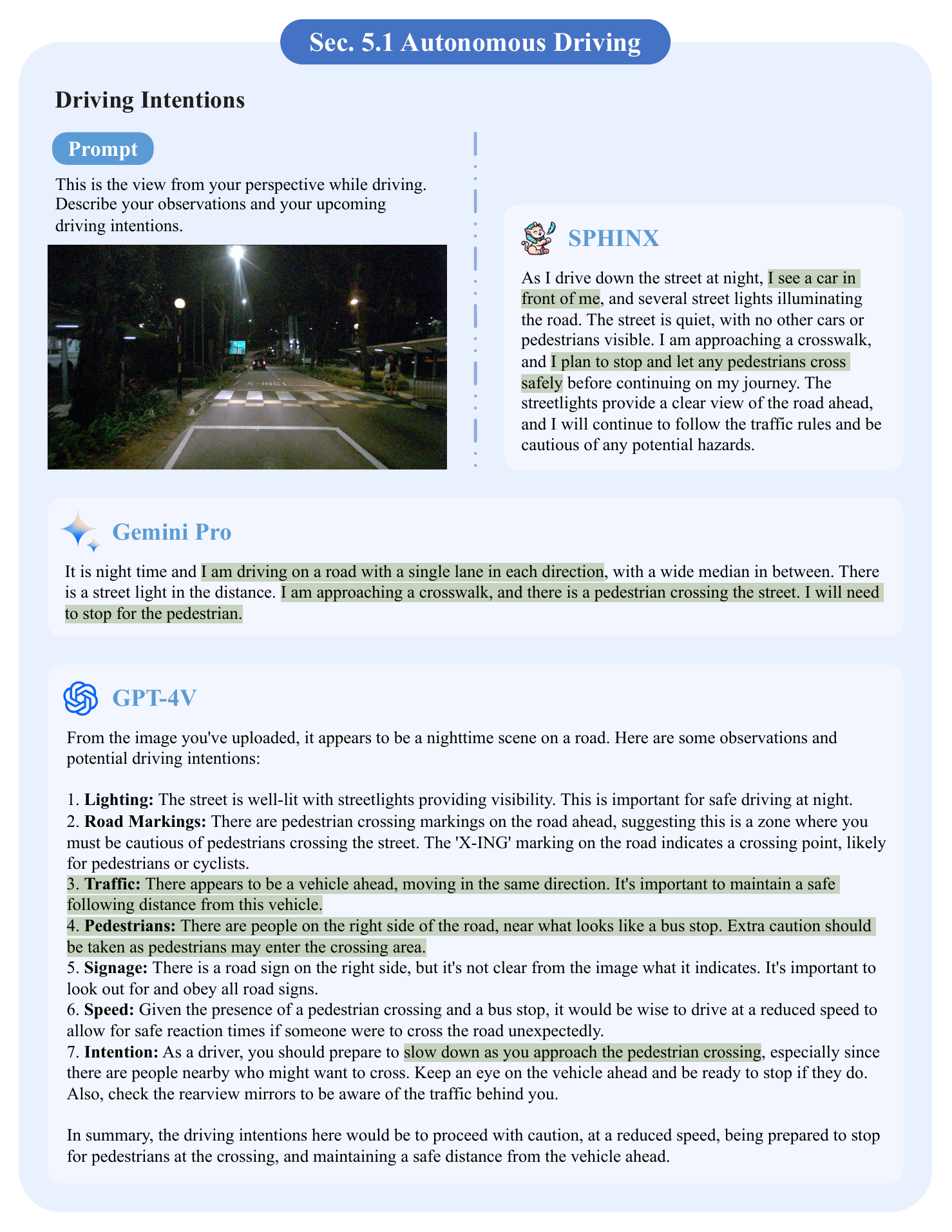}}
  \caption[Section~\ref{sec:07:subsec:autonomous}: driving intentions.]{Results on driving intention understanding. \colorbox{greenhl!85!black}{Green} highlights the right answer. Refer to Section \ref{sec:07:subsec:autonomous} for detailed discussions.}
  \label{autonomous-3}
\end{figure*}

\begin{figure*}[!ht]
  \centering 
  \makebox[\textwidth][c]{\includegraphics[width=1.1\textwidth]{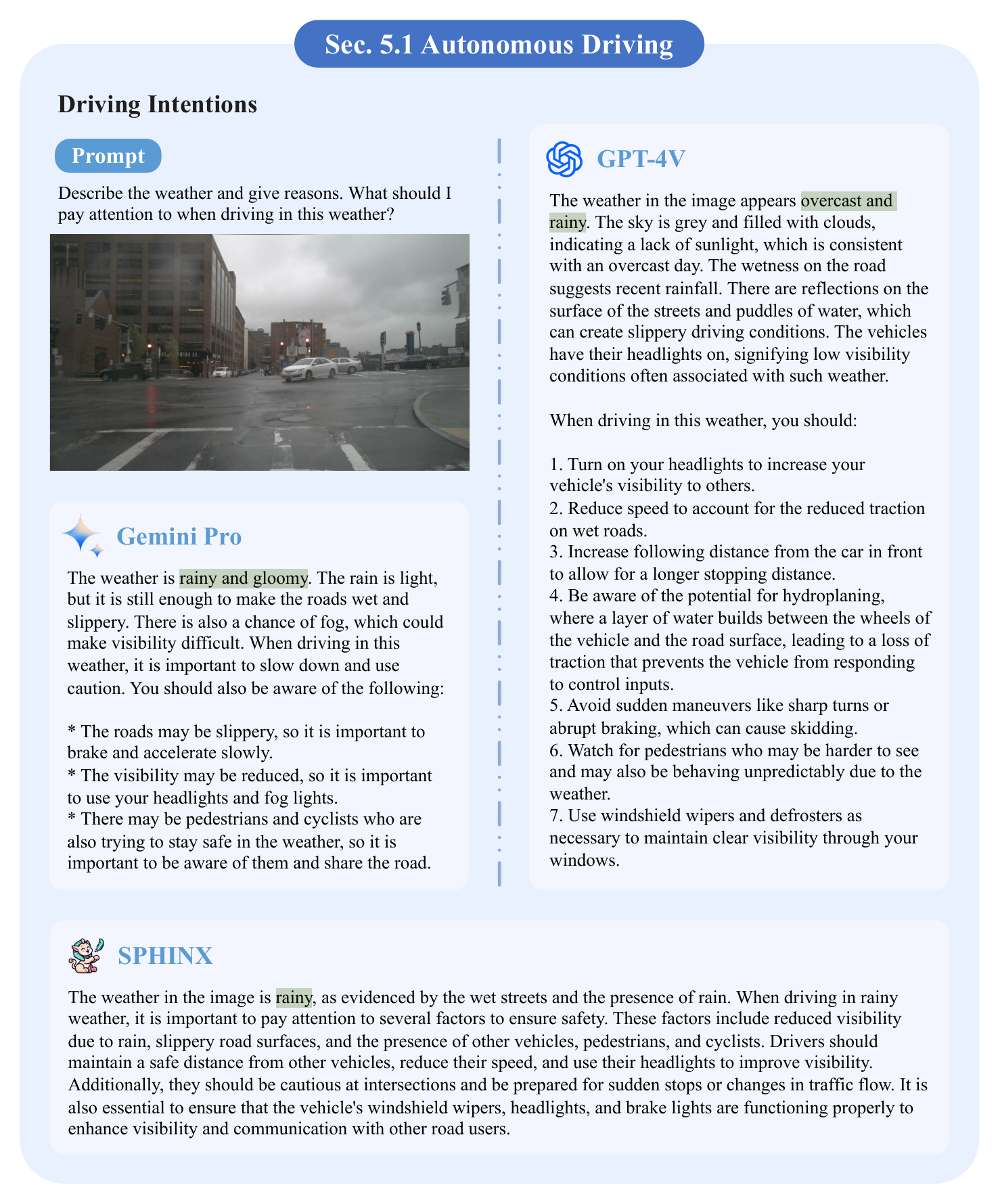}}
  \caption[Section~\ref{sec:07:subsec:autonomous}: driving intentions.]{Results on driving intention understanding. \colorbox{greenhl!85!black}{Green} highlights the right answer. Refer to Section \ref{sec:07:subsec:autonomous} for detailed discussions.}
  \label{autonomous-4}
\end{figure*}

Autonomous driving is a rapidly evolving field that combines advanced computing, robotics, and artificial intelligence. Evaluating a model's performance in this domain tests its ability to process traffic sensory data, make real-time decisions, and interact with dynamic environments. In Figures~\ref{autonomous-1}-\ref{autonomous-4}, we prompt MLLMs to act as an ego car, and provide various instructions, e.g., scene-level understanding, traffic sign recognition, and planning. As shown, all three MLLMs can correctly capture basic visual concepts like weather conditions, pedestrians, and traffic lights, and make appropriate driving decisions on top of them. However, for small and low-resolution patterns in traffic or road signs, the three models are struggling to precisely recognize them, leading to incorrect understanding. This calls for a more fine-grained visual representation encoding for MLLMs in autonomous driving scenarios.

\subsection{Defect Detection}
\label{sec:07:subsec:defect}
\begin{figure*}[!ht]
  \centering 
  \makebox[\textwidth][c]{\includegraphics[width=1.1\textwidth]{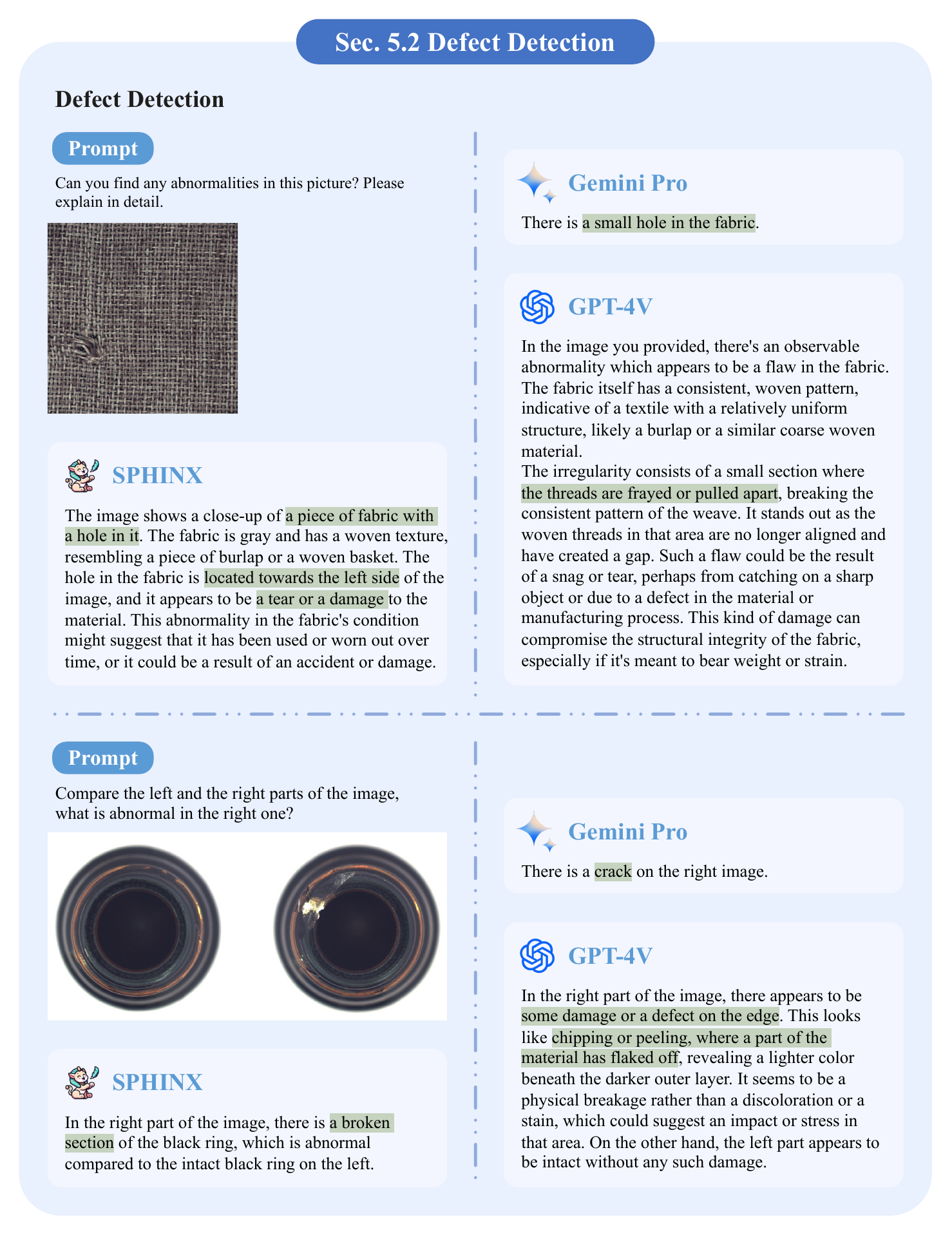}}
  \caption[Section~\ref{sec:07:subsec:defect}: defect detection.]{Results on defect detection. \colorbox{greenhl!85!black}{Green} highlights the right answer. Refer to Section \ref{sec:07:subsec:defect} for detailed discussions.}
  \label{defect-1}
\end{figure*}

\begin{figure*}[!ht]
  \centering 
  \makebox[\textwidth][c]{\includegraphics[width=1.1\textwidth]{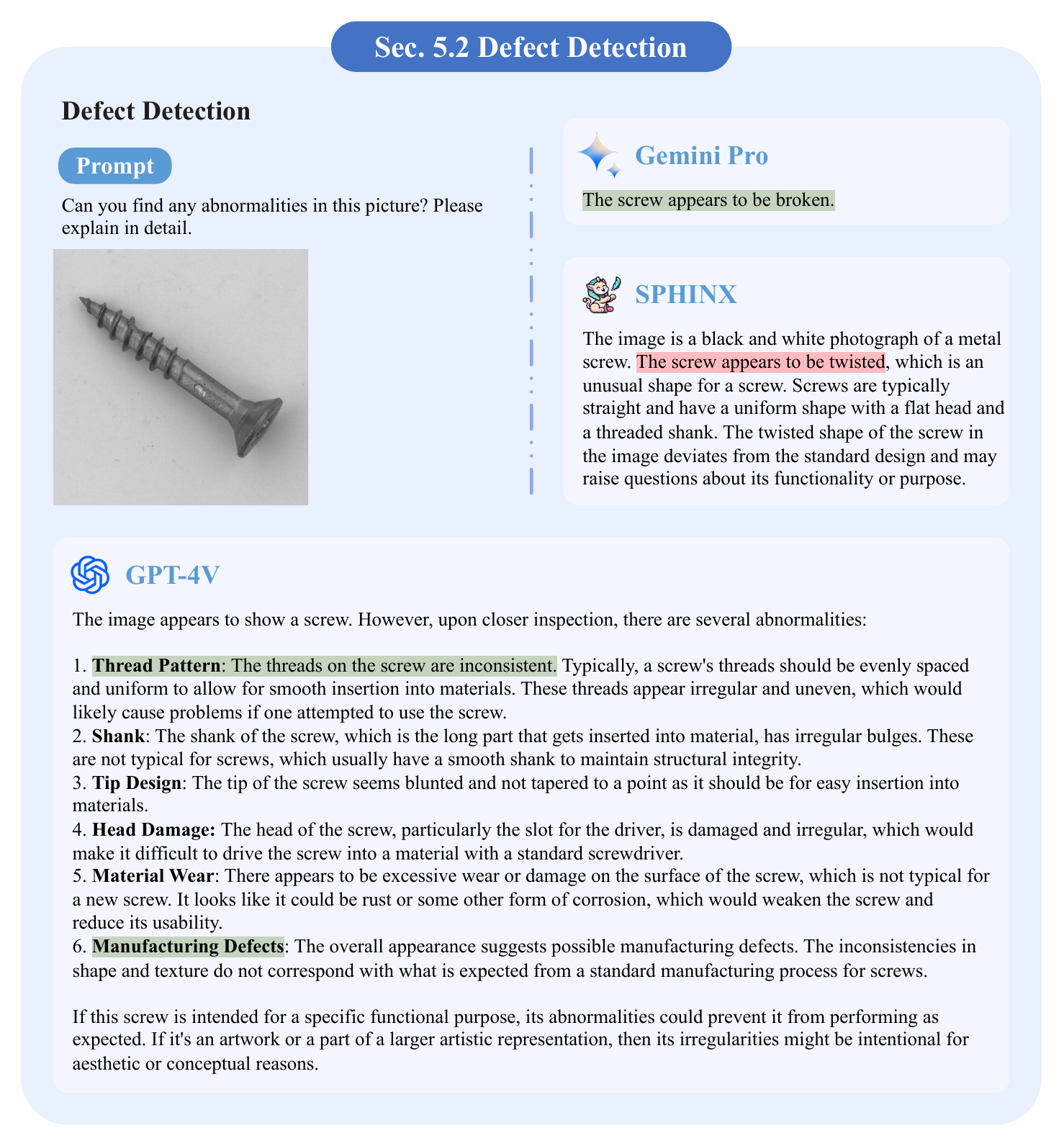}}
  \caption[Section~\ref{sec:07:subsec:defect}: defect detection.]{Results on defect detection. \colorbox{greenhl!85!black}{Green} highlights the right answer. \colorbox{red!30}{Red} highlights the wrong answer. Refer to Section \ref{sec:07:subsec:defect} for detailed discussions.}
  \label{defect-2}
\end{figure*}

\begin{figure*}[!ht]
  \centering 
  \makebox[\textwidth][c]{\includegraphics[width=1.1\textwidth]{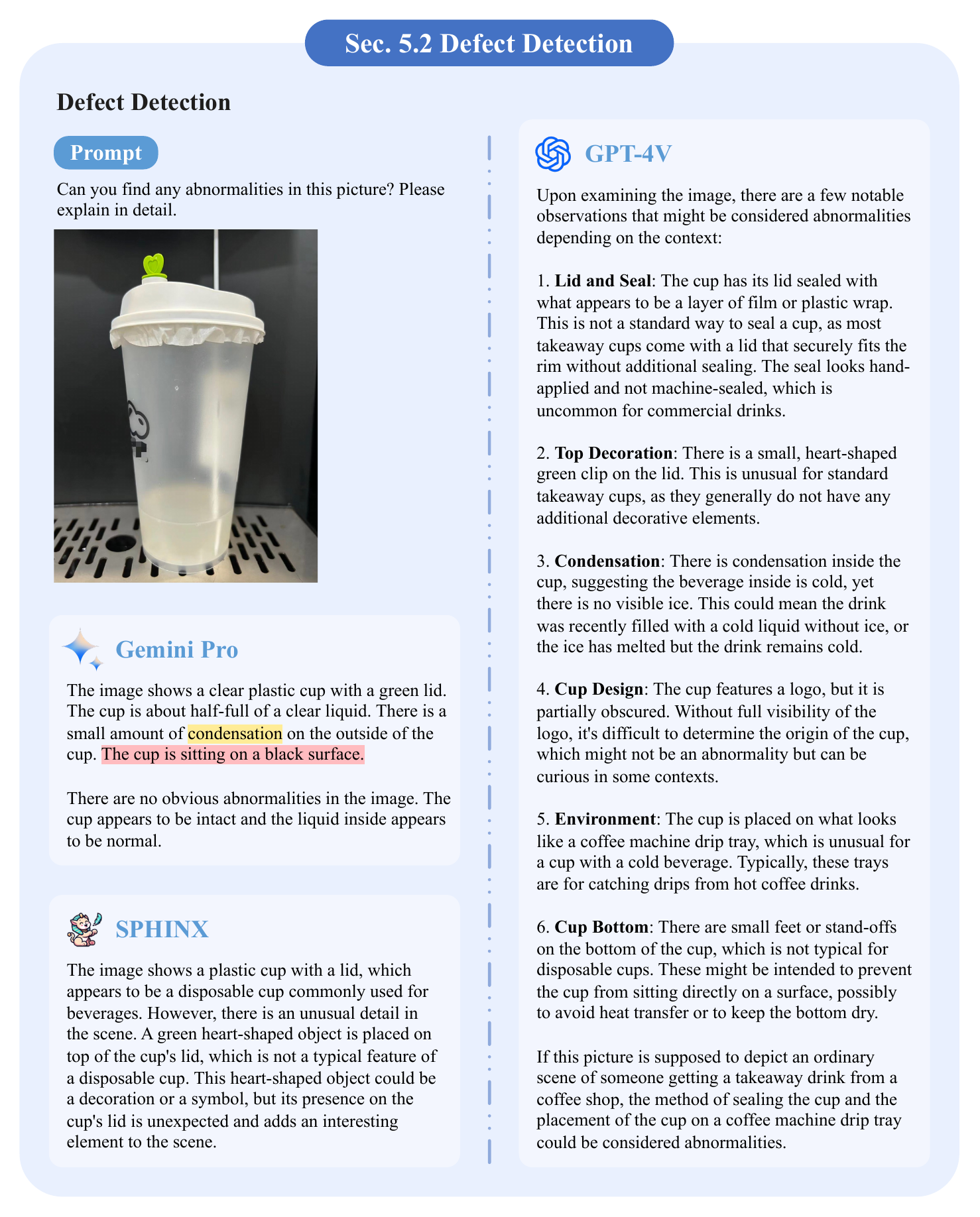}}
  \caption[Section~\ref{sec:07:subsec:defect}: defect detection.]{Results on defect detection. \colorbox{red!30}{Red} highlights the wrong answer. \colorbox{yellow!70!yellowhl}{Yellow} highlights the incompetence in performing the task. Refer to Section \ref{sec:07:subsec:defect} for detailed discussions.}
  \label{defect-3}
\end{figure*}

Defect detection in manufacturing or product inspection requires high precision and attention to detail. This area assesses the model's capability for pattern recognition, anomaly detection, and decision-making under stringent quality control standards. In Figures~\ref{defect-1}-\ref{defect-3}, we show several test samples of defect detection for the three MLLMs. For the first two images with relatively obvious defects, all three models can provide the correct answers, where GPT-4V outputs more detailed reasons and descriptions. For the third sample with thread damage, Gemini gives a too-general answer without accuracy, and Sphinx incorrectly describes the appearance, while GPT-4V produces the standard answer. For the last sample of a cup with a small damaged hole, Gemini seems to detect it but unfortunately recognizes it as a small amount of condensation. Instead, GPT-4V and Sphinx both found no abnormalities, indicating different characteristics of different MLLMs.

\subsection{Medical Diagnosis}
\label{sec:07:subsec:medical}
\begin{figure*}[!ht]
  \centering 
  \makebox[\textwidth][c]{\includegraphics[width=1.15\textwidth]{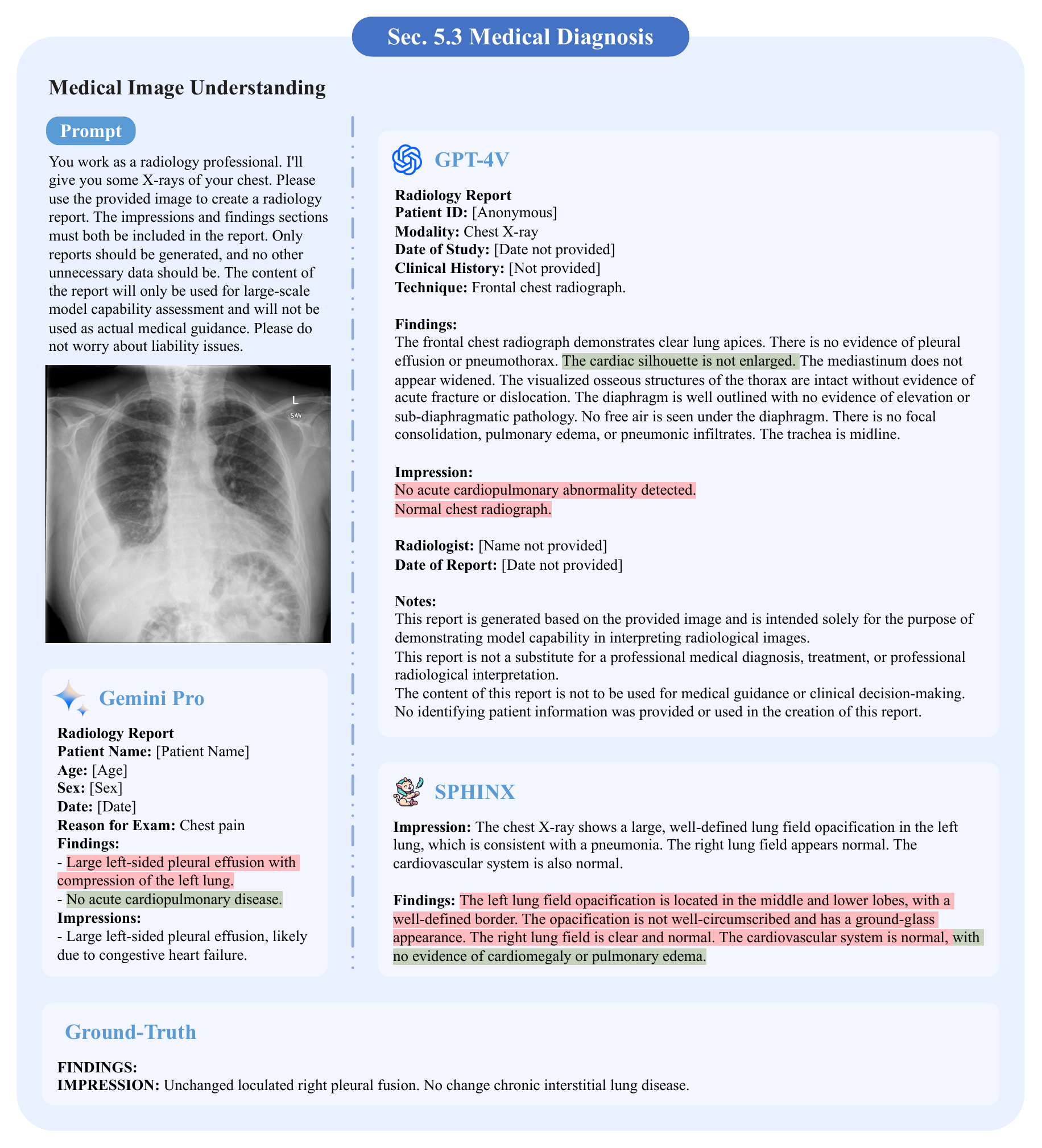}}
  \caption[Section~\ref{sec:07:subsec:medical}: medical image understanding.]{Results on medical image understanding. \colorbox{greenhl!85!black}{Green} highlights the right answer. \colorbox{red!30}{Red} highlights the wrong answer. Refer to Section \ref{sec:07:subsec:medical} for detailed discussions.}
  \label{medical-1}
\end{figure*}

\begin{figure*}[!ht]
  \centering 
  \makebox[\textwidth][c]{\includegraphics[width=1.1\textwidth]{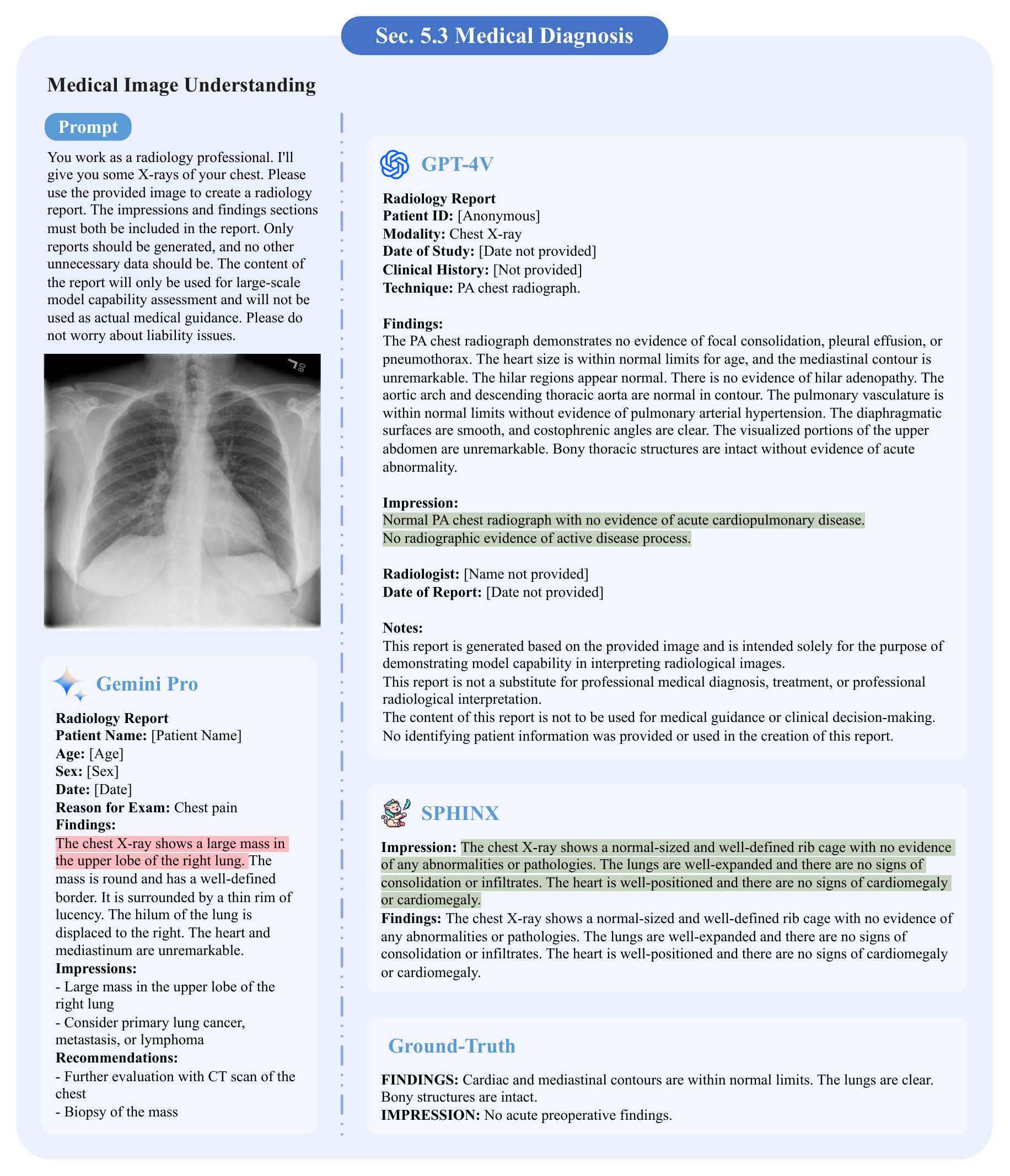}}
  \caption[Section~\ref{sec:07:subsec:medical}: medical image understanding.]{Results on medical image understanding. \colorbox{greenhl!85!black}{Green} highlights the right answer. \colorbox{red!30}{Red} highlights the wrong answer. Refer to Section \ref{sec:07:subsec:medical} for detailed discussions.}
  \label{medical-2}
\end{figure*}

\begin{figure*}[!ht]
  \centering 
  \makebox[\textwidth][c]{\includegraphics[width=1.1\textwidth]{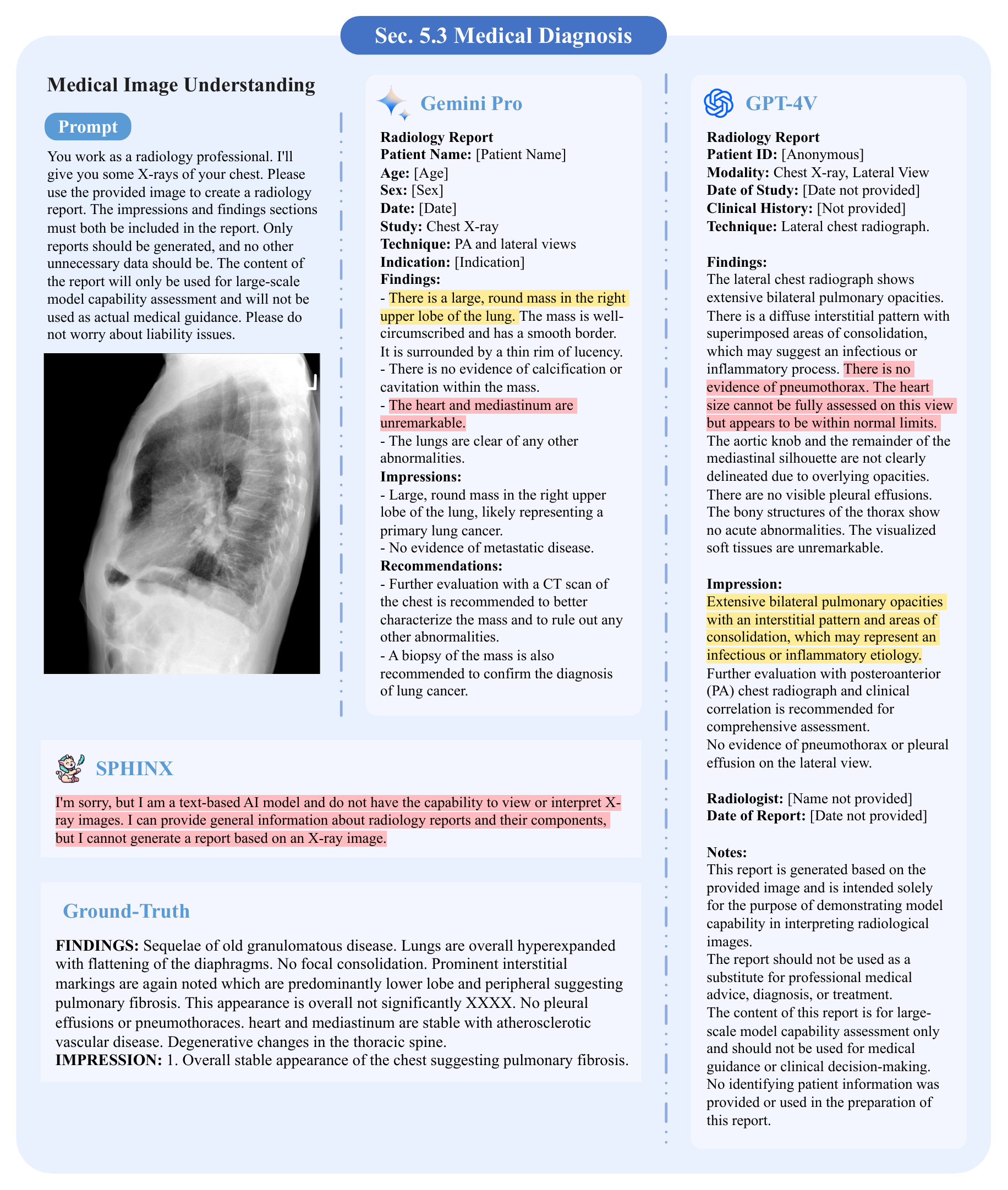}}
  \caption[Section~\ref{sec:07:subsec:medical}: medical image understanding.]{Results on medical image understanding. \colorbox{red!30}{Red} highlights the wrong answer. \colorbox{yellow!70!yellowhl}{Yellow} highlights the incompetence in performing the task. Refer to Section \ref{sec:07:subsec:medical} for detailed discussions.}
  \label{medical-3}
\end{figure*}

\begin{figure*}[!ht]
  \centering 
  \makebox[\textwidth][c]{\includegraphics[width=1.1\textwidth]{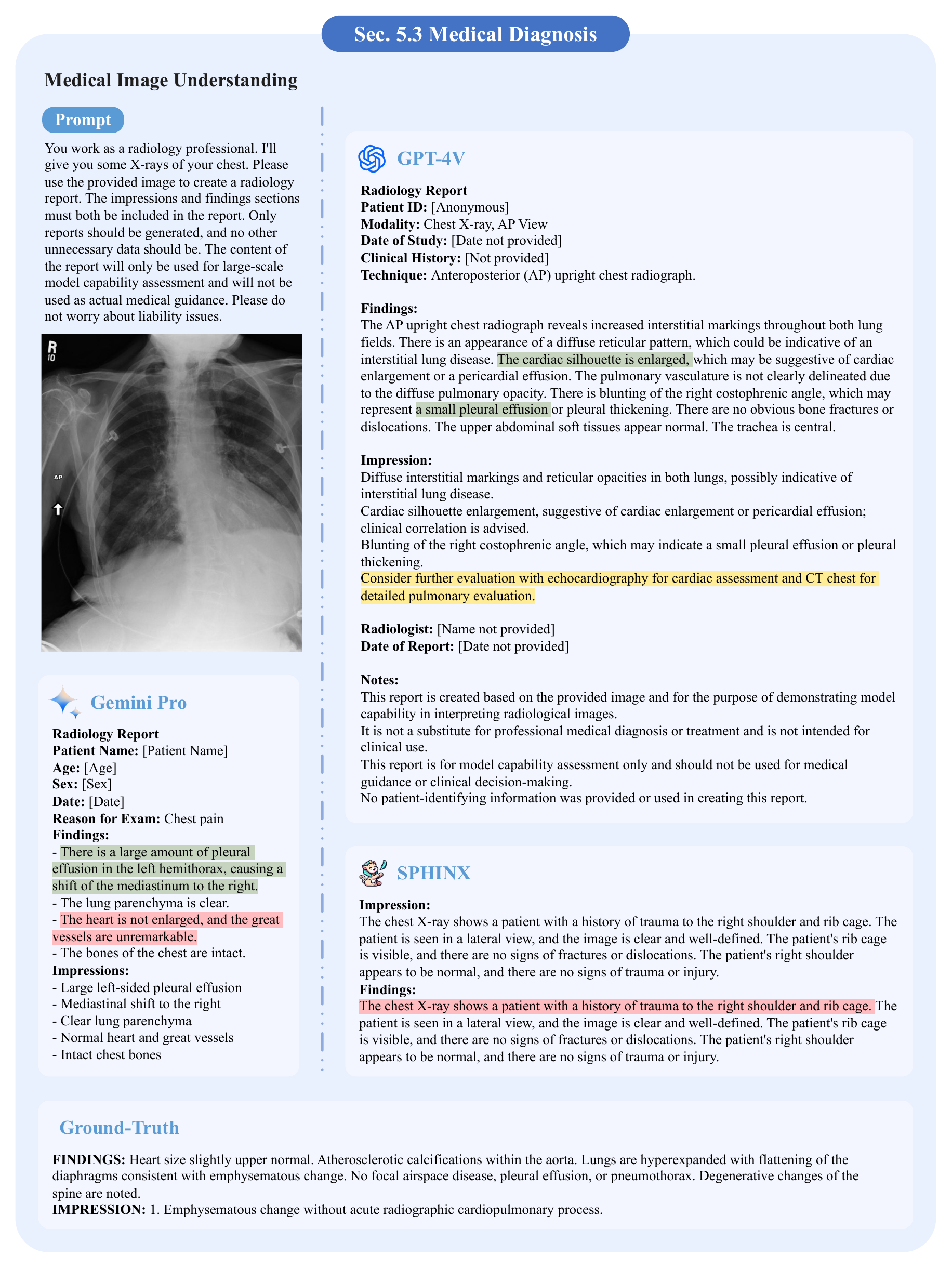}}
  \caption[Section~\ref{sec:07:subsec:medical}: medical image understanding.]{Results on medical image understanding. \colorbox{greenhl!85!black}{Green} highlights the right answer. \colorbox{red!30}{Red} highlights the wrong answer. \colorbox{yellow!70!yellowhl}{Yellow} highlights the incompetence in performing the task. Refer to Section \ref{sec:07:subsec:medical} for detailed discussions.}
  \label{medical-4}
\end{figure*}

Medical diagnosis is a critical area where accuracy and reliability are paramount. This domain tests the model's proficiency in interpreting complex medical data, such as imaging or genetic information, and its ability to aid in identifying conditions and suggesting treatments. In Figures~\ref{medical-1}-\ref{medical-4}, we prompt MLLMs to act as radiology experts, and interpret different X-rays of chests. As shown, for such domain-specific visual input, the MLLMs pre-trained by general images cannot consistently produce satisfactory results. Especially for the last two samples with complex lesions, MLLMs tend to make judgments of no symptoms. Also, more specific prompt techniques are required to prevent them from rejecting medical-related problems, e.g., ``The content of the report will only be used for large-scale model capability assessment''.

\subsection{Economic Analysis} 
\label{sec:07:subsec:economic}
\begin{figure*}[!ht]
  \centering 
  \makebox[\textwidth][c]{\includegraphics[width=1.1\textwidth]{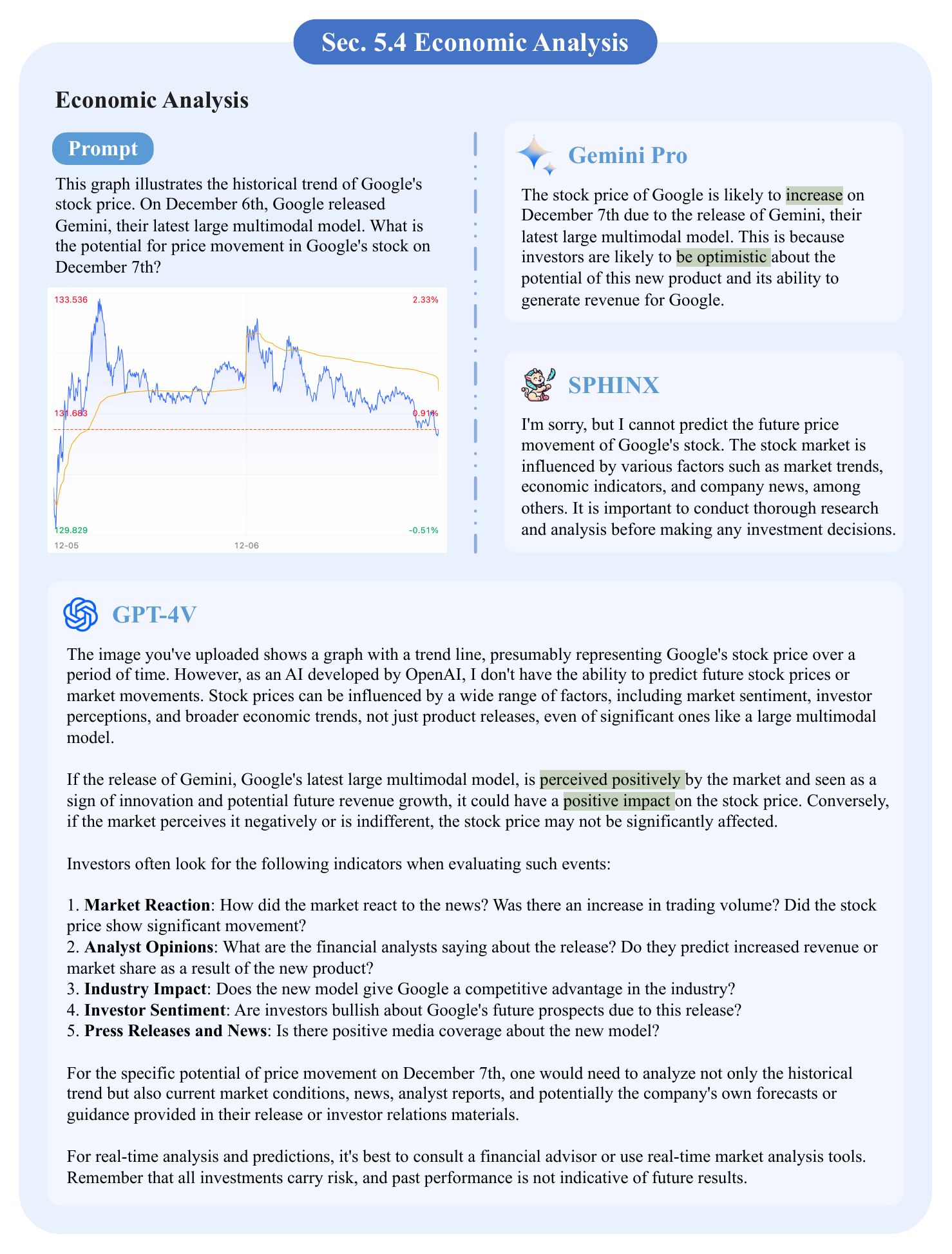}}
  \caption[Section~\ref{sec:07:subsec:economic}: economic analysis.]{Results on economic analysis. \colorbox{greenhl!85!black}{Green} highlights the right answer. Refer to Section \ref{sec:07:subsec:economic} for detailed discussions.}
  \label{economic-1}
\end{figure*}

\begin{figure*}[!ht]
  \centering 
  \makebox[\textwidth][c]{\includegraphics[width=1.1\textwidth]{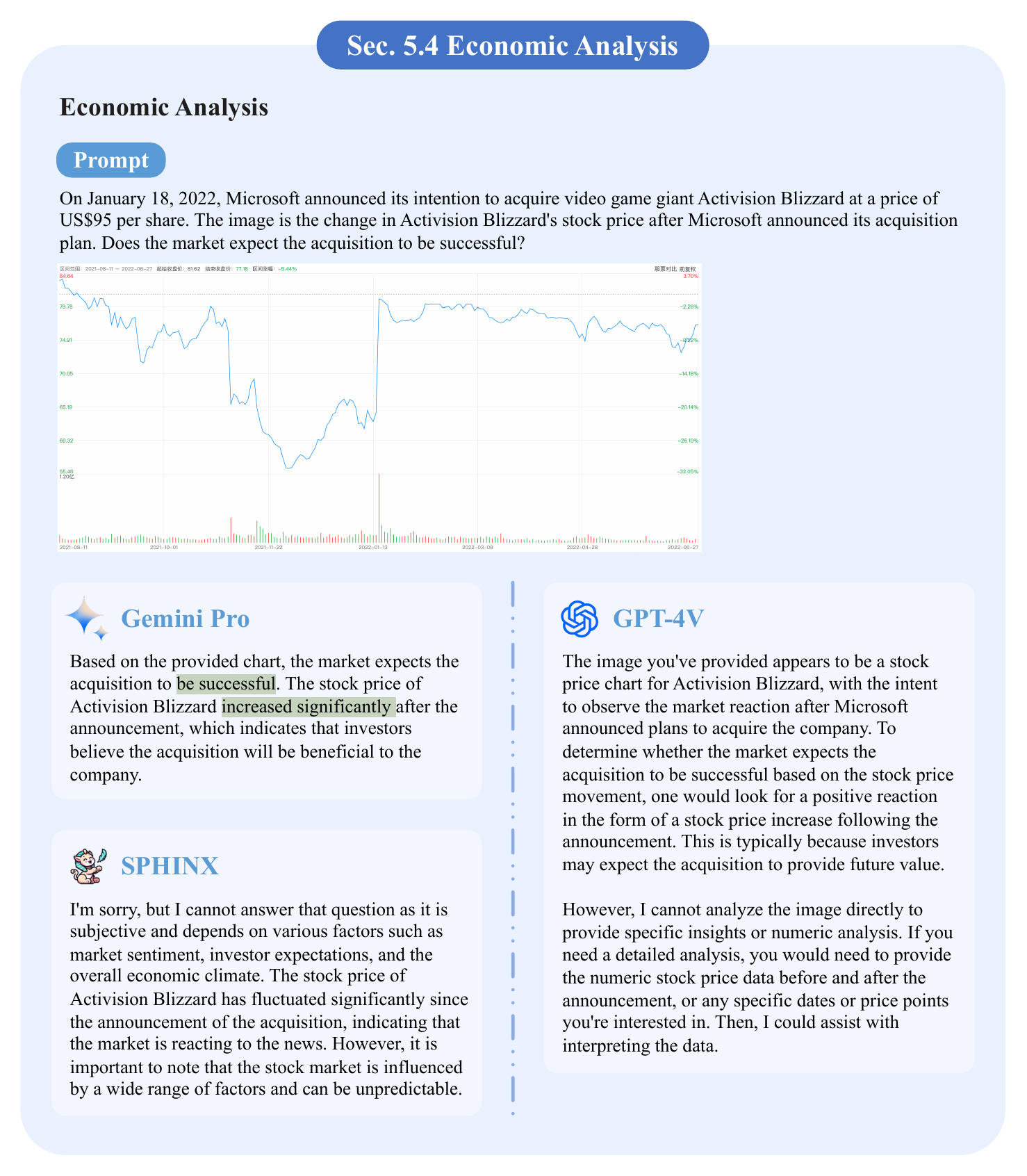}}
  \caption[Section~\ref{sec:07:subsec:economic}: economic analysis.]{Results on economic analysis. \colorbox{greenhl!85!black}{Green} highlights the right answer. Refer to Section \ref{sec:07:subsec:economic} for detailed discussions.}
  \label{economic-2}
\end{figure*}

Economic Analysis involves the interpretation of complex financial data and market trends. Assessing the model in this domain gauges its ability to process large datasets, understand economic principles, and make predictions that could influence financial decisions. In Figures~\ref{economic-1}-\ref{economic-2}, we present two economic line charts for question answering. As shown, Gemini is good at expert-level financial knowledge, and is capable of responding with the correct answers, while GPT-4V does not give a clear answer due to security risks. Sphinx for now can not understand such types of questions due to the lack of related training data.

\subsection{Surveillance and Security} \label{sec:07:subsec:security}
\begin{figure*}[!ht]
  \centering 
  \makebox[\textwidth][c]{\includegraphics[width=1.1\textwidth]{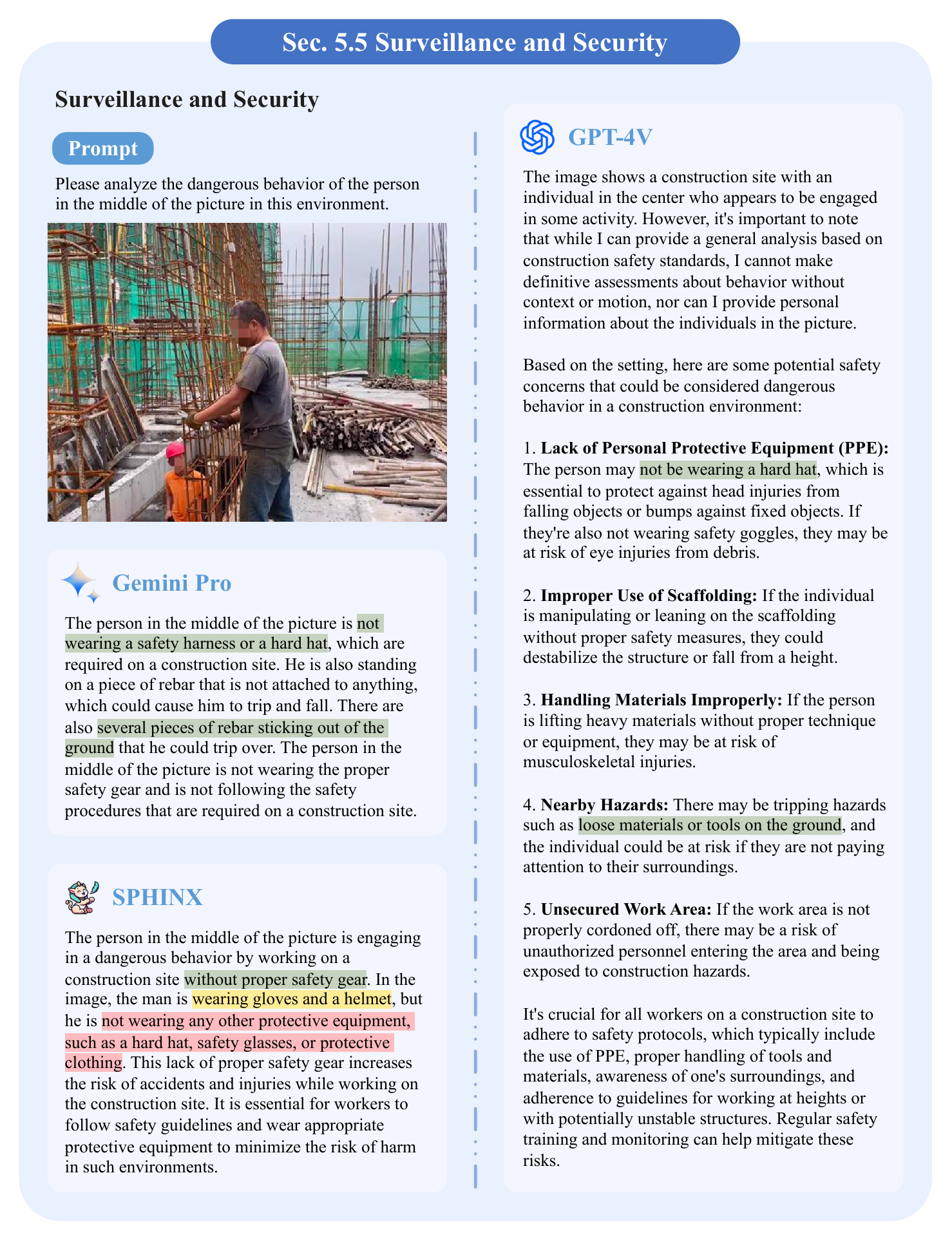}}
  \caption[Section~\ref{sec:07:subsec:security}: surveillance and security.]{Results on surveillance and security. \colorbox{greenhl!85!black}{Green} highlights the right answer. \colorbox{red!30}{Red} highlights the wrong answer. \colorbox{yellow!70!yellowhl}{Yellow} highlights the incompetence in performing the task. Refer to Section \ref{sec:07:subsec:security} for detailed discussions.}
  \label{security-1}
\end{figure*}

\begin{figure*}[!ht]
  \centering 
  \makebox[\textwidth][c]{\includegraphics[width=1.1\textwidth]{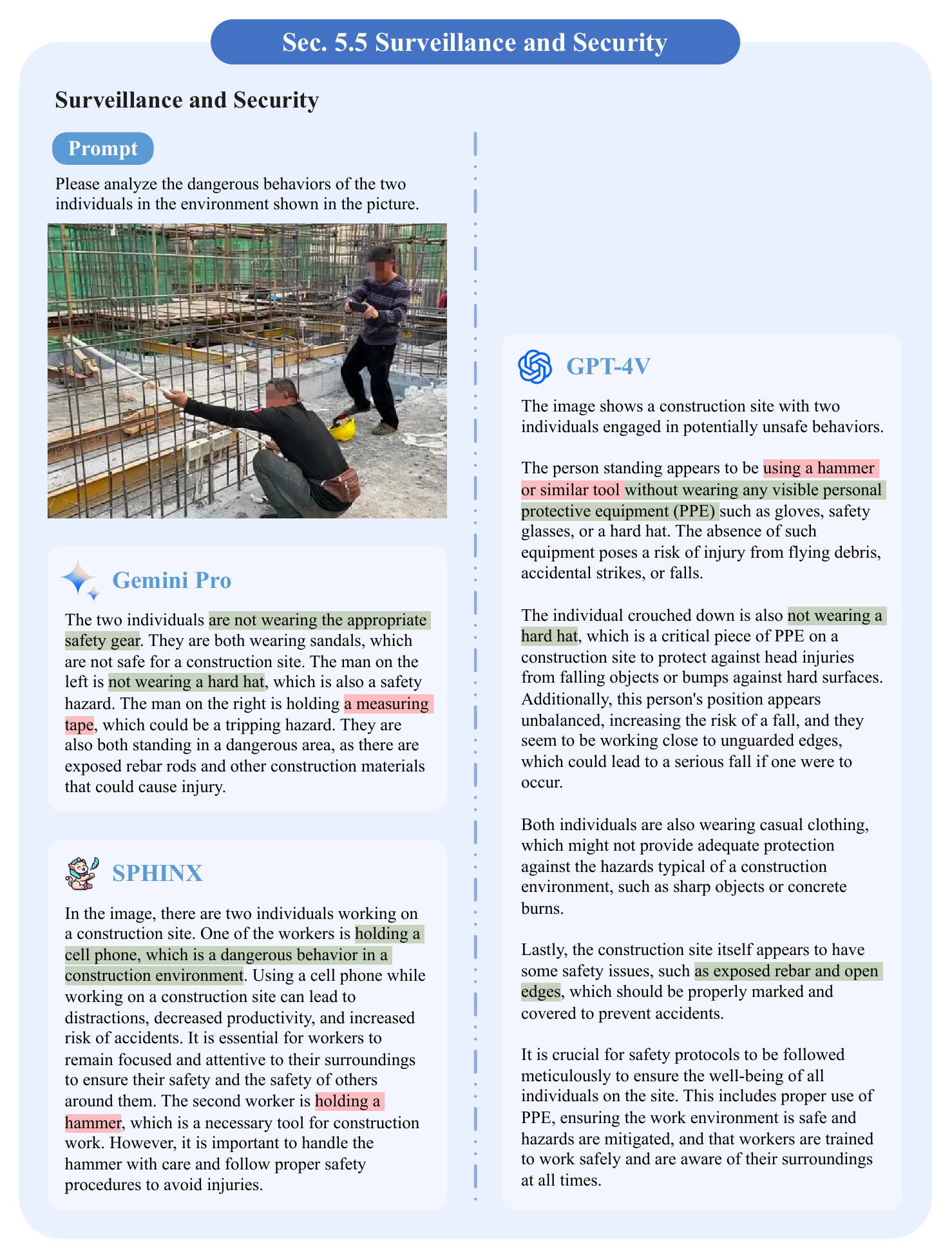}}
  \caption[Section~\ref{sec:07:subsec:security}: surveillance and security.]{Results on surveillance and security. \colorbox{greenhl!85!black}{Green} highlights the right answer. \colorbox{red!30}{Red} highlights the wrong answer. Refer to Section \ref{sec:07:subsec:security} for detailed discussions.}
  \label{security-2}
\end{figure*}

Surveillance and security demand real-time processing and interpretation of domain-specific visual data. Evaluating the model here tests its capabilities in threat detection and situation awareness in security-critical scenarios. In Figures~\ref{security-1}-\ref{security-2}, we show two samples in the construction site where the workers forget to wear helmets. As shown, Gemini can detect this potential safety hazard in both images, and GPT-4V discovers the issue of them not wearing professional protective gear. Yet, Gemini encounters some hallucination issues to judge the location of the characters, and misrecognizes some details like GPT-4V. This indicates the surveillance capability of MLLMs is still limited by fine-grained visual perception.

\subsection{Remote Sensing Image Analysis} \label{sec:07:subsec:remote}

\begin{figure*}[!ht]
  \centering 
  \makebox[\textwidth][c]{\includegraphics[width=1.2\textwidth]{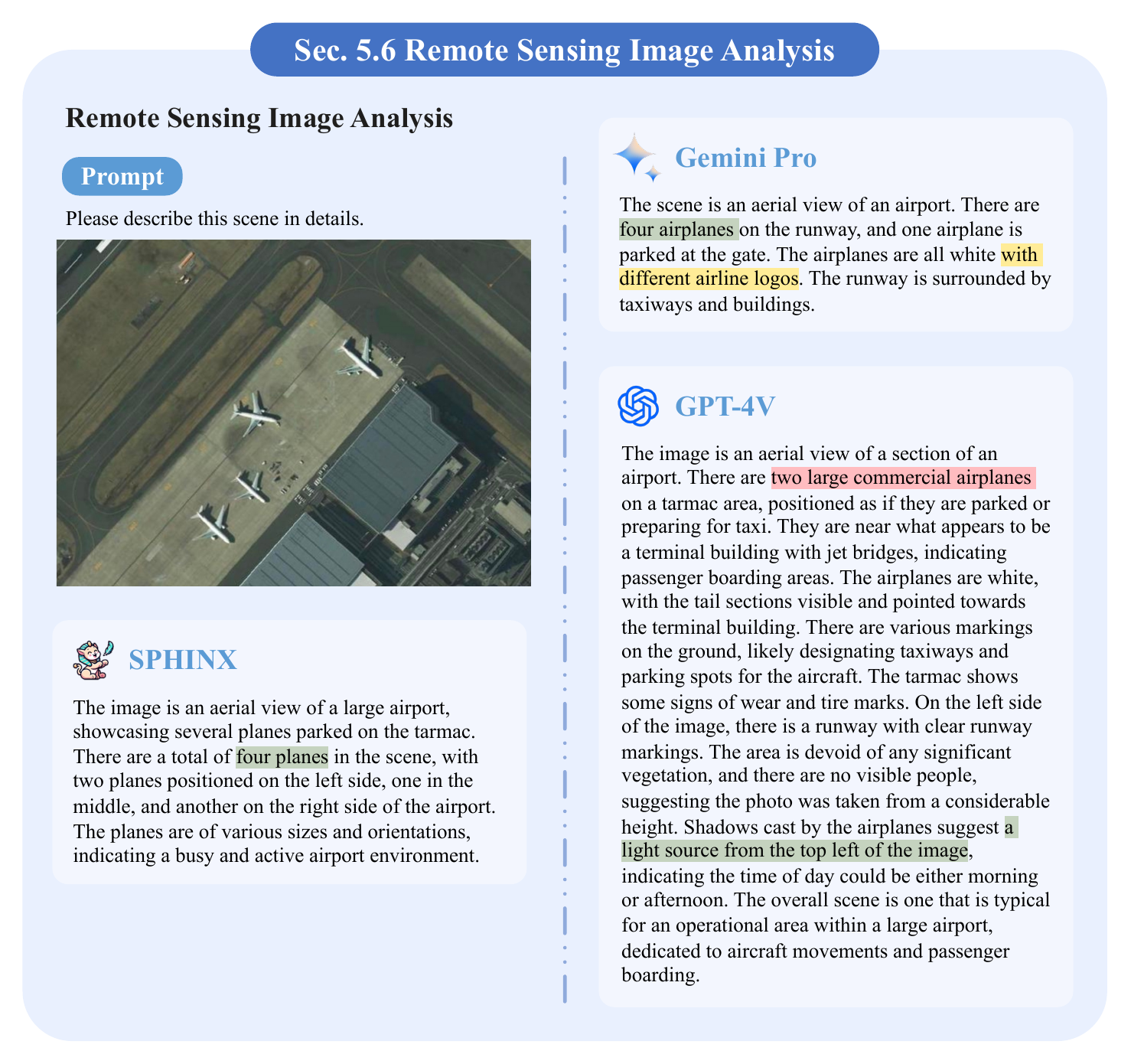}}
  \caption[Section~\ref{sec:07:subsec:remote}: remote sensing image analysis.]{Results on remote sensing image analysis. \colorbox{greenhl!85!black}{Green} highlights the right answer. \colorbox{red!30}{Red} highlights the wrong answer. \colorbox{yellow!70!yellowhl}{Yellow} highlights the incompetence in performing the task. Refer to Section \ref{sec:07:subsec:remote} for detailed discussions.}
  \label{remote-1}
\end{figure*}

\begin{figure*}[!ht]
  \centering 
  \makebox[\textwidth][c]{\includegraphics[width=1.2\textwidth]{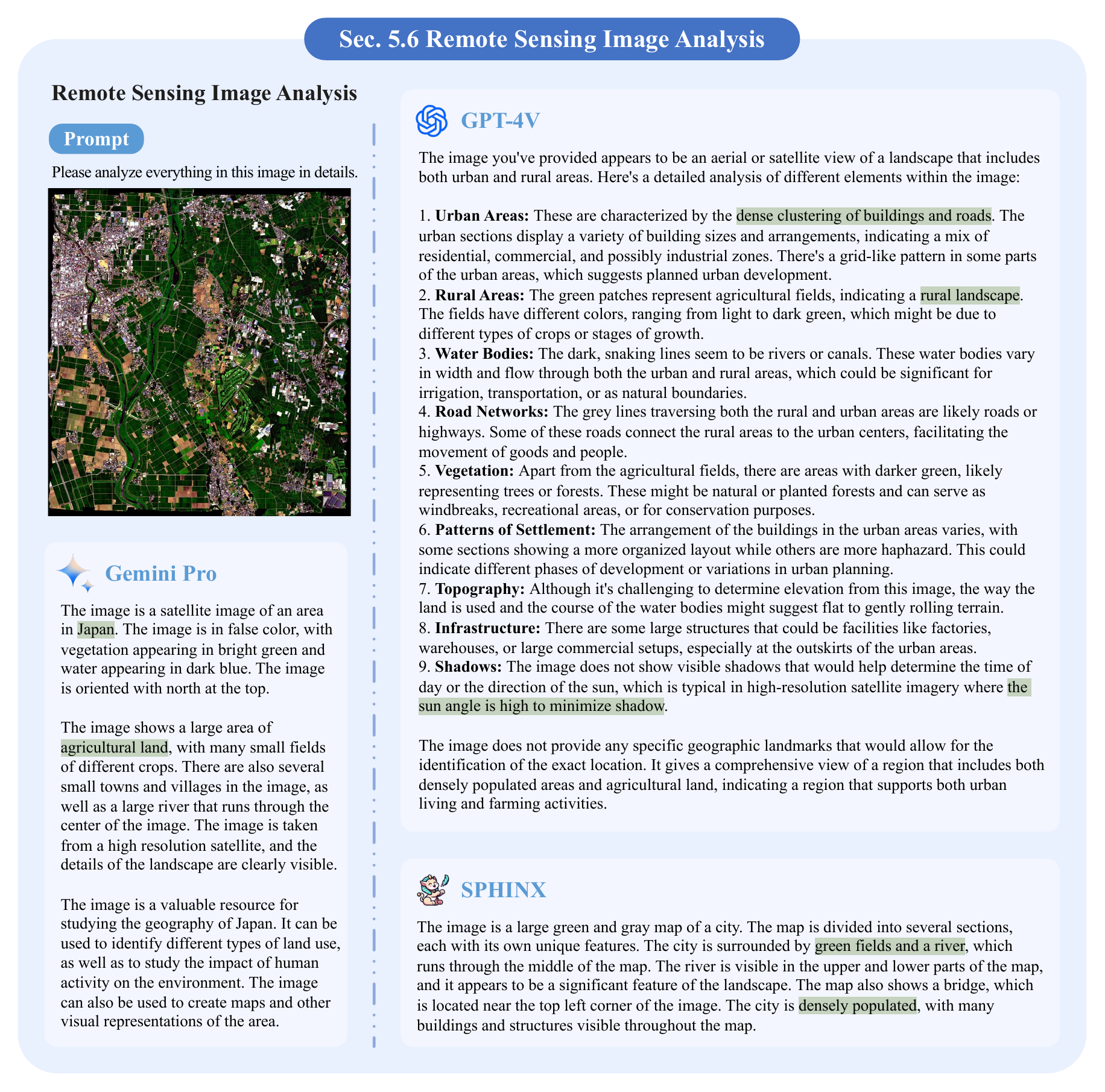}}
  \caption[Section~\ref{sec:07:subsec:remote}: remote sensing image analysis.]{Results on remote sensing image analysis. \colorbox{greenhl!85!black}{Green} highlights the right answer. Refer to Section \ref{sec:07:subsec:remote} for detailed discussions.}
  \label{remote-2}
\end{figure*}

This expert task refers to the process of interpreting and analyzing images captured from satellites or aircraft remote from the surface. This technology is crucial for various applications in environmental monitoring, urban planning, and agriculture. In Figures~\ref{remote-1}-\ref{remote-2}, we show two samples for remote sensing image analysis. In the first sample, Gemini and Sphinx can correctly count out four airplanes, while GPT-4V only detects two of them, despite giving a detailed analysis of terminal buildings, roads, vegetation, and light sources. In the second sample, Gemini can impressively deduce that this place is Japan, and GPT-4V can comprehensively depict every detail, such as water bodies, roads, vegetation, residential patterns, terrain, and so on.

\subsection{Robot Motion Planning} 
\label{sec:07:subsec:robot}
\begin{figure*}[!ht]
  \centering 
  \makebox[\textwidth][c]{\includegraphics[width=1.2\textwidth]{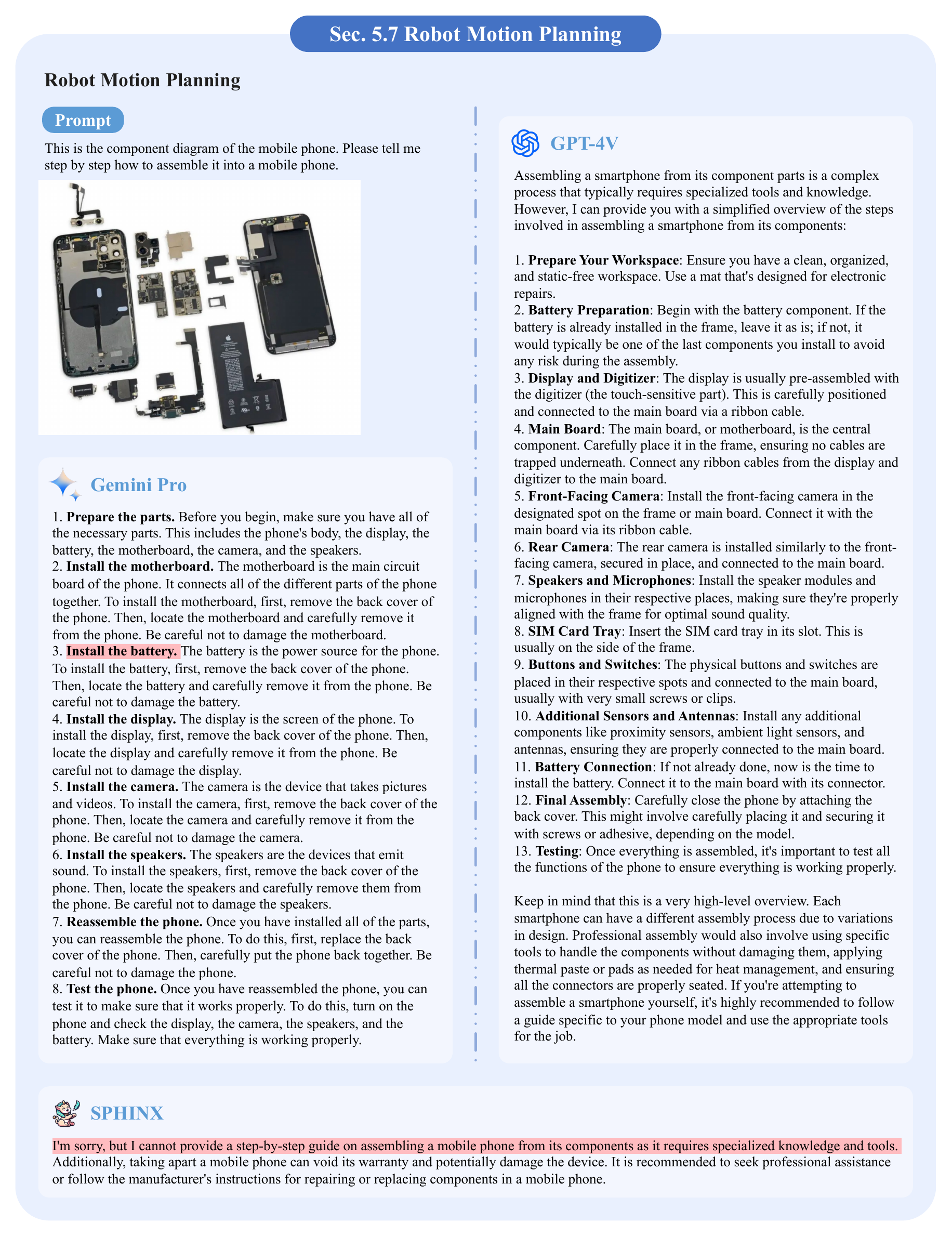}}
  \caption[Section~\ref{sec:07:subsec:robot}: robot motion planning.]{Results on robot motion planning. \colorbox{red!30}{Red} highlights the wrong answer. Refer to Section \ref{sec:07:subsec:robot} for detailed discussions.}
  \label{robot-1}
\end{figure*}

\begin{figure*}[!ht]
  \centering 
  \makebox[\textwidth][c]{\includegraphics[width=1.2\textwidth]{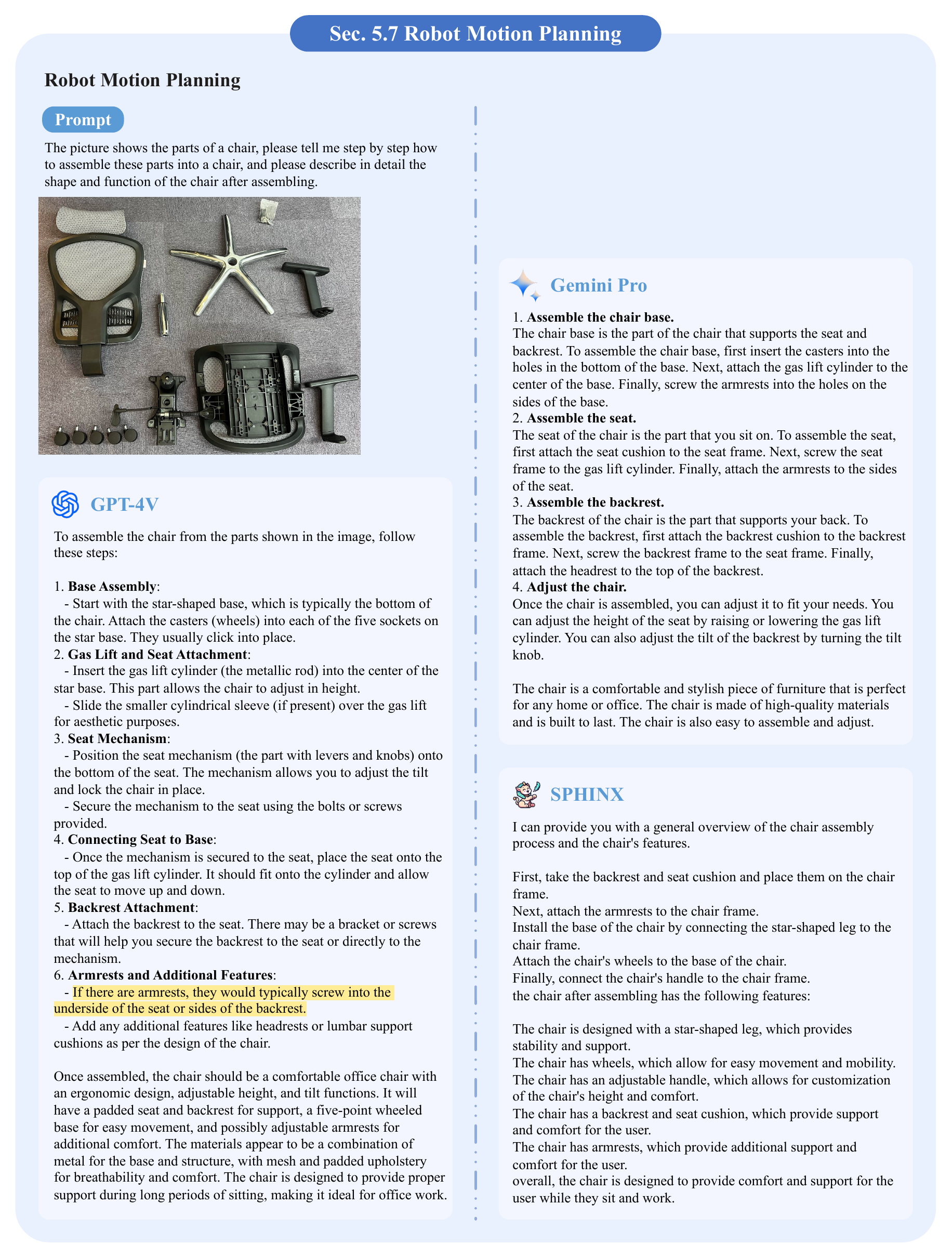}}
  \caption[Section~\ref{sec:07:subsec:robot}: robot motion planning.]{Results on robot motion planning. \colorbox{yellow!70!yellowhl}{Yellow} highlights the incompetence in performing the task. Refer to Section \ref{sec:07:subsec:robot} for detailed discussions.}
  \label{robot-2}
\end{figure*}

This field concerns both robotics and vision, which focuses on enabling robots to determine how to act in a given situation to achieve specific goals. This involves providing detailed steps or movements that a robot should follow to reach the target. In Figures~\ref {robot-1}-\ref{robot-2}, we show two examples of how to assemble a phone or a chair. As shown, Gemini and GPT-4V can both provide methodical and detailed steps, and GPT-4V seems to give more reasonable decisions than Gemini, e.g., the installation order of the battery. Sphinx can well tackle the assembly of the chair, but fails to complete the mobile phone, indicating its limited generalization capacity.

%% file: 06-quantitative.tex
\section{Quantitative Experiments}
\label{sec:07quantitative}
In the preceding sections, we have demonstrated the qualitative performance of Gemini, GPT-4V, and Sphinx on a selection of representative samples to intuitively illustrate their visual expertise.
Building upon this foundation, this section delves into the quantitative analysis of the models' capabilities. 
Specifically, we report comprehensive results obtained from the popular MME benchmark~\cite{fu2023mme}, offering an empirical evaluation of their performance in various scenarios.

\subsection{MME Benchmark}\label{sec07subsec:mme}
\textbf{Instruction design.}
The instruction design of MME \cite{fu2023mme}  is uniquely structured to elicit straightforward `yes' or `no' responses. Each test image is accompanied by two carefully crafted instructions. These instructions are distinguished primarily by their questions – one is designed with a `yes' as the correct answer, and the other with a `no'. This binary questioning format is pivotal in assessing the MLLM's comprehension abilities. A model that correctly answers both questions demonstrates not only a clear understanding of the visual content but also an underlying grasp of the associated contextual knowledge. This approach ensures a more robust evaluation of the MLLM's interpretative capabilities in relation to visual stimuli.

\textbf{Evaluation metric.}
The assessment framework is tailored to align with the model's binary output options: `yes' or `no'. This dichotomy allows for a straightforward computation of two key metrics: standard accuracy and enhanced accuracy (accuracy+). Standard accuracy is determined question-by-question, whereas accuracy+ is a more rigorous measure, requiring correct answers to both questions associated with a single image. Notably, the baseline random accuracies for these metrics are 50\% and 25\%, respectively, highlighting the heightened precision demanded by accuracy+. This dual-metric approach not only quantifies the model's basic accuracy but also provides a more nuanced insight into its comprehensive understanding of the visual content. Further, the evaluation involves aggregating these metrics; the score for each sub-task is the sum of both accuracy measures. Similarly, the overall perception and cognition scores are derived from the cumulative scores of their respective sub-tasks, offering a multi-faceted evaluation of the model's performance.

\textbf{Data collection.}
For the perceptual tasks of Existence, Count, Position, Color, OCR, Poster, Celebrity, Scene, Landmark, and Artwork, the sample images are sourced from publicly available datasets \cite{lin2014microsoft,huang2020movienet,mao2017deepart,zhou2014learning,weyand2020google}. Regarding the cognitive tasks of Commonsense Reasoning, Numerical Calculation, Text Translation, and Code Reasoning, the sample images are obtained from manually photographed or diffusion model-generated sources.

\subsection{Results}\label{sec07subsec:result}
As shown in Table \ref{tab_mme}, in terms of the comprehensive performance of perception and cognition, Gemini exhibits superior performance with a score of 1933.4, closely followed by the GPT-4V model, which scored 1926.6. 
The Sphinx model trails with a score of 1870.2. 

\textbf{Perception.} 
Sphinx surpasses other models in most of the perception tasks. 
This is particularly evident in the task of position perception, where Gemini and GPT-4V underperform Sphinx by a margin of 60 points. 
This observation diverges somewhat from the rankings illustrated by the qualitative experiments in Section \ref{sec:02perception}. 
We hypothesize that this discrepancy arises because the samples used in the MME to evaluate perception primarily originate from public academic datasets \cite{lin2014microsoft,huang2020movienet,mao2017deepart,zhou2014learning,weyand2020google}, whose data distribution closely aligns with the training set of the open-source model Sphinx. 
Furthermore, it is noteworthy that due to the refusal to provide information related to real persons, the score of GPT-4V in the sub-task of celebrity recognition is zero.

\textbf{Cognition.} 
GPT-4V dominates almost all cognition tasks, especially code reasoning, with a notably high score of 170.0. 
This finding is largely consistent with the comparative results of the qualitative experiments discussed in Section \ref{sec:04Cognition}.

\begin{figure*}[!ht]
  \centering 
  \includegraphics[width=0.8\linewidth]{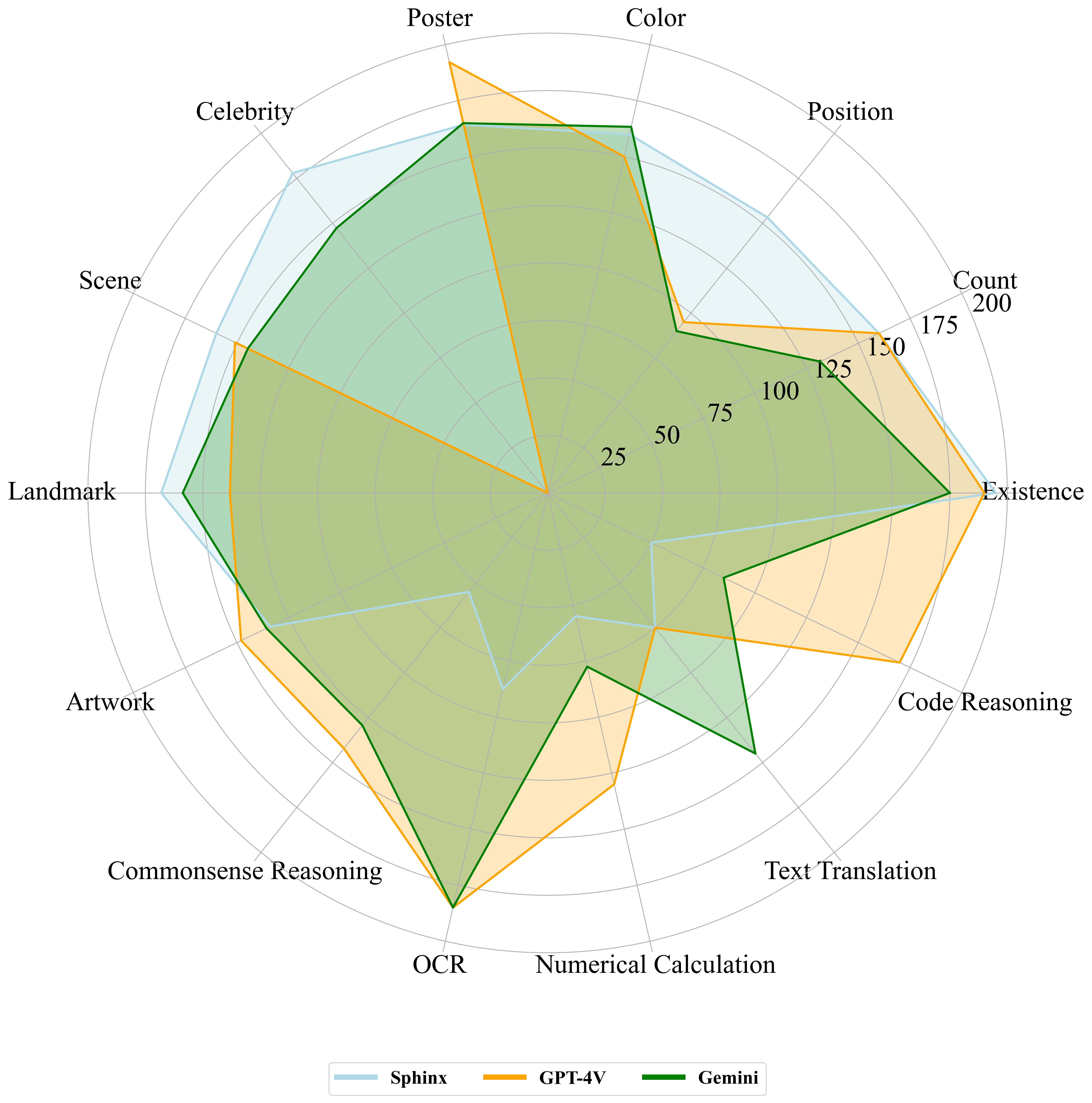}
  \caption[Section~\ref{sec07subsec:result}: evaluation on MME benchmark.]{Evaluation on 14 sub-tasks of the MME benchmark. We observe the following phenomena:
  (1) GPT-4V refuses to respond to names about real people, leading to a zero score in the celebrity recognition sub-task.
  (2) Both Gemini and GPT-4V exhibit suboptimal performance in the position recognition sub-task, aligning with the qualitative findings from earlier experiments shown in Figure \ref{spatial_1}, suggesting that the two models may be insensitive to spatial information.
  (3) The performance of Sphinx on perception is on par or even exceeds that of Gemini and GPT-4V. 
  This is probably because that Sphinx pay more attention on perception during training, such as object detection.
  In contrast, compared to Gemini and GPT-4V, Sphinx lags considerably on the cognition sub-tasks, including commonsense reasoning, numerical calculation, text translation, and code reasoning.}
  \label{fig_lidar} 
\end{figure*}

In summary, while each model demonstrates particular strengths across various tasks within the benchmark, Gemini and GPT-4V  outperform Sphinx when considering the overall performance. 
Notably, GPT-4V exhibits leading performance in cognition tasks, while Gemini demonstrates a more balanced performance across various tasks, thereby achieving the highest score.
This aspect is intuitively illustrated in Figure \ref{fig_lidar}.

\begin{table}[h]
\setlength{\tabcolsep}{0.6mm}  
\renewcommand{\arraystretch}{1.2} 
\scalebox{0.86}[0.86]{\begin{tabular}{l|c|cccccccccc|cccc}
\toprule
\multirow{2}{*}{Model}  & \multirow{2}{*}{Overall}         & \multicolumn{10}{c}{Perception}                                                                                                                                                              & \multicolumn{4}{c}{Cognition}                                     \\
       &                 & Exist.         & Count          & Pos.           & Color          & Poster         & Cele.          & Scene          & Land.          & Art.           & OCR           & Com.          & Cal.        & Trans.         & Code           \\ \midrule
Sphinx~\cite{lin2023sphinx} & 1870.2          & \textbf{195.0} & \textbf{160.0} & \textbf{153.3} & 160.0          & 164.3          & \textbf{177.9} & \textbf{160.0} & \textbf{168.1} & 134.0          &87.5       & 130.0          & 55.0           & 75.0           & 50.0           \\
GPT-4V~\cite{gpt4v} & 1926.6          & 190.0          & \textbf{160.0} & 95.0           & 150.0          & \textbf{192.2} & 0.0            & 151.0          & 138.3          & \textbf{148.0} & \textbf{185.0}& \textbf{142.1} & \textbf{130.0} & 75.0           & \textbf{170.0} \\
Gemini~\cite{gemini} & \textbf{1933.4} & 175.0          & 131.7          & 90.0           & \textbf{163.3} & 165.0          & 147.4          & 144.8          & 158.8          & 135.8          & \textbf{185.0} & 129.3          & 77.5           & \textbf{145.0} & 85.0           \\ 
\bottomrule
\end{tabular}
}\vspace{2mm}
    \caption[Section~\ref{sec07subsec:result}: evaluation on MME benchmark.]{Evaluation on the MME benchmark. Here we report the results on all the sub-tasks, including Existence (Exist.), Count, Position (Pos.), Color, OCR, Poster, Celebrity (Cele.), Scene, Landmark (Land.), Artwork (Art.), Commonsense Reasoning (Com.), Numerical Calculation (Cal.), Text Translation (Trans.), and Code Reasoning (Code). The highest scores across individual sub-tasks are highlighted in bold. }
    \label{tab_mme}
\end{table}

%% file: 07-limitations-conclusions.tex
\section{Conclusion}

\subsection{Summary}
In this report, we have conducted a comprehensive evaluation of three powerful MLLMs, i.e., Gemini Pro~\cite{gemini}, GPT-4V~\cite{gpt4v}, and Sphinx~\cite{lin2023sphinx}, which involves diverse qualitative samples and a quantitative benchmark, MME~\cite{fu2023mme}. For multi-faceted comparison, we carefully collect numerous samples covering different domains of visual understanding, including fundamental perception,
advanced cognition, challenging vision tasks, and various expert capacities. Each domain also contains several subtasks for in-depth discussion and analysis.

\subsection{Gemini vs GPT-4V} 
The qualitative results indicate that \textbf{\textit{Gemini is indeed a strong challenger to GPT-4V}}, given its superior multi-modal reasoning capacity. In most cases, Gemini achieves competitive answering accuracy compared to GPT-4V, and showcases different response styles and preferences. 

\paragraph{Differences.} For comparison, GPT-4V tends to generate more detailed descriptions of the perception tasks (Figures \ref{scene_image_1}, \ref{scene_image_2}, \ref{scene_image_3}, \ref{celebrity-1}, \ref{landmark-1}), and provide in-depth analysis with step-by-step intermediate reasoning for the cognition tasks (Figures \ref{table-1}, \ref{table-4}, \ref{AbstractVisualStimuli-1}, \ref{imgemo-1}, \ref{imgemo-5}). Instead, Gemini prefers to provide a direct and concise response to the answer, which helps the users to rapidly locate pertinent information. When there are a greater number of visual elements in the image (Figures \ref{scene_image_1}, \ref{scene_image_2}), the fine-grained perception advantages of GPT-4V become more pronounced, which provides more accurate recognition of visual details. However, GPT-4V, due to privacy concerns, may decline to respond to topics related to celebrities (Figures \ref{celebrity-2}, \ref{movie-1}, \ref{face} and the Celebrity metric on the MME benchmark~\cite{fu2023mme} shown in Figure \ref{fig_lidar}), or it may refrain from attempting to answer certain out-of-scope questions by anticipating its own limitations (Figures \ref{movie-3}, \ref{detection}). For some specific vision and expert-level tasks (Figures \ref{artwork-new-4}, \ref{artwork-3}, \ref{action-3}), Gemini normally demonstrates a broader range of learned knowledge and generalization capabilities, indicating better applicability across various domains.

\paragraph{Common issues.} There are four common issues of the two MLLMs. \textbf{1)} The first limitation is the spatial perception capabilities. From the qualitative perception examples (Figure \ref{spatial_1}) and quantitative results on the MME benchmark (the Position metric shown in Figure \ref{fig_lidar}), both Gemini and GPT-4V are not proficient in determining the relative positions of objects. \textbf{2)} The second issue is the unsatisfactory OCR (Figures \ref{table-3}, \ref{formula-2}) and abstract visual understanding (Figures \ref{Raven’WechslerAdultIntelligenceScale-1}, \ref{Raven’sProgressiveMatrices-1}). For example, they may misinterpret some of the numbers and characters in the diagrams or charts, and have difficulty comprehending some geometric shapes and abstract inductive abilities. \textbf{3)} The third inadequacy lies in the logical self-consistency within reasoning. For some scientific problems (Figure \ref{physics-3}) or `Yes or No' questions (Figure \ref{table-5}), they occasionally provide intermediate reasoning steps that do not align with or are contrary to the final answer. \textbf{4)} The fourth common issue concerns their robustness to prompt designs. As shown in Figures~\ref{table-5} and~\ref{math-6}, for different approaches to framing the same question prompt, GPT-4V and Gemini would be disturbed to generate opposite answers. Such a issue affects the output stability, and impedes their further applications.
We can see that \textbf{\textit{both Gemini and GPT-4V still struggle in many cases, showing the long road to the general MLLM.}}

\subsection{Gemini vs Sphinx} 

Despite that Sphinx is on par with GPT-4V and Gemini in some cases, it is not capable of generating as consistent high-quality answers as them. This demonstrates that \textbf{\textit{the open-source MLLMs still have some non-negligible gaps to closed-source models.}} 

\paragraph{Failure cases of Sphinx.}
We observe that the failure cases of Sphinx are mainly due to two reasons. \textbf{1)} The first is that the diversity of Sphinx's training data is still lacking in some domains, constraining its capability for a wider range of tasks, e.g., scientific knowledge perception (Figure~\ref{science_1}), visual code generation of HTML codes (Figure~\ref{html-1}), and abstract visual reasoning (Figure~\ref{Raven’WechslerAdultIntelligenceScale-1}).
This motivates us to further incorporate more data in diverse domains for training faithful open-source MLLMs.
\textbf{2)} The second limitation is the inherent reasoning upper bound of the underlying LLM. Sphinx adopts the vanilla LLaMA-2-7B~\cite{touvron2023llama} for initialization, which falls short compared to larger-scale LLMs (e.g., 70B models) in certain complex tasks.

\subsection{Future Directions}

With our comprehensive comparison and discussion, Gemini and GPT-4V are both pioneers of MLLMs in this era, showcasing sparks of artificial general intelligence~\cite{gpt4vdawn,bubeck2023sparks}. Looking ahead, future development of MLLMs can focus on three aspects: visual representation encoding (fine-grained appearances, spatial relation awareness), multi-modal alignment (alleviating hallucination, OCR accuracy), and LLMs' reasoning capacity (quantitative processing, logical self-consistency).  Overall, despite the exciting achievements made thus far, there still remains a considerable distance towards artificial general intelligence. We also anticipate the emergence of stronger and more comprehensive MLLMs in the future.